\documentclass{article}

\usepackage[final]{neurips_2025}

\newcommand{\ve}{\mathbf{v}}
\newcommand{\z}{\mathbf{z}}

\newcommand{\x}{\mathbf{x}}

\newcommand{\R}{\mathbb{R}}

\newcommand{\G}{\mathbf{G}}
\newcommand{\M}{\mathbf{M}}

\newcommand{\I}{\mathbf{I}}
\newcommand{\D}{\mathbf{D}}

\newcommand{\0}{\boldsymbol{0}}
\newcommand{\F}{\mathbf{F}}
\newcommand{\N}{\mathcal{N}}
\newcommand{\bepsilon}{\boldsymbol{\epsilon}}
\newcommand{\bmu}{\boldsymbol{\mu}_0}
\newcommand{\bSigma}{\boldsymbol{\Sigma}}
\newcommand{\bSigmaZ}{\boldsymbol{\Sigma}_0}
\newcommand{\bLambda}{\boldsymbol{\Lambda}_0}
\newcommand{\balpha}{\bar{\alpha}}
\newcommand{\bbalpha}{\bar{\boldsymbol{\alpha}}}

\newcommand{\DKL}{D_{\mathrm{KL}}}

\usepackage[dvipsnames]{xcolor}
\definecolor{darkgreen}{RGB}{0,100,0}

\newcommand{\evmatrix}{\mathbf{U}}
\newcommand{\T}{\mathsf{T}}
\newcommand{\evmatrixT}{\mathbf{U}^\T}
\newcommand{\projectedx}{\mathbf{v}}
\newcommand{\projectedmu}{\boldsymbol{\mu}_0^{\mathbf{u}}}

\newcommand{\projectedzt}{\boldsymbol{\z}_t^{\mathbf{u}}}
\newcommand{\projectedzi}{\boldsymbol{\z}_i^{\mathbf{u}}}

\usepackage{graphicx}
\usepackage{caption}
\usepackage{subcaption}
\usepackage{hyperref}
\usepackage{url}
\usepackage{wrapfig}
\usepackage{amsmath}
\usepackage{amssymb}
\usepackage{mathtools}
\usepackage{amsthm}

\usepackage{float}

\usepackage[utf8]{inputenc} % allow utf-8 input
\usepackage[T1]{fontenc}    % use 8-bit T1 fonts
\usepackage{hyperref}       % hyperlinks
\usepackage{url}            % simple URL typesetting
\usepackage{booktabs}       % professional-quality tables
\usepackage{amsfonts}       % blackboard math symbols
\usepackage{nicefrac}       % compact symbols for 1/2, etc.
\usepackage{microtype}      % microtypography
\usepackage{xcolor}         % colors

\theoremstyle{plain}
\newtheorem{theorem}{Theorem}[section]

\newtheorem{lemma}[theorem]{Lemma}

\theoremstyle{definition}

\theoremstyle{remark}

\title{Spectral Analysis of Diffusion Models with Application to Schedule Design}

\author{%
Roi Benita \\
Department of Electrical and Computer Engineering\\
Technion, Haifa, Israel \\
roibenita@campus.technion.ac.il
\And
Michael Elad \\
Department of Computer Science \\
Technion, Haifa, Israel \\
elad@cs.technion.ac.il
\And
Joseph Keshet \\
Department of Electrical and Computer Engineering\\
Technion, Haifa, Israel \\
jkeshet@technion.ac.il
}

\begin{document}
\maketitle
\begin{abstract}
Diffusion models (DMs) have emerged as powerful  tools for modeling complex data distributions and generating realistic new samples. Over the years, advanced architectures and sampling methods have been developed to make these models practically usable. However, certain synthesis process decisions still rely on heuristics without a solid theoretical foundation. 

In our work, we offer a novel analysis of the DM's inference process, introducing a comprehensive frequency response perspective. Specifically, by relying on Gaussianity assumption, we present the inference process as a closed-form spectral transfer function, capturing how the generated signal evolves in response to the initial noise. We demonstrate how the proposed analysis can be leveraged to design a noise schedule that aligns effectively with the characteristics of the data. The spectral perspective also provides insights into the underlying dynamics and sheds light on the relationship between spectral properties and noise schedule structure. Our results lead to scheduling curves that are dependent on the spectral content of the data, offering a theoretical justification for some of the heuristics taken by practitioners. 
\end{abstract}
% revealing
% \begin{abstract}
% Diffusion models (DMs) have emerged as powerful  tools for modeling complex data distributions and generating realistic new samples. Over the years, advanced architectures and sampling methods have been developed to make these models practically usable. However, certain synthesis process decisions still rely on heuristics without a solid theoretical foundation. 
% In our work, we offer a novel analysis of the DM's inference process, introducing a comprehensive frequency response perspective. Specifically, by relying on Gaussianity and shift-invariance assumptions, we present the inference process as a closed-form spectral transfer function, capturing how the generated signal evolves in response to the initial noise. 
% We demonstrate how the proposed analysis can be leveraged for optimizing the noise schedule, ensuring the best alignment with the original dataset's characteristics. Our results lead to scheduling curves that are dependent on the frequency content of the data, 
% offering a theoretical justification for some of the heuristics taken by practitioners. 
% \end{abstract}
\section{Introduction}
\label{Introduction}
Diffusion Models (DMs) have become powerful tools for generating high-quality, diverse signals, with applications in image, audio, and video synthesis. Alongside their practical success and the ability to handle complex distributions, some aspects of the diffusion processes still rely on heuristics rooted in empirical experimentation. A key example is choosing an appropriate noise schedule for the inference phase. Developing theoretical foundations for these heuristics may provide valuable insights into the diffusion process itself, and enable greater adaptation to different setups. Our work aims to provide such a theoretical backbone from a spectral perspective, as outlined below. 

% Diffusion Models (DMs) have become powerful tools for generating high-quality and diverse signals,  with applications ranging from image generation to audio, video synthesis, and more.
% Diffusion Models (DMs) have become powerful tools for generating high-quality and diverse signals, with applications such as image generation, % time-dependent domains including 
% audio and video synthesis, and more. Alongside their practical success and the ability to handle complex distributions, some aspects of the diffusion processes still rely on heuristics rooted in empirical experimentation. A key example is choosing an appropriate noise schedule for the inference phase. Developing theoretical foundations for these heuristics may provide valuable insights into the diffusion process itself, and enable greater adaptation to different setups. Our work aims to provide such a theoretical backbone, as outlined below. 
% Our starting point lies in the continuous-time formulation of DMs via SDEs and associated ODE flows \citep{song2020score}, which are mathematically well-founded, yet
% continuous-time
% its associated flow
% \textcolor{red}{Our starting point is the fact that,}
Our starting point lies in a well-established observation; while the continuous description of DMs via SDE or the associated flow ODE \citep{song2020score} are mathematically well-founded, their practice necessarily deviates from these theoretical foundations, introducing various errors \citep{chen2022sampling, pierret2024diffusion}. 
% A major source of error is discretization, which arises from replacing the analytical formulations with their discrete-time counterparts.
A major source of error is discretization, which replaces the analytical formulations by their discrete-time counterparts.
Another source of error is the approximation error, originating from the gap between the ideal denoiser and its neural network realization.

% other words for countparts: alternatives, substitues?

% Our starting point is the fact that, while the continuous-time description of DMs via SDE or ODE \citep{song2020score} may be mathematically sound and well-founded, their practice necessarily deviates from these theoretical foundations, introducing various errors \citep{chen2022sampling, pierret2024diffusion}. 
% A major source of this error is discretization,
% which replaces the DMs' SDE/ODE  formulations by their discrete-time approximations. 
% Another source of error is the approximation error, originating from the gap between the ideal denoiser and its neural network realization.

% in recent work 

 % and adapt DMs \textcolor{red}{for real-world applications}
% importance

Significant efforts have been made to minimize these associated errors, with advanced numerical solvers \citep{song2020score, jolicoeur2021gotta, zhang2022fast, liu2022pseudo, lu2022dpm, zheng2023dpm, zhao2024unipc} offering various methods for better treating the discretization of the differential equations. A fundamental aspect is the decision on the time point discretization, which directly affects the synthesis quality. We refer to these anchor points as the \emph{noise schedule}, highlighting their role in setting the noise variance introduced at each diffusion step. Realizing their importance, researchers have recently shifted their focus from custom-tailored heuristics \citep{ho2020denoising, nichol2021improved, karras2022elucidating, chen2023importance} to the development of optimized noise schedules \citep{sabour2024align,tong2024learning,watson2022learning, wang2023learning, xia2023towards,chen2024trajectory, xue2024accelerating, williams2024score}. More on these methods and their relation to this paper's contributions is detailed in Section \ref{Sec:Related Work}.

In this work, we analyze the reverse diffusion process in the spectral domain, presenting the generated output signal as the outcome of a linear transfer function operating on the i.i.d. Gaussian input noise.
This analysis is enabled by assuming that the destination distribution to sample from is Gaussian, providing an explicit spectral expression of the generated signal and insight on the evolving dynamics throughout the process. Our analysis, applied to the discretized inference procedure, encompasses both DDPM \cite{ho2020denoising} and DDIM \cite{song2020denoising} frameworks, thereby revealing the effects of deviating from the continuous formulation. In addition, we extend our study for both variance preserving (VP) and variance exploding (VE) numerical schemes \citep{song2020score}.

Posing the derived explicit expressions of the transfer systems as functions of the noise scheduling parameters, we can optimize a noise schedule tailored to a given dataset, its resolution, and the specified number of sampling steps required. We demonstrate how effectively solving these optimization problems numerically yields a noise schedule that accounts for these data characteristics, discuss the relation between the noise schedule structure and spectral properties, and validate our approach with existing works.
Finally, we apply our method to publicly available datasets, including CIFAR-10 \cite{krizhevsky2009learning}, AFHQv2 \cite{choi2020starganv2}, MUSIC \cite{moura2020music} and SC09 \citep{warden2018speech}, examining the relation to heuristic choices in past work.
% \red{and their improvement}.

% Finally, we demonstrate our method using publicly available datasets, 
% discuss their impact on the resulting schedule 
% and validate \textcolor{red}{our approach by comparing it with existing works. }

% Finally, we apply the found scheduling to real-world scenarios

% Posing the derived explicit expressions of the transfer systems as functions of the noise scheduling parameters, we may optimize a noise schedule tailored to a given dataset, its resolution, and the specified number of sampling steps required. We demonstrate how effectively solving these optimization problems numerically yields a noise schedule that accounts for these data characteristics, discuss their impact on the resulting schedule and validate our approach by comparing it with existing works. Finally, we apply the found scheduling to real-world scenarios using publicly available datasets, such as MUSIC  \cite{moura2020music} and SC09 \citep{warden2018speech}, and demonstrate the relation to heuristic choices in past work and their improvement.

In summary, our contributions are the following:
(i) Assuming the target signal is drawn from a Gaussian distribution, we present a novel spectral analysis on the discrete diffusion reverse process and derive a closed-form expression for its spectral transfer function. (ii) We formulate an optimization problem to find an optimal noise schedule that aligns with the data characteristics. Our approach provides an effective solution without relying on bounds or constraints on the number of diffusion steps.
(iii) We analyze the evolving spectral properties of the signal throughout the diffusion process and at its output, demonstrating how handcrafted noise scheduling decisions and related diffusion phenomena are often well-predicted by our approach.
(iv) Our spectral analysis examines various setups, including DDIM and DDPM procedures, VP and VE formulations, the selection of loss functions, and additional features such as expectancy drift.

% (iii) We compare our approach to existing work, showing that handcrafted noise scheduling decisions and related phenomena in the diffusion processes are often well-predicted by our approach.

% diffusion dynamics and their connection to frequency components and feature 

 % (iii) We compare our approach to existing work, showing that handcrafted noise scheduling decisions and related phenomena in the diffusion processes are often well-predicted by our approach.

% We demonstrate how our method allows to practically examine the evolving properties of the signal throughout and 

% In summary, our contributions are the following:
% (i) Assuming a Gaussian distributed dataset, we present a novel spectral perspective on the discrete diffusion reverse process and derive a closed-form expression for its frequency transfer function. (ii) We formulate an optimization problem to find an optimal noise schedule that aligns with the dataset's characteristics. Our approach provides an effective solution without relying on bounds or constraints on the number of diffusion steps. 
% (iii) We compare our approach to existing work, showing that handcrafted noise scheduling decisions and related phenomena in the diffusion processes are often well-predicted by our approach. (iv) Our spectral analysis examines various setups, including DDIM and DDPM procedures, VP and VE formulations, the selection of loss functions, and additional features such as expectancy drift.  

\section{Background}
\label{Sec:Background}

We introduce the notations and the framework of diffusion probabilistic models, which are designed to generate samples $\x_0\in\R^d$ from an underlying, unknown probability distribution $p(\x_0)$. While the sampling procedure can be described as a Stochastic Differential Equation (SDE) or by its associated flow, an Ordinary Differential Equation (ODE), these formulations lack a general analytical solutions and are instead discretized and solved using numerical methods \cite{song2020score}.
% do not have
 % While the diffusion process can be described as a Stochastic Differential Equation (SDE) or Ordinary Differential Equation (ODE), these formulations do not have general analytical solutions and are instead discretized and solved using numerical methods \cite{song2020score}.
Accordingly, we turn to describe the discrete formulations  -- DDPM \cite{ho2020denoising} and DDIM \cite{song2020denoising} -- which stand as the basis for our work.

The diffusion process is a generative procedure constructed from two stochastic paths: a \emph{forward} and a \emph{reverse} trajectories in which data flows \cite{ho2020denoising}. Each process is defined as a fixed Markovian chain composed of $T$ latent variables. 
During the forward process, a signal instance is gradually contaminated with white additive Gaussian noise as
follows:\footnote{Here we adopt the variance-preserving (VP) setting; see Appendix \ref{sec:appendix_VP_VE} for the variance-exploding (VE) form.}
% \footnote{Our analysis is focused here on the variance preserving approach. We refer the reader to a similar analysis of the variance exploding approach in Appendix \ref{sec:appendix_VP_VE}.}
% \footnote{Our analysis focuses on the variance-preserving case; for the variance-exploding counterpart, see Appendix~\ref{sec:appendix_VP_VE}.}
\begin{equation}\label{eq:forward_DDPM}
    \x_t = \sqrt{\alpha_t} \x_{t-1} + \sqrt{1-\alpha_t} \bepsilon_t ~,
\end{equation}
where $\alpha_t $ for $t\in[1,T]$ is referred to as the incremental noise schedule and $\bepsilon_t \sim \N(\mathbf{0},\I)$. Under the assumption that $\alpha_T$ is close to zero, we get that the final latent variable becomes \(x_T \sim \mathcal{N}(\mathbf{0}, \I)\). 
A direct consequence of the above equation is an alternative relation of the form: 
\begin{equation}\label{eq:marginal_dist}
    \mathbf{x}_t = \sqrt{\balpha_t} \mathbf{x}_0 + \sqrt{1-\balpha_t} \boldsymbol{\epsilon} ~~~~~  \boldsymbol{\epsilon} \sim \N(\mathbf{0},\I) ~,
\end{equation}
for $\balpha_t = \prod_{i=1}^{t} \alpha_i$.
Based on the above relationships, the reverse process  aims to reconstruct $\x_0$ from the noise $\x_T$ by progressively denoising it. This can be written as
\begin{equation}
\mathbf{x}_{t-1} = \frac{1}
{\sqrt{{\alpha}_t}}\left(\mathbf{x}_t - \frac{1-\alpha_t}{\sqrt{1-\balpha_t}}\boldsymbol{\epsilon}_\theta(\mathbf{x}_t, t)\right) + \sigma_t \mathbf{z}_t ~, 
\end{equation}
where $\bepsilon_{\theta}$ is the estimator of $\bepsilon$ for a given $\x_t$ at time $t$ with a neural network parameterized by $\theta$, $\sigma_t=\sqrt{\frac{1-\bar{\alpha}_{t-1}}{1-\bar{\alpha}_{t}}(1-\alpha_t)}$ ~ and ~ $\mathbf{z}_t \sim \N(\mathbf{0},\I)$.
% . \red{In the above expression}
% Alongside this stochastic formulation, \citet{song2020denoising} provides a deterministic framework for the sampling process, which can be utilized to enable faster sampling. 
% DDIM models the forward process as a non-Markovian one, while preserving the same marginal distribution as in \eqref{eq:marginal_dist}. As a result, the reverse process can be expressed by\footnote{We follow here the DDIM notations that replaces $t$ with $s$.}
Alongside this stochastic formulation, The authors of \cite{song2020denoising} introduce DDIM, a deterministic framework for the inference process, which can be utilized to enable faster sampling. 
% \textcolor{blue}{While DDIM retains the same Markovian forward process as DDPM, preserving the marginal distribution in \eqref{eq:marginal_dist}, it introduces a 
 % deterministic and Non-Markovian reverse process.  }
While DDIM relies on a non-Markovian forward process, it preserves the marginal distribution as in \eqref{eq:marginal_dist}.
% DDIM models the reverse process as a non-Markovian one, while preserving the same marginal distribution as in \eqref{eq:marginal_dist}. 
As a result, the sampling trajectory can be expressed by
% \footnote{We follow here the DDIM notations that replaces $t$ with $s$.}
% Alongside this stochastic formulation, \citet{song2020denoising} propose 
% % introduce DDIM,
% a deterministic sampling framework, which enables faster generation. While DDIM retains the same Markovian forward process as in DDPM, thereby preserving the marginal distribution in \eqref{eq:marginal_dist}, it introduces a deterministic and non-Markovian reverse process. As a result, the sampling trajectory can be expressed by\footnote{We follow here the DDIM notations that replace $t$ with $s$.}
\begin{equation}
\label{eq:DDIM_sampling_procedure}
    \x_{s-1} = \sqrt{\bar{\alpha}_{s-1}} \left( 
    \frac{\x_{s} - \sqrt{1 - \bar{\alpha}_{s}} \cdot \bepsilon_\theta(\x_{s},s)}{\sqrt{\bar{\alpha}_{s}}}\right)
     \\ + \sqrt{1 - \bar{\alpha}_{s-1}} \cdot \bepsilon_\theta(\x_{s},s),
\end{equation}
% \begin{multline}\label{eq:DDIM_sampling_procedure}
%     \x_{s-1} = \sqrt{\bar{\alpha}_{s-1}} \left( 
%     \frac{\x_{s} - \sqrt{1 - \bar{\alpha}_{s}} \cdot \bepsilon_\theta(\x_{s},s)}{\sqrt{\bar{\alpha}_{s}}}\right)
%      \\ + \sqrt{1 - \bar{\alpha}_{s-1}} \cdot \bepsilon_\theta(\x_{s},s),
% \end{multline}

% where $s \in \left[0, S\right]$. 
where $s \in \{0, 1, \dots, S\}$ is the time index in the DDIM formulation.
Throughout the rest of the paper, we denote by $\bbalpha$ the set of noise schedule parameters, $\left\{ {{\balpha}_t} \right\}_{t=0}^T$ for DDPM and $\left\{ {{\balpha}_s} \right\}_{s=0}^S$ for DDIM.

\section{Analysis of Diffusion processes} \label{seq:mathematical_derivative}
% frequency
We consider the reverse process as a system that takes as input a noisy signal $\x_T$ and outputs 
$\x_0$. In this section, we develop the transfer function, which characterizes the relationship between these inputs and outputs in the spectral domain. To do so, we assume that the output signals are vectors drawn from a Gaussian distribution, 
\begin{equation}
\x_0 \sim \N(\bmu, \bSigmaZ)~,      
\end{equation}
where $\bmu \in \mathbb{R}^{d}$ and $\bSigmaZ \in \mathbb{R}^{d\times d}$. A similar assumption was used in previous work \citep{pierret2024diffusion, sabour2024align, wang2024unreasonable}. 
% \textcolor{red}{While this model greatly simplifies the signal's distribution, we will demonstrate that it allows us to design a noise scheduling mechanism for different objectives.}
% While this model greatly simplifies the signal's distribution, we will demonstrate that it facilitates spectral analysis of the signal's properties throughout the diffusion process and enables the design of noise scheduling mechanisms for different objectives.
While this model simplifies the signal distribution, we will demonstrate that it facilitates spectral analysis throughout the diffusion process and allows us to design a noise scheduling mechanism for different objectives.
% \textcolor{blue}{While this model greatly simplifies the signal distribution, we will demonstrate that it facilitates the analysis of the signal's spectral characteristics throughout the diffusion process and allows us to design a noise scheduling mechanism for different objectives.}
\subsection{The optimal denoiser for a Gaussian input}
The following theorem states a well-known fact: under the above Gaussianity assumption, the Minimum Mean-Squared Error (MMSE) denoiser operating on $\x_t$ to recover $\x_0$ is linear. It is the Wiener Filter 
\cite{wiener1949} and can be expressed in a closed form. 
 % ; it is the Wiener Filter \cite{wiener1949}, given as a closed-form expression
 
% \begin{theorem} \label{theorem:optimal_demoiser}
% Let $\x_0 \sim \N(\bmu, \bSigmaZ)$ and let $\x_t$ %the marginal distribution 
% be defined by \eqref{eq:marginal_dist}. Then, the denoised signal obtained from the MMSE (and the MAP) denoiser is given by:
% \begin{equation}
% \label{eq:wiener_filter}
% \x^*_0  =  \left(\balpha_t \bSigmaZ + (1-\balpha_t)\I \right)^{-1}\left(\sqrt{\balpha_t}\bSigmaZ \x_t + \left(1-\balpha_t\right) \bmu \right).
% \end{equation}
% % \begin{eqnarray}\label{eq:wiener_filter}
% % \x^*_0 & = & \left(\balpha_t \bSigmaZ + (1-\balpha_t)\I \right)^{-1} \\ \nonumber && \hspace{0.4in}
% % \left(\sqrt{\balpha_t}\bSigmaZ \x_t + \left(1-\balpha_t\right) \bmu \right).
% % \end{eqnarray}
% \end{theorem}
% A detailed proof is given in Appendix~\ref{sec:appendix_Map_est}. Here, we outline the main steps. The MAP estimator seeks to maximize the posterior probability:
% \begin{equation}\nonumber
% \max_{\x_0} \, \log p(\x_0 | \x_t) =  \min_{\x_0} \,  -\log p(\x_t | \x_0) -\log p(\x_0).
% \end{equation}
% By substituting the explicit density functions $p(\x_0)$ and $p(\x_t | \x_0)$ according to \eqref{eq:marginal_dist} into the above, and differentiating with respect to $\x_0$ we obtain the desired result. Under the assumption of Gaussian distributions, applying the MAP estimator is equivalent to minimizing the MSE, as both yield the same solution.

% ####

\begin{theorem} \label{theorem:optimal_demoiser}
Let $\x_0 \sim \N(\bmu, \bSigmaZ)$ and let $\x_t$ %the marginal distribution 
be defined by \eqref{eq:marginal_dist}. Then, the denoised signal obtained from the MMSE (and the MAP) denoiser is given by:
\begin{equation}\label{eq:wiener_filter}
\x^*_0  =  \left(\balpha_t \bSigmaZ + (1-\balpha_t)\I \right)^{-1} \left(\sqrt{\balpha_t}\bSigmaZ \x_t + \left(1-\balpha_t\right) \bmu \right).
\end{equation}
\end{theorem}
A detailed proof is given in Appendix~\ref{sec:appendix_Map_est}. Here, we outline the main steps under the assumption of Gaussian distributions, where applying the Maximum A Posteriori (MAP) estimator is equivalent to minimizing the MSE.
% Here, we outline the main \red{steps:} Under the assumption of Gaussian distributions, applying the Maximum A Posteriori (MAP) estimator is equivalent to minimizing the MSE. 
The MAP estimator seeks to maximize the posterior probability:
\begin{equation}
\max_{\x_0} \, \log p(\x_0 | \x_t) =  \min_{\x_0} \,  -\log p(\x_t | \x_0) -\log p(\x_0).
\end{equation}
By substituting the explicit density functions $p(\x_0)$ and $p(\x_t | \x_0)$ according to \eqref{eq:marginal_dist} into the above, and differentiating with respect to $\x_0$ we obtain the desired result.

% #####3

% maximizing the MAP estimator is equivalent to minimizing the Mean Squared Error (MSE), as both yield the same solution.

 % the MAP estimator is equivalent to MMSE, as both yield the same solution.

 % We now turn to analyze the discrete sampling procedures, as introduced by \citet{ho2020denoising} and \citet{song2020denoising}, and described in Section \ref{Sec:Background}. 
 
\subsection{The reverse process in the time domain}
 % presented 
We now turn to analyze the discrete sampling procedures, as introduced in \cite{ho2020denoising, song2020denoising} and described in Section \ref{Sec:Background}. We begin by focusing on the DDIM formulation presented in \cite{song2020denoising}, as it highlights the fundamental principles more clearly and facilitates the analysis of faster sampling techniques. 
The lemma below describes the relationship between two adjacent time steps during the inference process.
\begin{lemma}\label{lemma:ddim_time_infer}

Assume $\x_0 \sim \N(\bmu, \bSigmaZ)$ and let $\bbalpha$ be the noise schedule parameters, we have 
\begin{equation}
\label{eq:ddim_time_infer}
\mathbf{x}_{s-1} = \left(a_s\I + b_s\sqrt{\balpha_s} \bar{\bSigma}^{-1}_{0,s}\bSigma_0 \right) \x_s \\ + 
b_s(1-\balpha_s) \bar{\bSigma}^{-1}_{0,s}\boldsymbol{\mu_0},
\end{equation}
% \begin{multline}\label{eq:ddim_time_infer}
% \mathbf{x}_{s-1} = \left(a_s\I + b_s\sqrt{\balpha_s} \bar{\bSigma}^{-1}_{0,s}\bSigma_0 \right) \x_s \\ + 
% b_s(1-\balpha_s) \bar{\bSigma}^{-1}_{0,s}\boldsymbol{\mu_0},
% \end{multline}
where $\bar{\boldsymbol{\Sigma}}_{0,s} = \balpha_s {\boldsymbol{\Sigma}}_0 + (1-\balpha_s)\I$, and the coefficients are
\begin{equation}\nonumber
a_s = \frac{\sqrt{1 - \balpha_{s-1}}}{\sqrt{1-\balpha_s}} ~~,  ~~ b_s =   \sqrt{\balpha_{s-1}} - \frac{\sqrt{{\balpha}_s}\sqrt{1 - \balpha_{s-1}}}{\sqrt{1 - \balpha_{s}}}~.
\end{equation}
\end{lemma}
The above is obtained by plugging the optimal denoiser into the DDIM reverse process in \eqref{eq:DDIM_sampling_procedure}. The derivation is given in Appendix \ref{sec:appendix_inference_Time}. This lemma establishes an explicit connection between adjacent diffusion steps, incorporating the characteristics of the destination signal density function as expressed by $\bmu$ and $\bSigma_0$, along with the chosen noise schedule parameters  $\bbalpha$.

\subsection{Migrating to the spectral domain} \label{sec:migrating_to_the_spectral_domain}
Analyzing the diffusion process in the time domain can be mathematically and computationally challenging, particularly in high-dimensional spaces $(d\gg 1)$. To address this, we leverage the spectral decomposition of the covariance matrix, simplifying the analysis and enabling the examination of the signal behavior along its eigen-directions.
% \textcolor{blue}{we leverage the spectral (PCA) domain representation, simplifying the analysis and enabling the examination of the signal behavior along its principal components.}

% Consider a destination signal $\mathbf{x}_0$ drawn from a multivariate Gaussian distribution with a fixed mean and covariance matrix $\Sigma$. Let $\mathbf{U}$ be the orthonormal matrix whose columns are the eigenvectors (principal components) of $\Sigma$. Then, the projection of $\mathbf{x}_0$ onto this basis, denoted $\mathbf{z}_0 = \mathbf{U}^\top \mathbf{x}_0$, also follows a Gaussian distribution.

Consider a destination signal $\mathbf{x}_0$ drawn from a multivariate Gaussian distribution with a given mean $\bmu$ and a covariance matrix $\boldsymbol{\Sigma}_0$, and  let $\mathbf{U} \in \mathbb{R}^{d\times d}$ be a unitary matrix whose columns are the eigenvectors of $\boldsymbol{\Sigma}_0$. The projection of $\mathbf{x}_0$ onto this eigenbasis, denoted by $\mathbf{v}_0 = \mathbf{U}^\top \mathbf{x}_0$, also follows a Gaussian distribution. Specifically, $\mathbf{v}_0 \sim \mathcal{N}(\boldsymbol{\mu}_0^{\mathbf{u}}, \boldsymbol{\Lambda_0})$, where $\boldsymbol{\mu}_0^{\mathbf{u}} = \mathbf{U}^\top \boldsymbol{\mu}_0\in\mathbb{R}^{d}$ is the transformed mean vector, and $\bLambda \in \mathbb{R}^{d\times d}$ is a positive semi-definite  diagonal matrix, containing the eigenvalues of $\bSigma_0$, denoted $\{ \lambda_i \}_{i=1}^{d}$.
Based on Lemma  \ref{lemma:ddim_time_infer}, we derive the following property:
% \begin{lemma}\label{lemma:ddim_joint_diagonalization}
% Assume $\x_0 \sim \mathcal{N}(\bmu, \bSigma_0)$, and let $\{\x_s\}_{s=0}^S$ denote the latent variables of the diffusion process defined by a given noise schedule $\bbalpha$. Then, the covariance matrices of all intermediate variables $\x_s$ are jointly diagonalizable using the eigenbasis $\mathbf{U}$ of $\bSigma_0$.
% \end{lemma}
% \red{and let $\bbalpha$ be the noise schedule parameters.}
\begin{lemma}\label{lemma:ddim_joint_diagonalization}
Assume $\x_0 \!\sim \!\N(\bmu,\! \bSigmaZ)$. The covariance matrices associated with all intermediate steps  $\{\x_s\}_{s=0}^S$ in the diffusion process are jointly diagonalizable using the eigenbasis $\mathbf{U}$ of $\boldsymbol{\Sigma}_0$.
\end{lemma}
This lemma holds since all the diffusion steps in \eqref{eq:ddim_time_infer} involve only $\boldsymbol{\Sigma}_0$, its inverse, or the addition of the identity matrix, leading to covariance matrices that share the same eigenbasis. We now turn to describe the diffusion reverse process in the spectral domain. By projecting both sides of \eqref{eq:ddim_time_infer} onto the eigenbasis $\evmatrix$, we obtain the following result.
% where $\bSigmaZ$ is a circulant matrix
% frequency
\begin{lemma}\label{Lemma:migrating_spectral_domain}
Assume $\x_0 \sim \N(\bmu, \bSigmaZ)$  and let $\bbalpha$ be the noise schedule parameters. The subsequent step in the reverse process can be expressed in the spectral domain via
% \begin{equation}\label{eq:ddim_frequency_infer}
% \mathbf{x}_{s-1}^{\mathcal{F}} = \G(s) \mathbf{x}_s^{\mathcal{F}} + \M(s)
% \boldsymbol{\mu_0}^{\mathcal{F}},
% \end{equation}
% where $\x_s^\mathcal{F}$ denotes the DFT of the signal $\x_s$, 
\begin{equation}\label{eq:ddim_frequency_infer}
\mathbf{v}_{s-1} = \G(s)\mathbf{v}_s + \M(s)
\boldsymbol{\mu^{u}_0},
\end{equation}
where $\ve_s$ denotes the projection of the signal $\x_s$ onto the eigenbasis $\mathbf{U}$, 
$$
\G(s) = \left[ a_s\I  + b_s \sqrt{\balpha_s}\left[\balpha_s \mathbf{\Lambda}_0 + (1-\balpha_s) \I\right]^{-1} \mathbf{\Lambda}_0 \right] 
~ \text{and} ~~
\M(s) = b_s(1-\balpha_s)\left[\balpha_s \mathbf{\Lambda}_0 + (1-\balpha_s)\I\right]^{-1}.
$$
% \begin{equation}
% \G(s) = \left[ a_s  + b_s \sqrt{\balpha_s}\left[\balpha_s \mathbf{\Lambda}_0 + (1-\balpha_s) \I\right]^{-1} \mathbf{\Lambda}_0 \right]  
% \end{equation}
% and
% \begin{equation}
% \M(s) = b_s(1-\balpha_s)\left[\balpha_s \mathbf{\Lambda}_0 + (1-\balpha_s)\I\right]^{-1} ~.
% \end{equation}
% frequency
\end{lemma}
The lemma is proven in Appendix \ref{sec:appendix_inference_spectral}. Equation \eqref{eq:ddim_frequency_infer} describes the relationship between two consecutive steps in the reverse process. Note that both matrices, $\G(s)$ and $\M(s)$, are diagonal, and thus, the reverse process in the spectral domain turns into a system of $d$ independent scalar equations. Based on the above relationship, % between two adjacent diffusion steps in \eqref{eq:ddim_frequency_infer},
% frequency
% $\mathbf{\hat{x}}_{0}^{\mathcal{F}}$.
we may derive an expression for the generated  signal in the spectral domain, denoted as $\mathbf{\hat{\ve}}_{0}$. The complete derivation of the following result can be found in Appendix \ref{sec:appendix_inference_spectral} as well.
% where $\bSigmaZ$ is a circulant matrix
% $\hat{\x}_0^{\mathcal{F}}$
% $\mathbf{x}_S^{\mathcal{F}}$
\begin{theorem}
Assume $\x_0 \sim \N(\bmu, \bSigmaZ)$  and let $\bbalpha$ be the noise schedule parameters.  The generated signal in the frequency domain $\hat{\ve}_0$ can be described as a function of $\ve_S$ via
% \begin{align}\label{eq:D_1,D_2}
% \mathbf{\hat{x}}_{0}^{\mathcal{F}} &= \D_1 \mathbf{x}_S^{\mathcal{F}} + \D_2\boldsymbol{\mu_0}^{\mathcal{F}},\end{align}
% $\text{where} ~~ \mathbf{x}_S^{\mathcal{F}} \sim \N(\0, \I) ~~ , ~~
%  \D_1 =  \prod_{k=1}^{S}\G(k)  ~~      ~~ \text{and}~~\D_2 =\sum_{i=1}^{S}\left(\prod_{j=1}^{i-1}\G(j)\right)\M(i)  ~.$
% \\ Moreover, $\mathbf{\hat{x}}_{0}^{\mathcal{F}}$ follows Gaussian distribution:
% \begin{equation}\label{eq:DDIM_dist_freq}
% \mathbf{\hat{x}}_{0}^{\mathcal{F}} \sim \N(\D_2 \boldsymbol{\mu_0}^{\mathcal{F}}, \D_1^2), \quad \mathbf{\hat{x}}_{0}^{\mathcal{F}} \in \mathbb{R}^{d}.
% \end{equation} 
% \end{theorem}
\begin{align}\label{eq:D_1,D_2}
\mathbf{\hat{\ve}}_{0} &= \D_1 \ve_S + \D_2\boldsymbol{\mu^{u}_0},\end{align}
$\text{where} ~~ \ve_S \sim \N(\0, \I) ~~ , ~~
 \D_1 =  \prod_{k=1}^{S}\G(k)  ~~      ~~ \text{and}~~\D_2 =\sum_{i=1}^{S}\left(\prod_{j=1}^{i-1}\G(j)\right)\M(i)  ~.$
% \begin{equation}\label{eq:D_1,D_2}
% \mathbf{\hat{x}}_{0}^{\mathcal{F}} = \D_1 \mathbf{x}_S^{\mathcal{F}} + \D_2\boldsymbol{\mu_0}^{\mathcal{F}}, 
% \end{equation}
% where $\mathbf{x}_S^{\mathcal{F}} \sim \N(\0, \I)$ ~~ and
% \begin{equation}\nonumber
%  \D_1 =  \prod_{k=1}^{S}\G(k)  \quad\text{    ,   }\quad \D_2 =\sum_{i=1}^{S}\left(\prod_{j=1}^{i-1}\G(j)\right)\M(i)  ~.
% \end{equation} 
\\ Moreover, $\hat{\ve}_{0}$ follows Gaussian distribution:
\begin{equation}\label{eq:DDIM_dist_freq}
\hat{\ve}_{0} \sim \N(\D_2 \boldsymbol{\mu^{u}_0}, \D_1^2), \quad \hat{\ve}_{0} \in \mathbb{R}^{d}.
\end{equation} 
\end{theorem}
% frequency
Equation \eqref{eq:D_1,D_2} provides a novel view of the generated signal in the spectral domain. Specifically, we can view \eqref{eq:D_1,D_2} as a \emph{transfer function} which models the relationship between the input signal $\ve_S$ and the output $\hat{\ve}_{0}$. Furthermore, since the matrices $\D_1$ and $\D_2$ are diagonal, the expression simplifies to a set of $d$ scalar \emph{transfer functions}, with the only parameters being the noise schedule, $\bbalpha$. 

So far, we examined DDIM. Similarly, a closed-form expression for the stochastic DDPM method \citep{ho2020denoising} is presented in the following Theorem and is proven in Appendix \ref{sec:appendix_DDPM_Formulation}.
% where $\bSigmaZ$ is a circulant matrix 
% \begin{equation}\label{eq:D1D2-DDPM}
% \mathbf{\hat{x}}_{0}^{\mathcal{F}} = \D_1 \mathbf{x}_T^{\mathcal{F}} + \sum_{i=1}^{T}\left(\prod_{j=1}^{i-1}\G(j)\right)c_i \mathbf{z}^{\mathcal{F}}_i  + \D_2\boldsymbol{\mu_0}^{\mathcal{F}}
% \end{equation}
\begin{theorem}\label{theorem:ddap_spectral_eq}
Assume $\x_0 \sim \N(\bmu, \bSigmaZ)$ and let $\bbalpha$ be the noise schedule parameters. The signal $\hat{\ve}_{0}$ generated by DDPM can be expressed as a function of $\ve_T$ via
\begin{equation}\label{eq:D1D2-DDPM}
\hat{\ve}_{0} = \D_1 \ve_T + \sum_{i=1}^{T}\bigg(\prod_{j=1}^{i-1}\G(j)\bigg)c_i \mathbf{z}^{\boldsymbol{u}}_i  + \D_2\boldsymbol{\mu^{u}_0}
\end{equation}
where~ 
$\ve_T \sim \N(\0, \I)$,  $c_i = \sqrt{\frac{1-\bar{\alpha}_{i-1}}{1-\bar{\alpha}_i}(1-\alpha_i)}~~$. The terms $\D_1, \D_2$ and the matrices $\G(j)$ are defined in Appendix \ref{sec:appendix_DDPM_Formulation}.
Moreover, $\hat{\ve}_{0}$ follows a Gaussian distribution:
\begin{equation}
\hat{\ve}_{0}\sim \N\bigg( \D_2\boldsymbol{\mu^{u}_0} ~,~ \D_1^2 + \sum_{i=1}^{T}\Big(\prod_{j=1}^{i-1}\G^2(j)\Big)c^2_i \I \bigg).
\end{equation} 
\end{theorem}

\section{Optimal spectral schedules}\label{subsec:establish_optimal_spectral_noise_schedule}

With the closed-form expressions in \eqref{eq:D_1,D_2} and \eqref{eq:D1D2-DDPM}, we can now explore different aspects of the diffusion process and examine how subtle changes in its design affect the output distribution. More specifically, a key aspect in this design is the choice of the noise schedule. In the discussion that follows we demonstrate how the proposed scheme enables optimal scheduler design.

% We define the probability density function of the output of the diffusion process in the spectral domain as $p(\hat{\ve}_{0};
% circulant
We start by focusing on the direct dependence between the generated distribution and the noise schedule coefficients, $\bbalpha$. We define the spectral domain probability density function of the diffusion process output as $p(\hat{\ve}_{0}; \bbalpha)$. Our objective is to bring this distribution to become as close as possible to the original distribution, $p(\ve_0)$. 
Specifically, given a dataset with a covariance matrix, $\bSigma_0$, defined by the eigenvalues $\{\lambda_i\}_{i=1}^d$ and $S$ diffusion steps, our goal is to identify the coefficients $\bbalpha$ that minimize some distance $\mathcal{D}$ between these two distributions. This results in the following optimization problem with a set of specified constraints:
\begin{align}
\label{eq:optimization_problem}
\bbalpha^* = &\arg\min_{\bbalpha}  \mathcal{D}\left(p(\hat{\ve}_{0}; \bbalpha),p(\ve_0)\right)  \\ \nonumber
&~\text{subject to} ~~~~ \balpha_0 = 1-\varepsilon_0, ~~~ \balpha_S =\varepsilon_S,  \\ \nonumber
% {{\balpha}_S} = 4.035 \times 10^{-5}, \quad {{\balpha}_0} =9.99 \times 10^{-1},\\
& ~~~~~~~~~~~~~~~~~~~~~\balpha_{s-1} \geq \balpha_s ~~ \text{for} ~~ s = 1, \ldots, S.
\end{align}

% \begin{align}
% \label{eq:optimization_problem}
% \bbalpha^* = &\arg\min_{\bbalpha} \mathcal{D}\left(p(\mathbf{\hat{x}_{0}^{\mathcal{F}}}; \bbalpha),p(\mathbf{{x}_{0}^{\mathcal{F}}})\right)  \\ \nonumber
% &~\text{subject to} ~~~~ \balpha_0 = 1-\varepsilon_0, ~~~ \balpha_S =\varepsilon_S,  \\ \nonumber
% % {{\balpha}_S} = 4.035 \times 10^{-5}, \quad {{\balpha}_0} =9.99 \times 10^{-1},\\
% & ~~~~~~~~~~~~~~~~~~~~~\balpha_{s-1} \geq \balpha_s ~~ \text{for} ~~ s = 1, \ldots, S.
% \end{align}

% \begin{multline}
% \label{eq:optimization_problem}
% \bbalpha^* = \arg\min_{\bbalpha} \mathcal{D}\left(p(\mathbf{\hat{x}_{0}^{\mathcal{F}}}; \bbalpha),p(\mathbf{{x}_{0}^{\mathcal{F}}})\right)  \\
% \text{subject to} ~~~ \balpha_0 = 1-\varepsilon_0, ~~~ \balpha_S =\varepsilon_S,  \\
% % {{\balpha}_S} = 4.035 \times 10^{-5}, \quad {{\balpha}_0} =9.99 \times 10^{-1},\\
% \balpha_{s-1} \geq \balpha_s ~ \text{for} ~ s = 1, \ldots, S.
% \end{multline}
The equality constraints ensure compatibility between the training and the synthesis processes. This involves ending the diffusion process with white Gaussian noise and starting it with very low noise to capture fine details in the objective distribution accurately \cite{lin2024common}. The inequality constraints align with the core principles of diffusion models and their gradual denoising process \cite{ho2020denoising}. 
The distance $\mathcal{D}$ between the probabilities can be chosen depending on the specific characteristics of the task. In this work, we consider the \emph{Wasserstein-2} and \emph{Kullback-Leibler divergence}, but other distances can also be used. The theorems presented below are detailed in Appendix \ref{sec:appendix_loss_functions}. 

The \emph{Wasserstein-2} distance (or Earth Mover's Distance) measures the minimal cost of transporting mass to transform one probability distribution into another. In the case of measuring a distance between two Gaussians, this has a closed-form expression. 
\begin{theorem}\label{thm:wasserstein}
The \emph{Wasserstein-2} distance $\mathcal{D}_{W_2}$ between $p(\hat{\ve}_{0}; \bbalpha)$ and $p(\ve_0)$  is given by:
\begin{equation}
% \begin{empheq}[box=\fbox]{align}
\mathcal{D}_{W_2}^2 \big({p}(\hat{\ve}_{0}; \bbalpha), p(\ve_0)\big) = 
\sum_{i=1}^{d} \left( \sqrt{\lambda_i} - 
 [\D_1]_i  \right)^2 \\ + \sum_{i=1}^{d} [\boldsymbol{\mu^{u}_0}]^2_i \, \left( [\D_2]_i-1\right)^2,   
\end{equation}
% \begin{theorem}\label{thm:wasserstein}
% The \emph{Wasserstein-2} distance $\mathcal{D}_{W_2}$ between the distributions $P(\mathbf{\hat{x}_{0}^{\mathcal{F}}}; \bbalpha)$ and $P(\mathbf{{x}_{0})^{\mathcal{F}}}$  is given by:
% % \setlength{\fboxsep}{15pt} 
% \begin{equation}
% % \begin{empheq}[box=\fbox]{align}
% \mathcal{D}_{W_2}^2 \big(P(\mathbf{\hat{x}_{0}^{\mathcal{F}}}; \bbalpha), P(\mathbf{{x}_{0}^{\mathcal{F}})}\big) = 
% \sum_{i=1}^{d} \left( \sqrt{\lambda_i} - 
%  [\D_1]_i  \right)^2 \\ + \sum_{i=1}^{d} [\boldsymbol{\mu}_0^\mathcal{F}]^2_i \, \left( [\D_2]_i-1\right)^2,   
% \end{equation}
% \begin{multline}
% % \begin{empheq}[box=\fbox]{align}
% \mathcal{D}_{W_2}^2 \big(P(\mathbf{\hat{x}_{0}^{\mathcal{F}}}; \bbalpha), P(\mathbf{{x}_{0}^{\mathcal{F}})}\big) = 
% \sum_{i=1}^{d} \left( \sqrt{\lambda_i} - 
%  [\D_1]_i  \right)^2 \\ + \sum_{i=1}^{d} [\boldsymbol{\mu}_0^\mathcal{F}]^2_i \, \left( [\D_2]_i-1\right)^2,   
% \end{multline}
where $\{{\lambda}_i\}_{i=1}^d$ denote the $d$ eigenvalues of $\bSigma_0$. 
\end{theorem}

The KL divergence $\DKL(P\|Q)$ assesses how much a model probability distribution $Q$ differs from a reference probability distribution $P$. Note that this divergence is not symmetric. As with the Wasserstein-2 case, here as well we obtain a closed-form expression for the two Gaussians considered. 

\begin{theorem}\label{eq:Dkl_loss}
The Kullback-Leibler divergence between the generated distribution $p(\hat{\ve}_{0}; \bbalpha)$ and the true distribution $p(\ve_0)$  is given by
% \begin{equation}
% \DKL(P(\mathbf{{x}_{0}^{\mathcal{F}})}\| P(\mathbf{\hat{x}_{0}^{\mathcal{F}}}; \bbalpha))=\sum_{i=1}^{d} \log{[\D_1]_i} -  \frac{1}{2} \sum_{i=1}^{d} \log{{\lambda}_i} \  - \frac{d}{2}  + \frac{1}{2}\sum_{i=1}^{d}\frac{ {\lambda}_i + ([\D_2]_i-1)^2   {(\boldsymbol{\mu_0}  ^\mathcal{F})  }_i^2 }{    [\D_1]_i^2 }  .
% \end{equation}
\begin{equation}
\DKL( p(\ve_0)\| p(\hat{\ve}_{0}; \bbalpha))\!=\!\!\sum_{i=1}^{d} \log{[\D_1]_i} -  \!\frac{1}{2}\!\Bigg( \!\sum_{i=1}^{d} \log{{\lambda}_i} - \!d  \!+ \!\!\!\sum_{i=1}^{d}\frac{ {\lambda}_i\!+\! ([\D_2]_i-1)^2   {(\boldsymbol{\mu^{u}_0})  }_i^2 }{    [\D_1]_i^2 } \Bigg)
\end{equation}
% \begin{equation}
% \DKL(P(\mathbf{{x}_{0}^{\mathcal{F}})}\| P(\mathbf{\hat{x}_{0}^{\mathcal{F}}}; \bbalpha))\!=\!\!\sum_{i=1}^{d} \log{[\D_1]_i} -  \!\frac{1}{2}\!\left( \!\sum_{i=1}^{d} \log{{\lambda}_i} - \!d  \!+ \!\!\!\sum_{i=1}^{d}\frac{ {\lambda}_i\!+\! ([\D_2]_i-1)^2   {(\boldsymbol{\mu_0}  ^\mathcal{F})  }_i^2 }{    [\D_1]_i^2 } \right)
% \end{equation}
% \begin{multline}
% \DKL(P(\mathbf{{x}_{0}^{\mathcal{F}})}\| P(\mathbf{\hat{x}_{0}^{\mathcal{F}}}; \bbalpha))=\sum_{i=1}^{d} \log{[\D_1]_i} -  \frac{1}{2} \sum_{i=1}^{d} \log{{\lambda}_i} \  - \frac{d}{2}  + \frac{1}{2}\sum_{i=1}^{d}\frac{ {\lambda}_i + ([\D_2]_i-1)^2   {(\boldsymbol{\mu_0}  ^\mathcal{F})  }_i^2 }{    [\D_1]_i^2 }  .
% \end{multline}

\end{theorem}
For solving the resulting optimization problems, we have employed the Sequential Least SQuares Programming (SLSQP) method \citep{kraft1988software}, a well-suited method for minimization problems with boundary conditions, and equality and inequality constraints. 

Before turning into the empirical evaluation of the proposed optimization, and the implications of the obtained noise scheduling on the various schemes, we pause to describe related work in this field.

\section{Related work}
\label{Sec:Related Work}

% ## Numeric ##

Recent work has acknowledged the importance of the noise scheduling in diffusion models, and the need to shift the focus from custom-tailored heuristics \cite{ho2020denoising, nichol2021improved, karras2022elucidating, chen2023importance} to the development of optimized alternatives. For instance, the authors of \cite{sabour2024align} introduced the KL-divergence Upper Bound (KLUB), which minimizes the mismatch between the continuous
reverse-time SDE and its linearized approximation.
However, this approach, along with others \citep{watson2022learning, wang2023learning, xia2023towards, tong2024learning}, aim to minimize the estimation error as well, hence requires retraining a denoiser or entailing substantial computation time and resources when solving the optimization problem.

 % over short intervals. 
% a student ODE solver with learnable discretized time steps was trained in \cite{tong2024learning} by minimizing the KL-divergence to mimic a teacher ODE solver. However, these approaches, along with others \citep{watson2022learning, wang2023learning, xia2023towards}, aim to minimize the estimation error as well, which requires retraining a denoiser or entailing substantial computation time and resources when solving the optimization problem.

% Recent work has acknowledged the importance of the noise scheduling in diffusion models, and the need to shift the focus from custom-tailored heuristics \cite{ho2020denoising, nichol2021improved, karras2022elucidating, chen2023importance} to the development of optimized alternatives. For instance, the KL-divergence Upper Bound (KLUB), introduced in \cite{sabour2024align}, minimizes the mismatch between the continuous SDE and its linearized approximation over short intervals.
% Subsequently, a student ODE solver with learnable discretized time steps was trained in \cite{tong2024learning} by minimizing the KL-divergence to mimic a teacher ODE solver. However, these approaches, along with others \citep{watson2022learning, wang2023learning, xia2023towards}, aim to minimize the estimation error as well, which requires retraining a denoiser or entailing substantial computation time and resources when solving the optimization problem.

While pursuing the same goal of optimizing the noise schedule, the work reported in \cite{chen2024trajectory, xue2024accelerating, williams2024score} made notable strides in simplifying the induced optimization problem. Such efficiency is enabled in \cite{xue2024accelerating}
by introducing an upper bound on the truncation error and assuming the data-dependent component to be negligible during optimization.
% In a related effort, the authors of \cite{williams2024score} proposed a predictor-corrector update approach and minimized the Stein divergence for deriving the noise schedule. 
Yet, despite this improved scalability, a direct relationship between data characteristics and the derived noise schedule still remains vague.

% address the dataset
% the dataset's characteristics 
% The application of frequency representation 
 A potential approach to bridge this gap is spectral analysis, a fundamental tool in signal processing, which frequently serves to associate design choices and properties of diffusion models with data characteristics \cite{rissanen2022generative, biroli2024dynamical, crabbe2024time, yang2023diffusion, corvi2023intriguing}. The application of frequency representation is diverse, ranging from introducing the \emph{coarse-to-fine} behavior in the reverse process \cite{rissanen2022generative}, to uncover unique architectural fingerprints and identify biases in modeling different frequency ranges \cite{corvi2023intriguing,yang2023diffusion}. However, to the best of our knowledge, no existing approach has connected spectral analysis with noise scheduling design.
 
 % for improved
% relating properties like spectral localization for improved distribution modeling \cite{crabbe2024time}. Recently, the spectral perspective has also been used

 % A closely related work is \cite{pierret2024diffusion}, assuming a centered Gaussian distribution and commutativity properties to derive an exact solution to the reverse SDE. This is later used to evaluate ODE and SDE solvers via the Wasserstein distance.
 % Another related works is \cite{wang2024unreasonable} in which the authors derive a closed-form solution to the probability flow ODE for a Gaussian distribution and demonstrate the relationship between the learned neural score and its single Gaussian score approximation.  

% From a complementary perspective, several works have explored  diffusion models by assuming a Gaussian target distribution. In 
% \cite{pierret2024diffusion} the authors used the centered gausian distribution and commutativity properties to derive an exact solution to the reverse SDE, which is then used to evaluate ODE and SDE solvers via the Wasserstein distance. Similarly, A closed-form solution to the probability flow ODE
% provided in \cite{wang2024unreasonable}, demonstraiting the learned neural scores can be well-approximated by a single Gaussian score at high noise levels.

From a complementary perspective, several works have explored  diffusion models by assuming a Gaussian target distribution \cite{pierret2024diffusion, wang2024unreasonable}. In \cite{pierret2024diffusion}, the authors leveraged a centered distribution and commutativity properties to derive an exact solution for both the reverse SDE and ODE. 
% enabling  evaluation via the Wasserstein distance. 
Similarly, a closed-form solution to the probability flow ODE provided in \cite{wang2024unreasonable}, which also demonstrated the relation between learned neural scores and their single Gaussian score approximation. Although these analytical solutions provide valuable insights into diffusion dynamics, they are derived in the continuous domain via integration, potentially neglecting discretization effects and missing an explicit connection to the full traversal of the noise schedule.

\section{Experiments}
\label{sec:Experiments}
% We turn to empirically validate the schedules obtained by solving the optimization problem, referring to them as the \emph{spectral schedule} or \emph{spectral recommendation}. We present three main scenarios, gradually progressing from the strict assumptions to more realistic conditions. 

% We turn to empirically validate the schedules obtained by solving the optimization problem,  referred to as the \emph{spectral schedule} or \emph{spectral recommendation}, and investigate spectral phenomenon emerging from the diffusion process and their relation to the schedule structure.
We turn to empirically validate the schedules obtained by solving the optimization problem,  referred to as the \emph{spectral schedule} or \emph{spectral recommendation}. In addition, we examine spectral phenomena arising from the diffusion process and their relation to the schedule structure.
% We turn to evaluate our theoretical analysis. We empirically validate the schedules derived from the optimization problem, which we refer to as the spectral schedule or spectral recommendation. In addition, and further examine phenomena that are closely related to, and in some cases go beyond, the theoretical framework.
% \subsection{Scenario 1: Synthetic  Gaussian Distribution} 
\subsection{Synthetic  Gaussian distribution} 

\label{subsec:Scenario_1}
In the first set of experiments, we assume a Gaussian data distribution, $\x_0 \sim \N(\bmu, \bSigmaZ)$, where $\x_0 \in \mathbb{R}^{d}$ and $\bSigma_0$ is a circulant covariance matrix. Although circularity is not essential, it facilitates a direct examination of the frequency components, as 
% which are crucial in time-dependent signals. In this framework,
$\bSigma_0$ is diagonalized by the DFT, with eigenvalues  $\left\{\lambda_i \right\}_{i=1}^d$ corresponding to its DFT coefficients \cite{Davis1970}.
Specifically, the covariance is chosen to satisfy $\bSigmaZ \!= \!A^TA$  where $A$ is a circulant matrix whose first row is 
$a = [ -l, -l + 1/(d \!- \!1), \ldots, l - 1/(d \!-\! 1), l ]$. The mean vector, $\bmu$, is chosen to be a constant-value vector, following the stationarity assumption.

% % \textbf{Defining the dataset:}
% In the first set of experiments, we assume a Gaussian data distribution, $\x_0 \sim \N(\bmu, \bSigmaZ)$, where $\x_0 \in \mathbb{R}^{d}$ and $\bSigma_0$ is a circulant matrix.
% The covariance is chosen to satisfy $\bSigmaZ = A^TA$  where $A$ is a circulant matrix whose first row is 
% $a = [ -l, -l + 1/(d - 1), \ldots, l - 1/(d - 1), l ]$.
% The mean vector, $\bmu$, is chosen to be a constant-value vector, following the stationarity assumption.

\begin{figure}[t]
    \centering
        \begin{subfigure}{0.32\textwidth}
        \centering
        \includegraphics[width=\columnwidth]{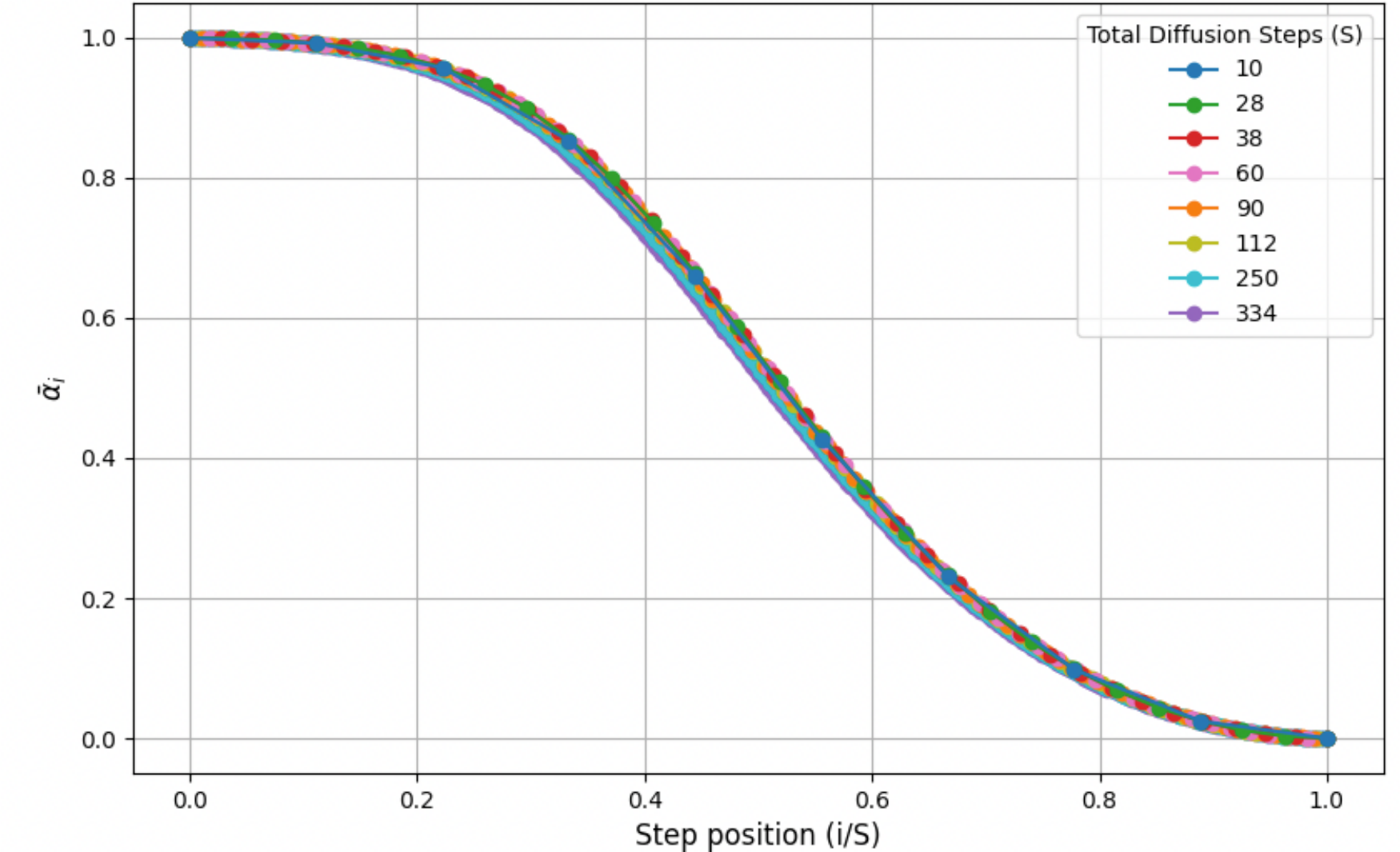}
        \caption{}
    \label{fig:Exp_1_Spectral_reccomandation_wasserstein}
    \end{subfigure}
        \hspace{1mm}
    \begin{subfigure}{0.32\textwidth}
        \centering\includegraphics[width=\columnwidth]{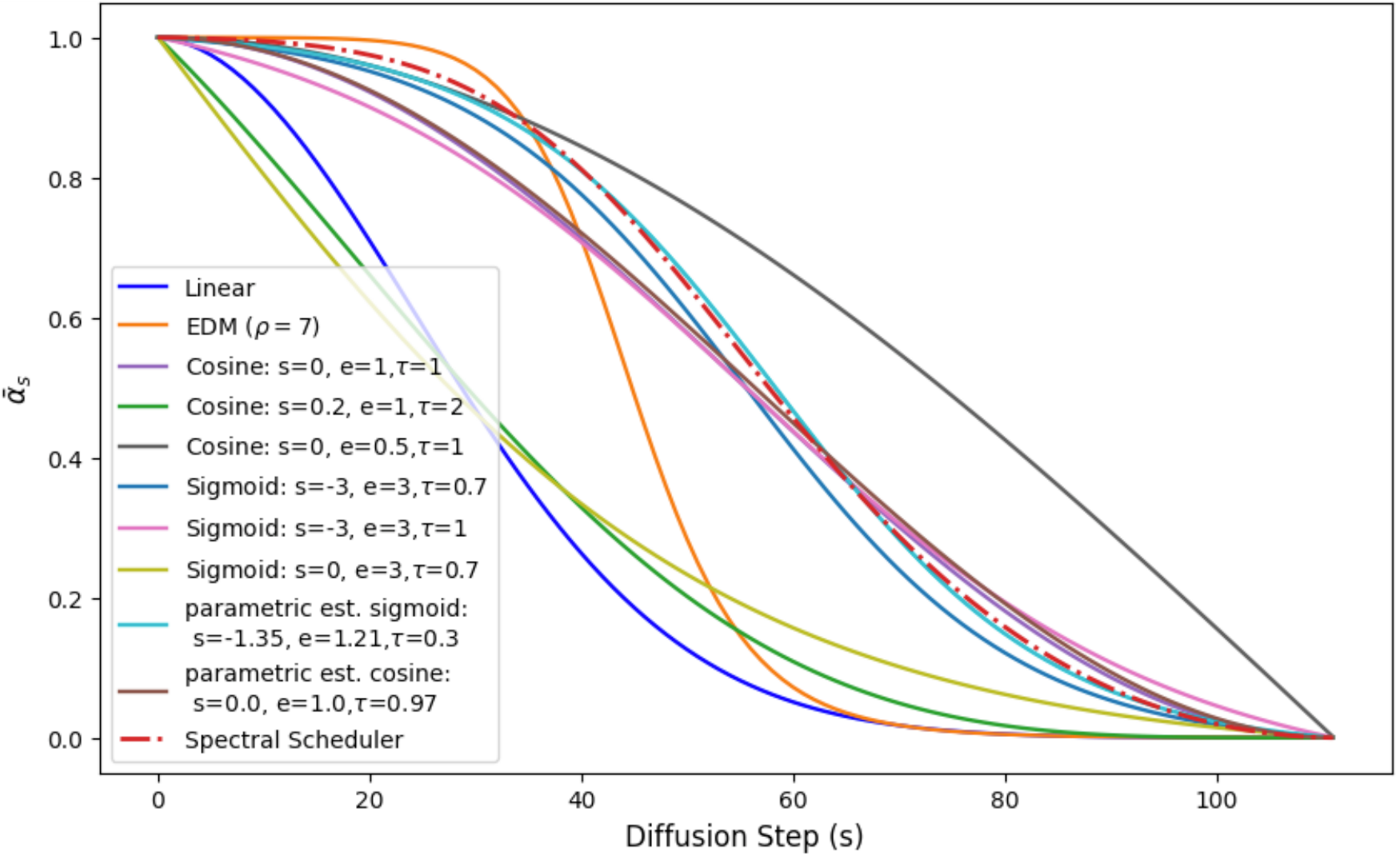}
        \caption{}\label{subfig:Spectral_scheduler_comparison}
    \end{subfigure}
    \hspace{1mm}
    % \hfill
    \begin{subfigure}{0.32\textwidth}
        \centering
        \includegraphics[width=\columnwidth]{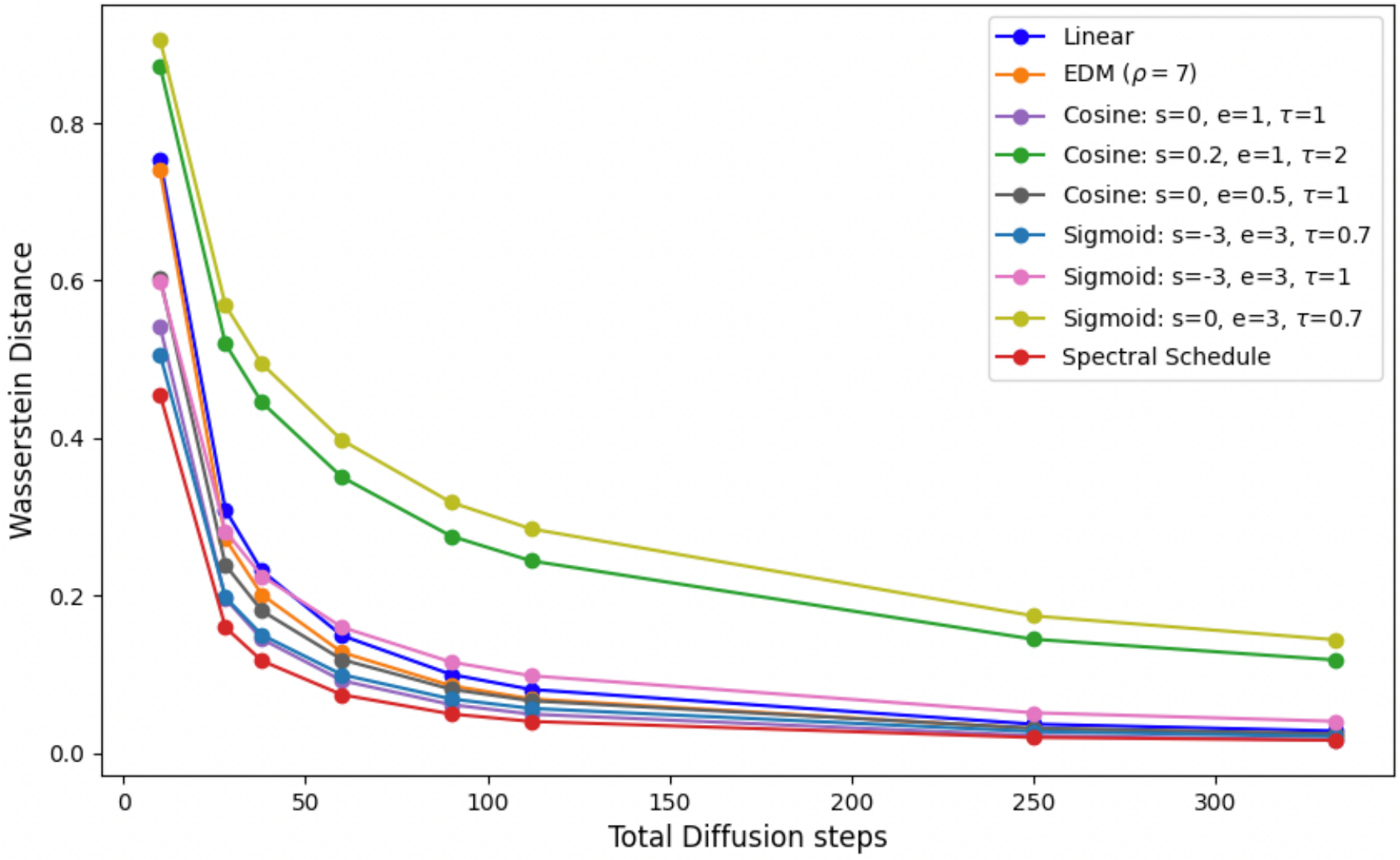}
        \caption{}
        \label{fig:Exp_1_loss_comparison_wasserstein}
    \end{subfigure}
\caption{
% Optimized spectral schedules for $\bSigmaZ = A^TA$ with $d = 50$, $l = 0.1$, and $\boldsymbol{\mu_0} = 0.05 \cdot \mathbf{1}_d$, obtained by minimizing the \emph{Wasserstein-2} distance for various numbers of diffusion steps $[10, 28, 38, 60, 90, 112, 250, 334]$ The spectral schedule (dotted gray) for $S=112$ diffusion steps is compared against various heuristic noise schedules. These include linear, EDM $(\rho = 7)$ , and Cosine-based schedules such as \emph{Cosine} ($s=0$,$e=1$,$\tau=1$) as in \citep{nichol2021improved,chen2023importance}. Additionally, Sigmoid-based schedules like \emph{Sigmoid} ($s=-3$,$e=3$,$\tau=1$) from \citep{jabri2022scalable,chen2023importance} are included. Parametric estimations for the Cosine and Sigmoid functions are shown in red and brown, respectively.  \textcolor{red}{\emph{Wasserstein-2}} distance comparison between the spectral recommendation and \emph{Cosine}, \emph{Sigmoid}, EDM and linear schedules where $d = 50$, $l = 0.1$, $\boldsymbol{\mu_0} = 0.05 \!\cdot\! \mathbf{1}_d$, and the number of diffusion steps considered are $\{10,28,38,60,90,112,250,334\}$.
Figure \ref{fig:Exp_1_Spectral_reccomandation_wasserstein} presents the optimized spectral schedules for $\bSigmaZ = A^TA$ with $d = 50$, $l = 0.1$, and $\boldsymbol{\mu_0} = 0.05 \cdot \mathbf{1}_d$, obtained by minimizing the \emph{Wasserstein-2} distance for various numbers of diffusion steps $S \in[10, 28, 38, 60, 90, 112, 250, 334]$. Figure \ref{subfig:Spectral_scheduler_comparison} compares the spectral schedule (dotted red)  for $S=112$  with various heuristic alternatives, including linear, EDM $(\rho = 7)$ , Cosine-based schedules such as \emph{Cosine} ($s=0$,$e=1$,$\tau=1$) as in \citep{nichol2021improved,chen2023importance} and Sigmoid-based schedules like \emph{Sigmoid} ($s=-3$,$e=3$,$\tau=1$) from \citep{jabri2022scalable,chen2023importance}. Parametric estimations for the Cosine and Sigmoid schedules appear in brown and cyan, respectively. Figure \ref{fig:Exp_1_loss_comparison_wasserstein} compares the \emph{Wasserstein-2} distance of the spectral recommendation with that of the baseline schedules across different step counts.
}
\vspace{-0.35cm}
\label{fig:Exp_1}
\end{figure}

Finding the optimal noise-schedule scheme  $\bbalpha^*$
depends on the target signal characteristics $\left\{\lambda_i \right\}_{i=1}^d$, the resolution $d$, and the number of diffusion steps applied $S$. Figure \ref{fig:Exp_1_Spectral_reccomandation_wasserstein} shows the resulting noise schedules for $d = 50$, $l = 0.1$ and $\boldsymbol{\mu_0} = 0.05 \!\cdot\! \mathbf{1}_d$, obtained by minimizing the \emph{Wasserstein-2} distance
 for different diffusion steps.\footnote{This follows the principles outlined in \cite{lin2024common}, ensuring a fair comparison with other noise schedules.}
 % $[10, 28, 38, 60, 90, 112, 250, 334]$.\footnote{This follows the principles outlined in \citet{lin2024common}, ensuring a fair comparison with other noise schedules.}
 Further examples involving different forms of $\bSigmaZ$ and $\bmu$, as well as the use of the \emph{KL divergence}, are provided in Appendix \ref{sec:appendix_Supplementary_Experiments_Scenario_1}.
% Appendix \ref{sec:appendix_Supplementary_Experiments_Scenario_1} provides additional examples with different forms of \bSigmaZ\bSigmaZ and \bmu\bmu, along with the use of \emph{KL divergence}."
% For additional examples with different forms of $\bSigmaZ$ and $\bmu$, see Appendix
% \ref{sec:appendix_Supplementary_Experiments_Scenario_1}
% Scenario 1

% #### right now I make the figures together: ####
% \begin{wrapfigure}{t}{0.4\textwidth}
%   \centering
% \includegraphics[width=0.4\textwidth]{figures_spectral/Exp_1/Dim_50_line_0_1_mu_0_5_wasserstein/optimal_sched_wasserstein_wider.pdf}
%  \caption{Optimized spectral schedules for $\bSigmaZ = A^TA$ with $d = 50$, $l = 0.1$, and $\boldsymbol{\mu_0} = 0.05 \cdot \mathbf{1}_d$, obtained by minimizing the \emph{Wasserstein-2} distance for various numbers of diffusion steps $[10, 28, 38, 60, 90, 112, 250, 334]$.}
% \label{fig:Exp_1_Spectral_reccomandation_wasserstein}
% \end{wrapfigure}
% ###############

% \begin{figure}[h]
%   \centering
% \includegraphics[width=0.45\textwidth]{figures_spectral/Exp_1/Dim_50_line_0_1_mu_0_5_wasserstein/optimal_sched_wasserstein_wider.pdf}
%  \caption{Optimized spectral schedules for Scenario 1 with $d = 50$, $l = 0.1$, and $\boldsymbol{\mu_0} = 0.05 \cdot \mathbf{1}_d$, obtained by minimizing the \emph{Wasserstein-2} distance for various numbers of diffusion steps.}
% \label{fig:Exp_1_Spectral_reccomandation_wasserstein}
% \end{figure}

At first glance, the optimization-based solution produces a noise schedule that aligns with the principles of diffusion models. Specifically, it exhibits a monotonically decreasing behaviour, with linear drop in $\bbalpha^*$ in the middle of the process and minimal variation near the extremes.
Although each schedule was independently optimized for a specified number of diffusion steps, it can be observed that the overall structure remains the same.

Interestingly, solving the optimization problem in \eqref{eq:optimization_problem} while altering the initial conditions or removing the inequality constraints yields the same optimal solution. This suggests that these constraints are passive, and that known characteristics of noise schedules, such as monotonicity, naturally arise from the problem formulation, as demonstrated in Appendix \ref{sec:appendix_clarifications_and_validations_constrains_omission}.

 % optimal spectral noise schedule
A key aspect is how the spectral schedule aligns with the existing heuristics. Figure \ref{subfig:Spectral_scheduler_comparison} provides a comparison with the Cosine \cite{nichol2021improved}, the Sigmoid \cite{jabri2022scalable}, the linear \cite{ho2020denoising}
and the EDM $(\rho = 7)$ \cite{karras2022elucidating} schedules, along with a parametric approximation of the spectral recommendation. To achieve this, Cosine and Sigmoid functions were fitted to the optimal solution by minimizing the $l_2$ loss, identifying the closest match.
An interesting outcome from Figure \ref{subfig:Spectral_scheduler_comparison} is that the spectral recommendation provides a partial retrospective justification for existing noise schedule heuristics, as the parametric estimation resembles Cosine and Sigmoid functions, when their parameters are properly tuned.
% This was done by fitting the cosine and sigmoid functions to the optimal solution, minimizing the $l_2$ loss to find the closest match.

% In both cases, the optimization-based solution produces a noise schedule that aligns with the principles of existing heuristics.
% Specifically, these schedules show a linear drop-off of ${\balpha}_s$ in the middle of the process, with minimal variation near the extremes ($s=0, s=S$).

% \textbf{Comparison to Heuristics:} 
% A key aspect is how the optimal spectral solution aligns with existing heuristics. \autoref{fig:Spectral_scheduler_comparison} makes such a comparison, along with a parametric approximation of the spectral recommendation. This was done by fitting the cosine and sigmoid functions to the optimal solution, minimizing the $l_2$ loss to find the closest match. In both cases, the optimization-based solution produces a noise schedule that aligns with the principles of existing heuristics.
% Specifically, these schedules show a linear drop-off of ${\balpha}_s$ in the middle of the process, with minimal variation near the extremes ($s=0, s=S$). To the best of our knowledge, this is the first time the solution to a simple optimization problem can be used to retrospectively justify the existing noise schedule heuristics.

% ################Reoved for now ###########

% \begin{figure}[h]
%   \centering
% \includegraphics[width=0.45\textwidth]{figures_spectral/Exp_1/Dim_50_line_0_1_mu_0_5_wasserstein/Diff_alpha_bar_time_steps=112_cosine_sigmoid_with_cosine_and_sigomid_est_change_sigmoid.pdf}
%   \caption{The spectral schedule (dotted gray) for $S=112$ diffusion steps is compared against various heuristic noise schedules. These include  linear, EDM $(\rho = 7)$ , and Cosine-based schedules such as \emph{Cosine} ($s=0$,$e=1$,$\tau=1$) as in \citep{nichol2021improved,chen2023importance}. Additionally, Sigmoid-based schedules like \emph{Sigmoid} ($s=-3$,$e=3$,$\tau=1$) from \citep{jabri2022scalable,chen2023importance} are included. Parametric estimations for the Cosine and Sigmoid functions are shown in red and brown, respectively.}
% \label{fig:Spectral_scheduler_comparison}
% \end{figure}
% obtained

To validate the optimization procedure, Figure \ref{fig:Exp_1_loss_comparison_wasserstein} compares the \emph{Wasserstein-2} distance of various noise schedules with that of the spectral recommendation across different diffusion steps. While the spectral schedule consistently achieves the lowest \emph{Wasserstein-2} distance, the optimization is most effective 
% effect is most pronounced
with fewer diffusion steps, where discretization errors are higher. As the number of steps increases, the gap between the different noise schedules narrows.

% \begin{figure}[t]
%   \centering
% \includegraphics[width=0.45\textwidth]{figures_spectral/Exp_1/Dim_50_line_0_1_mu_0_5_wasserstein/wassestein_l_0_1_mu_0_5_loss_comparison_change_sigmoid_and_tau.pdf}
%   \caption{  \emph{Wasserstein-2} distance comparison between the spectral recommendation and \emph{Cosine}, \emph{Sigmoid}, EDM and linear schedules where $d = 50$, $l = 0.1$, $\boldsymbol{\mu_0} = 0.05 \!\cdot\! \mathbf{1}_d$, and the number of diffusion steps considered are $\{10,28,38,60,90,112,250,334\}$.}\label{fig:Exp_1_loss_comparison_wasserstein}
% \end{figure}

\begin{wrapfigure}{t}{0.64\textwidth}
    \centering
    \hspace{-.2cm}
    \begin{minipage}{0.32\textwidth}
        \centering
        \begin{subfigure}[b]{1.0\textwidth}        \includegraphics[width=\textwidth]{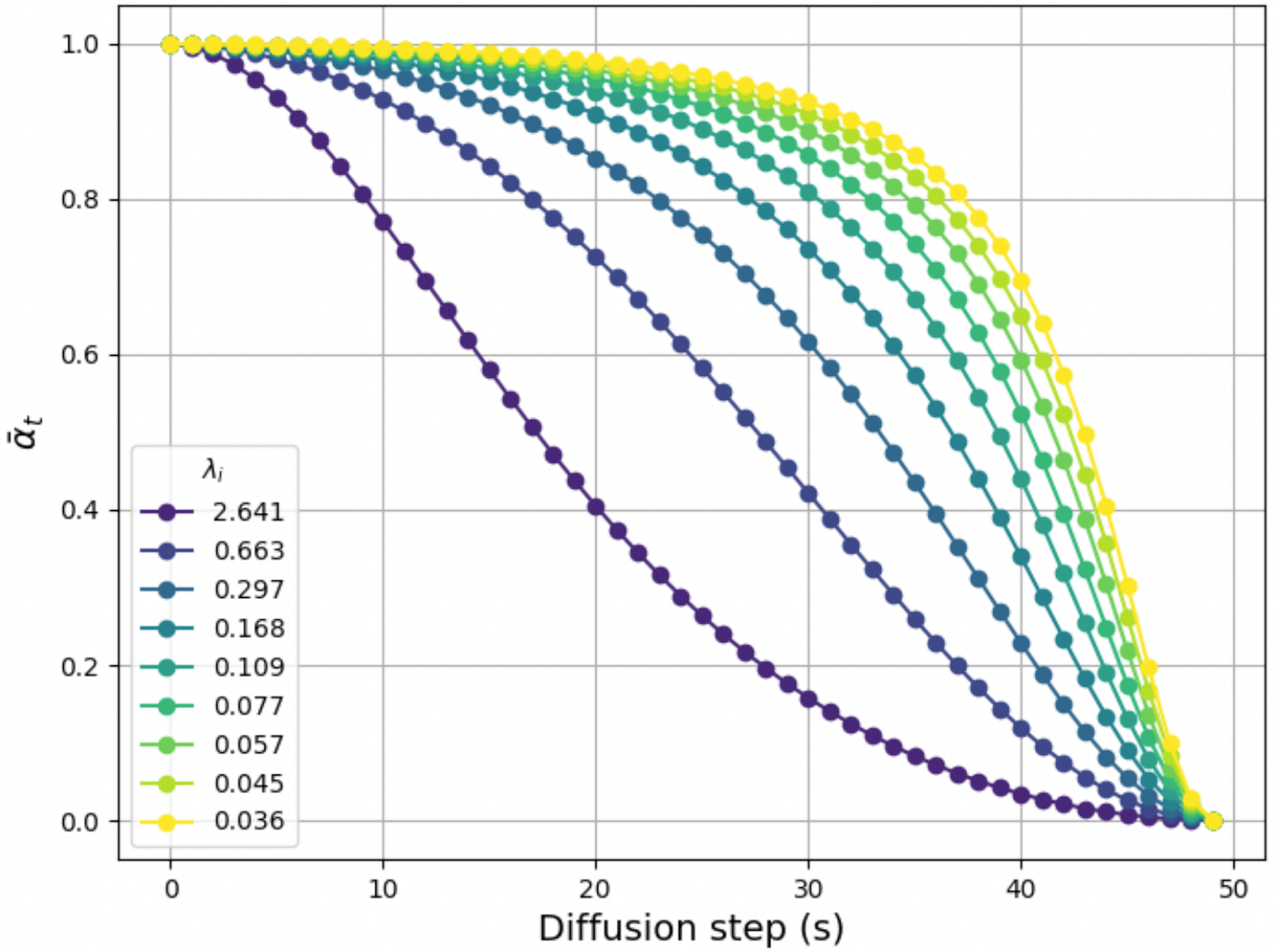}
            \caption{}
\label{subfig:Each_eigenvalue_at_a_time_paper} 
        \end{subfigure}
    \end{minipage}
    \hspace{-.1cm}
    \begin{minipage}{0.32\textwidth}
        \centering
        \begin{subfigure}[b]{1.0\textwidth}
            \includegraphics[width=\textwidth]{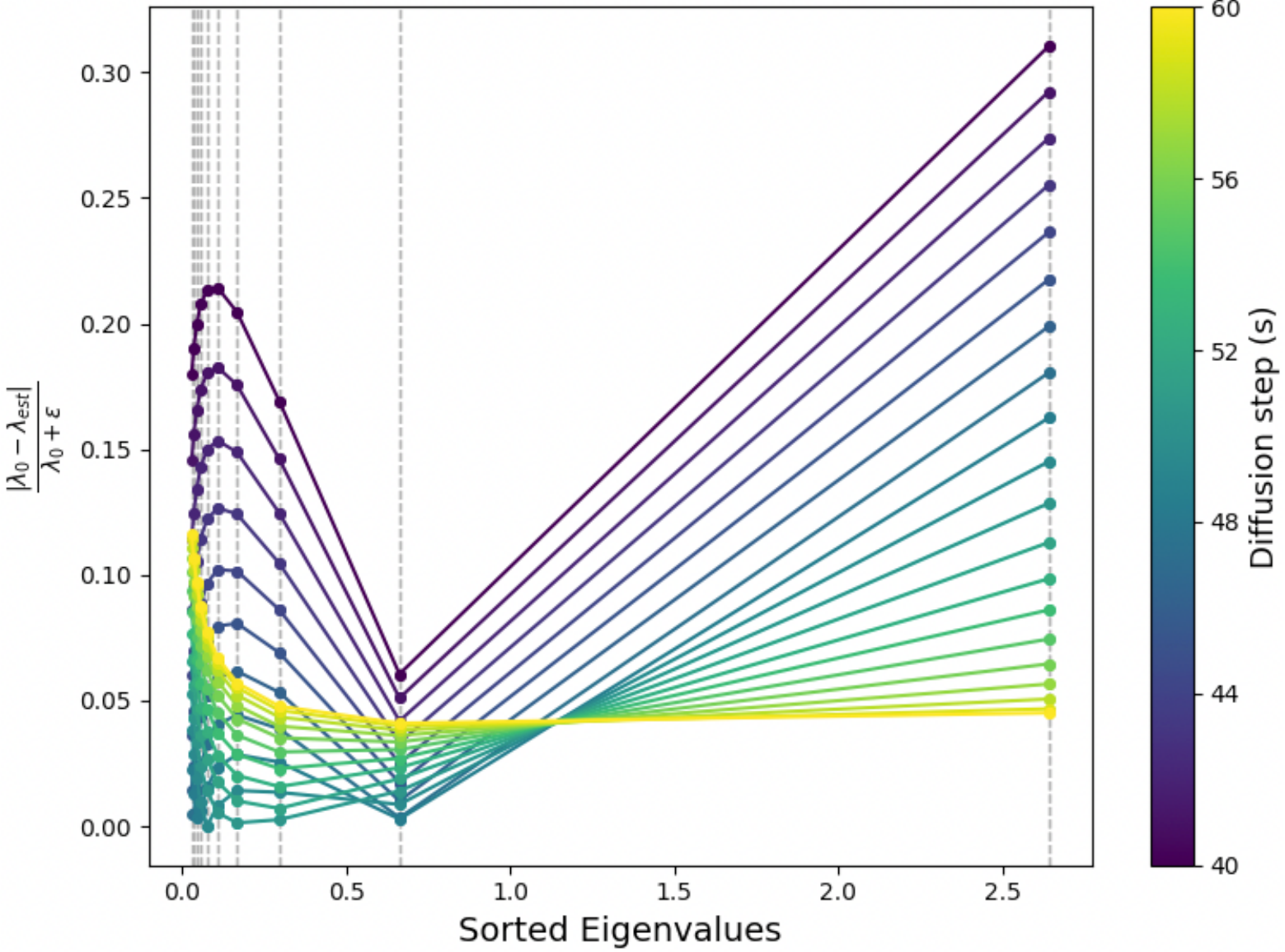}
        \caption{}
        \label{fig:Eagenvalues_Relative_error_cosine}
        \end{subfigure}
    \end{minipage}
    \caption{Figure \ref{subfig:Each_eigenvalue_at_a_time_paper} shows the spectral schedules obtained by solving the optimization problem individually for each eigenvalue, with other contributions set to zero (note that the reverse process proceeds from right to left). Figure \ref{fig:Eagenvalues_Relative_error_cosine} illustrates the relative error of the 10 largest eigenvalues over the final 20 steps of a 60-step diffusion process using the Cosine $(0,1,1)$ schedule.}
    % \vspace{-1.5cm}
\end{wrapfigure}

We also compare our derived schedules with those proposed in prior work. Specifically, the authors in \cite{sabour2024align} present a closed-form solution for the optimal noise schedule in a \emph{simplified setting}, assuming an initial isotropic Gaussian distribution with standard deviation $C$, i.e $ ~ \x_0 \sim \N(\0, C^2\I)$. 
To enable a fair comparison, we frame our optimization problem using the $\DKL$ loss \eqref{eq:Dkl_loss} and adopt the Variance-Exploding (VE) framework , following the setup in \cite{sabour2024align}. Appendix \ref{sec:appendix_clarifications_and_validations_AYS_comparison} demonstrates the equivalence between the solutions and highlight their relation to commonly used heuristics.
\textbf{Frequencies components and schedule structure:} We turn to explore the relationship between the eigenvalue magnitudes and the structure of the derived schedule. To do so, we solved the optimization problem for each eigenvalue individually, with the contributions from the
other eigenvalues set to zero. It can be observed from \autoref{subfig:Each_eigenvalue_at_a_time_paper} that the solution becomes more concave as the magnitude of the eigenvalue decreases (yellow curve) and more convex as the magnitude increases (blue curve).

 % frequency
 % (e.g., the $1/f$ behavior observed in speech \citep{voss1975f})
Notably, under the shift-invariance conditions, the eigenvalues directly correspond to the system's frequency components. In scenarios where this components follow a monotonically decreasing distribution (e.g., the $1/f$ trend in speech \citep{voss1975f}), the first eigenvalues correspond to the low frequencies, having larger amplitudes, while the last  correspond to high frequencies and smaller amplitudes. This pattern, along with the previous observation, aligns with the well-known coarse-to-fine signal construction behavior of diffusion models. farther details are provided in Appendix \ref{sec:Relationship_Noise_Schedules_Eigenvalues}

\textbf{Relative error evolution:} While the optimization problem primarily focuses on the output signal $\hat{\ve}_0$, each diffusion step $\ve_s$ can be also expressed as a function of the initial noise as described in  \eqref{eq:itermidate_step_diffusion_process} in Appendix \ref{sec:appendix_inference_spectral}. This enables the analysis of spectral properties across all diffusion steps, rather than only at the output signal.
 % was designed for 
% presents the relative error of each eigenvalue over the final 20 steps of a 60-step diffusion process using the \emph{Cosine} ($0$,$1$,$1$) 
Figure \ref{fig:Eagenvalues_Relative_error_cosine} presents the relative error of the 10 largest eigenvalues (sorted, largest on the right) over the final 20 steps of a 60-step diffusion process using the \emph{Cosine} ($0$,$1$,$1$) \cite{nichol2021improved, chen2023importance} schedule. While the error consistently decreases for high-amplitude eigenvalues, it increases near the final diffusion steps for those with smaller magnitudes.  Additionally, the final relative error (yellow) tends to be smaller for large eigenvalues (i.e., low frequencies). This behavior aligns with the observed bias in mid-to-high frequencies reported in prior works \cite{yang2023diffusion, corvi2023intriguing}. Further discussion, including an illustration of the eigenvalue and \emph{Wasserstein-2} error dynamics, appears in App. \ref{sec:Dynamics_throughout_the_diffusion_process}.
% towards
% A further discussion including ilustration of the eigenvalues dynamics appeard in Appendix \ref{sec:Dynamics_throughout_the_diffusion_process}

% \cite{nichol2021improved}
% \red{\textbf{Eigenvalues dynamics:} \autoref{fig:Eagenvalues_convergence} illustrates the dynamics of the eigenvalues across $60$ diffusion steps using the spectral schedule. While lower eigenvalues tend to stabilize earlier in the process, though this does not necessarily indicate the values they ultimately reach (see Appendix \ref{} for details).}

% ##### original ######
% \begin{wrapfigure}{h}{0.3\textwidth}
%   \centering
%     \includegraphics[width=0.3\textwidth]{figures_spectral/synthetic_Gausian_distribution/eagenvalues_colormap_by_rows_without_original_spectral_0_1.pdf}
% \caption{}
% \label{fig:Eagenvalues_convergence}
% \end{wrapfigure}

% \begin{wrapfigure}{t}{0.3\textwidth}
%   \centering
%     \includegraphics[width=0.3\textwidth]{figures_spectral/synthetic_Gausian_distribution/eagenvalues_colormap_weighted_diff_epsilon_0_last_steps_20_sorted.pdf}
% \caption{}
% \label{fig:Eagenvalues_Relative_error_cosine}
% \end{wrapfigure}
% ###############

% ######## original - before change 13.5!!!! ######
% ###########
\textbf{Mean drift:}
The explicit expressions in \eqref{eq:D_1,D_2} and \eqref{eq:D1D2-DDPM} offer a further insight into the diffusion process. A notable consideration is whether this process introduces a bias, i.e. drifting the mean component during synthesis. To study this, we analyze the mean bias expression for DDIM, 
$(\D_2 - \I)\boldsymbol{\mu_0^u}$ derived from the difference between  $\mathbb{E} \left[ \mathbf{\ve}_{0} \right]$ and $\mathbb{E}  \left[\mathbf{\hat{\ve}}_{0} \right]$.
In Appendix \ref{sec:appendix_Mean_Bias}, we further explore the relationship between the target signal characteristics $\left\{\lambda_i \right\}_{i=1}^d$, the noise schedule $\bbalpha$, and the expression $|\D_2 - \I|$. It appears that different choices of the noise schedule influence the bias value, with some choices effectively mitigating it. Additionally, as the depth of the diffusion process increases, the bias value tends to grow, regardless of the selected noise schedule. 
\textbf{DDPM vs DDIM:}
 The explicit formulations of DDPM and DDIM in \eqref{eq:D_1,D_2} and \eqref{eq:D1D2-DDPM}  enable a comparison of their losses across varying diffusion depths and noise schedules. In Appendix \ref{sec:appendix_DDPM_vs_DDIM}, we present such an evaluation using the \emph{Wasserstein-2} distance. The results clearly show that DDIM sampling is faster and yields lower loss values, aligning with the empirical observations in \cite{song2020denoising}.

% \begin{figure}[H]
%   \centering
% \includegraphics[width=0.45\textwidth]{figures_spectral/DDPM_DDIM/loss_DDPM_DDIM_log_short_8_0_1_0_5_with_sigma_1_total_of_diffusion_steps.pdf}
% \caption{Comparison of the \emph{Wasserstein-2} distance between DDPM and DDIM for different noise schedules, including the spectral recommendation, across various diffusion steps.}\label{fig:DDPM_DDIM_several_schedules_334}
% \end{figure}

% \subsection{Scenario 2: Practical considerations}

\subsection{Empirical Gaussian distribution}
\label{subsec:scenario_2}

% ################## need before change!!! ########
\begin{wrapfigure}{h}{0.4\textwidth}
% \begin{wrapfigure}{h}{0.4\textwidth}
  \centering
  % \vspace{-0.1cm}
    \vspace{-1.4cm}
\centering\includegraphics[width=0.4\textwidth]{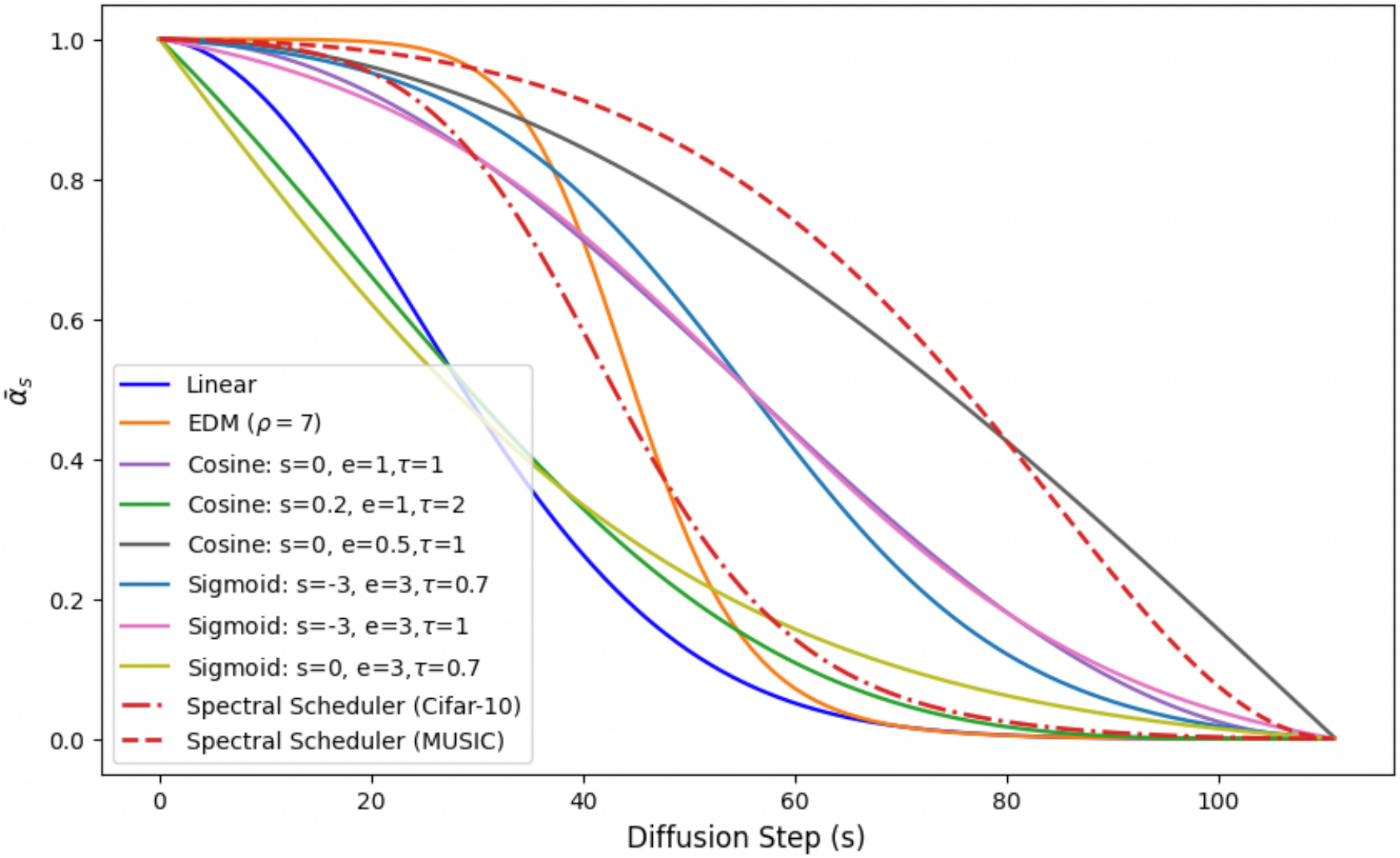}
\caption{Comparison of the spectral schedules for CIFAR-10 and MUSIC Datasets with various heuristic noise schedules, using $112$ diffusion steps.}
% \caption{Comparison of the spectral schedules for \emph{Gaussian CIFAR-10 Dataset} and \emph{Gaussain MUSIC piano Dataset} with various heuristic noise schedules, using $112$ diffusion steps.}
\label{fig:Exp_2_Spectral_scheduler_comparison_MUSIC_CIFAR}
% \vspace{-1.3cm}
\vspace{-0.5cm}
\end{wrapfigure}
% ##########
% A detailed discussion of the estimation procedure for temporal signals, including the role of the window size ($d$) and the silence threshold ($th$)  is provided in Appendix \ref{sec:scenario_2_Analysis_of_Different_Aspects}.

% 
We now shift towards a more practical scenario in which we refer to real data, while still maintaining the Gaussianity assumption. To assess the generality of our approach, we consider different signal types. Specifically, we use CIFAR-10, and the MUSIC dataset \cite{moura2020music}, which comprises recordings of various musical instruments down-sampled to $16$kHz. We fit a Gaussian distribution for each dataset by estimating their mean and covariance matrix, while for the MUSIC dataset, we extracted piano-only recordings. These datasets are referred to hereafter as the \emph{Gaussian MUSIC piano dataset} and the \emph{Gaussian CIFAR-10 dataset} \cite{pierret2024diffusion}. The estimation procedure for temporal signals, including the role of the window size ($d$) and silence threshold ($th$), is detailed in Appendix \ref{sec:scenario_2_Analysis_of_Different_Aspects}.

% Figure \ref{fig:Exp_2_Spectral_scheduler_comparison_MUSIC_CIFAR} presents the spectral recommendations obtained by minimizing the \emph{Wasserstein-2} distance using the estimated covariance matrices, alongside several heuristic noise schedules.
Figure \ref{fig:Exp_2_Spectral_scheduler_comparison_MUSIC_CIFAR} presents the spectral recommendations obtained by minimizing the \emph{Wasserstein-2} distance for each estimated covariance matrix, alongside several heuristic noise schedules.
While both spectral schedules retain some resemblance to the hand-crafted approaches, the one derived for the  MUSIC dataset, introduces a somewhat different design of slower decay, adapting to the unique data properties. Consequently, adopting a spectral analysis perspective enables the design of noise schedules tailored to specific needs and data characteristics. 
The spectral recommendations for the SC09 \citep{warden2018speech} dataset, with the \emph{Wasserstein-2} evaluations, are provided in Appendix \ref{sec:Scenario_2_experiments}.
\subsection{Empirical distribution}
% Expanding to Realistic Settings
% Toward Real-World Conditions
\label{subsec:scenario_3}
% In the previous experiments we assumed a Gaussian data distribution; however, in real-world scenarios, the dataset may exhibit a more complex structure. When the Gaussianity assumption does not hold, the optimal denoiser may be nonlinear and may not converge to the MAP estimator. In such cases, training a neural denoiser becomes necessary to facilitate the diffusion process. Consequently, alongside the discretization error, an additional estimation error arises due to the potential suboptimality of the denoiser.

We now aim to evaluate whether the optimized noise schedule from Sec. \ref{subsec:scenario_2} remains effective when the Gaussianity assumption is removed and a trained neural denoiser is employed within the diffusion process. Notably, In this setting, approximation error arises due to suboptimality of the denoiser \cite{pierret2024diffusion}.
% Since our focus is on errors introduced by the discretization procedure,

Our approach towards performance assessment of different noise schedules is the following: We run the diffusion process using a trained denoiser, and the DDIM inference procedure \eqref{eq:DDIM_sampling_procedure} to generate a large corpus of samples. We then compare the statistics of the synthesized signals with those of real data by measuring the distance between their second moments, applying the \emph{Wasserstein-2} distance to the empirical covariance matrices. For CIFAR-10, we also use the standard FID score.
% score as a perceptual metric.
% \footnote{\textcolor{red}{FAD calculation is less reliable for the MUSIC dataset due to its short frame lengths, and less informative for SC09, given its nature as a speech dataset.}}} additionally

% ompares the spectral recommendations with various 

\begin{figure}[t]
    \centering
    \begin{subfigure}{0.32\textwidth}
        \centering\includegraphics[width=\columnwidth]{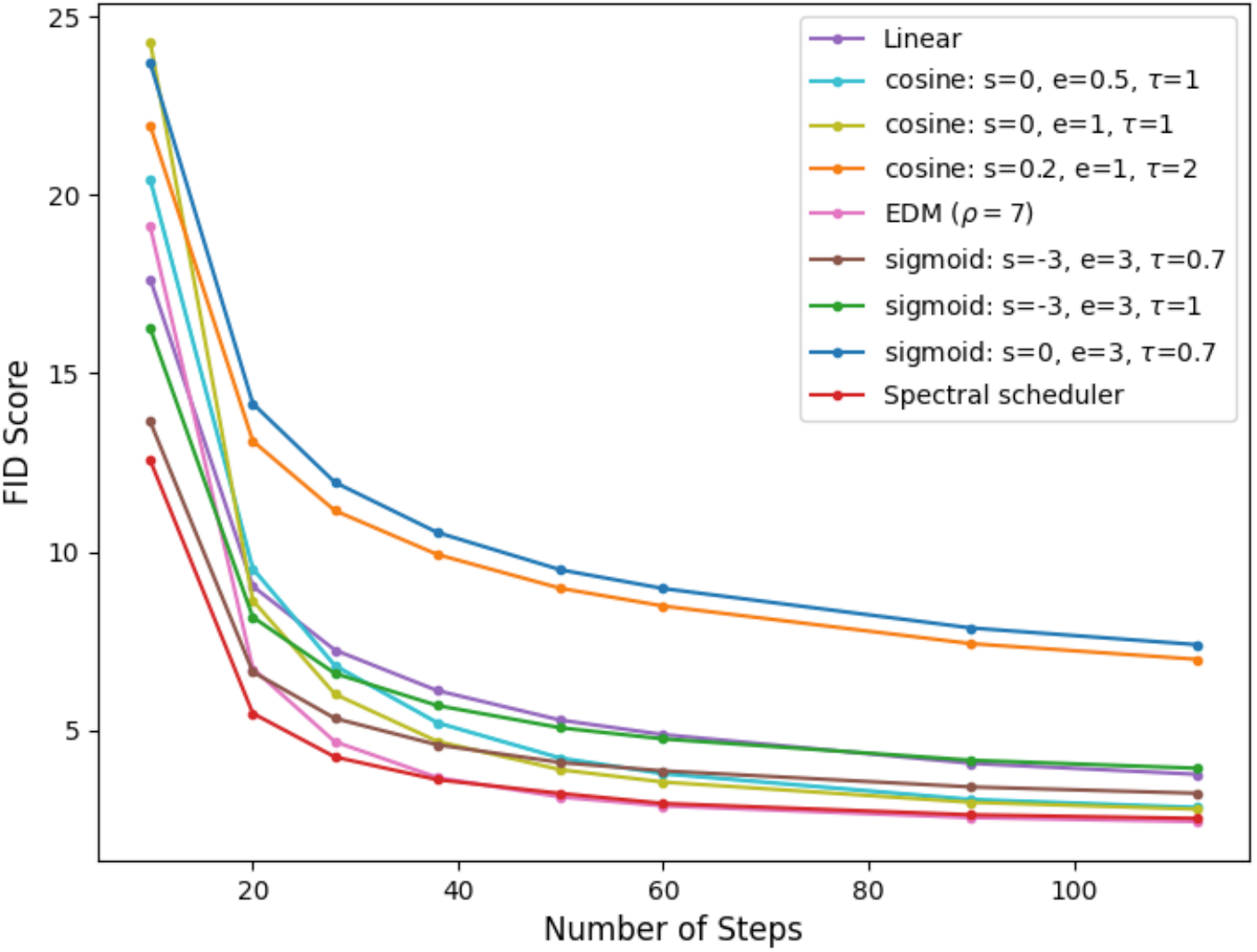}
        \caption{}\label{subfig:Exp_3_FID_Cifar}
    \end{subfigure}
    \hspace{1mm}
        \begin{subfigure}{0.315\textwidth}
        \centering\includegraphics[width=\columnwidth]{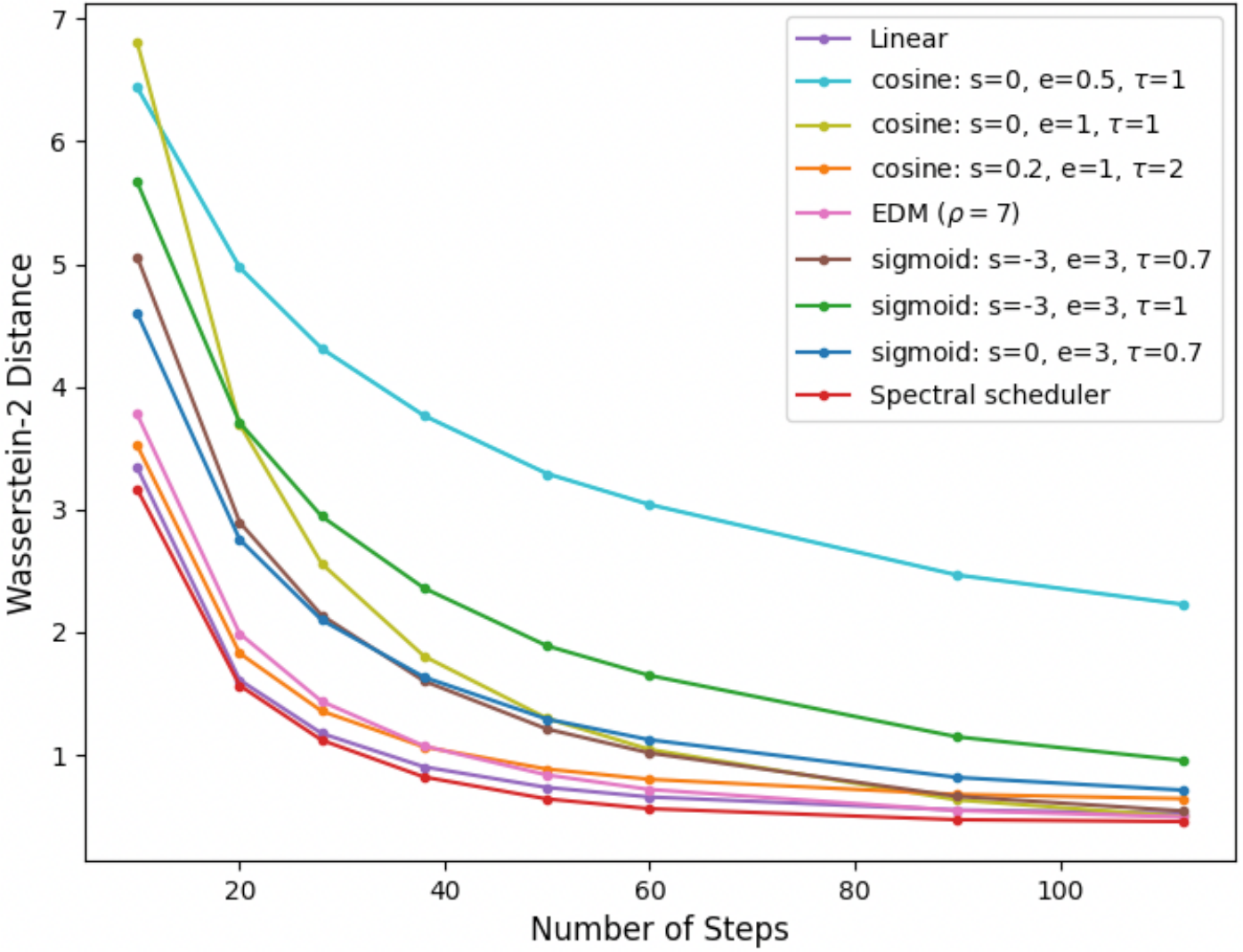}
        \caption{}\label{subfig:Exp_3_Wasserstein_Cifar}
    \end{subfigure}
    \hspace{1mm}
    % \hfill
    \begin{subfigure}{0.32\textwidth}
        \centering
        \includegraphics[width=\columnwidth]{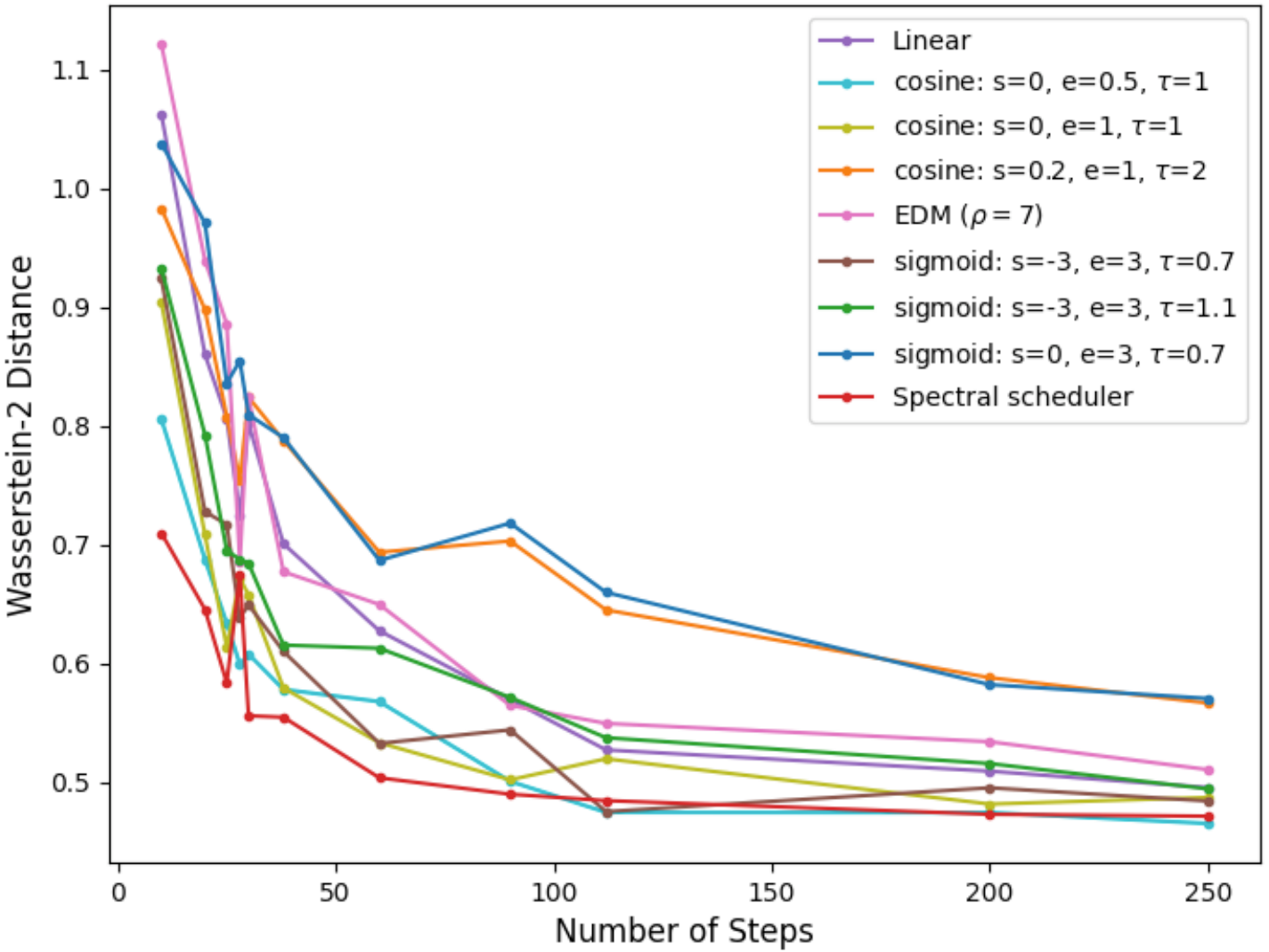}
        \caption{}
        \label{subfig:Exp_3_wasserstein_2_distance_}
    \end{subfigure}
\caption{Figures \ref{subfig:Exp_3_Wasserstein_Cifar} and \ref{subfig:Exp_3_wasserstein_2_distance_} show the \emph{Wasserstein-2 distance} of the proposed spectral noise schedule (in red) compared to existing heuristics, evaluated across various diffusion step sizes, for CIFAR-10 and MUSIC. Figure \ref{subfig:Exp_3_FID_Cifar} presents the corresponding FID scores for CIFAR-10. Across all comparisons, the spectral schedule generally outperforms the heuristics, with a wider gap at lower step counts. For CIFAR-10, the approximation error is less pronounced, with results showing greater stability.}
% evaluated across the following diffusion steps: ${10, 20, 28, 30, 38, 60, 90, 112, 200, 250}$,
% at fewer steps
% \caption{Figures \ref{subfig:Exp_3_Wasserstein_Cifar} and \ref{subfig:Exp_3_wasserstein_2_distance_} visualize the \emph{Wasserstein distance} of the spectral noise schedule (in red) and compared to existing heuristics, considering the following diffusion steps: $\{10, 20, 28, 30, 38, 60, 90, 112, 200, 250\}$ for the MUSIC and Cifar-10 datasets. Figure \ref{subfig:Exp_3_FID_Cifar} compares the FID values for the Cifar-10.
% In all comparisons, the spectral schedule achieve almost consistetly better results, While, for Cifar-10, the approximation error appears to be less pronounced, as the results demonstrate greater stability.
%  }
 % \emph{Frobenius norm} and the
\label{fig:Exp_3_metrics_comparisons_Cifar_MUSIC}
\end{figure}

Figure \ref{fig:Exp_3_metrics_comparisons_Cifar_MUSIC} performs such a comparison with various heuristic noise schedules. For each schedule and number of diffusion steps tested, 50,000 signals are synthesized using a trained model. For CIFAR-10, we used a pretrained denoiser from \cite{karras2022elucidating} which is based on a continuous noise schedule. For the MUSIC dataset, we trained a model based on the architecture presented in \cite{kong2020diffwave, benita2023diffar}, which employs a linear noise schedule with $T=1000$ diffusion steps during training.

Figure \ref{fig:Exp_3_metrics_comparisons_Cifar_MUSIC} demonstrates that the spectral schedules consistently outperform most heuristic alternatives across both metrics, with the performance gap narrowing as the number of diffusion steps increases and discretization error diminishes. Notably, the Cosine ($0$,$0.5$,$1$) and EDM heuristics, whose structures closely resemble the spectral schedules derived for the CIFAR-10 and MUSIC datasets in Figure \ref{fig:Exp_2_Spectral_scheduler_comparison_MUSIC_CIFAR}, also perform competitively (Figures \ref{subfig:Exp_3_FID_Cifar}, \ref{subfig:Exp_3_wasserstein_2_distance_}). These suggests that the proposed spectral recommendation effectively preserves the core data properties and may support the broader design of scheduling strategies. A more detailed comparison, with results for SC09, appears in Appendix \ref{sec:Scenario_3_experiments}.  
\section{Discussion}
\label{further_discussion}
By recognizing how the eigenvalues shape the noise schedule, we can refine the loss term to better align with our needs. In Appendix \ref{sec:Noise_schedule_loss_function}, we introduce a \emph{weighted-l1} loss, which promotes low frequencies while sacrificing high ones, ultimately yielding a solution consistent with the well-known Cosine ($0$,$1$,$1$) heuristic. This relationship can be further explored and may help in mitigating phenomena such as bias in certain frequency ranges \cite{corvi2023intriguing,yang2023diffusion}. Another potential research direction is leveraging the evolution of the stepwise distribution  for enabling accelerated sampling \eqref{eq:itermidate_step_diffusion_process}.

 We analyze the optimization times of our method. For small step counts, where discretization error is more pronounced, the solutions are typically obtained within a few seconds. However, for higher step counts or high-resolution signals, the computation time may vary. By prioritizing the most significant eigenvalue components during schedule design or initializing the optimization with solutions from smaller step counts, the runtime can be reduced to a few tens of seconds, as detailed in Appendix \ref{sec:appendix_optimization_time_analysis}.

 \textbf{Limitations:} While our approach demonstrates the benefits of a principled scheduler design and outperforms heuristic methods in FID across multiple datasets, a natural next step is to extend the framework to settings closer to real data distributions, such as Gaussian mixture models.

This paper presents a spectral perspective on the inference process in diffusion models. Under the assumption of Gaussianity, we establish a direct link between the input white noise and the output signal. Our approach enables noise schedule design based on data characteristics, diffusion steps count, and sampling methods. Effective across synthetic and realistic settings,
the optimized schedules resemble existing heuristics, offering insights
on handcrafted design choices. By leveraging our method to explore various aspects of the reverse process, we hope this work encourages further exploration of diffusion models and their dynamics from a spectral perspective.
% shedding light 
% \section*{References}
% \bibliographystyle{plainnat}
\bibliographystyle{plain}
\bibliography{references}

\clearpage

\newpage
\appendix
\onecolumn

% \section{Wiener Estimation}
% \label{sec:appendix_Map_est}
%
\section{The Optimal Denoiser for a Gaussian Input}
\label{sec:appendix_Map_est}
This appendix provides the derivation and explanation of Theorem \ref{theorem:optimal_demoiser}.

% Let $\x_0 \in \N(\boldsymbol{\mu}_0, \boldsymbol{\Sigma}_0)$  
Let $\x_0 \sim \N(\bmu, \bSigmaZ)$  represent the distribution of the original dataset, where $\x_0 \in \mathbb{R}^{d}$ . The probability density function $f(\x_0)$ can be written as:
$$
f(\x_0) =  \frac{1}{\sqrt{\left(2\pi\right)^d|\bSigmaZ|}} \cdot \exp\left\{ -\frac{1}{2} \left( \bar{\x}_0 - \bar{\boldsymbol{\mu}}_0 \right)^T {\boldsymbol{\Sigma}}_0^{-1} \left( \bar{\x}_0 - \bar{\boldsymbol{\mu}}_0 \right) \right\}
$$
% Where $ ~C_1 = \frac{1}{\sqrt{2\pi|\bSigmaZ|}}$ is a normalization constant

Through the diffusion process, the signal undergoes noise contamination, leading to the following marginal expression for $\x_t$:
\begin{equation}\label{eq:1}
    \x_t = \sqrt{\balpha_t} \x_0 + \sqrt{1-\balpha_t} \bepsilon ~~~~~  \bepsilon \sim \N(\mathbf{0},\I)
\end{equation}
% Using the following notation:
% $$
% k_{1,t} = \sqrt{\balpha_t}, \quad k_{2,t}= \sqrt{1-\balpha_t}
% $$
% We obtain:
% $$
% \x_t = k_{1,t} \x_0 + k_{2,t} \bepsilon ~~~~~  \bepsilon \sim \N(\mathbf{0},\I)
% $$
For the Maximum A Posteriori (MAP) estimation, we seek to maximize the posterior distribution: 
$$
\underset{x_0}{\max} \log p(\x_0 | \x_t)
$$
Using Bayes' rule, this can be written as:
\begin{equation}\label{minimization_problem}
\underset{x_0}{\min} \, -\log \left[ \frac{p(\x_t | \x_0) p(\x_0)}{p(\x_t)} \right] =  \underset{x_0}{\min} \, -\log p(\x_t | \x_0) -\log p(\x_0)
\end{equation}

The conditional log-likelihood $ \log{p(\x_t | \x_0)} $ is given by:

$$p(\x_t|\x_0) = \frac{1}{\sqrt{\left(2\pi\right)^d|\Sigma_1|}} \exp \left\{-\frac{1}{2}\left(\x_t-\sqrt{\balpha_t}\x_0\right)^T\left( (1-\balpha_t)\I\right)^{-1}\left(\x_t-\sqrt{\balpha_t}\x_0\right)\right\}$$

$$\log \, p(\x_t|\x_0) = -\frac{1}{2}\log{\left(2\pi\right)^d|\Sigma_1|} - \frac{1}{2(1-\balpha_t)}\left(\x_t-\sqrt{\balpha_t} \x_0\right)^T\left(\x_t-\sqrt{\balpha_t} \x_0\right)
$$

The log-likelihood $ \log{p(\x_0)} $ is given by:
$$p(\x_0) = \frac{1}{\sqrt{\left(2\pi\right)^d|\bSigmaZ|}} \exp \left\{-\frac{1}{2}\left(\x_0-\bmu\right)^T\bSigmaZ^{-1}\left(\x_0-\bmu\right)\right\}$$

$$\log{p(\x_0)} = -\frac{1}{2}\log{\left(2\pi\right)^d|\bSigmaZ|} - \frac{1}{2}\left(\x_0-\bmu\right)^T\bSigmaZ^{-1}\left(\x_0-\bmu\right)
$$

We will differentiate the given expression in \eqref{minimization_problem} with respect to $x_0$ and equate it to zero:

% $$\frac{d\log{p(\x_t|\x_0)}}{d\x_0} = - - \frac{2\sqrt{\balpha_t} \left(\x_t-\sqrt{\balpha_t} \x_0\right)}{2(1-\balpha_t)}
% $$

$$\frac{d\log{p(\x_t|\x_0)}}{d\x_0} =  \frac{2\sqrt{\balpha_t} \left(\x_t-\sqrt{\balpha_t} \x_0\right)}{2(1-\balpha_t)}
$$

$$\frac{d\log{p(\x_0)}}{d\x_0} = - \frac{2\bSigmaZ^{-1}\left(\x_0-\bmu\right)}{2}
$$

$$
\frac{-2\sqrt{\balpha_t} \left(\x_t -\sqrt{\balpha_t} \x_0\right)}{2(1-\balpha_t)} + \frac{2{\boldsymbol{\Sigma}}_0^{-1} \left(\x_0 - \boldsymbol{\mu}_0\right)}{2} = 0
$$
This simplifies to:
$$
\frac{-\sqrt{\balpha_t} \left(\x_t - \sqrt{\balpha_t} \x_0\right)}{(1-\balpha_t)} + {\boldsymbol{\Sigma}}_0^{-1} \left(\bar{\x}_0 - \boldsymbol{\mu}_0\right) = 0
$$
Resulting in:
$$
-\sqrt{\balpha_t} {\boldsymbol{\Sigma}}_0 \x_t + \balpha_t {\boldsymbol{\Sigma}}_0 \x_0 + (1-\balpha_t) \x_0 - (1-\balpha_t) \boldsymbol{\mu}_0 = 0
$$
Thus:
$$
\left( \balpha_t  {\boldsymbol{\Sigma}}_0 + (1-\balpha_t)\I \right) \x_0 = \sqrt{\balpha_t} {\boldsymbol{\Sigma}}_0 \x_t + (1-\balpha_t) \boldsymbol{\mu}_0 
$$
Finally:

\begin{equation}
\x_0^* = \left(\balpha_t {\boldsymbol{\Sigma}}_0 + (1-\balpha_t)\I \right)^{-1} \left(\sqrt{\balpha_t}{\boldsymbol{\Sigma}}_0 \x_t + (1-\balpha_t) \boldsymbol{\mu}_0 \right)
\end{equation}

% \section{Evaluating the Inference Process in the Time Domain}\label{sec:appendix_inference_Time}
\section{ The Reverse Process in the Time Domain}
\label{sec:appendix_inference_Time}
% \textbf{The Inference Process in the Time Domain:}

% \subsection{The Inference Process in the Time Domain:}\label{sec:appendix_inference_spectral_subsec_1}
Here, we present the reverse process in the time domain for the DDIM \cite{song2020denoising}, as outlined in Lemma \ref{lemma:ddim_time_infer}.

Let $\x_0$ follow the distribution:
$$ 
\x_0 \sim \mathcal{N}(\bmu, \bSigmaZ), \quad \x_0 \in \R^{d}
$$
Using the procedure outline in \cite{song2020denoising}, the diffusion process begins with $ \x_S\sim \mathcal{N}(\boldsymbol{0}, \I)$, where $ \x_S \in \R^{d} $ and progresses through an iterative denoising process described as follows:\footnote{We follow here the DDIM notations that replaces $t$ with $s$, where the steps \(\left[1, \ldots, S\right]\) form a subsequence of \(\left[1, \ldots, T\right]\) and \(S = T\).}
\begin{equation}
    \x_{s-1}(\eta) = \sqrt{\balpha_{s-1}} \left( 
    \frac{\x_{s} - \sqrt{1 - \balpha_{s}} \cdot \bepsilon_\theta(\x_{s},s)}{\sqrt{\balpha_{s}}}\right)
    + \sqrt{1 - \balpha_{s-1} - \sigma_{s}^2(\eta)} \cdot \bepsilon_\theta(\x_s,s) + \sigma_{s}(\eta)\mathbf{z}_s
\end{equation}

where
\begin{equation}\label{eq:sgma_eta}
    \sigma_{s}(\eta) = \eta \sqrt{\frac{1 - \balpha_{{s-1}}}{1 - \balpha_s}} \sqrt{1 - \frac{\balpha_{s}}{\balpha_{{s-1}}}}
\end{equation}  
Substituting the marginal property from \eqref{eq:marginal_dist}:
\begin{equation}
\boldsymbol{\epsilon}_\theta(\x_s, s) = \frac{\x_s-\sqrt{{\balpha}_s}\hat{\x}_0}{\sqrt{1-\balpha_s}} \quad \quad \hat{\x}_0 = \frac{\x_{s} - \sqrt{1 - \balpha_{s}} \cdot \bepsilon_\theta(\x_{s},s)}{\sqrt{\balpha_{s}}}
\end{equation}

\begin{equation}
    \x_{s-1}(\eta) = \sqrt{\balpha_{s-1}} \hat{\x}_0
    + \sqrt{1 - \balpha_{s-1} - \sigma_{s}^2(\eta)} \left(\frac{\x_s-\sqrt{{\balpha}_s}\hat{\x}_0}{\sqrt{1-\balpha_s}}\right) + \sigma_{s}(\eta)\mathbf{z}_s
\end{equation}
For the deterministic scenario, we choose \(\eta = 0\) in \eqref{eq:sgma_eta} and obtain $ \sigma_{s}(\eta=0) = 0$. Therefore:
\begin{align}\nonumber
    \x_{s-1}(\eta=0) &= \sqrt{\balpha_{s-1}} \hat{\x}_0
    + \sqrt{1 - \balpha_{s-1}} \left(\frac{\x_s-\sqrt{{\balpha}_s}\hat{\x}_0}{\sqrt{1-\balpha_s}}\right) \\
     &= \frac{\sqrt{1 - \balpha_{s-1}}}{\sqrt{1-\balpha_s}}\x_s + \left[  \sqrt{\balpha_{s-1}} - \frac{\sqrt{{\balpha}_s}\sqrt{1 - \balpha_{s-1}}}{\sqrt{1 - \balpha_{s}}}\right]\hat{\x}_0
\end{align}
We denote the following:
$$a_s = \frac{\sqrt{1 - \balpha_{s-1}}}{\sqrt{1-\balpha_s}} \quad \quad b_s =   \sqrt{\balpha_{s-1}} - \frac{\sqrt{{\balpha}_s}\sqrt{1 - \balpha_{s-1}}}{\sqrt{1 - \balpha_{s}}}$$
Therefore we get the following equation:
$$
\x_{s-1} = a_s\x_s +b_s\x_0^* ~.
$$

% ##########

Using the result from the MAP estimator:
$$\x_0^* = \left(\balpha_s \bSigmaZ + (1-\balpha_s) \I \right)^{-1} \left(\sqrt{\balpha_s} \bSigmaZ \x_t + (1-\balpha_s) \bmu \right)$$
we get: 
$$
\x_{s-1} = a_s\x_s +b_s \left(\balpha_s {\boldsymbol{\Sigma}}_0 + (1-\balpha_s)\I \right)^{-1} \left(\sqrt{\balpha_s} {\boldsymbol{\Sigma}}_0 \x_s + (1-\balpha_s) \boldsymbol{\mu}_0 \right ) 
$$

 \begin{equation}
\x_{s-1} =  \left(a_s+ b_s \left(\balpha_s {\boldsymbol{\Sigma}}_0 + (1-\balpha_s)\I \right)^{-1}\sqrt{\balpha_s} {\boldsymbol{\Sigma}}_0 \right)\x_s + \left(b_s \left(\balpha_s {\boldsymbol{\Sigma}}_0 + (1-\balpha_s)\I \right)^{-1}(1-\balpha_s) \boldsymbol{\mu}_0 \right ) 
\end{equation}
Introduce the notation:
$$\bar{\boldsymbol{\Sigma}}_{0,s} = \balpha_s {\boldsymbol{\Sigma}}_0 + (1-\balpha_s)\I$$
We can rewrite the equation as:

\begin{equation}
\x_{s-1} = \left(a_s\I + b_s \sqrt{\balpha_s}\left(\bar{\boldsymbol{\Sigma}}_{0,s} \right)^{-1}{\boldsymbol{\Sigma}}_0 \right)\x_s + b_s\left(1-\balpha_s\right)\left(\bar{\boldsymbol{\Sigma}}_{0,s}\right)^{-1}    \boldsymbol{\mu_0}
\end{equation}

\newpage

% ###########

\section{Migrating to the Spectral Domain}\label{sec:appendix_inference_spectral}
Here, we demonstrate the application of projecting both sides of Eq.~\eqref{eq:ddim_time_infer} onto the eigenbasis of $\bSigma_0$ as outlined in Lemma \ref{Lemma:migrating_spectral_domain}. Since $\bSigma_0$ is symmetric, its eigenvalue decomposition can written as $\bSigma_0 = \evmatrix \bLambda \evmatrix^{\T}$,
where the columns of $\evmatrix$ represent the eigenvectors, and $\bLambda$  is the diagonal matrix containing the corresponding eigenvalues.\footnote{In case the covariance matrix $\bSigma_0$ is circulant, it can be diagonalized using the Discrete Fourier Transform (DFT) denoted by \( \mathcal{F}\{\cdot\} \). In this case, the projection exposes the frequency components, enabling their explicit analysis.}

\begin{equation}
    \evmatrixT \x_{s-1} =   \evmatrixT \left[ \left(a_s\I + b_s\sqrt{\balpha_s} \bar{\bSigma}^{-1}_{0,s}\bSigma_0 \right) \x_s + 
b_s(1-\balpha_s) \bar{\bSigma}^{-1}_{0,s}\boldsymbol{\mu_0}\right]
\end{equation}

\begin{equation}\nonumber
 \evmatrixT \x_{s-1} = a_s \evmatrixT \x_{s} + \evmatrixT b_s \sqrt{\balpha_s}\bar{\boldsymbol{\Sigma}}_{0,s} ^{-1}\evmatrix \evmatrixT  {\boldsymbol{\Sigma}}_0\evmatrix\evmatrixT \mathbf{x}_s  + \evmatrixT b_s (1-\balpha_s)  \bar{\boldsymbol{\Sigma}}_{0,s}^{-1} \evmatrix   \evmatrixT\boldsymbol{\mu_0}  \end{equation}

 By introducing the notations $\ve = \evmatrixT \x$ and $\projectedmu = \evmatrixT \bmu$, we obtain:

\begin{equation}\nonumber
\projectedx_{s-1} = a_s  \projectedx_{s} +  b_s \sqrt{\balpha_s}\evmatrixT\bar{\boldsymbol{\Sigma}}_{0,s} ^{-1}\evmatrix \evmatrixT  {\boldsymbol{\Sigma}}_0\evmatrix \projectedx_{s} + b_s (1-\balpha_s)  \evmatrixT\bar{\boldsymbol{\Sigma}}_{0,s}^{-1} \evmatrix\projectedmu    \end{equation}

By applying eigenvalue decomposition, the symmetric matrix  $\bSigma_0$ can be diagonalized using its eigenbasis $\evmatrix$, where $\evmatrix^{-1}$ = $\evmatrixT$ Specifically:
\begin{itemize}
\item $\bSigma_0  = \evmatrix \mathbf{\Lambda}_0 \evmatrixT$, \quad  $ \mathbf{\Lambda}_0 = \evmatrixT \bSigma_0 \evmatrix$
\item $a \bSigma_0 + b \I = a \evmatrix \mathbf{\Lambda}_0 \evmatrixT + b \evmatrix \I \evmatrixT = \evmatrix \left(a \mathbf{\Lambda}_0 + b\I\right) \evmatrixT$
\item $ \bSigma_0^{-1} = \evmatrix \mathbf{\Lambda}_0^{-1} \evmatrixT $  , \quad $\mathbf{\Lambda}_0^{-1} = \evmatrixT\boldsymbol{\Sigma}_0^{-1} \evmatrix $
\end{itemize}

Therefore, we obtain:

$$
 \bar{\boldsymbol{\Sigma}}_{0,s}^{-1}  =  \left[\balpha_s \boldsymbol{\Sigma}_0 + (1-\balpha_s)\I\right]^{-1} = \left[\balpha_s  \evmatrix \mathbf{\Lambda}_0 \evmatrixT + (1-\balpha_s) \evmatrix \I \evmatrixT\right]^{-1} =  \left[\evmatrix\left(\balpha_s   \mathbf{\Lambda}_0 + (1-\balpha_s) \I   \right)\evmatrixT\right]^{-1}
$$
$$  \bar{\boldsymbol{\Sigma}}_{0,s}^{-1}  = \evmatrix\left[\balpha_s \mathbf{\Lambda}_0 + (1-\balpha_s)\I\right]^{-1}\evmatrixT$$

Incorporating these elements into the main equation, we obtain:

% Including those elements in the main equation,
% While inducing the following notations $\ve = \evmatrixT \x$ and $\projectedmu = \evmatrixT \bmu$ we obtain:

\begin{equation}\nonumber
\projectedx_{s-1} = a_s \projectedx_{s} +  b_s\sqrt{\balpha_s} \left[\balpha_s \mathbf{\Lambda}_0 + (1-\balpha_s)\I\right]^{-1}  \bLambda \projectedx_{s} + b_s (1-\balpha_s)  \left[\balpha_s \mathbf{\Lambda}_0 + (1-\balpha_s)\I\right]^{-1} \projectedmu   \end{equation}

\begin{equation}\nonumber 
\projectedx_{s-1} = \left[ a_s\I  + b_s\sqrt{\balpha_s} \left[\balpha_s \mathbf{\Lambda}_0 + (1-\balpha_s)\I\right]^{-1}  \mathbf{\Lambda}_0 \right] \projectedx_{s} + b_s(1-\balpha_s)\left[\balpha_s \mathbf{\Lambda}_0 + (1-\balpha_s)\I\right]^{-1}\projectedmu
\end{equation}

We will denote the following:

$$\G(s) = \left[ a_s  + b_s \sqrt{\balpha_s}\left[\balpha_s \mathbf{\Lambda}_0 + (1-\balpha_s)\I\right]^{-1}  \mathbf{\Lambda}_0 \right] $$

$$\M(s) = b_s(1-\balpha_s)\left[\balpha_s \mathbf{\Lambda}_0 + (1-\balpha_s)\I\right]^{-1} $$

and get:

\begin{equation}
\projectedx_{s-1} = \G(s) \projectedx_s + \M(s)\projectedmu
\end{equation}

We can then recursively obtain 
$\projectedx_l$ 
for a general $l$:
\begin{equation}
\label{eq:itermidate_step_diffusion_process}
\projectedx_{l} = \left[ \prod_{s^{'}=l+1}^{S}\G(s^{'})\right] \projectedx_S + \left[\sum_{i=l+1}^{S}\left(\prod_{j=l+1}^{i-1}\G(j)\right)\M(i)\right]\projectedmu
\end{equation}
specifically for $l=0$:
\begin{equation}
\label{eq:last_step_diffusion_process}
\hat{\projectedx}_{0}= \left[ \prod_{s^{'}=1}^{S}\G(s^{'})\right] \projectedx_S + \left[\sum_{i=1}^{S}\left(\prod_{j=1}^{i-1}\G(j)\right)\M(i)\right]\projectedmu
\end{equation}

We will denote the following:
\begin{equation}
 \D_1 =  \prod_{s=1}^{S}\G(s) = \prod_{s=1}^{S} \left[ a_s  + b_s \sqrt{\balpha_s} \left[\balpha_s \mathbf{\Lambda}_0 + \I(1-\balpha_s)\right]^{-1} \mathbf{\Lambda}_0\right]
\end{equation}

% \begin{equation}
% \projectedx_{s-1} = \G(s) \projectedx_s + \M(s)\projectedmu 
% \end{equation}

\begin{align}
 \D_2 &=\sum_{i=1}^{S}\left(\prod_{j=1}^{i-1}\G(j)\right)\M(i) \\ \nonumber &=  \sum_{i=1}^{S}\left[\left(\prod_{j=1}^{i-1}
 \left[ a_j  + b_j \sqrt{\balpha_j}\left[\balpha_j \mathbf{\Lambda}_0 + \I(1-\balpha_j)\right]^{-1} \mathbf{\Lambda}_0 \right]
 \right)
 b_i(1-\balpha_i)\left[\balpha_i\mathbf{\Lambda}_0 + \I(1-\balpha_i)\right]^{-1}
 \right]
\end{align}

Substitute $D_1$ and $D_2$ into the last equation, we get:
\begin{equation}\label{eq:X_0_func_of_x_T}
\hat{\projectedx}_{0} = \D_1 \projectedx_S + \D_2\projectedmu
\end{equation}

% ############################

The resulting vector from Equation \ref{eq:X_0_func_of_x_T} is a linear combination of Gaussian signals, therefore it also follows a Gaussian distribution.
% Since the resulting vector is a linear
% combination of Gaussian signals, it follows a Gaussian distribution.
We now aim to determine the mean and covariance of that distribution.
% at the output of the diffusion process.

\begin{equation}\nonumber
  \hat{\projectedx}_{0} \sim \mathcal{N}(\mathbb{E}  \left[\hat{\projectedx}_{0} \right], \mathbf{\Sigma}_{\hat{\projectedx}_{0}}), \quad \hat{\projectedx}_{0} \in \mathbb{R}^{d}  
\end{equation}
\textbf{Mean:}
\begin{equation}\nonumber
 \mathbb{E}  \left[\hat{\projectedx}_{0} \right]= \mathbb{E}  \left[  \D_1 \projectedx_S+ \D_2\projectedmu \right] = \D_1\F\mathbb{E} \left[  \projectedx_S\right] +  \mathbb{E}\left[\D_2\projectedmu \right] = \D_2\projectedmu 
\end{equation}
$$ \projectedx_S \sim  \mathcal{N}(\0,\I)$$
\begin{equation}\nonumber
\mathbb{E}  \left[\hat{\projectedx}_{0}  \right] = \D_2\projectedmu 
\end{equation}
% \begin{empheq}[box=\fbox]{align}
% \mathbb{E}  \left[\hat{\projectedx}_{0}  \right] = D_2\projectedmu
% \end{empheq}
% \[
%  \mathbb{E}  \left[ \mathbf{x}_{0}^{\mathcal{F}} \right]= F \mathbb{E}  \left[ \mathbf{x}_{0} \right] = F \boldsymbol{\mu_0}
% \]
\textbf{Covariance:}
\begin{equation}\nonumber
\mathbf{\Sigma}_{\hat{\projectedx}_{0} } = \mathbb{E}\left[\left(\D_1 \projectedx_S+ \D_2 \projectedmu - \mathbb{E}\left[\D_1 \projectedx_S+ \D_2 \projectedmu\right]\right)\left(\D_1 \projectedx_S+ \D_2 \projectedmu - \mathbb{E}\left[\D_1 \projectedx_S+ \D_2 \projectedmu\right]\right)^T \right]
\end{equation}
\begin{equation}\nonumber
= \mathbb{E}\left[\left(\D_1 \projectedx_S+ \D_2 \projectedmu - \D_2 \projectedmu\right)\left(\D_1 \projectedx_S+ \D_2 \projectedmu - \D_2 \projectedmu\right)^T \right]
\end{equation}
\begin{equation}\nonumber
= \mathbb{E}\left[\left(\D_1 \projectedx_S\right)\left(\D_1 \projectedx_S\right)^T \right] = \D_1 \mathbb{E}\left[\projectedx_S\left(\projectedx_S\right)^T\right] {\D_1}^T
\end{equation}
% \begin{equation}\nonumber
% \end{equation}
% \[
% \Sigma_{\hat{\projectedx}_{0} }  = D_1 {D_1}^T = {D_1}^2
% \]
% \begin{empheq}[box=\fbox]{align}
% \Sigma_{\hat{\projectedx}_{0} }  = D_1 {D_1}^T = {D_1}^2
% \end{empheq}
% \begin{empheq}[box=\fbox]{align}
% \Sigma_{\mathbf{x}_{0}^{\mathcal{F}}} =  \bLambda
% \end{empheq}
\begin{equation}\nonumber
\mathbf{\Sigma}_{\hat{\projectedx}_{0} }  = \D_1 {\D_1}^T = {\D_1}^2
\end{equation}
% with regard to the data distribution $\x_0 \sim \mathcal{N}(\bmu, \bSigmaZ), \quad \x_0 \in \R^{d}$, its first and second moments in the frequency domain are defined as follows:
% It is worth noting that,

\begin{equation}
\label{fig:v_o_hat_distribution}
\hat{\projectedx}_{0}  \sim \N(\D_2 \projectedmu, \D_1^2), \quad \hat{\projectedx}_{0}  \in \mathbb{R}^{d}.
\end{equation}

% It should be noted that 

The same procedure used to derive the distribution of $\hat{\projectedx}_{0}$ in \eqref{fig:v_o_hat_distribution} from \eqref{eq:last_step_diffusion_process} can similarly be applied to any intermediate step $\projectedx_l$ in \eqref{eq:itermidate_step_diffusion_process}, enabling exploration of the spectral dynamics throughout the inference process.

% It should be noted that the same procedure we made for deriving the distribution of $\hat{\projectedx}_{0}$ in \eqref{fig:v_o_hat_distribution}
% from 
% \eqref{eq:last_step_diffusion_process} could be done to any intermediate step $\projectedx_l$ in \eqref{eq:itermidate_step_diffusion_process}, enabling the exploration of the spectral dynamics through the inference process.

For the data distribution  $\x_0 \sim \mathcal{N}(\bmu, \bSigmaZ), ~~ \text{where} ~~   \x_0 \in \R^{d}$, its first and second moments in the frequency domain are given by:
\begin{equation}\nonumber
 \mathbb{E}  \left[ \projectedx_0 \right]= \evmatrixT \mathbb{E}  \left[ \mathbf{\x}_{0} \right] = \projectedmu 
\end{equation}
\begin{equation}\nonumber
\Sigma_{\projectedx_0} =  \bLambda
\end{equation}

\begin{equation}
\ \projectedx_0 \sim \mathcal{N}(\projectedmu ,\bLambda)
\end{equation}

\newpage
\newcommand{\projectedxhat}{{\hat{\projectedx}}_0}
\newcommand{\projectedxz}{\boldsymbol{\projectedx}_0}

\section{Evaluating loss functions expressions}\label{sec:appendix_loss_functions}

% Here, we define selected loss functions using the terms defined in Appendix \ref{seq:mathematical_derivative}.

Here, we present selected loss functions based on the derivations provided in Section \ref{seq:mathematical_derivative}.

\subsection{Wasserstein-2 Distance:}

The Wasserstein-2 distance between two Gaussian distributions with means  $\mu_1$ and $\mu_2$, and covariance matrices $\Sigma_1$ and $\Sigma_2$,  and the corresponding eigenvalues $\{\lambda_1^{(i)}\}_{i=1}^d$ and $\{\lambda_2^{(i)}\}_{i=1}^d$  is given by:

\begin{equation}
W_2(\mathcal{N}_1, \mathcal{N}_2) = \sqrt{(\boldsymbol{\mu}_1 - \boldsymbol{\mu}_2)^T (\boldsymbol{\mu}_1 - \boldsymbol{\mu}_2) + \sum_i \left( \sqrt{\lambda_1^{(i)}} - \sqrt{\lambda_2^{(i)}} \right)^2}
\end{equation}

$$ \text{Since } \quad \projectedxhat \sim \mathcal{N}(\D_2 \projectedmu, \D_1^2), \quad \projectedxhat \in \mathbb{R}^{d} \quad \text{and} \quad \projectedxz \sim \mathcal{N}( \projectedmu,\bLambda), \quad \projectedxz \in \mathbb{R}^{d}\quad $$

we obtain:

\begin{equation}\nonumber
\left(\boldsymbol{\mu}_1 - \boldsymbol{\mu}_2\right)^T(\boldsymbol{\mu}_1 - \boldsymbol{\mu}_2) = \left(\D_2\projectedmu - \projectedmu\right)^T\left(\D_2\projectedmu - \projectedmu\right) =  \left((\projectedmu)^T\D_2^T - (\projectedmu)^T\right)\left(\D_2\projectedmu - \projectedmu\right)
\end{equation}

\begin{equation} \nonumber
= (\projectedmu)^T \projectedmu - 2 (\projectedmu)^T \D_2  \projectedmu + (\projectedmu)^T \D_2^T \D_2 \boldsymbol{\mu_0} ^\mathcal{F}
\end{equation}

\begin{equation} \nonumber
= \sum_{i=1}^{d}
{\left(\projectedmu\right)}_i
^2   - 2\sum_{i=1}^{d}  {\left(\projectedmu\right) }_i^2  {\D_2}^{(i)}     +    \sum_{i=1}^{d} {\left(\projectedmu\right)  }_i^2 \left({\D_2}^{(i)}\right)^2  
\end{equation}

\begin{equation} \nonumber
=\sum_{i=1}^{d}\left({\D_2}^{(i)}-1\right)^2  {\left(\projectedmu\right)  }_i^2
\end{equation}

\begin{equation}\nonumber
 \left(\boldsymbol{\mu}_1 - \boldsymbol{\mu}_2\right)^T(\boldsymbol{\mu}_1 - \boldsymbol{\mu}_2) =\sum_{i=1}^{d}\left({\D_2}^{(i)}-1\right)^2  {\left(\projectedmu\right)  }_i^2
\end{equation}

\begin{equation}\nonumber
\sum_i \left( \sqrt{\lambda_1^{(i)}} - \sqrt{\lambda_2^{(i)}} \right)^2 = \sum_i \left( \sqrt{\lambda_0^{(i)}} - \sqrt{\left(  {\D_1}^{(i)}  \right)^2} \right)^2
\end{equation}

\begin{equation}
W_2( \projectedxz , \projectedxhat ) = \sqrt{\sum_{i=1}^{d}\left({d_2}^{(i)}-1\right)^2  {\left(\projectedmu\right)  }_i^2 + \sum_i \left( \sqrt{\lambda_0^{(i)}} - \sqrt{\left(  {\D_1}^{(i)}  \right)^2} \right)^2 }
\end{equation}
% \end{empheq}
\subsection{Kullback-Leibler divergence:}
\label{seq:dkl_derivative}
The Kullback-Leibler (KL) divergence between two Gaussian distributions with means $\mu_1$ and $\mu_2$, and covariance matrices $\Sigma_1$ and $\Sigma_2$,  and the corresponding eigenvalues $\{\lambda_1^{(i)}\}_{i=1}^d$ and $\{\lambda_2^{(i)}\}_{i=1}^d$  is given by:

\begin{equation} \nonumber
D_{\text{KL}} \left(\mathcal{N}(\boldsymbol{\mu}_1, \bSigma_1) \parallel \mathcal{N}(\boldsymbol{\mu}_2, \bSigma_2) \right) = \frac{1}{2} \left( \log \frac{|\bSigma_2|}{|\bSigma_1|} - d + \operatorname{tr}\left( \bSigma_2^{-1} \bSigma_1 \right) + (\boldsymbol{\mu}_2 - \boldsymbol{\mu}_1)^T \bSigma_2^{-1} (\boldsymbol{\mu}_2 - \boldsymbol{\mu}_1) \right)
\end{equation}

Given: $$\quad \projectedxhat \sim \mathcal{N}(\D_2 \projectedmu, \D_1^2), \quad \projectedxhat \in \mathbb{R}^{d} \quad  \text{and} \quad \projectedxz \sim \mathcal{N}( \projectedmu,\bLambda), \quad \projectedxz \in \mathbb{R}^{d}\quad$$

 The \emph{KL divergence} is given by:

\begin{equation} \nonumber
 D_{\text{KL}}\left(\projectedxz \parallel \projectedxhat \right) = D_{\text{KL}} \left( \mathcal{N}\left(  \mathbb{E}  \left[ \projectedxz \right] ,  \bSigma_{\projectedxz}  \right), \, \mathcal{N}\left(\mathbb{E}  \left[ \projectedxhat \right], \bSigma_{\projectedxhat} \right) \right)   
\end{equation}

\begin{equation}\nonumber
= D_{\text{KL}} \left( \mathcal{N}\left( \projectedmu ,   \mathbf{\Lambda}_0 \right), \, \mathcal{N}\left(\D_2\ , \projectedmu {\D_1}^2 \right) \right)
\end{equation}

By decomposing the KL divergence elements, we obtain the following terms:

\begin{itemize}
    \item $|\bSigma_2| = |\D_1^T\D_1| = |\D_1^2| =  \prod_{i=1}^{d} {\D_1^{(i)}}^2 
    % =\prod_{i=1}^{d} \left[\left( \prod_{s=1}^{S} \left[ a_s + b_s \left( k_{1,s}^2 \lambda_0^{(i)} + I k_{2,s}^2 \right)^{-1} k_{1,s} \lambda_0^{(i)} \right] \right)^2 \right]
    $
   
    \item $|\bSigma_1| = \prod_{i=1}^{d} \lambda_0^{(i)}$

    \item $\operatorname{tr}\left(\bSigma_2^{-1} \bSigma_1\right) = \sum_{i=1}^{d}\frac{ \lambda_0^{(i)}}{    {\D_1^{(i)}}^2} $
    
    % \item $\operatorname{tr}\left(\Sigma_2^{-1} \Sigma_1\right) = \sum_{i=1}^{d}\frac{ \lambda_0^{(i)}}{ \prod_{s=1}^{S} \left[ a_s + b_s \left( k_{1,s}^2 \lambda_0^{(i)} + I k_{2,s}^2 \right)^{-1} k_{1,s} \lambda_0^{(i)} \right]} $
    \item $ (\boldsymbol{\mu}_2 - \boldsymbol{\mu}_1)^T \bSigma_2^{-1} (\boldsymbol{\mu}_2 - \boldsymbol{\mu}_1) = (\D_2\projectedmu-\projectedmu)^T \left({{\D_1}^2}\right)^{-1} (\D_2\projectedmu-\projectedmu)=$
    % $$ ((\projectedmu)^T\D_2^T-(\projectedmu)^T) \left({{\D_1}^2}\right)^{-1}(\D_2\projectedmu-F\boldsymbol{\mu_0})$$
    
    $$= \left(\projectedmu\right)^T\left(\D_2^T-I\right)\left({{\D_1}^2}\right)^{-1} \left(\D_2-I\right)\projectedmu  = \sum_{i=1}^{d} \frac{\left({
  \D_2}^{(i)}-1\right)^2}{{\D_1^{(i)}}^2} {\left({\projectedmu}_i\right)}^2  $$

    % $(\boldsymbol{\mu_0}^T\left(D_2^T-I\right) \left({{D_1}^2}\right)^{-1} \left(D_2-I\right)\boldsymbol{\mu_0}) = \sum_{i=1}^{d} \frac{\left({d_2}^{(i)}-1\right)^2}{{d_1^{(i)}}^2}{{\mu_0}^{(i)}}^2 $

  % $$(\boldsymbol{\mu}_0^T\left(\D_2^T-I\right) \left({{\D_1}^2}\right)^{-1} \left(\D_2-I\right)\boldsymbol{\mu_0})$$
  
  % $$
  % = \sum_{i=1}^{d} \frac{\left({
  % \D_2}^{(i)}-1\right)^2}{{\D_1^{(i)}}^2} {\left(\projectedmu_i\right)}^2 $$
 
\end{itemize}

Applying the substitution, the term results in:

\begin{equation}\nonumber
D_{\text{KL}} \left( \mathcal{N}\left( \projectedmu ,   \mathbf{\Lambda}_0 \right), \, \mathcal{N}\left(\D_2\projectedmu, {\D_1}^2 \right) \right) = 
\end{equation}

% $$
% = \frac{1}{2} \left[ \log \frac{|\Sigma_2|}{|\Sigma_1|} - d + \operatorname{tr}\left(\Sigma_2^{-1} \Sigma_1\right) + (\mu_2 - \mu_1)^T \Sigma_2^{-1} (\mu_2 - \mu_1) \right]
% $$

% \sum_{i=1}^{d} \frac{\left({d_2}_i-1\right)^2}{{d_1^2}_i}{{\mu_0}_i}^2

\begin{equation}\nonumber
= \frac{1}{2} \left[ \sum_{i=1}^{d} \log{\D_1^{(i)}}^2 -  \sum_{i=1}^{d} \log{\lambda_0^{(i)}} \ - d +\sum_{i=1}^{d}\frac{ \lambda_0^{(i)}}{    {\D_1^{(i)}}^2 }  + \sum_{i=1}^{d} \frac{\left({\D_2}^{(i)}-1\right)^2}{{\D_1^{(i)}}^2} {\left({\projectedmu}_i\right)  }^2  \right]
\end{equation}

% \[
% = \frac{1}{2} \left[ \sum_{i=1}^{d} \log{d_1^{(i)}}^2 -  \sum_{i=1}^{d} \log{\lambda_0^{(i)}} \ - d +\sum_{i=1}^{d}\frac{ \lambda_0^{(i)}}{    {d_1^{(i)}}^2 }  + \sum_{i=1}^{d} \frac{\left({d_2}^{(i)}-1\right)^2}{{d_1^{(i)}}^2}{{\mu_0}^{(i)}}^2 \right]
% \]

% \begin{empheq}[box=\fbox]{align}
% D_{\text{KL}}  = \frac{1}{2} \left[ \sum_{i=1}^{d} 2\log{d_1^{(i)}} -  \sum_{i=1}^{d} \log{\lambda_0^{(i)}} \ - d + \sum_{i=1}^{d}\frac{ \lambda_0^{(i)} + \left({d_2}^{(i)}-1\right)^2{{\mu_0}^{(i)}}^2 }{    {d_1^{(i)}}^2 }  \right]
% \end{empheq}
% \begin{empheq}[box=\fbox]{align}
% D_{\text{KL}}  = \frac{1}{2} \left[ \sum_{i=1}^{d} 2\log{d_1^{(i)}} -  \sum_{i=1}^{d} \log{\lambda_0^{(i)}} \ - d + \sum_{i=1}^{d}\frac{ \lambda_0^{(i)} + \left({d_2}^{(i)}-1\right)^2   {\left(\projectedmu\right)  }_i^2 }{    {d_1^{(i)}}^2 }  \right]
% \end{empheq}
\begin{equation}
 D_{\text{KL}}\left(\projectedxz \parallel \projectedxhat \right)  = \frac{1}{2} \left[ \sum_{i=1}^{d} 2\log{\D_1^{(i)}} -  \sum_{i=1}^{d} \log{\lambda_0^{(i)}} \ - d + \sum_{i=1}^{d}\frac{ \lambda_0^{(i)} + \left({\D_2}^{(i)}-1\right)^2   {\left(\projectedmu\right)  }_i^2 }{    {\D_1^{(i)}}^2 }  \right]
\end{equation}
\section{DDPM Formulation:}\label{sec:appendix_DDPM_Formulation}

Here, we apply an equivalent procedure to the DDPM scenario, as we did for the DDIM, as outlined in Theorem \ref{theorem:ddap_spectral_eq}.
% \subsection{Evaluating the Inference Process in the Time Domain}
\subsection{The Reverse Process in the Time Domain}

Using the procedure outline in \cite{ho2020denoising}, the diffusion process begins with $ \x_T\sim \mathcal{N}(\boldsymbol{0}, \I)$, where $ \x_T \in \R^{d} $, and progresses through an iterative denoising process described as follows:
\begin{equation}\label{eq:DDPM_inference_time}
\x_{t-1} = \frac{1}
{\sqrt{{\alpha}_t}}\left(\x_t - \frac{1-\alpha_t}{\sqrt{1-\bar{\alpha}_t}}\boldsymbol{\epsilon}_\theta(\x_t, t)\right) + \sigma_t \mathbf{z}_t \quad
\mathbf{z}_t \sim \mathcal{N}(0, I)
\end{equation}

Where $ \sigma_t = \sqrt{\frac{1-\bar{\alpha}_{t-1}}{1-\bar{\alpha}_t}(1-\alpha_t)}$.

Given the marginal property from \eqref{eq:marginal_dist}:
\begin{equation}\nonumber
\boldsymbol{\epsilon}_\theta(\x_t, t) = \frac{\x_t-\sqrt{{\bar{\alpha}}_t}\hat{\x}_0}{\sqrt{1-\bar{\alpha}_t}}
\end{equation}
we can incorporate it into \eqref{eq:DDPM_inference_time}:
\begin{eqnarray}\nonumber
\x_{t-1} &=& \frac{1}{\sqrt{{\alpha}_t}}\left(\x_t - \frac{1-\alpha_t}{\sqrt{1-\bar{\alpha}_t}}\left(\frac{\x_t-\sqrt{{\bar{\alpha}}_t}\hat{\x}_0}{\sqrt{1-\bar{\alpha}_t}}\right)\right) + \sigma_t \mathbf{z}_t \\\nonumber
\x_{t-1} &=& \frac{1}{\sqrt{{\alpha}_t}}\left(\x_t\left(1-\frac{1-\alpha_t}{1-\bar{\alpha}_t}\right)+ \frac{\left(1-\alpha_t\right)\sqrt{\bar{\alpha}_t}}{1-\bar{\alpha}_t}\hat{\x}_0\right) + \sigma_t \mathbf{z}_t \\
\x_{t-1} &=& \frac{1}{\sqrt{{\alpha}_t}} \left(\frac{\alpha_t - \bar{\alpha}_t}{1-\bar{\alpha}_t}\right)\x_t + \sqrt{\frac{\bar{\alpha}_t}{\alpha_t}}\frac{\left(1-\alpha_t\right)}{1-\bar{\alpha}_t} \hat{\x}_0 + \sigma_t \mathbf{z}_t
\end{eqnarray}

We denote the following, where the final term in each equation is represented by $\bar{\alpha}_t$ and $\bar{\alpha}_{t-1}$:
\begin{equation}\nonumber
a_t = \frac{1}{\sqrt{{\alpha}_t}} \left(\frac{\alpha_t - \bar{\alpha}_t}{1-\bar{\alpha}_t}\right)  = \frac{\sqrt{\bar{\alpha}_t}}{1-\bar{\alpha}_t}\left[ \frac{1}{\sqrt{\bar{\alpha}_{t-1}}} - \sqrt{\bar{\alpha}_{t-1}}\right] 
\end{equation}

\begin{equation}\nonumber
b_t = \sqrt{\frac{\bar{\alpha}_t}{\alpha_t}}\frac{\left(1-\alpha_t\right)}{1-\bar{\alpha}_t} = \sqrt{\bar{\alpha}_{t-1}}\left(\frac{1-\frac{\bar{\alpha}_t}{\bar{\alpha}_{t-1}}}{1-\bar{\alpha}_t}\right)
\end{equation}
% $$ c_t = \sigma_t = \frac{1-\bar{\alpha}_{t-1}}{1-\bar{\alpha}_t}\beta_t = \textcolor{blue}{\frac{1-\bar{\alpha}_{t-1}}{1-\bar{\alpha}_t} \left( 1-\frac{\bar{\alpha}_t}{\bar{\alpha}_{t-1}}\right)} $$

% \begin{equation}\nonumber
% c_t = \sigma_t = \sqrt{\frac{1-\bar{\alpha}_{t-1}}{1-\bar{\alpha}_t}\beta_t} = \sqrt{\frac{1-\bar{\alpha}_{t-1}}{1-\bar{\alpha}_t} \left( 1-\frac{\bar{\alpha}_t}{\bar{\alpha}_{t-1}}\right)}
% \end{equation}

\begin{equation}\nonumber
c_t = \sigma_t = \sqrt{\frac{1-\bar{\alpha}_{t-1}}{1-\bar{\alpha}_t}(1-\alpha_t)} = \sqrt{\frac{1-\bar{\alpha}_{t-1}}{1-\bar{\alpha}_t} \left( 1-\frac{\bar{\alpha}_t}{\bar{\alpha}_{t-1}}\right)}
\end{equation}

Therefore we get the following equation:
\begin{equation}\nonumber
\mathbf{x}_{t-1} = a_t\mathbf{x}_t +b_t\hat{\mathbf{x}}_0 + c_t\mathbf{z}_t
\end{equation}
Using the result for the MAP estimator from \eqref{eq:wiener_filter}:

\begin{equation}\nonumber
\x_0^* = \left(\balpha_t {\boldsymbol{\Sigma}}_0 +(1-\balpha_t)\I \right)^{-1} \left( \sqrt{\balpha_t} {\boldsymbol{\Sigma}}_0 \mathbf{x}_t + (1-\balpha_t) \boldsymbol{\mu}_0 \right)
\end{equation}
we get: 
\begin{equation}\nonumber
\mathbf{x}_{t-1} = a_t\x_t +b_t \left(\balpha_t {\boldsymbol{\Sigma}}_0 + (1-\balpha_t)\I \right)^{-1} \left( \sqrt{\balpha_t} {\boldsymbol{\Sigma}}_0 \mathbf{x}_t + (1-\balpha_t) \boldsymbol{\mu}_0 \right ) + c_t\mathbf{z}_t
\end{equation}

\begin{equation}\nonumber
\mathbf{x}_{t-1} =  \left(a_t+ b_t \left(\balpha_t {\boldsymbol{\Sigma}}_0 + (1-\balpha_t)\I \right)^{-1} \sqrt{\balpha_t} {\boldsymbol{\Sigma}}_0 \right)\mathbf{x}_t  + \left(b_t \left(\balpha_t {\boldsymbol{\Sigma}}_0 + (1-\balpha_t)\I \right)^{-1}(1-\balpha_t) \boldsymbol{\mu}_0 \right ) + c_t\mathbf{z}_t
\end{equation}
Using the notation from Appendix \ref{sec:appendix_inference_Time}:
$$\bar{\boldsymbol{\Sigma}}_{0,t} = \balpha_t {\boldsymbol{\Sigma}}_0 + (1-\balpha_t)\I$$
Thus, we can rewrite the equation as:
\begin{equation}\label{DDPM_time_infer}
\mathbf{x}_{t-1} =  \left(a_t+ b_t \sqrt{\balpha_t}  \left(\bar{\boldsymbol{\Sigma}}_{0,t} \right)^{-1}{\boldsymbol{\Sigma}}_0 \right)\mathbf{x}_t +b_t(1-\balpha_t) \left(\bar{\boldsymbol{\Sigma}}_{0,t}\right)^{-1} \boldsymbol{\mu_0} + c_t\mathbf{z}_t
\end{equation}

\subsection{Migrating to the Spectral Domain}

% Here, we demonstrate the application of projecting both sides of Eq.~\eqref{eq:ddim_time_infer} onto the eigenbasis of $\bSigma_0$ as outlined in Lemma \ref{Lemma:migrating_spectral_domain}. Since $\bSigma_0$ is symmetric, its eigenvalue decomposition can written as $\bSigma_0 = \evmatrix \bLambda \evmatrix^{\T}$,
% where the columns of $\evmatrix$ represent the eigenvectors, and $\bLambda$  is the diagonal matrix containing the corresponding eigenvalues. \footnote{In case the covariance matrix $\bSigma_0$ is circulant, it can be diagonalized using the Discrete Fourier Transform (DFT) denoted by \( \mathcal{F}\{\cdot\} \). In this case, the projection exposes the frequency components, enabling their explicit analysis.}

Next, we project both sides of the Eq.~\eqref{DDPM_time_infer} onto the spectral domain. As previously described in Appendix \ref{sec:appendix_inference_spectral}, this step relies on the eigenvalue decomposition of the covariance matrix $\bSigma_0 =  \evmatrix \bLambda \evmatrix^{\T}$ where the columns of $\evmatrix$ represent the eigenvectors, and $\bLambda$  is the diagonal matrix containing the corresponding eigenvalues.

% 
% ##########

\begin{equation}
    \evmatrixT\mathbf{x}_{t-1} =   \evmatrixT\left[\left(a_t+ b_t \sqrt{\balpha_t} \left(\bar{\boldsymbol{\Sigma}}_{0,t} \right)^{-1}{\boldsymbol{\Sigma}}_0 \right)\mathbf{x}_t + b_t(1-\balpha_t)\left(\bar{\boldsymbol{\Sigma}}_{0,t}\right)^{-1}  \boldsymbol{\mu_0} + c_t\mathbf{z}_t\right]
\end{equation} 

\begin{equation}\nonumber
 \evmatrixT\mathbf{x}_{t-1} = a_t \evmatrixT\mathbf{x}_{t} +\evmatrixT b_t \sqrt{\balpha_t} \left(\bar{\boldsymbol{\Sigma}}_{0,t} \right)^{-1} \evmatrix\evmatrixT {\boldsymbol{\Sigma}}_0 \evmatrix\evmatrixT\mathbf{x}_t  + \evmatrixT b_s (1-\balpha_t)\left(\bar{\boldsymbol{\Sigma}}_{0,s}\right)^{-1}\evmatrix\evmatrixT  \boldsymbol{\mu_0} + c_t\evmatrixT\mathbf{z}_t
\end{equation} 

 By introducing the notations $\ve = \evmatrixT \x$ and $\projectedmu = \evmatrixT \bmu$, we obtain:

\begin{equation} \nonumber
\projectedx_{t-1} = a_t \projectedx_{t} +  b_t \sqrt{\balpha_t} \evmatrixT\left(\bar{\boldsymbol{\Sigma}}_{0,t} \right)^{-1} \evmatrix\evmatrixT {\boldsymbol{\Sigma}}_0 \evmatrix \projectedx_{t}   + b_t (1-\balpha_t)\evmatrixT\left(\bar{\boldsymbol{\Sigma}}_{0,t}\right)^{-1}\evmatrix  \projectedmu  +c_t\projectedzt
\end{equation}

% \begin{equation}
%     \mathcal{F}\left\{\mathbf{x}_{t-1}\right\} =   \mathcal{F}\left\{\left(a_t+ b_t \sqrt{\balpha_t} \left(\bar{\boldsymbol{\Sigma}}_{0,t} \right)^{-1}{\boldsymbol{\Sigma}}_0 \right)\mathbf{x}_t + b_t(1-\balpha_t)\left(\bar{\boldsymbol{\Sigma}}_{0,t}\right)^{-1}  \boldsymbol{\mu_0} + c_t\mathbf{z}_t\right\}
% \end{equation} 

% \begin{equation}\nonumber
% \mathbf{x}^{\mathcal{F}}_{t-1} = a_t \mathbf{x}^{\mathcal{F}}_{t} + \F b_t \sqrt{\balpha_t} \left(\bar{\boldsymbol{\Sigma}}_{0,t} \right)^{-1} \F^T\F {\boldsymbol{\Sigma}}_0 \F^T\F\mathbf{x}_t  + \F b_s (1-\balpha_t)\left(\bar{\boldsymbol{\Sigma}}_{0,s}\right)^{-1}\F^T\F   \boldsymbol{\mu_0} +\F c_t\mathbf{z}_t
% \end{equation} 

% \begin{equation} \nonumber
% \mathbf{x}^{\mathcal{F}}_{t-1} = a_t \mathbf{x}^{\mathcal{F}}_{t} +  b_t \sqrt{\balpha_t} \F \left(\bar{\boldsymbol{\Sigma}}_{0,t} \right)^{-1} \F^T \F{\boldsymbol{\Sigma}}_0 \F^T\mathbf{x}^{\mathcal{F}}_{t}   + b_t (1-\balpha_t)\F\left(\bar{\boldsymbol{\Sigma}}_{0,t}\right)^{-1}\F^T  \projectedmu +c_t\projectedzt
% \end{equation} 

By applying eigenvalue decomposition, the symmetric matrix  $\bSigma_0$ can be diagonalized using its eigenbasis $\evmatrix$, where $\evmatrix^{-1}$ = $\evmatrixT$ . Therefore, we obtain:

$$
 \bar{\boldsymbol{\Sigma}}_{0,s}^{-1}  =  \left[\balpha_s \boldsymbol{\Sigma}_0 + (1-\balpha_s)\I\right]^{-1} = \left[\balpha_s  \evmatrix \mathbf{\Lambda}_0 \evmatrixT + (1-\balpha_s) \evmatrix \I \evmatrixT\right]^{-1} =  \left[\evmatrix\left(\balpha_s   \mathbf{\Lambda}_0 + (1-\balpha_s) \I   \right)\evmatrixT\right]^{-1}
$$
$$  \bar{\boldsymbol{\Sigma}}_{0,s}^{-1}  = \evmatrix\left[\balpha_s \mathbf{\Lambda}_0 + (1-\balpha_s)\I\right]^{-1}\evmatrixT$$

% Assuming circulancy, the matrix $\boldsymbol{\Sigma}_0$ can be diagonalized by the discrete Fourier transform (DFT) matrix.

% Therefore, we obtain:
% \begin{equation} \nonumber
% \F \left(\bar{\boldsymbol{\Sigma}}_{0,t}\right)^{-1} \F^T = \F \left[\balpha_t \boldsymbol{\Sigma}_0 + (1-\balpha_t)\I\right]^{-1} \F^T = \left[\balpha_t \mathbf{\Lambda}_0 + (1-\balpha_t)\I\right]^{-1}
% \end{equation} 

Incorporating these elements into the main equation, we obtain:

\begin{equation} \nonumber
\projectedx_{t-1} = a_t \projectedx_{t} + b_t \sqrt{\balpha_t} \left[\balpha_t \mathbf{\Lambda}_0 + (1-\balpha_t)\I\right]^{-1}  \mathbf{\Lambda}_0 \projectedx_{t} + b_t(1-\balpha_t)\left[\balpha_t \mathbf{\Lambda}_0 +(1-\balpha_t)\I\right]^{-1}\projectedmu + c_t \projectedzt
\end{equation} 

\begin{equation} \nonumber
\projectedx_{t-1} = \left[a_t \I  + b_t \sqrt{\balpha_t} \left[\balpha_t \mathbf{\Lambda}_0 + (1-\balpha_t)\I\right]^{-1}  \mathbf{\Lambda}_0\right] \projectedx_{t} + b_t(1-\balpha_t)\left[\balpha_t \mathbf{\Lambda}_0 + (1-\balpha_t)\I\right]^{-1}\projectedmu + c_t \projectedzt
\end{equation}

Including those elements in the main equation:
\begin{equation} \nonumber
\projectedx_{t-1} = a_t \projectedx_{t} + b_t \sqrt{\balpha_t} \left[\balpha_t \mathbf{\Lambda}_0 + (1-\balpha_t)\I\right]^{-1}  \mathbf{\Lambda}_0 \projectedx_{t} + b_t(1-\balpha_t)\left[\balpha_t \mathbf{\Lambda}_0 +(1-\balpha_t)\I\right]^{-1}\projectedmu + c_t \projectedzt
\end{equation} 

\begin{equation} \nonumber
\projectedx_{t-1} = \left[a_t \I  + b_t \sqrt{\balpha_t} \left[\balpha_t \mathbf{\Lambda}_0 + (1-\balpha_t)\I\right]^{-1}  \mathbf{\Lambda}_0\right] \projectedx_{t} + b_t(1-\balpha_t)\left[\balpha_t \mathbf{\Lambda}_0 + (1-\balpha_t)\I\right]^{-1}\projectedmu + c_t \projectedzt
\end{equation} 
We will denote the following:
\begin{equation} \nonumber
\G(t) = \left[ a_t  + b_t\sqrt{\balpha_t}  \left[\balpha_t \mathbf{\Lambda}_0 + (1-\balpha_t)\I\right]^{-1}  \mathbf{\Lambda}_0 \right] 
\end{equation}

\begin{equation} \nonumber
\M(t) = b_t(1-\balpha_t)\left[\balpha_t \mathbf{\Lambda}_0 + (1-\balpha_t)\I\right]^{-1}
\end{equation}
and get: 
\begin{equation}\nonumber
\projectedx_{t-1} = \G(t) \projectedx_{t} +  \M(t)\projectedmu+  c_t\projectedzt
\end{equation}

% #########
We can then recursively obtain
$\projectedx_l$ for a general $l$:
\begin{equation}\nonumber
\projectedx_l = \left[ \prod_{t^{'}=l+1}^{T}\G(t^{'})\right] \projectedx_{t} + \left[\sum_{i=l+1}^{T}\left(\prod_{j=l+1}^{i-1}\G(j)\right)\M(i)\right]\projectedmu + \left[\sum_{i=l+1}^{T}\left(\prod_{j=l+1}^{i-1}\G(j)\right)c_i \projectedzi \right]
\end{equation}
where the process iterates over all the steps: $\left[1, \ldots, T\right]$.
\begin{equation}\nonumber
\projectedxhat = \left[ \prod_{t^{'}=1}^{T}\G(t^{'})\right] \projectedx_{t}   + \left[\sum_{i=1}^{T}\left(\prod_{j=1}^{i-1}\G(j)\right)c_i \projectedzi \right] + \left[\sum_{i=1}^{T}\left(\prod_{j=1}^{i-1}\G(j)\right)\M(i)\right]\projectedmu
\end{equation}
We will denote the following:
\begin{equation}\nonumber
 \D_1 =  \prod_{s=1}^{S}\G(s)
\end{equation}

\begin{equation}\nonumber
 \D_2 =\sum_{i=1}^{S}\left(\prod_{j=1}^{i-1}\G(j)\right)\M(i) 
\end{equation}

Substitute $\D_1$ and $\D_2$ into the last equation, we get:
% \begin{empheq}[box=\fbox]{align}
% \projectedxhat = D_1 \projectedx_{t} + \left[\sum_{i=1}^{T}\left(\prod_{j=1}^{i-1}g(j)\right)c_i I \projectedzi \right] + D_2\projectedmu
% \end{empheq}

\begin{equation}\label{eq:appen_DDPM_X_0_X_T}
\projectedxhat = \D_1 \projectedx_{t} + \left[\sum_{i=1}^{T}\left(\prod_{j=1}^{i-1}\G(j)\right)c_i \projectedzi \right] + \D_2\projectedmu
\end{equation}

The resulting vector from Equation \ref{eq:appen_DDPM_X_0_X_T} is a linear combination of Gaussian signals, therefore it also follows a Gaussian distribution.
% Since the resulting vector is a linear
% combination of Gaussian signals, it follows a Gaussian distribution.
We now aim to determine the mean vector and the covariance matrix of that distribution.
\begin{equation}\nonumber
  \projectedxhat \sim \mathcal{N}(\mathbb{E}  \left[\projectedxhat \right], \mathbf{\Sigma}_{\projectedxhat}), \quad \projectedxhat \in \mathbb{R}^{d}  
\end{equation}
$$ \mathbf{\x}_T \sim  \mathcal{N}(\0,\I) \quad \text{and} \quad \mathbf{z}_i\sim  \mathcal{N}(\0,\I) , \quad \forall i$$
% $$\mathbf{\hat{x}_{0}^{\mathcal{F}}} \sim \mathcal{N}(\mathbb{E}  \left[\projectedxhat \right], \Sigma_{\projectedxhat}), \quad \projectedxhat \in \mathbb{R}^{d}  $$
\textbf{Mean:}
$$
 \mathbb{E}  \left[\projectedxhat \right]= \mathbb{E}  \left[ \D_1 \projectedx_{t} + \left[\sum_{i=1}^{T}\left(\prod_{j=1}^{i-1}\G(j)\right)c_i \projectedzi\I \right] + \D_2\projectedmu \right] = \D_2\projectedmu
$$
\begin{equation}\nonumber
\mathbb{E}  \left[\projectedxhat \right]= \D_2\projectedmu
\end{equation}

% \begin{empheq}[box=\fbox]{align}
% \mathbb{E}  \left[\projectedxhat \right]= D_2\projectedmu
% \end{empheq}
% \begin{empheq}[box=\fbox]{align}
%  \mathbb{E}  \left[ \mathbf{x}_{0}^{\mathcal{F}} \right]= F \mathbb{E}  \left[ \mathbf{x}_{0} \right] = F \boldsymbol{\mu_0}
% \end{empheq}

\textbf{Covariance:}
$$
\bSigma_{\projectedxhat} = \mathbb{E}[\left(\D_1 \projectedx_{t} +  \D_2\projectedmu + \left[\sum_{i=1}^{T}\left(\prod_{j=1}^{i-1}\G(j)\right)c_i \projectedzi \right]  - \mathbb{E}\left[\D_1 \projectedx_{t} +  \D_2\projectedmu + \left[\sum_{i=1}^{T}\left(\prod_{j=1}^{i-1}\G(j)\right)c_i \projectedzi \right] \right]\right)$$

$$
\left(\D_1 \projectedx_{t} +  \D_2\projectedmu + \left[\sum_{i=1}^{T}\left(\prod_{j=1}^{i-1}\G(j)\right)c_i \projectedzi \right] - \mathbb{E}\left[\D_1 \projectedx_{t} +  \D_2\projectedmu + \left[\sum_{i=1}^{T}\left(\prod_{j=1}^{i-1}\G(j)\right)c_i \projectedzi \right] \right]\right)^T ]
$$
\begin{equation}\label{eq:cov_DDPM}\nonumber
\bSigma_{\projectedxhat} = \mathbb{E}\left[\left(\D_1 \projectedx_{t}  + \left[\sum_{i=1}^{T}\left(\prod_{j=1}^{i-1}\G(j)\right)c_i \projectedzi \right] \right)
\left(\D_1 \projectedx_{t}  + \left[\sum_{i=1}^{T}\left(\prod_{j=1}^{i-1}\G(j)\right)c_i \projectedzi \right]\right)^T \right] 
\end{equation}

% \hdashrule[0.1ex]{16cm}{1pt}{3mm 2mm}
\begin{lemma}\label{lemma:covariance_linear_comb}
Let $\mathbf{x}_1, \mathbf{x}_2, \dots, \mathbf{x}_n$ be $n$ independent Gaussian random vectors with mean $\mathbb{E}[\mathbf{x}_i] = \mathbf{0}$ and covariance matrices $\text{Cov}(\mathbf{x}_i) = \mathbf{\Sigma}_i$, for $i = 1, 2, \dots, n$. Let the linear combination be defined as:
$$
\mathbf{y} = \sum_{i=1}^n a_i \mathbf{x}_i,
$$
where $a_1, a_2, \dots, a_n$ are constants.
The covariance matrix of $\mathbf{y}$, denoted as $\text{Cov}(\mathbf{y})$, is given by:
$$
\text{Cov}(\mathbf{y}) = \sum_{i=1}^n a_i^2 \mathbf{\Sigma}_i.
$$
\end{lemma}

Applying the result of Lemma \ref{lemma:covariance_linear_comb} to the expression in Equation \ref{eq:cov_DDPM}, where $ \mathbf{x}_T \sim \mathcal{N}(\0, \I) ~ \text{and} ~ \mathbf{z}_i \sim \mathcal{N}(\0, \I) $ are independent Gaussian noises for all $i$, we have:

\begin{equation} \nonumber
\mathbb{E}  \left[
\projectedx_{t} {\projectedx_{t}} ^T
\right] = \mathbb{E}  \left[
\evmatrixT\mathbf{x}_T \mathbf{x}_T^T\evmatrix
\right] =   \evmatrixT \mathbb{E}  \left[
\mathbf{x}_T \mathbf{x}_T^T
\right] \evmatrix = \I
\end{equation}

\begin{equation} \nonumber
 \mathbb{E}  \left[
\projectedzi {\projectedzi}^T \right]
= \I 
\end{equation}

Thus, the covariance is given by:

\begin{equation} \nonumber
\bSigma_{\projectedxhat} =\mathbb{E}  \left[  \projectedxhat \projectedxhat^T\right] =  \left[ \D_1\right]^2 + \sum_{i=1}^{T}\left[\left(\prod_{j=1}^{i-1}\G(j)\right)c_i\I \right]^2
\end{equation}

\begin{equation}\nonumber
\bSigma_{\projectedxhat} =  \left[ \D_1\right]^2 + \sum_{i=1}^{T}\left[\left(\prod_{j=1}^{i-1}\G(j)\right)c_i \I \right]^2
\end{equation}

\begin{equation}
\projectedxhat \sim \N\!\!\left( \D_2\projectedmu ~,~  \D_1^2 + \sum_{i=1}^{T}\Big(\prod_{j=1}^{i-1}\G^2(j)\Big)c^2_i \I \right)
\end{equation} 

% \ref{sec:appendix_Defining_the_requirements},

As discussed in Appendix \ref{sec:appendix_inference_spectral}, for a data distribution $\x_0 \sim \mathcal{N}(\bmu, \bSigmaZ),$ where $ ~~  \x_0 \in \R^{d}$, the first and second moments in the frequency domain are given by:

\begin{equation}\nonumber
 \mathbb{E}  \left[ \projectedx_0 \right]= \F \mathbb{E}  \left[ \mathbf{\x}_{0} \right] = \projectedmu 
\end{equation}
\begin{equation}\nonumber
\bSigma_{\mathbf{x}_{0}^{\mathcal{F}}} =  \bLambda
\end{equation}

\begin{equation}
\projectedx_0 \sim \mathcal{N}(\projectedmu ,\bLambda)
\end{equation}

\section{Variance preserving and Variance exploding theoretical analysis}\label{sec:appendix_VP_VE}

The paper \cite{song2020score} distinguishes between two sampling methods: \emph{Variance Preserving} (VP) and \emph{Variance Exploding} (VE). The primary difference lies in how variance evolves during the process. while VP maintains a fixed variance, VE results in an exploding variance as $t \rightarrow T$. Here, we focus on comparing these approaches within the context of our spectral noise schedule derivation for the DDIM procedure \cite{ho2020denoising}.
Throughout this paper, we described our methods based on the \emph{Variance Preserving} (VP) formulation, given by:

\begin{equation}
p(\x_t|\x_0) = \mathcal{N} \sim \left(\sqrt{\balpha_t}\x_0,\sqrt{1-\balpha_t}\I\right)
\end{equation}
where the only hyperparameters are the noise schedule parameters:
$\left\{ {{\balpha}_s} \right\}_{s=0}^S$ where ${{\balpha}_s} \in (0,1]$.

In contrast, under the Variance Exploding (VE) method, the hyperparameters are given by $\sigma_t$ where $\sigma_t \in [0,\infty)$, and the marginal distribution takes the form: 
\begin{equation}
p(\bar{\x}_t|\x_0) = \mathcal{N} \sim (\bar{\x}_0,\sigma_t^2\I)
\end{equation}
We used the notation $\bar{\x}_t$ to distinguish it from $\x_t$, except in the special case where $\x_0 = \bar{\x}_0$.
Applying the reparameterization trick, we obtain:

\begin{equation}
    \text{VP:} ~~\mathbf{x}_t = \sqrt{\bar{\alpha}_t} \mathbf{x}_0 + \sqrt{1-\bar{\alpha}_t} \boldsymbol{\epsilon} ~~~~~  \boldsymbol{\epsilon} \sim \mathcal{N}(\mathbf{0},\boldsymbol{I})
     \label{eq:diffusion_forward_VP}
\end{equation}

\begin{equation}
    \text{VE:}~~ \mathbf{\bar{\x}}_t =  \mathbf{\bar{\x}}_0 + \sigma_t \boldsymbol{\epsilon} ~~~~~  \boldsymbol{\epsilon} \sim \mathcal{N}(\mathbf{0},\boldsymbol{I})
     \label{eq:diffusion_forward_VE}
\end{equation}

A key relationship between the VP and VE formulations, as derived in \cite{kawar2022denoising}, is given by:

\begin{equation}\label{eq:VP_VE}
\bar{\x}_t = \frac{\x_t}{\sqrt{1+\sigma_t^2}} \quad \quad \balpha_t = \frac{1}{\left(1+\sigma_t^2\right)}
\end{equation}

\subsection{Determining the Optimal Denoiser:}

Following the derivation in \ref{sec:appendix_Map_est}, we obtained the expression for the optimal denoiser in the Gaussian case under the Variance Preserving (VP) formulation:

\begin{equation}\nonumber \mathbf{\hat{x}}_0^{\text{wiener,VP}} = \left({\bar{\alpha}_t}{\boldsymbol{\Sigma}}_0 + I\left(1-\bar{\alpha}_t\right) \right)^{-1} \left(\sqrt{\bar{\alpha}_t}{\boldsymbol{\Sigma}}_0 \mathbf{x}_t + \left(1-\bar{\alpha}_t\right) \boldsymbol{\mu}_0 \right). \label{eq:Wiener_filter} \end{equation}

Leveraging a similar approach, we derive the corresponding expression for the Variance Exploding (VE) scenario:

\begin{equation} \mathbf{\hat{x}}_0^{\text{wiener,VE}} = \left({\boldsymbol{\Sigma}}_0 + I \sigma_t^2\right)^{-1} \left({\boldsymbol{\Sigma}}_0 \bar{\mathbf{x}}_t + \sigma_t^2 \boldsymbol{\mu}_0 \right). \label{eq:Wiener_filter_VE} \end{equation}

\subsection{Evaluating the Inference Process in the time Domain:}
This part can be performed using two equivalent methods:

\textbf{Method 1:}

The ODE for the VE scenario in DDIM, as outlined in \cite{song2020denoising}, is given by:

\begin{equation}\label{eq:ODE_VE_DDIM}
d\bar{x} = -\frac{1}{2}g(t)^2\nabla_{\bar{x}}\log{p_t(\bar{x})}dt ~~~~ g(t) = \sqrt{\frac{d\sigma^2(t)}{dt}}
\end{equation}

Additionally, the score expression and the marginal equation are also derived in \cite{song2020denoising} as follows:

\begin{equation}\label{eq:score_func}
\nabla_{\bar{x}}\log{p_t(\bar{x})} = - \frac{\epsilon_\theta^{(t)}}{\sigma(t)}
\end{equation}
\begin{equation}\label{eq:marginal_vp}
\bar{x}_t=\bar{x}_0+\sigma_t\epsilon \quad where \quad \epsilon \sim \mathcal{N}(\0,\I)
\end{equation}
Substituting Equation \ref{eq:score_func} into Equation \ref{eq:ODE_VE_DDIM}:
\begin{equation}\nonumber
d\bar{x} = \frac{1}{2}\frac{d\sigma^2(t)}{dt} \frac{\epsilon_\theta^{(t)}}{\sigma(t)}{dt}
\end{equation}

\begin{equation}\label{eq:ODE_formulation_ve}
\frac{d\bar{x}}{dt} = \frac{d\sigma(t)}{dt} \epsilon_\theta^{(t)}
\end{equation}

Substituting Equation \ref{eq:marginal_vp} into Equation \ref{eq:ODE_formulation_ve}, we obtain:

\begin{equation}\nonumber
\bar{x}_t - \bar{x}_{t-1} = \left(\sigma_t-\sigma_{t-1}\right) \frac{{\bar{x}_t-\bar{x}_0}}{\sigma_t}
\end{equation}
\begin{equation} \nonumber
\bar{x}_{t-1} = \bar{x}_t + \left(\frac{\sigma_{t-1}}{\sigma_t}-1\right){\bar{x}_t-\bar{x}_0}
\end{equation}
\begin{equation} 
\label{eq:method_2}
\bar{x}_{t-1} =  \frac{\sigma_{t-1}}{\sigma_t}{\bar{x}_t+\left(1-\frac{\sigma_{t-1}}{\sigma_t}\right)\bar{x}_0}
\end{equation}
% $$ d\bar{x} = -\frac{1}{2}g(t)^2\nabla_{\bar{x}}\log{p_t(\bar{x})}dt ~~~~ g(t) = \sqrt{\frac{d\sigma^2(t)}{dt}} $$
\textbf{Method 2:}

given the inference process in the VP formulation \cite{song2020denoising}:\footnote{We follow here the DDIM notations that replaces $t$ with $s$, where the steps \(\left[1, \ldots, S\right]\) form a subsequence of \(\left[1, \ldots, T\right]\) and \(S = T\).}

$$
    \x_{s-1}(\eta=0) = \frac{\sqrt{1 - \bar{\alpha}_{s-1}}}{\sqrt{1-\bar{\alpha}_s}}\mathbf{x}_s + \left[  \sqrt{\bar{\alpha}_{s-1}} - \frac{\sqrt{{\bar{\alpha}}_s}\sqrt{1 - \bar{\alpha}_{s-1}}}{\sqrt{1 - \bar{\alpha}_{s}}}\right]\hat{\mathbf{x}}_0
$$

By leveraging the connections in Equation \ref{eq:VP_VE}
we can derive the following relationship between the two successive steps, 
$\x_{s-1}$ and $\x_s$, in the inference process:

\begin{equation}
\label{eq:method_1}
\bar{x}_{s-1} = \sqrt{\frac{{\sigma}^2_{s-1}}{{\sigma}^2_{s}}}\bar{x}_{s}+\left(1-\sqrt{\frac{{\sigma}^2_{s-1}}{{\sigma}^2_{s}}}\right){x}_{0}
\end{equation}
% \textcolor{red}{SDE}
% $$x_s = \frac{1}{\sqrt{{{\sigma}^2_s+1}}}\bar{x}_s \quad \bar{\alpha}_s=\frac{1}{{\sigma}^2_s+1}$$
\\
The Resulted expressions in \eqref{eq:method_1} and \eqref{eq:method_2} are identical.

By defining the following terms:\begin{equation} \nonumber 
\bar{a}_s =\frac{\sigma_{s-1}}{\sigma_s}
\end{equation}
\begin{equation} \nonumber
\bar{b}_s =  1- \frac{\sigma_{s-1}}{\sigma_s} = 1- \bar{a}_s
\end{equation}
we can express the relationship between $\x_{s-1} $ and $\x_s$ as:
\begin{equation}\label{eq:final_eq_time_inference_ve}
\mathbf{x}_{s-1} = a_s\mathbf{x}_s +b_s\hat{\mathbf{x}}_0 
\end{equation}

\subsection{Evaluating the Inference Process in the Spectral Domain}

Since a similar expression to Equation \ref{eq:final_eq_time_inference_ve} has already been discussed in Appendix \ref{sec:appendix_inference_spectral}, we can now describe the inference process in the spectral domain as follows:

\begin{equation}
\projectedx_{s-1} = \G(s) \projectedx_{s} + \M(s)\projectedmu
\end{equation}
where: 
$$\G(s) = \left[ \bar{a}_s  + \bar{b}_s \left[ \mathbf{\Lambda}_0 + \I\sigma_s^2\right]^{-1} \mathbf{\Lambda}_0 \right] $$
$$\M(s) = \bar{b}_s\left[ \mathbf{\Lambda}_0 + \I\sigma_s^2\right]^{-1}\sigma_s^2 $$

Following this and in alignment with the same methodology described in Appendix \ref{sec:appendix_inference_spectral} we obtain:

\begin{equation}
\label{eq:spectral_ve}
\projectedxhat = \D_1 \projectedx_{s} + \D_2\projectedmu
\end{equation}

\begin{equation}\nonumber
  \projectedxhat \sim \mathcal{N}(\mathbb{E}  \left[\projectedxhat \right], \mathbf{\Sigma}_{\projectedxhat}), \quad \projectedxhat \in \mathbb{R}^{d}  
\end{equation}
% $$ \mathbf{\x}_T \sim  \mathcal{N}(\0,\I)$$

\begin{equation}\nonumber
\mathbb{E}  \left[\projectedxhat \right] = \D_2\projectedmu, \quad \mathbf{\Sigma}_{\projectedxhat}  = \D_1 {\D_1}^T = {\D_1}^2 
\end{equation}

\begin{equation}
 \D_1 =  \prod_{s=1}^{S}\G(s) = \prod_{s=1}^{S} \left[ a_s  + b_s \sqrt{\balpha_s} \left[\balpha_s \mathbf{\Lambda}_0 + \I(1-\balpha_s)\right]^{-1} \mathbf{\Lambda}_0\right]
\end{equation}

% \begin{equation}
%  \D_2 =\sum_{i=1}^{S}\left(\prod_{j=1}^{i-1}\G(j)\right)\M(i) = \sum_{i=1}^{S}\left[\left(\prod_{j=1}^{i-1} 
%  \left[ a_j  + b_j \sqrt{\balpha_j}\left[\balpha_j \mathbf{\Lambda}_0 + \I(1-\balpha_j)\right]^{-1} \mathbf{\Lambda}_0 \right]
%  \right) 
%  b_i(1-\balpha_i)\left[\balpha_i\mathbf{\Lambda}_0 + \I(1-\balpha_i)\right]^{-1}
%  \right]
% \end{equation}

\begin{align}
 \D_2 &=\sum_{i=1}^{S}\left(\prod_{j=1}^{i-1}\G(j)\right)\M(i) \\&= \sum_{i=1}^{S}\left[\left(\prod_{j=1}^{i-1}
 \left[ a_j  + b_j \sqrt{\balpha_j}\left[\balpha_j \mathbf{\Lambda}_0 + \I(1-\balpha_j)\right]^{-1} \mathbf{\Lambda}_0 \right]
 \right)
 b_i(1-\balpha_i)\left[\balpha_i\mathbf{\Lambda}_0 + \I(1-\balpha_i)\right]^{-1}
 \right]
\end{align}

\newpage
\section{Clarifications and Validations:}\label{sec:appendix_clarifications_and_validations}

\subsection{Method Evaluations: Temporal and Spectral Domains}
\label{sec:appendix_clarifications_and_validations_Method_Evaluation}

\begin{figure}[H]
    \centering
    \begin{subfigure}{0.45\textwidth}
        \centering
        \includegraphics[width={\textwidth} ]{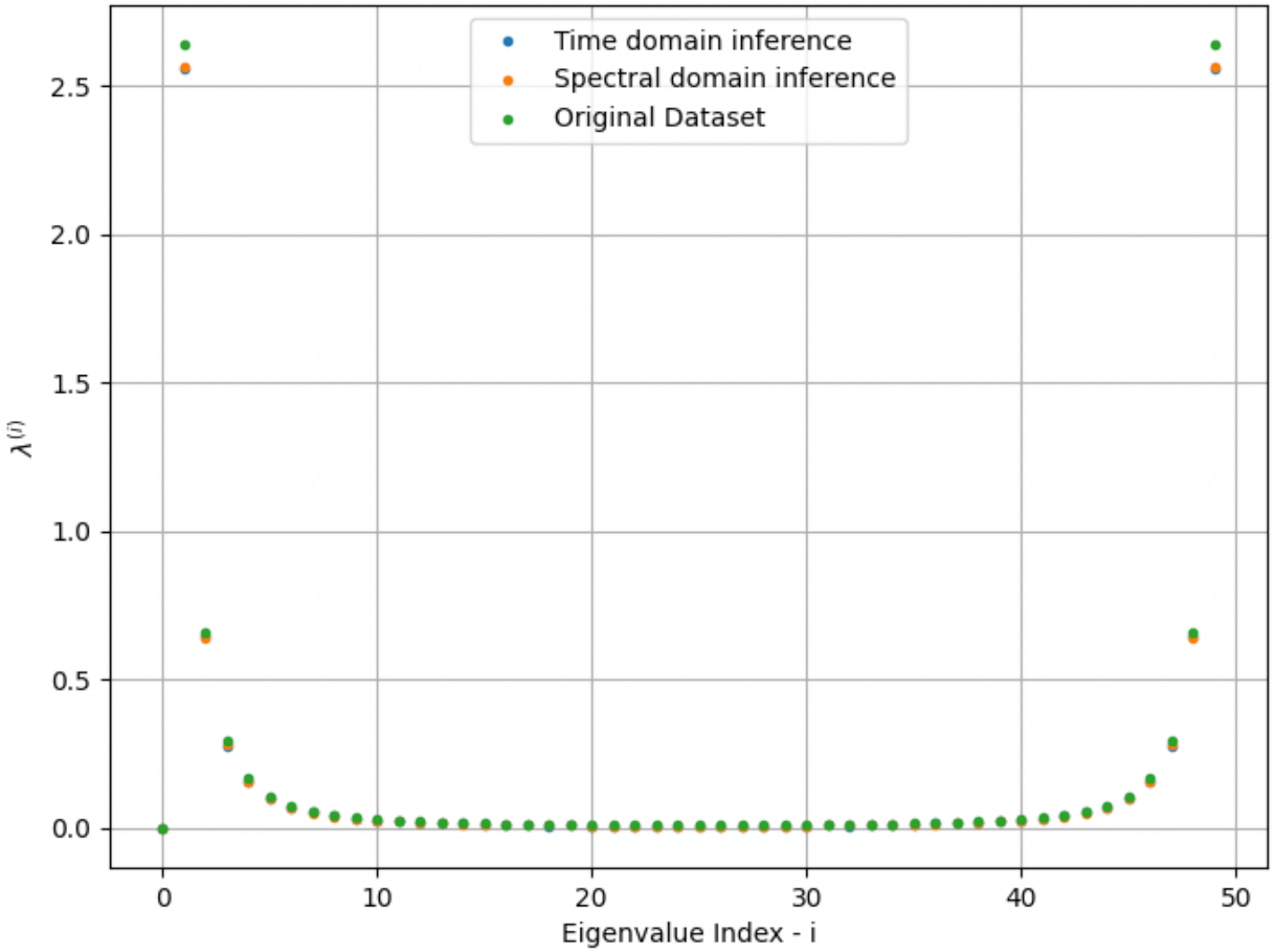}
        \caption{A comparison of the eigenvalues obtained from each method}
        \label{fig:Sanity_values}
    \end{subfigure}
     \hfill % Adjust horizontal space as needed
    \begin{subfigure}{0.45\textwidth}
        \centering
        \includegraphics[width={\textwidth} ]{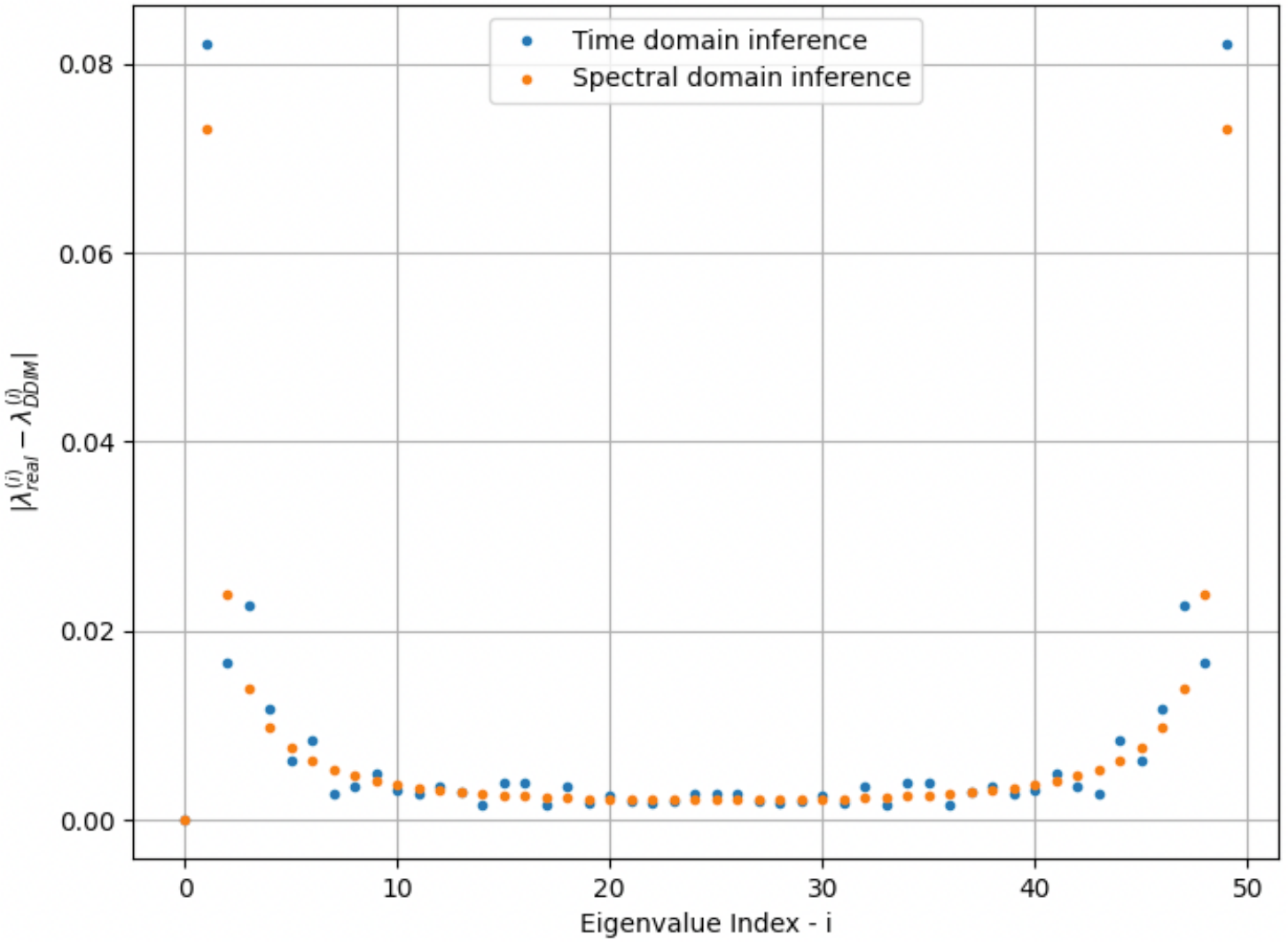}
        \caption{The absolute relative error between the estimated eigenvalues and the data eigenvalues.}
        \label{fig:Sanity_relative_Diff}
    \end{subfigure}
    \caption{Figure \ref{fig:Sanity_values} compares the eigenvalues derived from the spectral and time domain formulations of the DDIM method \cite{ho2020denoising}. The dataset, described in \ref{subsec:Scenario_1} with $l = 0.1$ and $\text{d} = 50$, is used for both approaches, involving $112$ diffusion steps and following the linear noise schedule proposed in \cite{song2020denoising}. Furthermore, \autoref{fig:Sanity_relative_Diff} illustrates the absolute error between the estimated and original eigenvalues for both methods.}
\label{fig:Sanity_check_time_and_frequency}
\end{figure}

We evaluated the compatibility between the diffusion process in the time domain, using the DDIM method \citep{ho2020denoising}, and its counterpart derived from Equation \ref{eq:D_1,D_2} in the spectral domain. Using an artificial covariance matrix, $\bSigma_0$, with parameters $l=0.1$ and $d=50$ from \ref{subsec:Scenario_1}, we estimated the covariance of $50,000$ signals that were denoised according to Equation \ref{eq:DDIM_sampling_procedure}, using the optimal denoiser from Equation \ref{eq:wiener_filter}, and computed their eigenvalues, denoted as  $\{ \lambda_i^{time} \}_{i=1}^{d}$. In the spectral domain, we applied the formulation from Equation \ref{eq:D_1,D_2} for deriving $D_1^2$ and extracted $\{ \lambda_i^{spectral} \}_{i=1}^{d}$ from its diagonal elements. The results are illustrated in Figure \ref{fig:Sanity_check_time_and_frequency}.

Figure \ref{fig:Sanity_values} shows that the derived eigenvalues from both procedures align with each other, thus verifying the transition from the time to spectral domain. However, they are not necessarily identical to the properties of the original dataset. Notably, as the number of steps increases, both processes converge toward the original dataset values.
Figure \ref{fig:Sanity_relative_Diff} allows for an examination of the absolute error in each process relative to the characteristics of the original dataset. It is evident that while the spectral equation exhibits stable behavior, the time-domain equation displays fluctuations that depend on the number of sampled signals. As the number of samples increases, these fluctuations diminish.
\newpage
\subsection{Method Evaluation: Comparison with Prior Works}
\label{sec:appendix_clarifications_and_validations_AYS_comparison}

We also compare our optimal solution with those from previous works. Specifically, The authors in \cite{sabour2024align} derive a closed-form expression for the optimal noise schedule under a \emph{simplified case}, where the initial distribution is an isotropic Gaussian with a standard deviation of $C$, $i.e  ~ \x_0 \sim \N(\0, C^2\I)$. To enable a proper comparison, we frame our optimization problem using the \emph{Kullback-Leibler divergence} $\DKL$ loss \eqref{eq:Dkl_loss} as done in \cite{sabour2024align}. 

\begin{figure}
  \centering
        \includegraphics[width=0.45\textwidth]{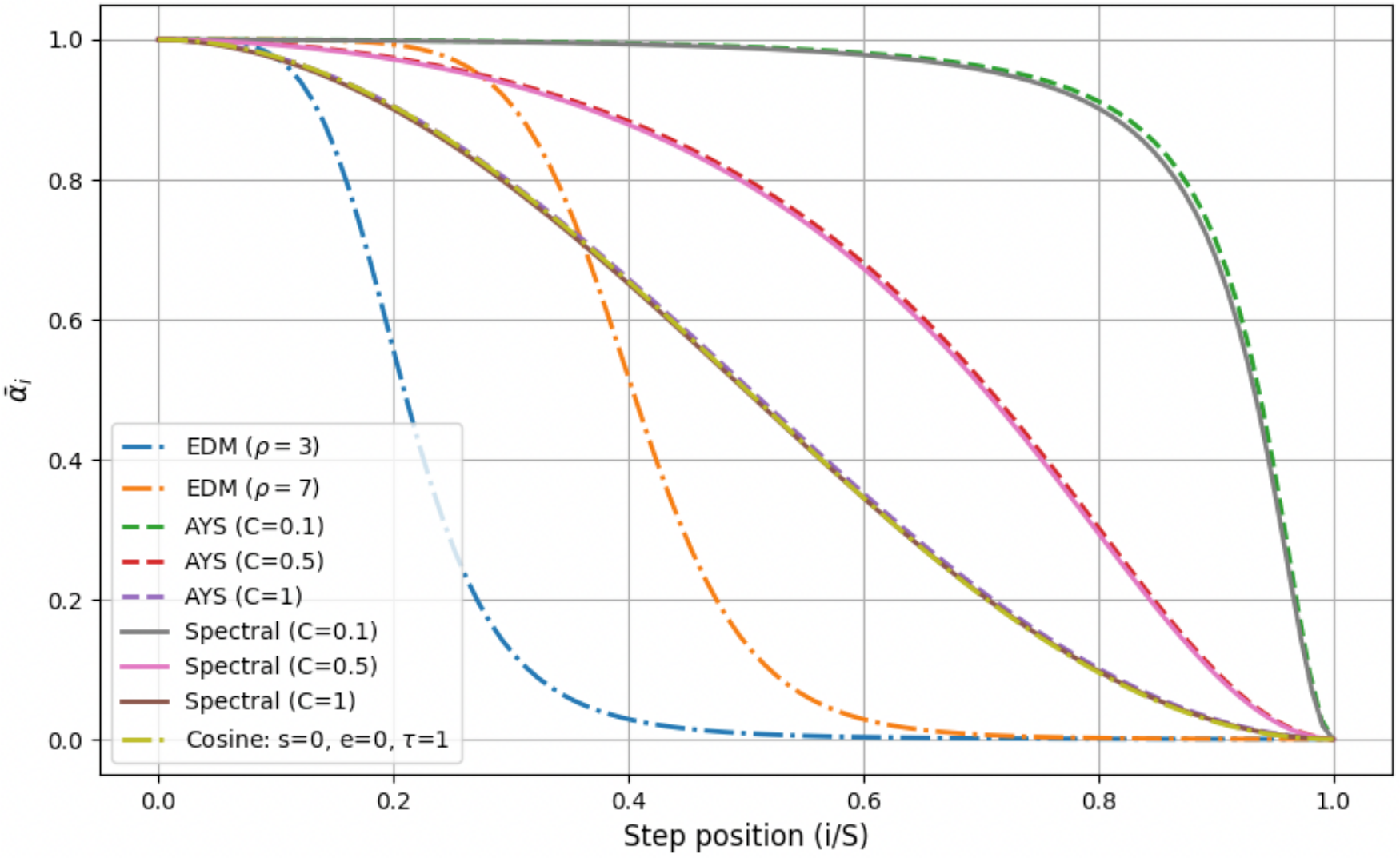}
       \caption{Comparison between the closed-form solution from AYS \cite{sabour2024align} and our numerical solution for the simplified case where $\x_0 \sim \N(\0, C^2\I)$ with $C=[0.1,0.5,1]$.}
        \label{fig:Comparison_to_AYS}
\end{figure}

Figure \ref{fig:Comparison_to_AYS} compares our optimal solution, obtained by numerically solving Equation \eqref{eq:optimization_problem}, with the closed-form solution from \cite{sabour2024align}. \footnote{Since the variance-exploding (VE) formulation of the diffusion process was employed in \cite{sabour2024align}, we used the corresponding relationship $\balpha_t = \frac{1}{1+{\sigma_t}^2}$ to transition the resulting noise schedule to the \emph{variance-preserving (VP)} formulation, as derived in Appendix \ref{sec:appendix_VP_VE}.} It can be observed that both methods align for arbitrary values of $C$. Notably, for $C = 1$, both noise schedules converge exactly to the \emph{Cosine} $(0, 1, 1)$ noise schedule, which was originally \emph{chosen heuristically} \cite{nichol2021improved}.

% \footnote{Since \citet{sabour2024align} employed the variance-exploding (VE) formulation of the diffusion process, we used the corresponding relationship $\balpha_t = \frac{1}{1+{\sigma_t}^2}$ to transition the resulting noise schedule to the \emph{variance-preserving (VP)} formulation, as derived in Appendix \ref{sec:appendix_VP_VE}.}

\subsection{Constraint Omission and Different Types of Initializations}
\begin{figure}[H]
    \centering
    \begin{subfigure}[b]{0.45\textwidth}
        \centering
       \includegraphics[width={\textwidth} ]{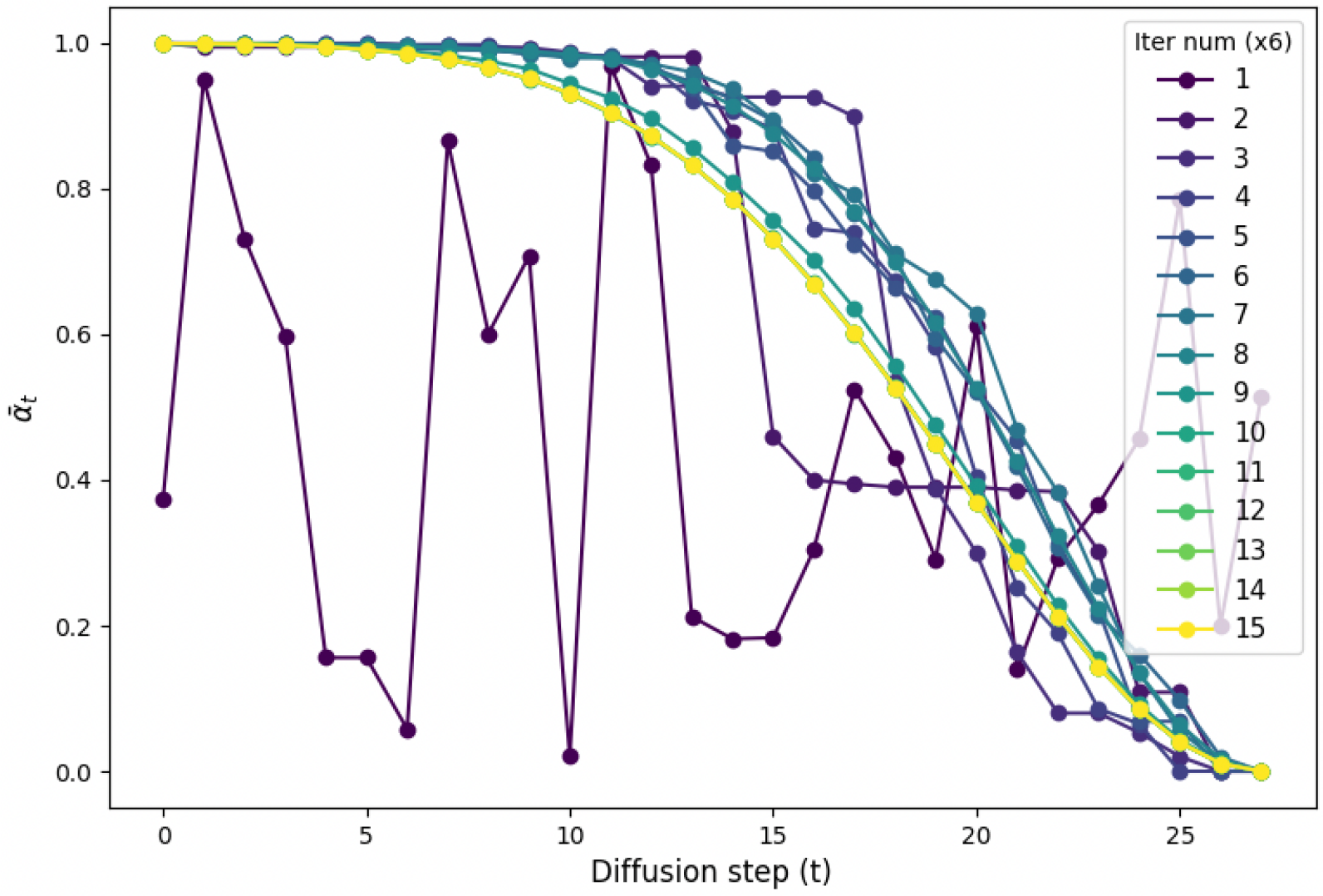}
        \caption{Optimization with inequality constraints.}
        \label{subfig:with_constraint_Uniformly}
    \end{subfigure}
         \hfill % Adjust horizontal space as needed
        \begin{subfigure}[b]{0.45\textwidth}
        \centering
        \includegraphics[width={\textwidth} ]{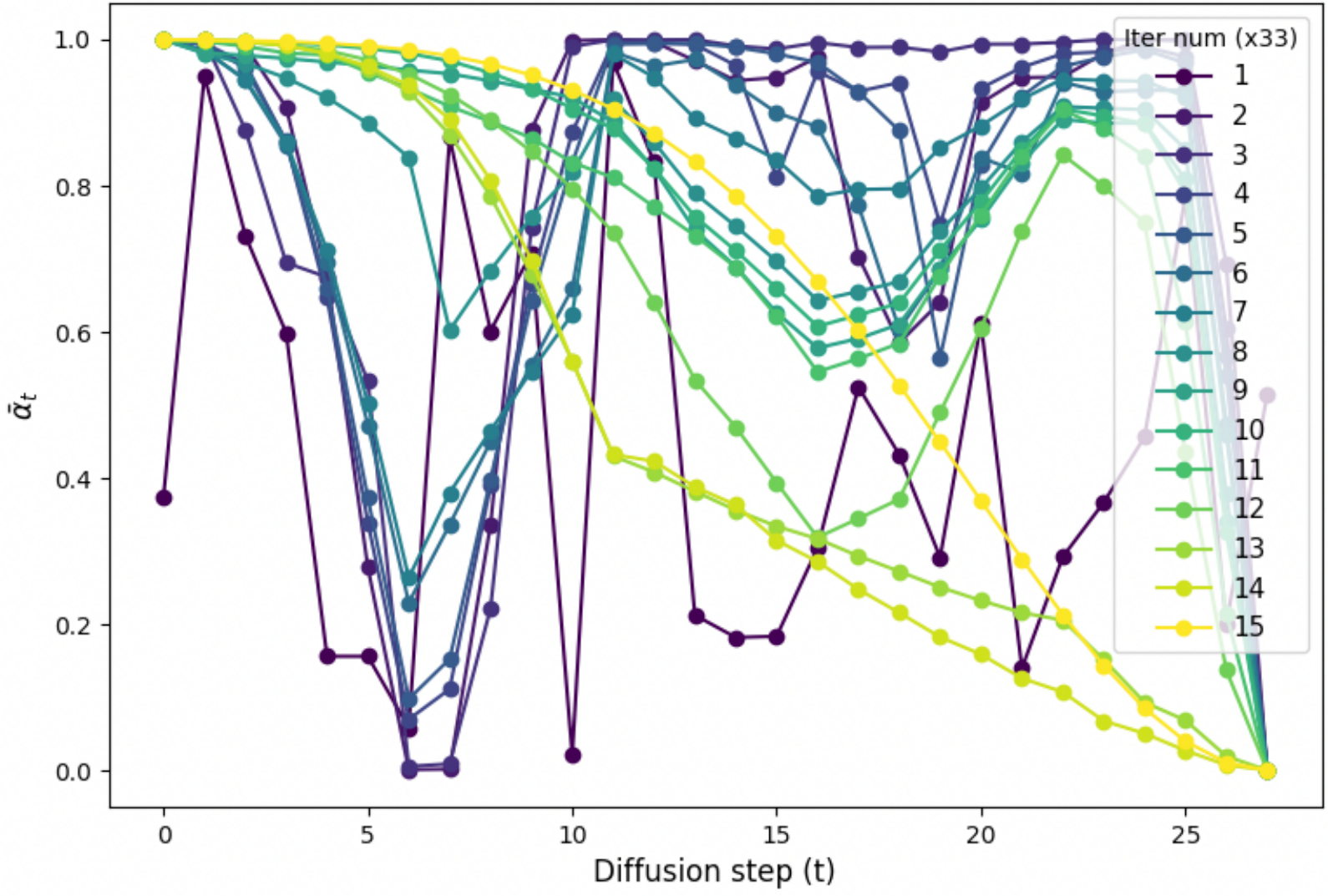}
        \caption{Optimization without inequality constraints.}
        \label{subfig:without_constraint_Uniformly}
    \end{subfigure}
\caption{A Comparison of the noise schedule parameters, $\{ \balpha_s \}_{s=0}^S$,  during the optimization process.
 The optimization was conducted over 28 diffusion steps, with a uniformly random initialization.
 Figure \ref{subfig:with_constraint_Uniformly} shows the results with inequality constraints, and Figure \ref{subfig:without_constraint_Uniformly} presents those without.
 % Figure \ref{subfig:with_constraint_Uniformly} with inequality constraints and Figure \ref{subfig:without_constraint_Uniformly} without inequality constraints.
 }
\label{fig:wasserstein_uniformly_constrains}
\end{figure}

\begin{figure}[H]
    \centering
        \begin{subfigure}[b]{0.45\textwidth}
        \centering
        \includegraphics[width={\textwidth}]
        {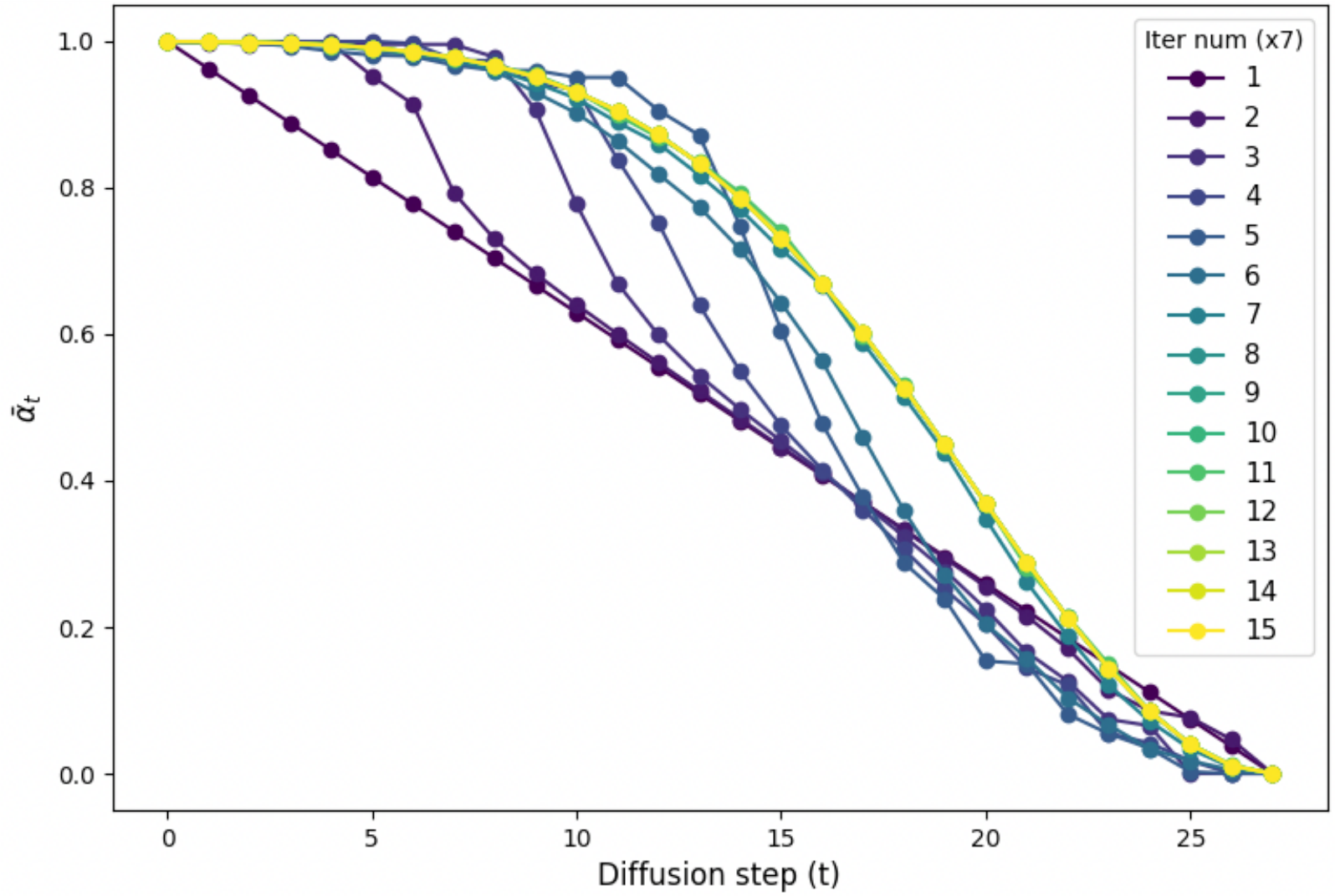}
        \caption{Optimization with inequality constraints.}
        \label{subfig:with_constraint_linear}
    \end{subfigure}
    % \hspace{1mm}
             \hfill % Adjust horizontal space as needed
        \begin{subfigure}[b]{0.45\textwidth}
        \centering
        \includegraphics[width={\textwidth}]
        {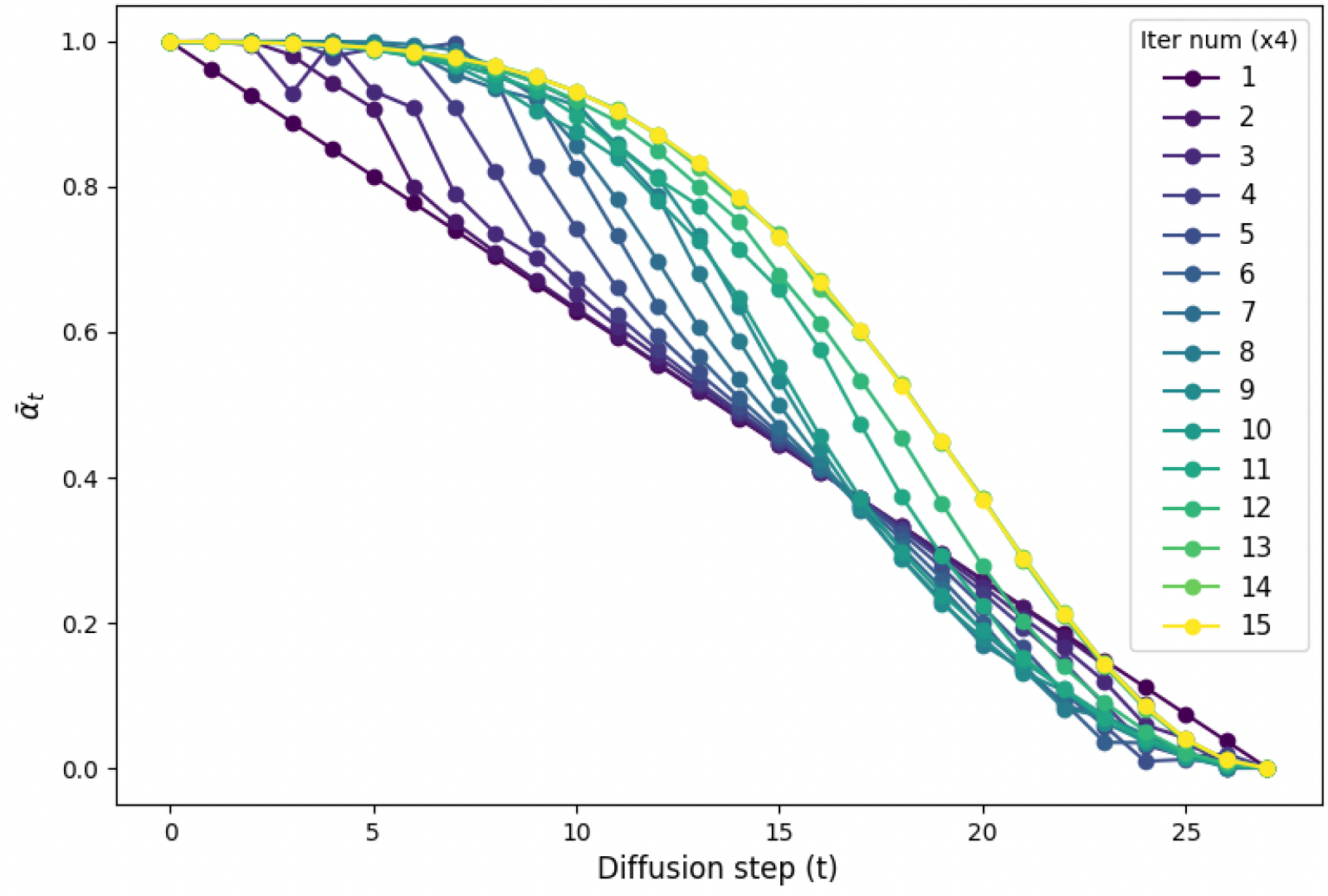}
        \caption{Optimization without inequality constraints.}
        \label{subfig:without_constraint_linear}
    \end{subfigure}
\caption{A Comparison of the noise schedule parameters, $\{ \balpha_s \}_{s=0}^S$,  during the optimization process.
 The optimization was conducted over 28 diffusion steps, with a Linearly decreasing  initialization.
 Figure \ref{subfig:with_constraint_linear} shows the results with inequality constraints, and Figure \ref{subfig:without_constraint_linear} presents those without.}
\label{fig:wasserstein_linear_constrains}
\end{figure}

% \subsection{Analysis of Constraints Omission}
\label{sec:appendix_clarifications_and_validations_constrains_omission}
We explore the influence of the initializations and the inequality constraints in the optimization problem.
Figures \ref{fig:wasserstein_uniformly_constrains} and \ref{fig:wasserstein_linear_constrains} illustrate the evolution of the noise schedule parameter, $\{ \balpha_s \}_{s=0}^S$, during the optimization process for two different initializations: \emph{uniformly random} and \emph{linearly decreasing} schedules, respectively. In both cases, the diffusion process consists of $28$ steps, and the \emph{Wasserstein-2} distance is used. Each scenario was conducted twice: once with the inequality constraints from \ref{subsec:establish_optimal_spectral_noise_schedule} and once without. The results are plotted at $15$ evenly spaced intervals throughout the process to avoid presenting each individual optimization step.


The results reveal that the optimized schedule is consistent across both initializations and independent of the inequality constraints. This suggests that known characteristics of noise schedules, such as monotonicity, naturally emerge from the  problem's formulation itself, even without an explicit demand for inequality constraints. Similar consistency is observed for other initializations, including linear and cosine schedules, demonstrating the stability of the optimization procedure.

\newpage
\section{Supplementary Experiments for Synthetic  Gaussian Distribution: }
\label{sec:appendix_Supplementary_Experiments_Scenario_1}

In the following sections, we demonstrate the received spectral recommendations for various alternative selections, applied to the matrix $\bSigma_0$ and the vector $\bmu$, differing from those presented in Section \ref{subsec:Scenario_1}. Additionally, we present the solutions obtained for defining the \emph{Wasserstein-2} distance and the \emph{KL-divergence}. Through this, we aim to provide a broader perspective on the behavior and applicability of the proposed approach.

\begin{figure}[H]
    \centering
\vskip 0.2in
    \begin{subfigure}{0.35\textwidth}
        \centering
     \includegraphics[width=\columnwidth]{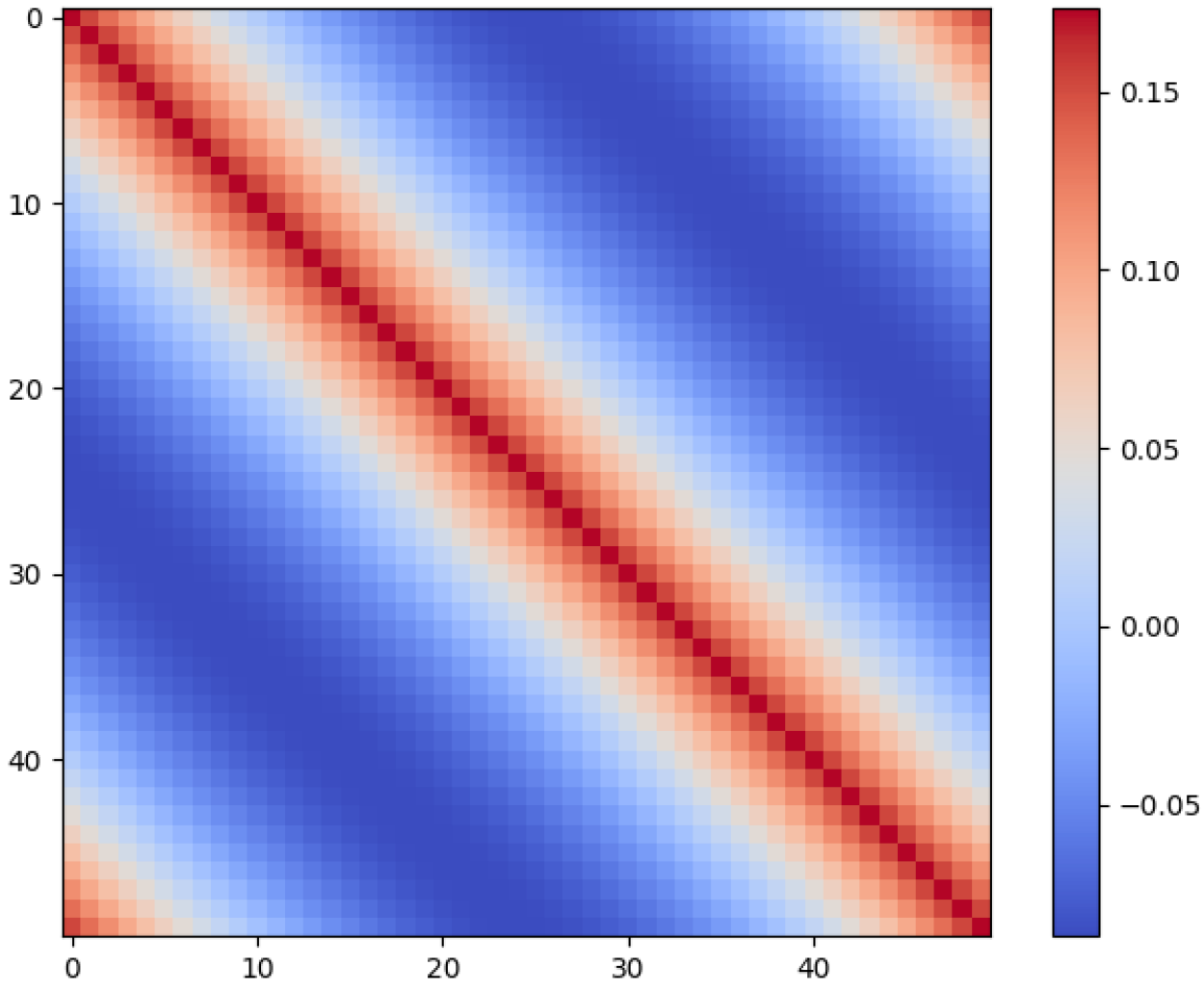}  
        \caption{$\bSigmaZ$}
        \label{fig:Exp_1_cov_mat_l_0_1_appendix}
        \end{subfigure}
    \begin{subfigure}{0.35\textwidth} % Define the width of the 
            \centering
\includegraphics[width=\columnwidth]{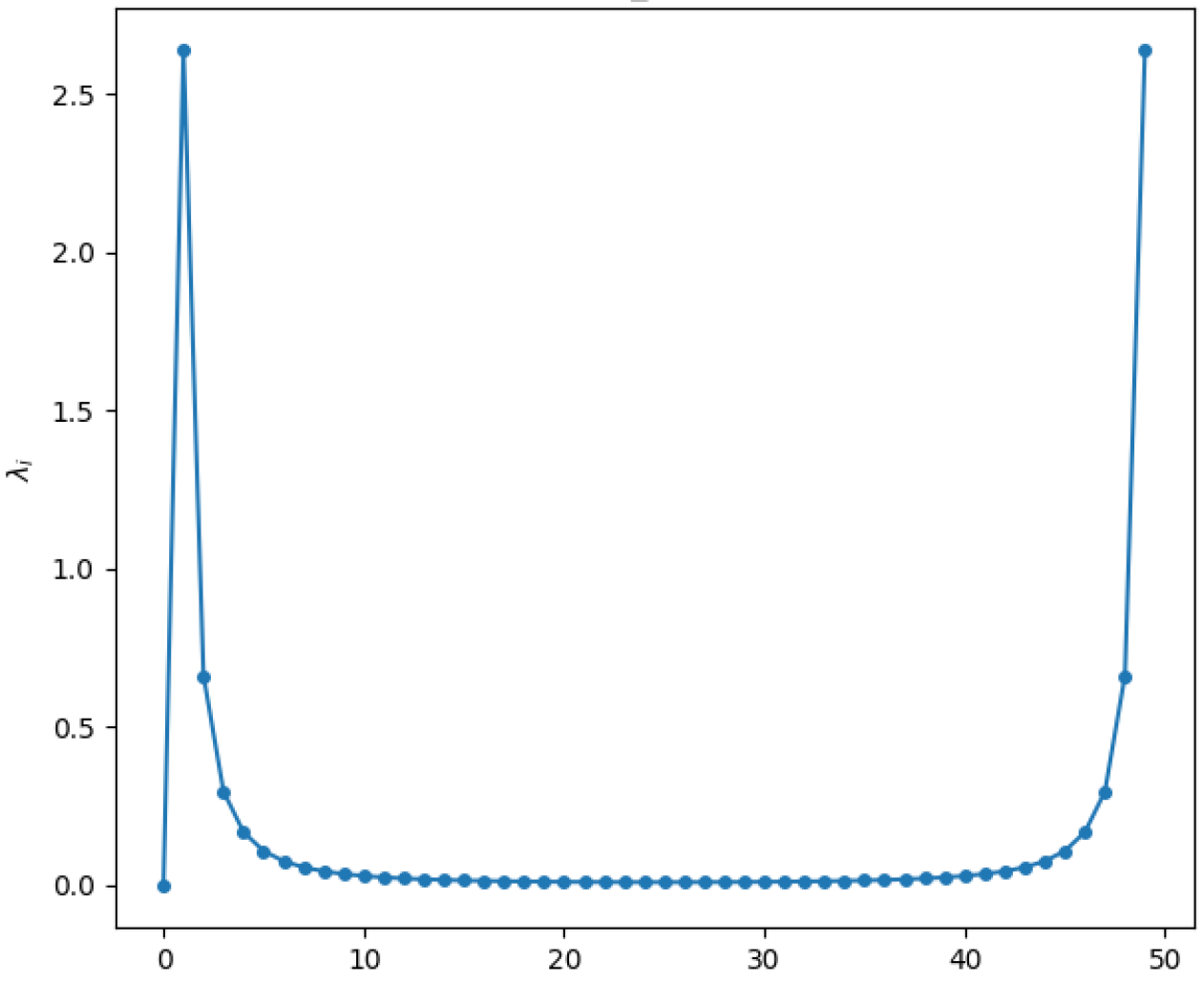}
 \caption{$tr(\mathbf{\Lambda}_0)$}
 \label{fig:Exp_1_lambda_mat_l_0_1_appendix}
    \end{subfigure}
% \vskip -0.2in
\caption{Visualization of $\bSigma_0$ and the trace of $\mathbf{\Lambda}_0$ for $d=50$ and $l=0.1$. The covariance matrix $\bSigma_0$ is circulant (\ref{fig:Exp_1_cov_mat_l_0_1_appendix}), while $\mathbf{\Lambda}_0$ is diagonal with symmetric diagonal elements (\ref{fig:Exp_1_lambda_mat_l_0_1_appendix}).}
\label{fig:exp_1_Sigma_0_and__Lambda_0}
\end{figure}

Figure \ref{fig:exp_1_Sigma_0_and__Lambda_0} visualizes the covariance matrices $\bSigma_0$ and $\bLambda$ as discussed in Sec. \ref{subsec:Scenario_1}.

\subsection{Wasserstein-2 distance}

\begin{figure}[H]
 \centering
     \begin{center}

    \centering
    \begin{subfigure}{0.45\textwidth}
        \centering
        \includegraphics[width=\columnwidth]{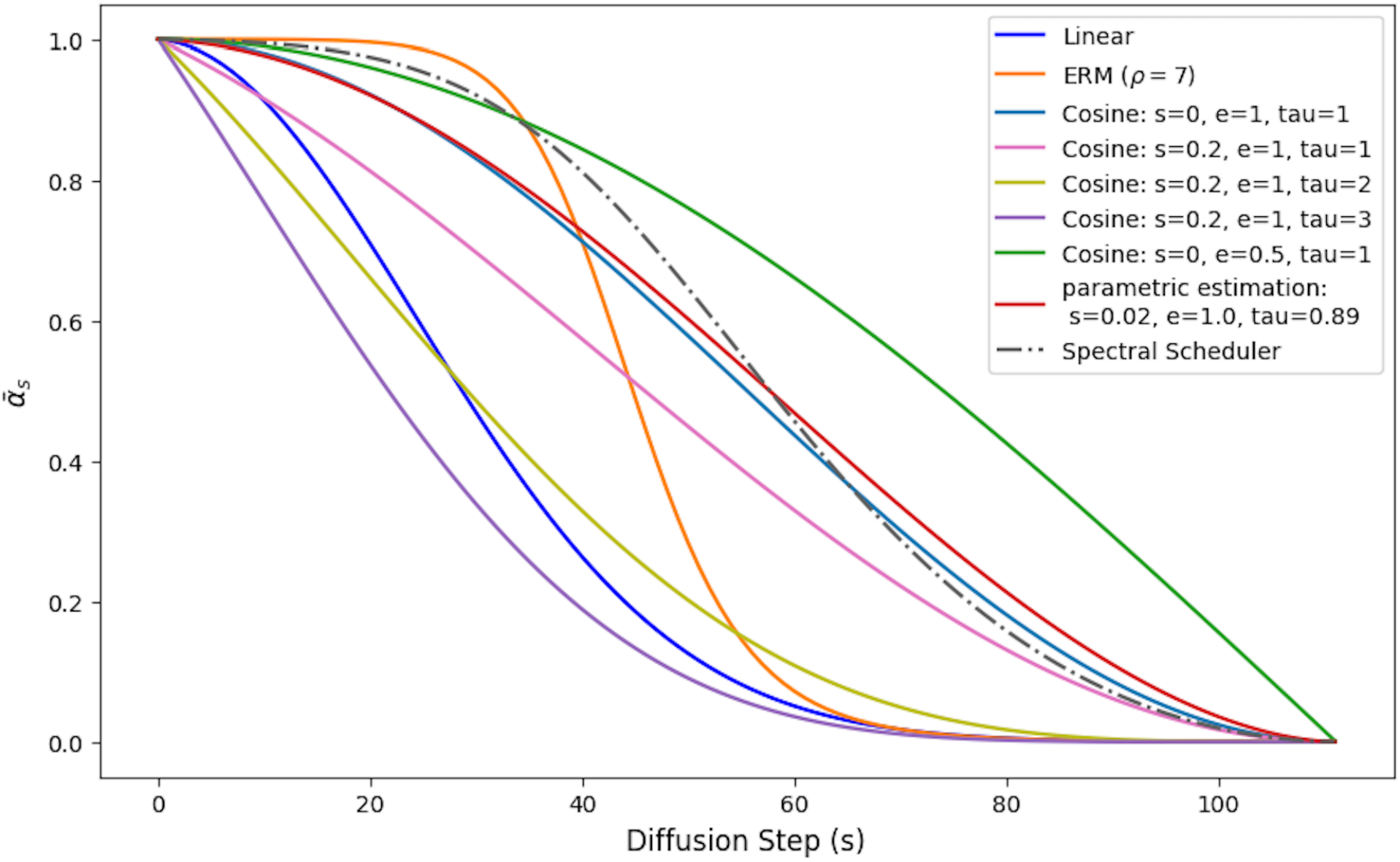}
        \caption{}
        \label{subfig:Exp_1_Cosine_Comparison_wasserstein_appendix}
    \end{subfigure}
    \hspace{2mm}
    \begin{subfigure}{0.45\textwidth}
        \centering
        \includegraphics[width=\columnwidth]{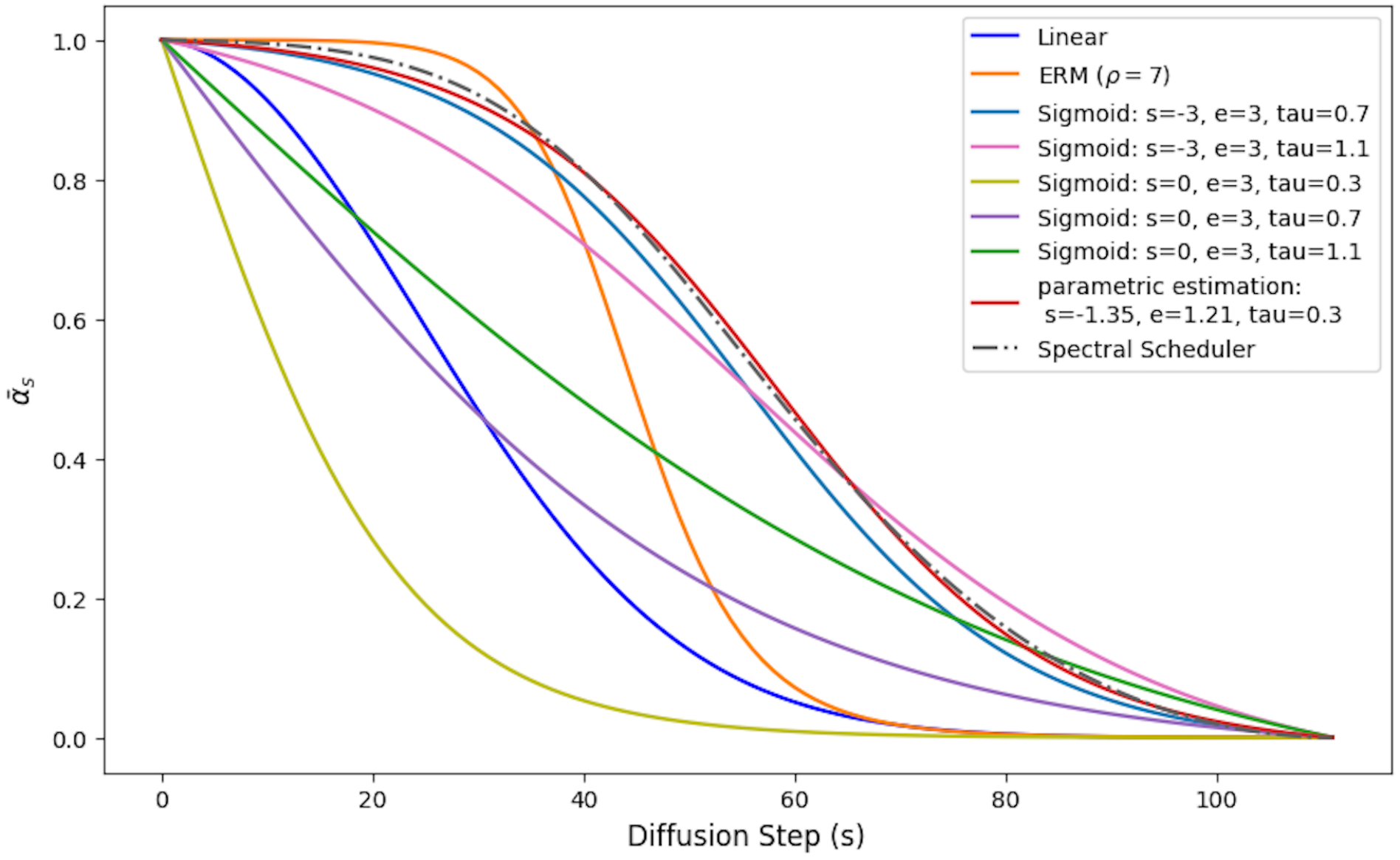}
        \caption{}
        \label{subfig:Exp_1_sigmoid_Comparison_wasserstein_appendix}
    \end{subfigure}
     \end{center}

% \caption{Figures \ref{subfig:Exp_1_Cosine_Comparison_wasserstein_appendix} and \ref{subfig:Exp_1_sigmoid_Comparison_wasserstein_appendix} compare the optimized noise schedules from Sec. \ref{subsec:Scenario_1} using $d=50$, $l=0.1$, $\bmu=0.05$ and the \emph{Wasserstein-2} distance as the loss function, with heuristic schedules for 112 diffusion steps. 
\caption{Figures \ref{subfig:Exp_1_Cosine_Comparison_wasserstein_appendix} and 
\ref{subfig:Exp_1_sigmoid_Comparison_wasserstein_appendix} compare the optimized noise schedules from Sec. \ref{subsec:Scenario_1}, using $d=50$, $l=0.1$, and $\bmu=0.05$, with heuristic schedules for 112 diffusion steps, where the optimization is based on the \emph{Wasserstein-2} distance. \autoref{subfig:Exp_1_Cosine_Comparison_wasserstein_appendix} examines the spectral schedule alongside the Linear \citep{ho2020denoising}, EDM \citep{karras2022elucidating} and Cosine-based schedules, including \emph{Cosine} ($s=0$, $e=1$, $\tau=1$) from \citep{nichol2021improved, chen2023importance}. Likewise, \autoref{subfig:Exp_1_sigmoid_Comparison_wasserstein_appendix} compares it to Sigmoid-based schedules \citep{jabri2022scalable}. The parametric estimations for the Cosine and Sigmoid functions are highlighted in red.}
    \label{fig:Scenario_1_wasserstein_distance_appendix}
\end{figure}

Figure \ref{fig:Scenario_1_wasserstein_distance_appendix} presents the resulting noise schedule based on the \emph{Wasserstein-2} distance.

\newpage
\subsection{KL-Divergence}

\begin{figure}[h]
 \centering

    \begin{subfigure}{0.45\textwidth}
  \centering
  \includegraphics[width=\textwidth]{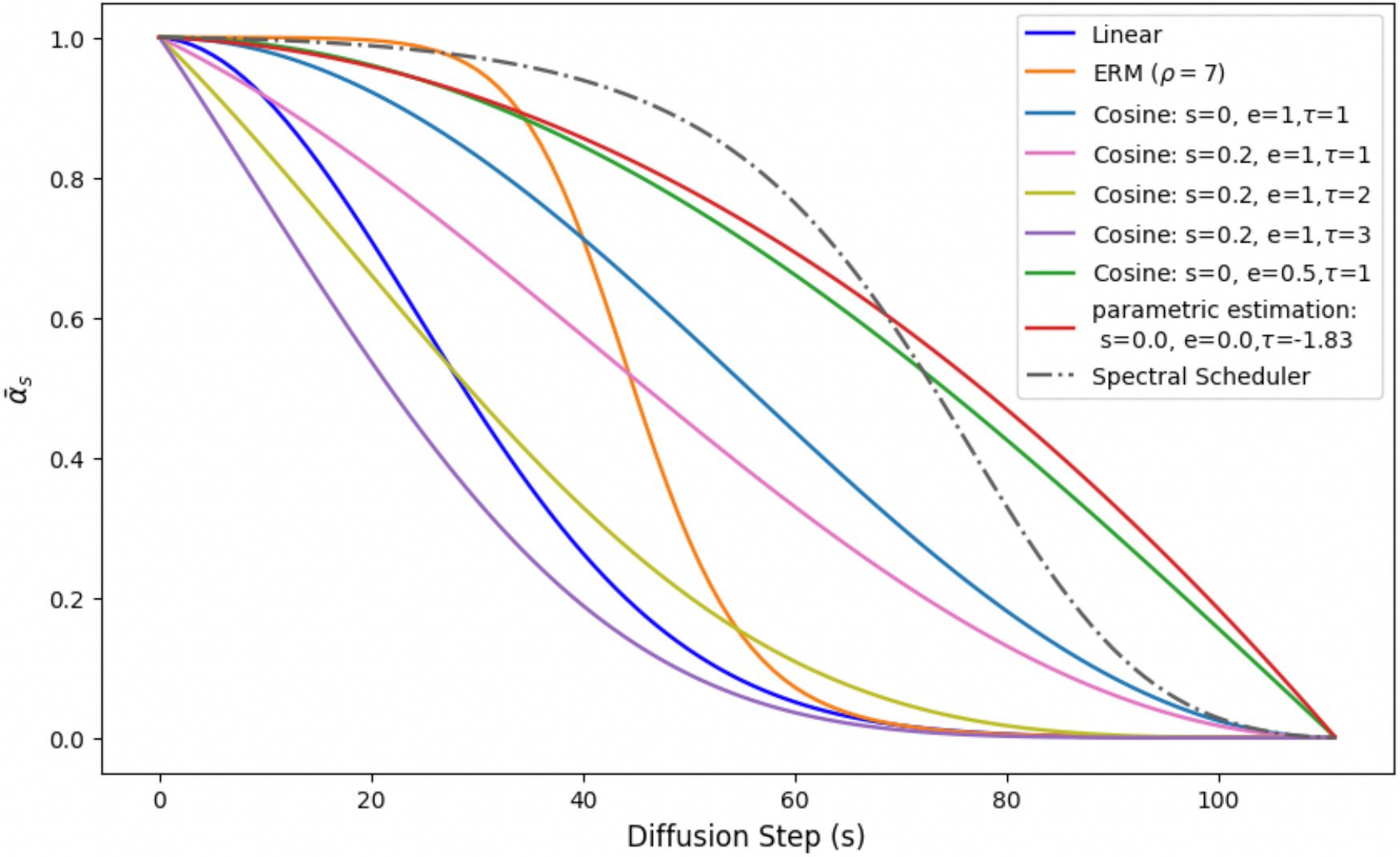}
  \caption{}
  \label{subfig:Exp_1_Cosine_Comparison_wasserstein_appendix_dkl}
  \end{subfigure}
 \centering
    \hspace{2mm}
\begin{subfigure}{0.45\textwidth}
  \centering
  \includegraphics[width=\textwidth]{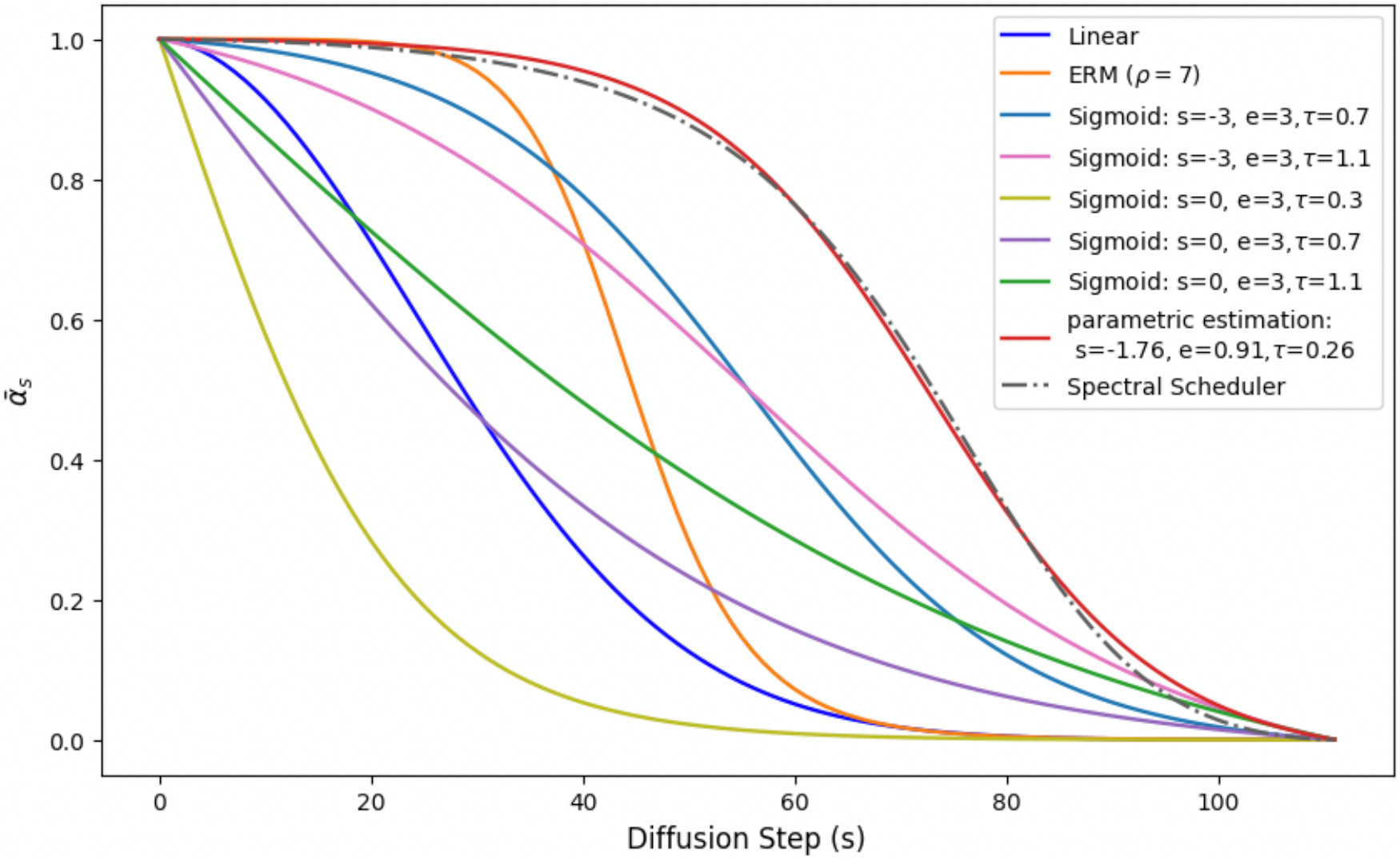}
 \caption{}
  \label{subfig:Exp_1_Sigmoid_Comparison_wasserstein_appendix_dkl}
\end{subfigure}

\caption{Figures \ref{subfig:Exp_1_Cosine_Comparison_wasserstein_appendix_dkl} and \ref{subfig:Exp_1_Sigmoid_Comparison_wasserstein_appendix_dkl} compare the optimized noise schedules from Sec. \ref{subsec:Scenario_1}, using $d=50$, $l=0.1$, and $\bmu=0.05$, with heuristic schedules for 112 diffusion steps, where the optimization is based on the  \emph{KL divergence}. \autoref{subfig:Exp_1_Cosine_Comparison_wasserstein_appendix_dkl} examines the spectral schedule alongside the Linear \citep{ho2020denoising}, EDM \citep{karras2022elucidating} and Cosine-based schedules, including \emph{Cosine} ($s=0$, $e=1$, $\tau=1$) from \citep{nichol2021improved, chen2023importance}. Likewise, \autoref{subfig:Exp_1_sigmoid_Comparison_wasserstein_appendix} compares it to Sigmoid-based schedules \citep{jabri2022scalable}. The parametric estimations for the Cosine and Sigmoid functions are highlighted in red.}

% \caption{\ref{subfig:Exp_1_Cosine_Comparison_wasserstein_appendix_dkl} and \ref{subfig:Exp_1_Sigmoid_Comparison_wasserstein_appendix_dkl} compare the optimized noise schedules from Sec. \ref{subsec:Scenario_1} ($d=50$, $l=0.1$, $\bmu=0.05$) using the \emph{KL-Divergence} as the loss function, with heuristic schedules for 112 diffusion steps. \autoref{subfig:Exp_1_Cosine_Comparison_wasserstein_appendix_dkl} examines the spectral schedule alongside Linear \citep{ho2020denoising} and Cosine-based schedules, including \emph{Cosine} ($s=0$, $e=1$, $\tau=1$) from \citep{nichol2021improved, chen2023importance}. Likewise, \autoref{subfig:Exp_1_sigmoid_Comparison_wasserstein_appendix} compares it to Sigmoid-based schedules \citep{jabri2022scalable}. The parametric estimations for the Cosine and Sigmoid functions are highlighted in red.}
  \label{fig:Scenario_1_dkl_distance_appendix}
\end{figure}

Figure \ref{fig:Scenario_1_dkl_distance_appendix} presents the resulting noise schedule based on the \emph{KL-Divergence}.

Notably, under the same conditions, the \emph{KL divergence} results in a more \emph{concave} spectral recommendation compared to the \emph{Wasserstein-2} distance.

\newpage
\subsection{Variations in Covariance Matrices and Mean Configurations}

\begin{figure}[H]
    \centering
        \begin{subfigure}[b]{0.3\textwidth}
        \centering
        \includegraphics[width={\textwidth}]{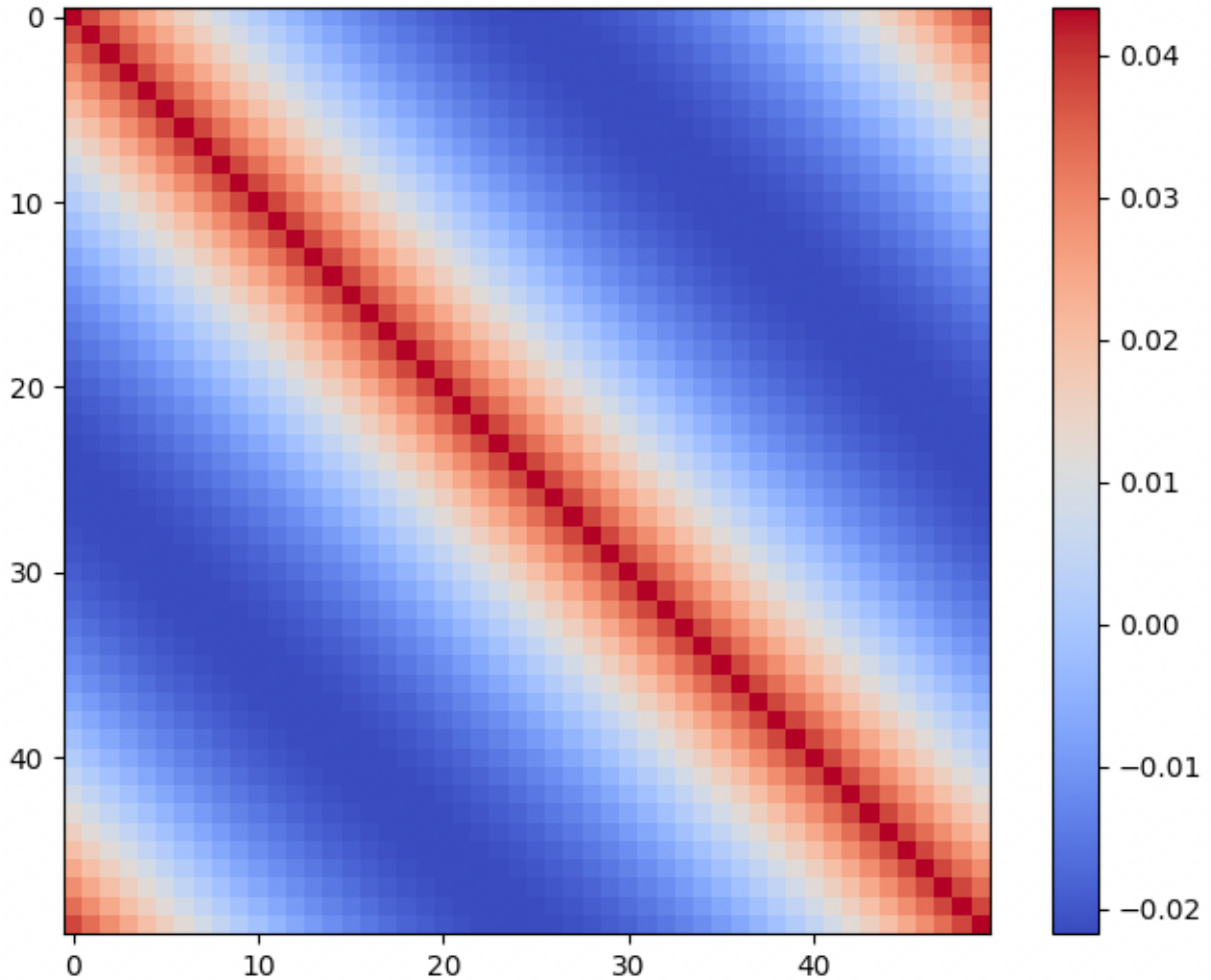}
         \caption{}
        % \caption{$\bSigma$}
        \label{subfig:cov_mat_1_none}
    \end{subfigure}
     \hfill
        \begin{subfigure}[b]{0.3\textwidth}
        \centering
        \includegraphics[width={\textwidth}]{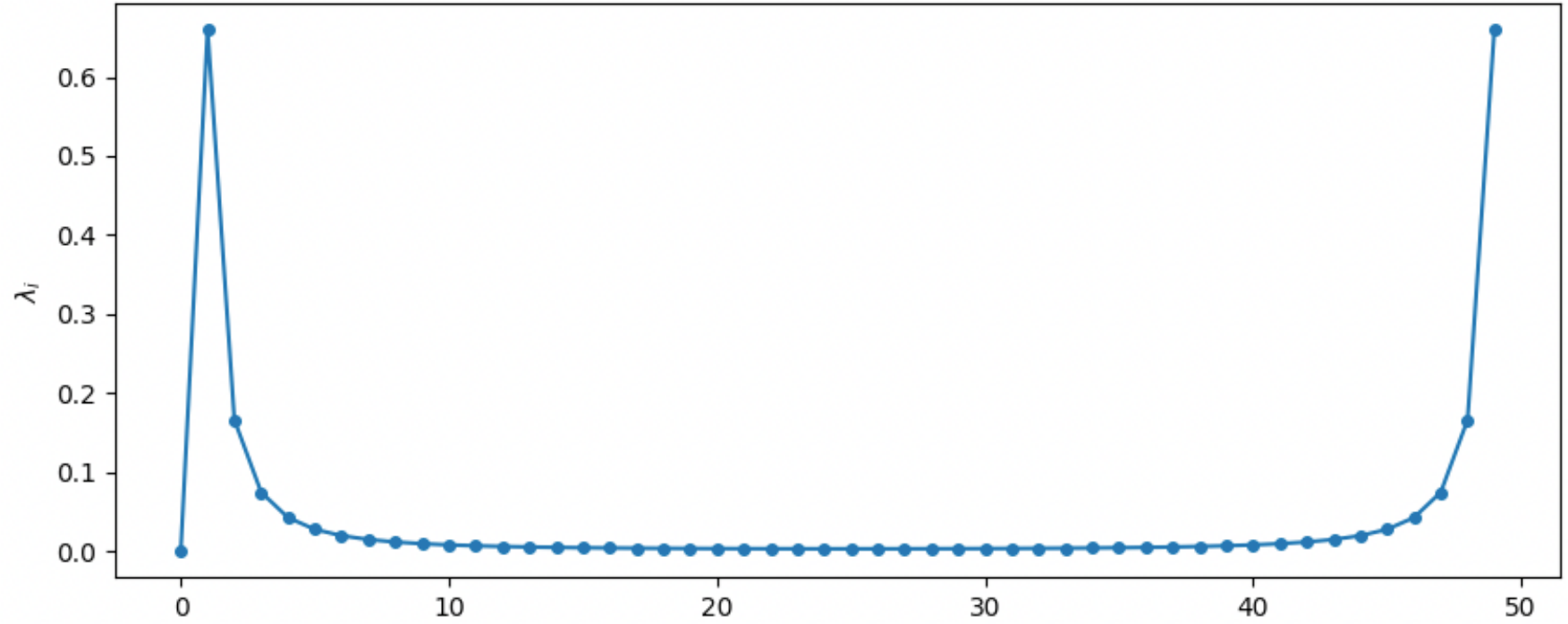}
         \caption{}
        % \caption{$\bSigma$}
        \label{subfig:lambda_0_mat_1_non}
    \end{subfigure}
         \hfill
    \begin{subfigure}[b]{0.3\textwidth}
        \centering
        \includegraphics[width={\textwidth}]{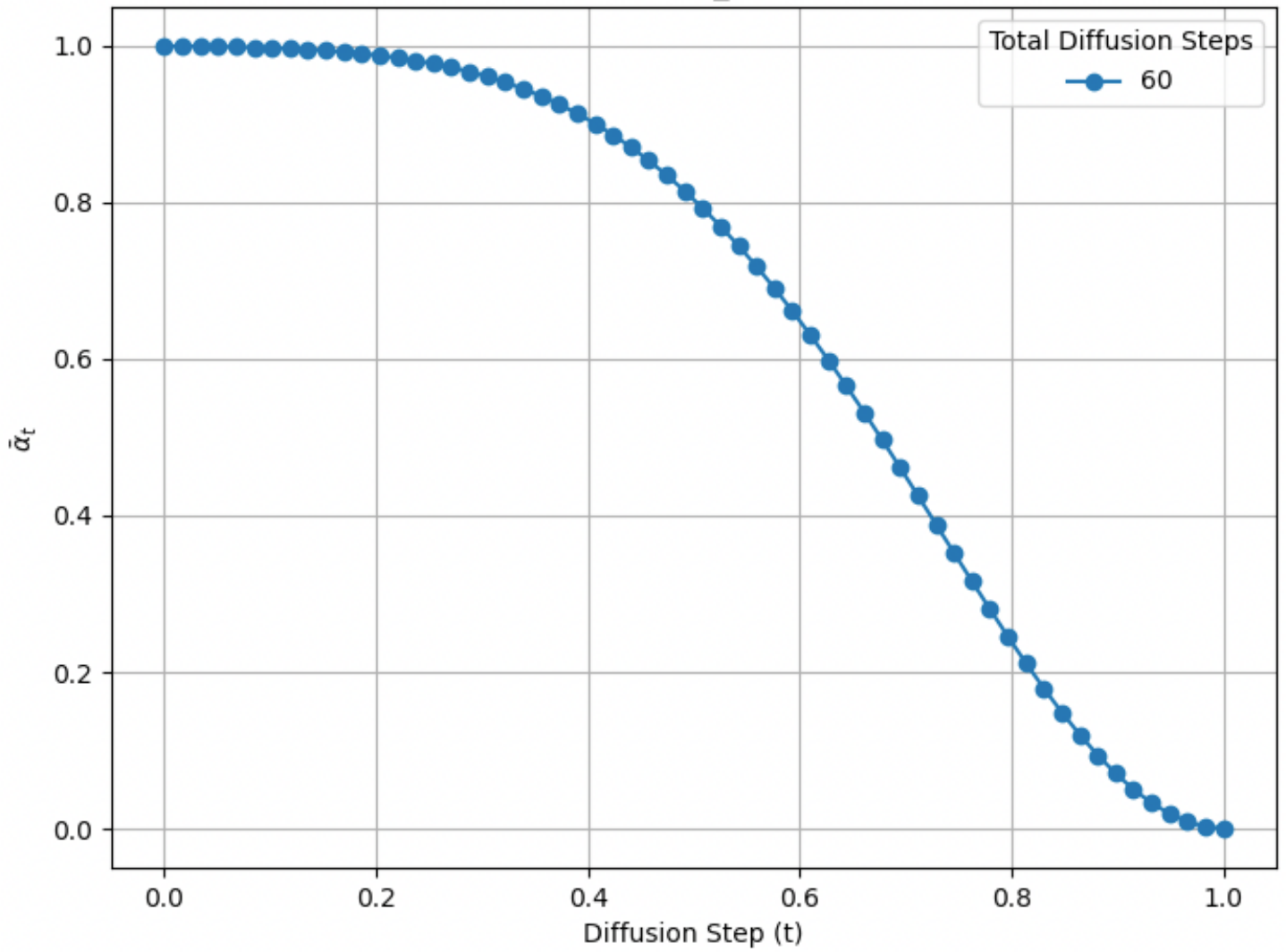}
         \caption{}
        % \caption{$\bSigma$}
        \label{subfig:alpha_bar_mat_1_None}
    \end{subfigure}
\hfill
    \hspace{1mm}
    \begin{subfigure}[b]{0.3\textwidth}
        \centering
        \includegraphics[width=\textwidth]{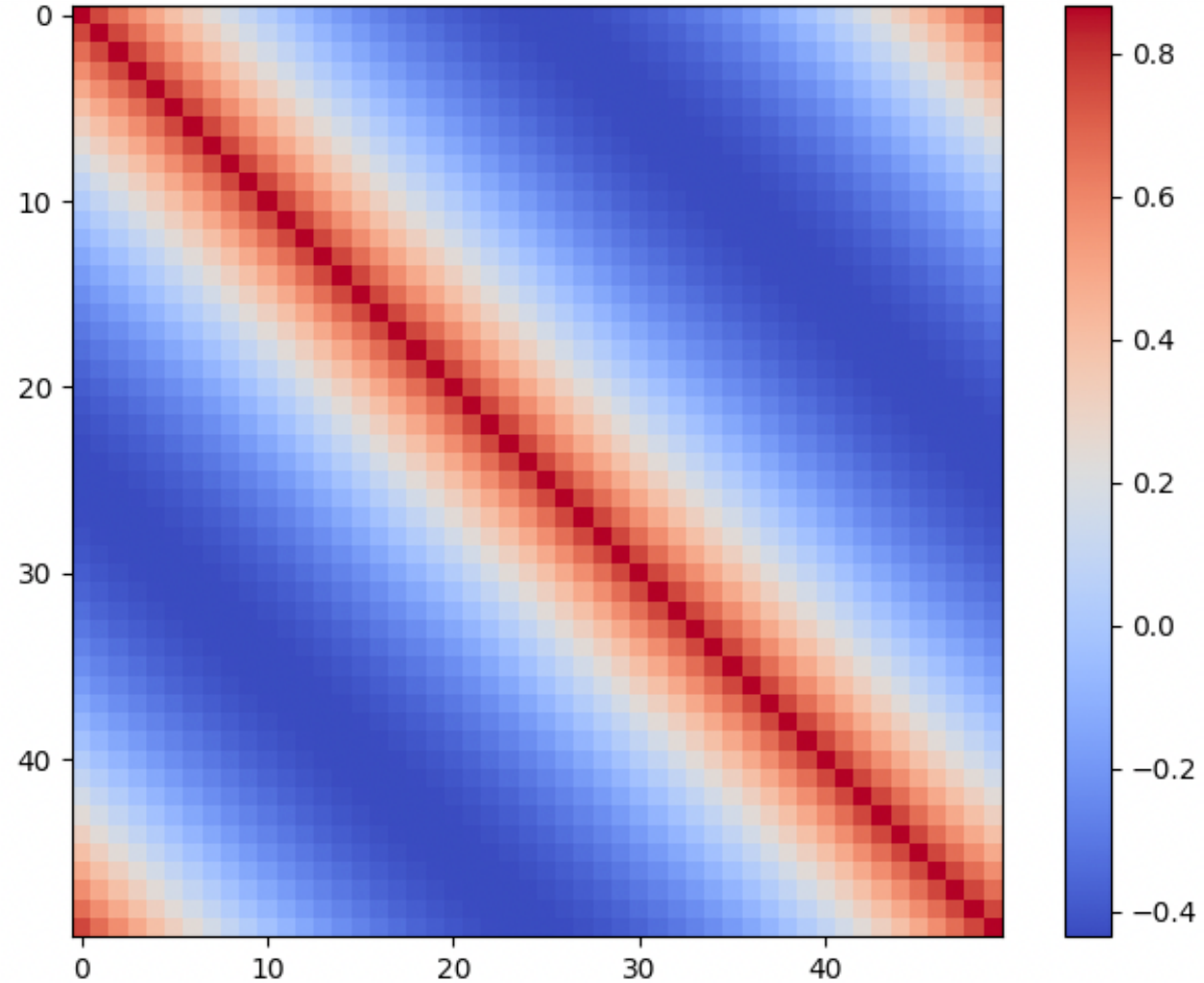}
         \caption{}
        % \caption{$\bSigma$}
        \label{subfig:cov_mat_1_mul_20}
    \end{subfigure}
         \hfill
    \begin{subfigure}[b]{0.3\textwidth}
        \centering
        \includegraphics[width=\textwidth]{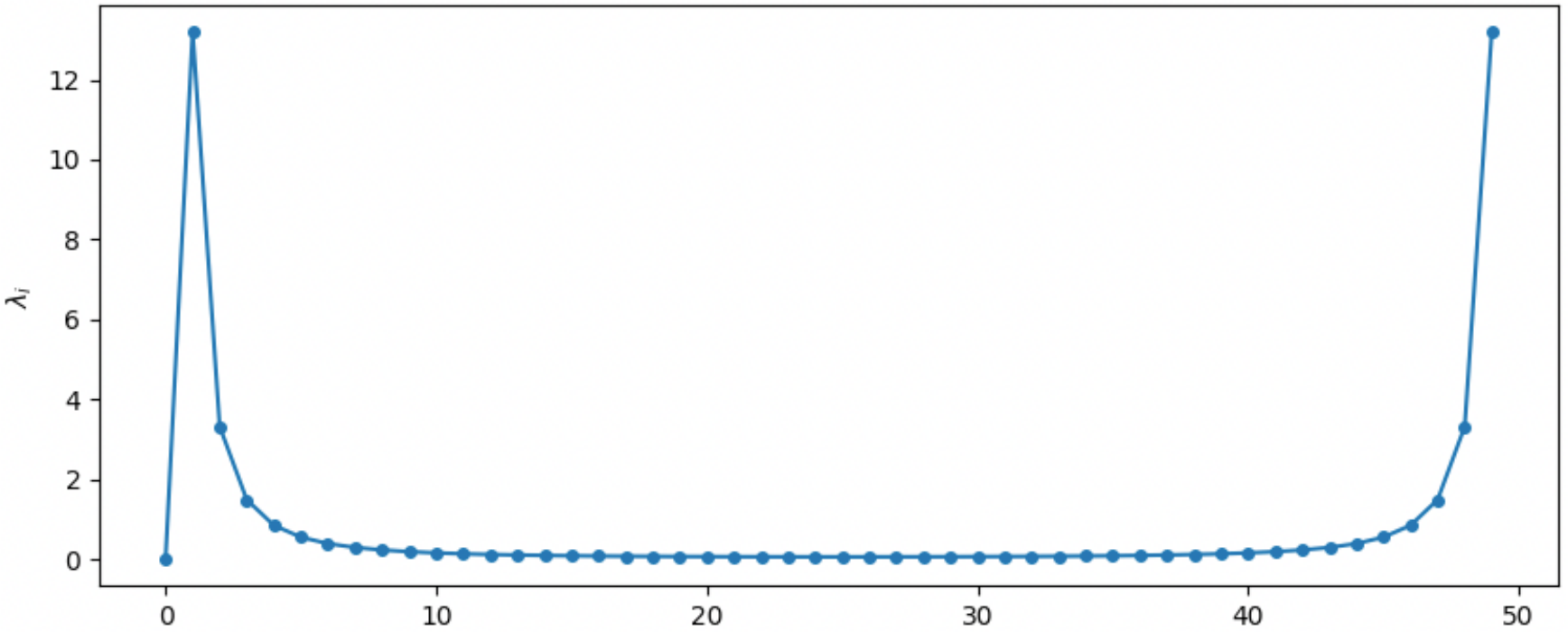}
         \caption{}
        % \caption{$\bSigma$}
        \label{subfig:lambda_0_cov_mat_1_mul_20}
    \end{subfigure}
      \hfill
        \begin{subfigure}[b]{0.3\textwidth}
        \centering
        \includegraphics[width=\textwidth]{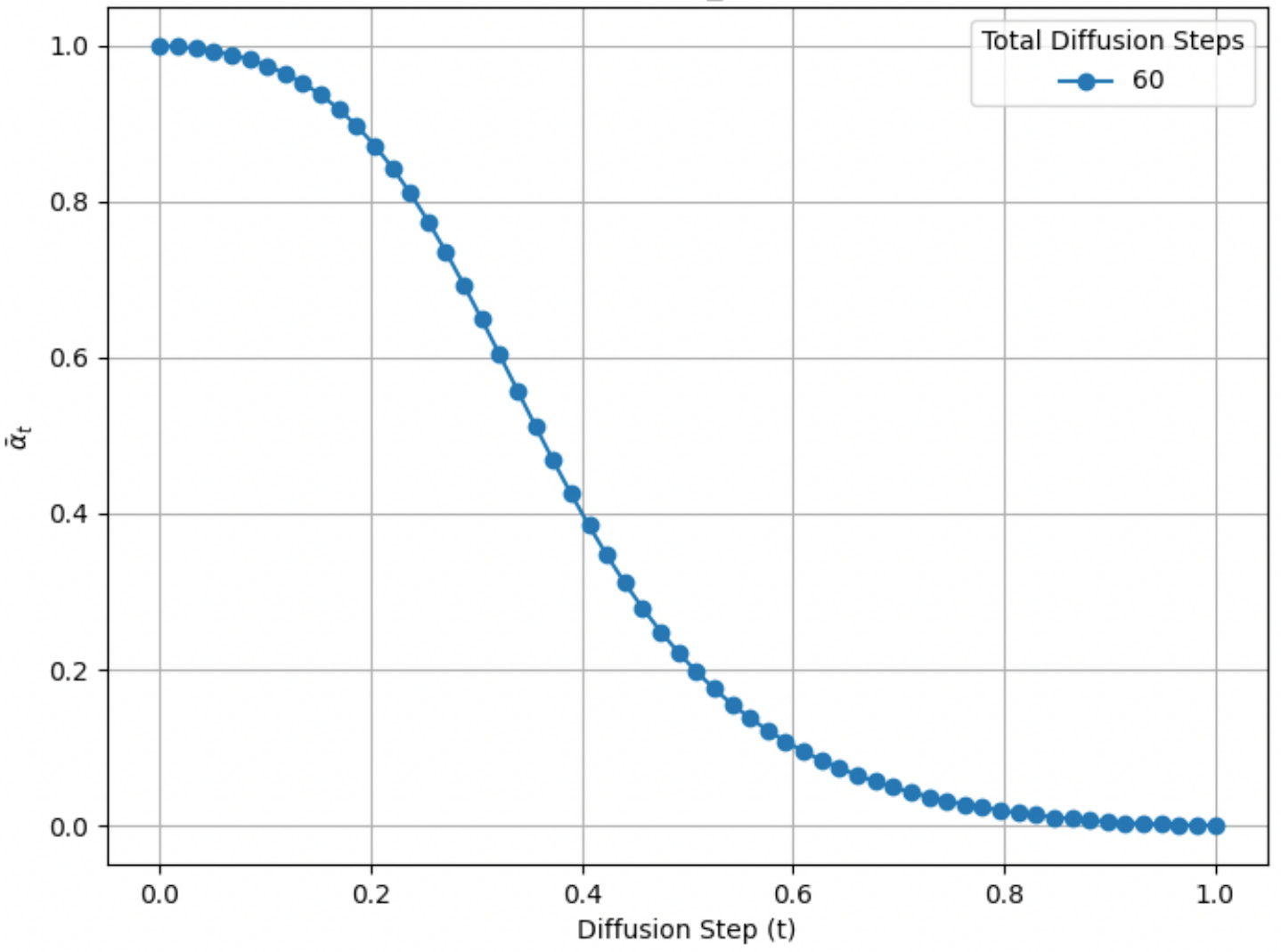}
         \caption{}
        % \caption{$\bSigma$}
        \label{subfig:alpha_bar_cov_mat_1_mul_20}
    \end{subfigure}
        \hspace{1mm}
    \begin{subfigure}[b]{0.3\textwidth}
        \centering
        \includegraphics[width=\textwidth]{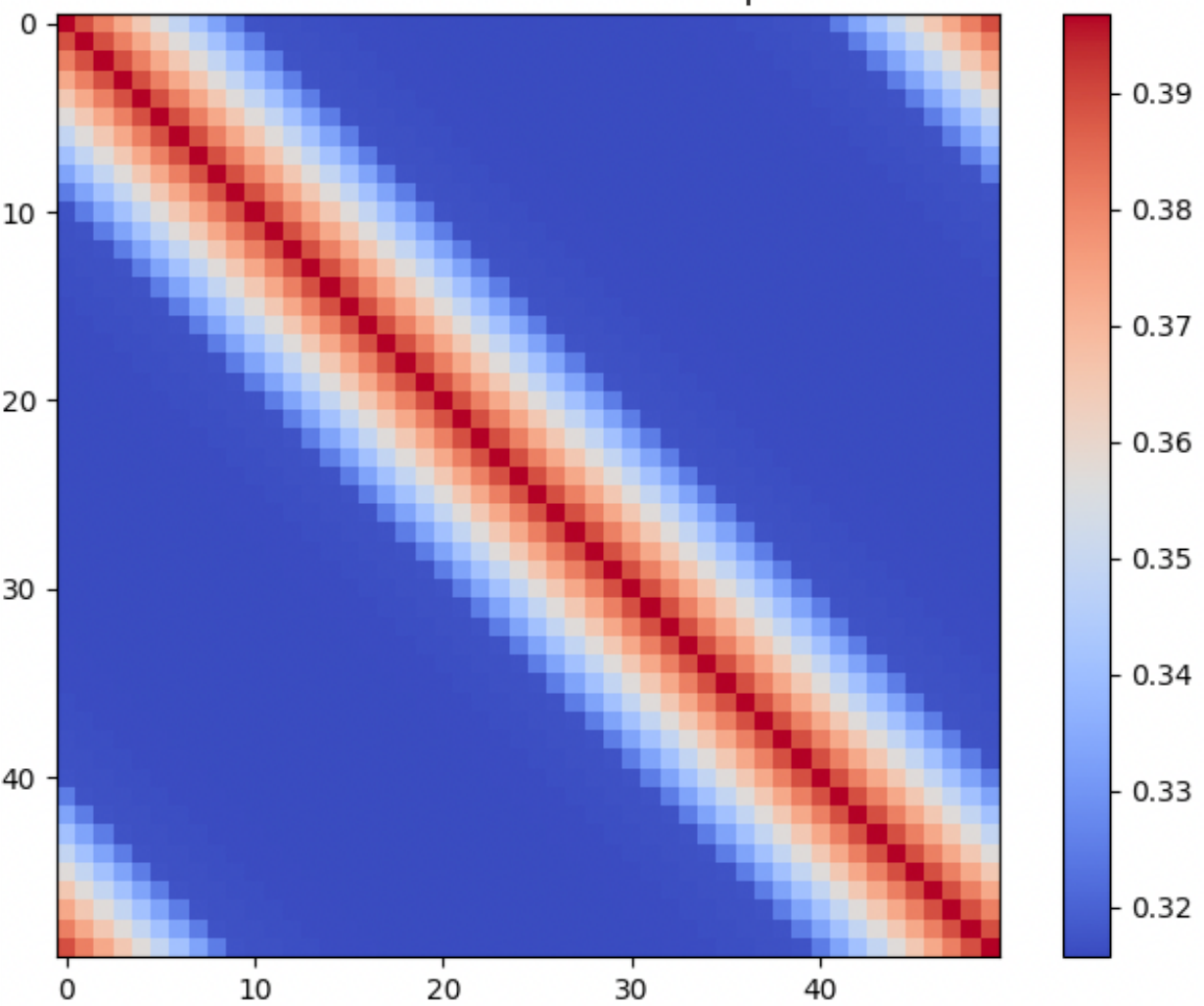}
         \caption{}
        % \caption{$\bSigma$}
        \label{subfig:cov_mat_5_mul_none}
    \end{subfigure}
        \hfill
        \begin{subfigure}[b]{0.3\textwidth}
        \centering
        \includegraphics[width=\textwidth]{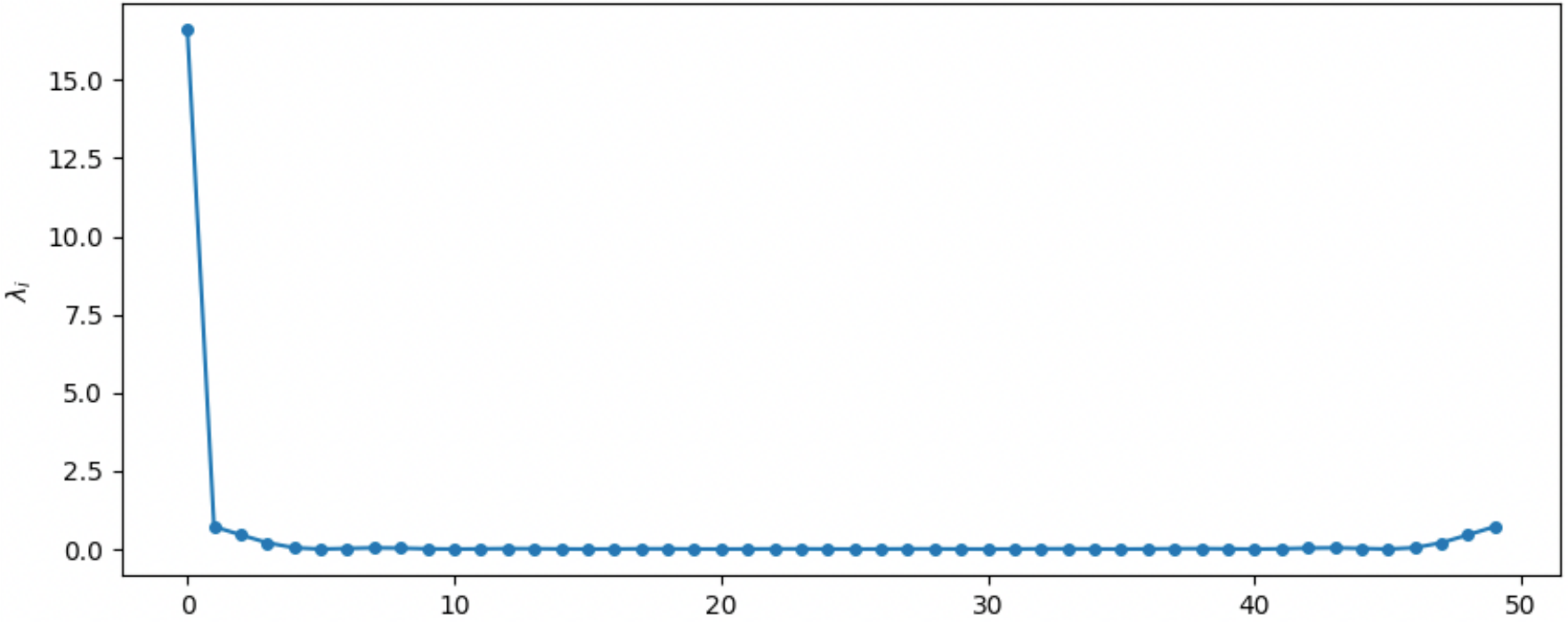}
         \caption{}
        % \caption{$\bSigma$}
        % \label{subfig:cov_mat_1_mul_20}
    \end{subfigure}
      \hfill
        \begin{subfigure}[b]{0.3\textwidth}
        \centering
        \includegraphics[width=\textwidth]{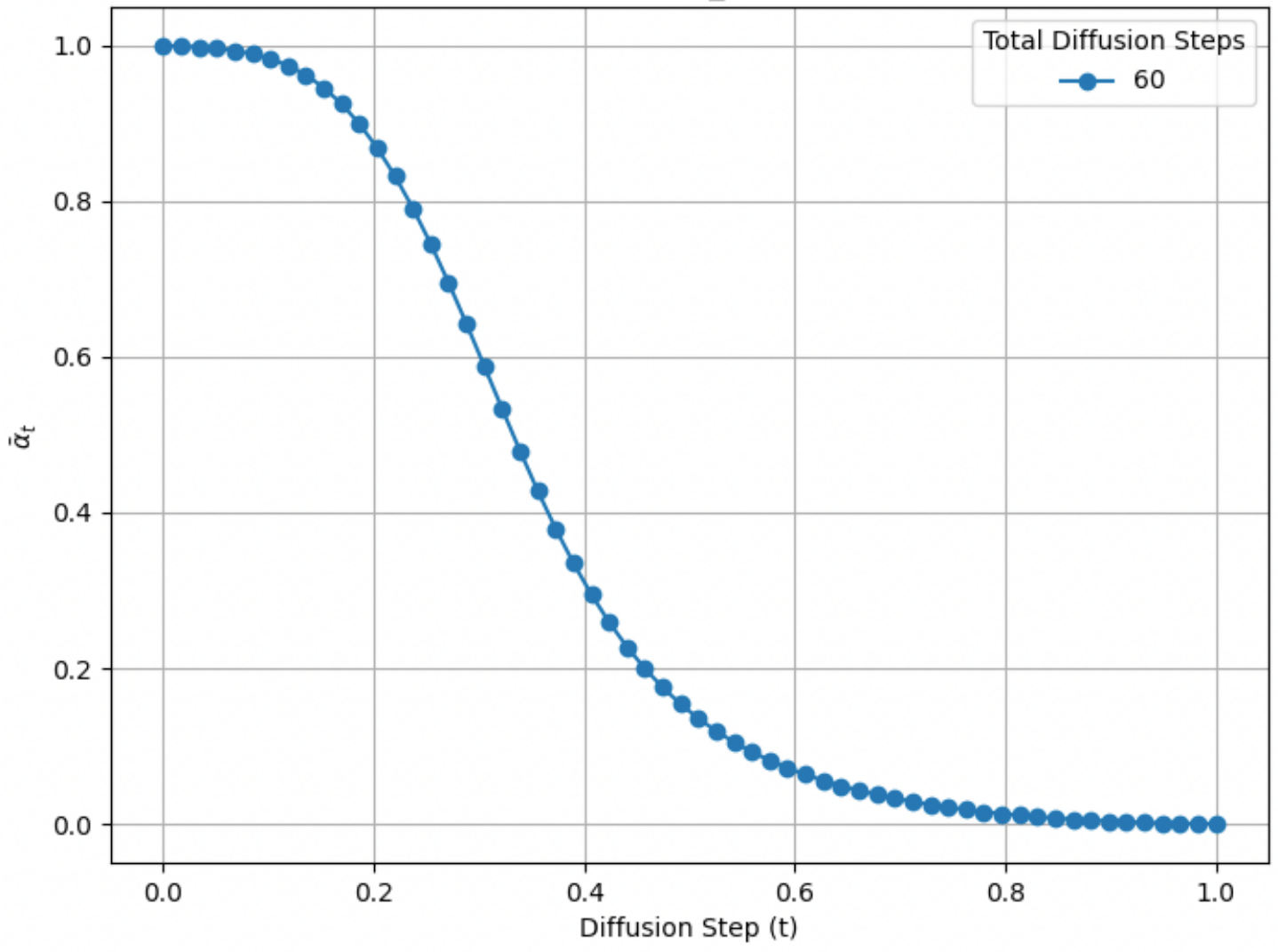}
         \caption{}
        % \caption{$\bSigma$}
        % \label{subfig:cov_mat_1_mul_20}
    \end{subfigure}
        \hspace{1mm}
    \begin{subfigure}[b]{0.3\textwidth}
        \centering
        \includegraphics[width=\textwidth]{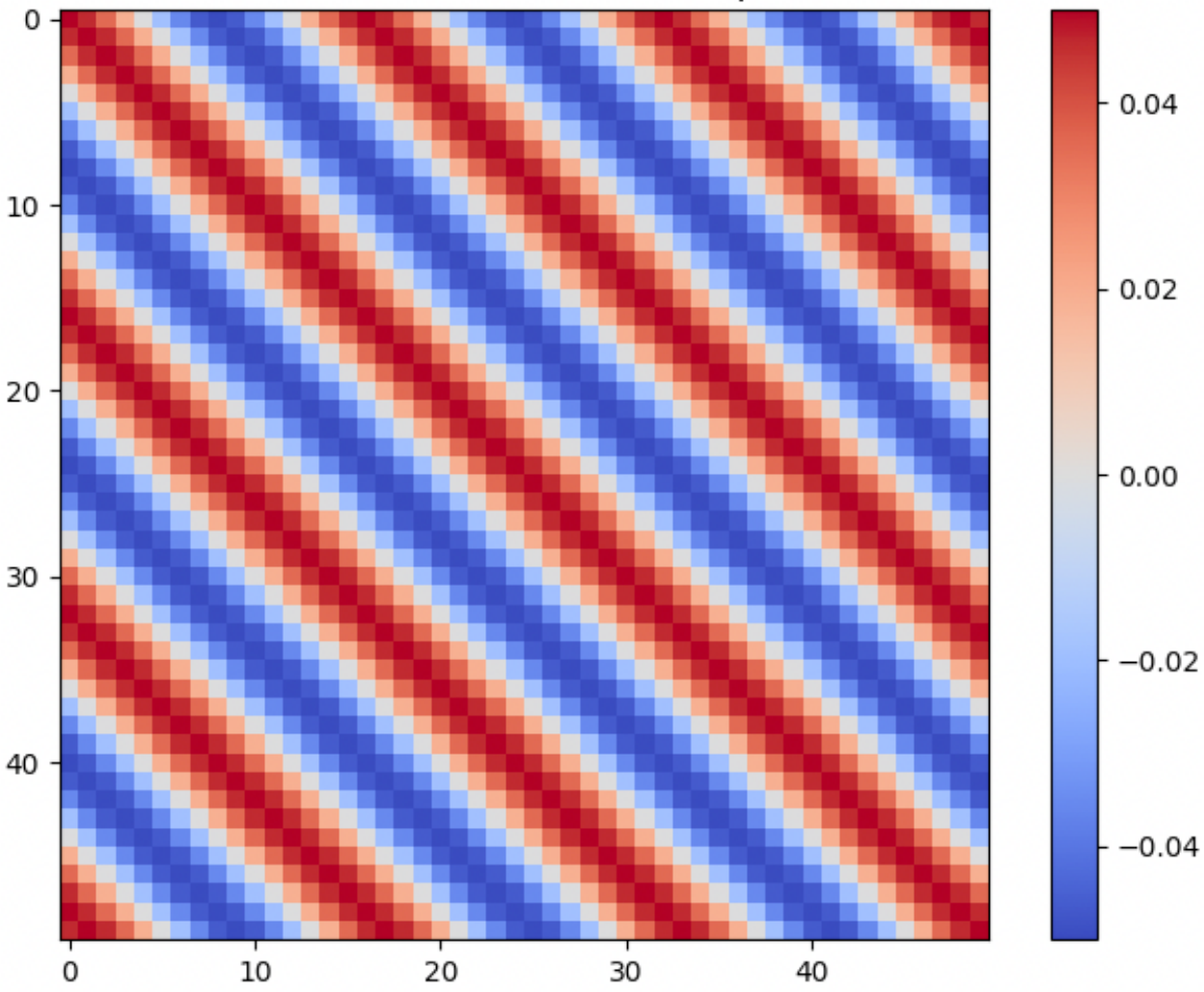}
         \caption{}
        % \caption{$\bSigma$}
        % \label{subfig:cov_mat_1_mul_20}
    \end{subfigure}
        \hfill
        \begin{subfigure}[b]{0.3\textwidth}
        \centering
        \includegraphics[width=\textwidth]{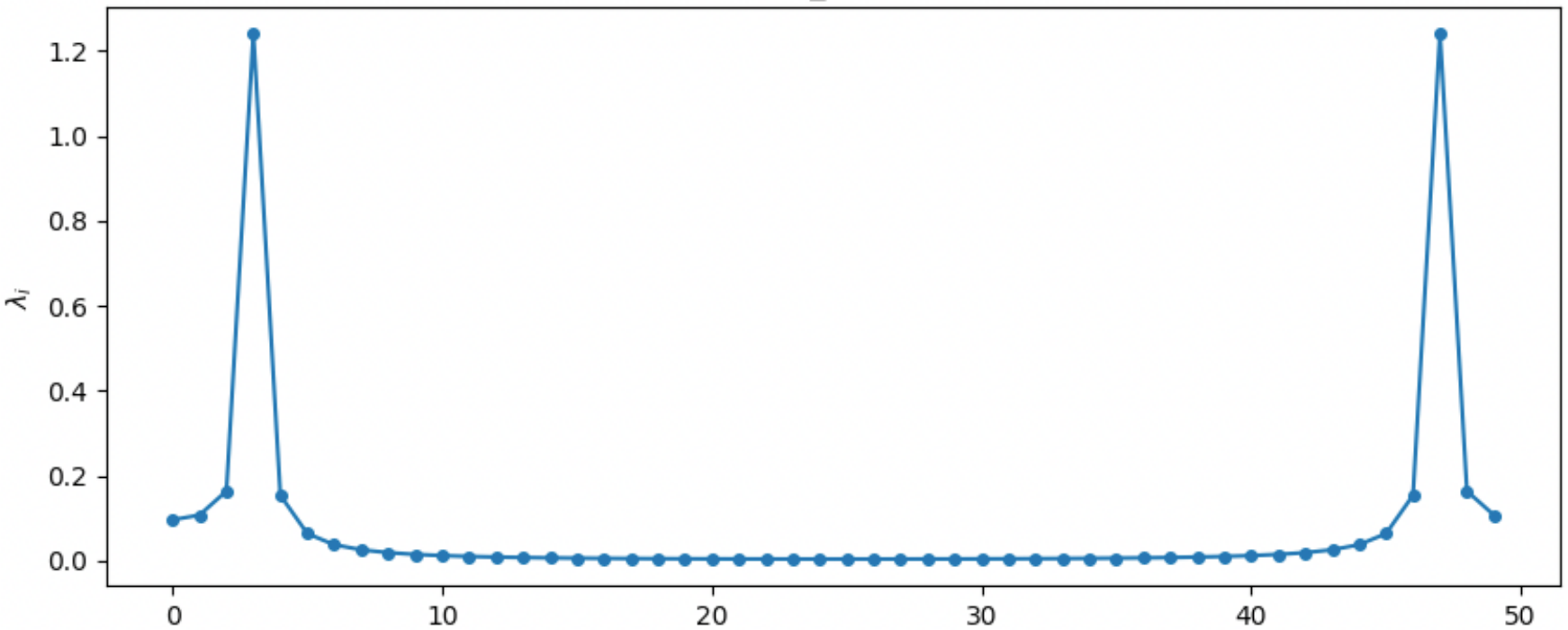}
         \caption{}
        % \caption{$\bSigma$}
        % \label{subfig:cov_mat_1_mul_20}
    \end{subfigure}
      \hfill
        \begin{subfigure}[b]{0.3\textwidth}
        \centering
        \includegraphics[width=\textwidth]{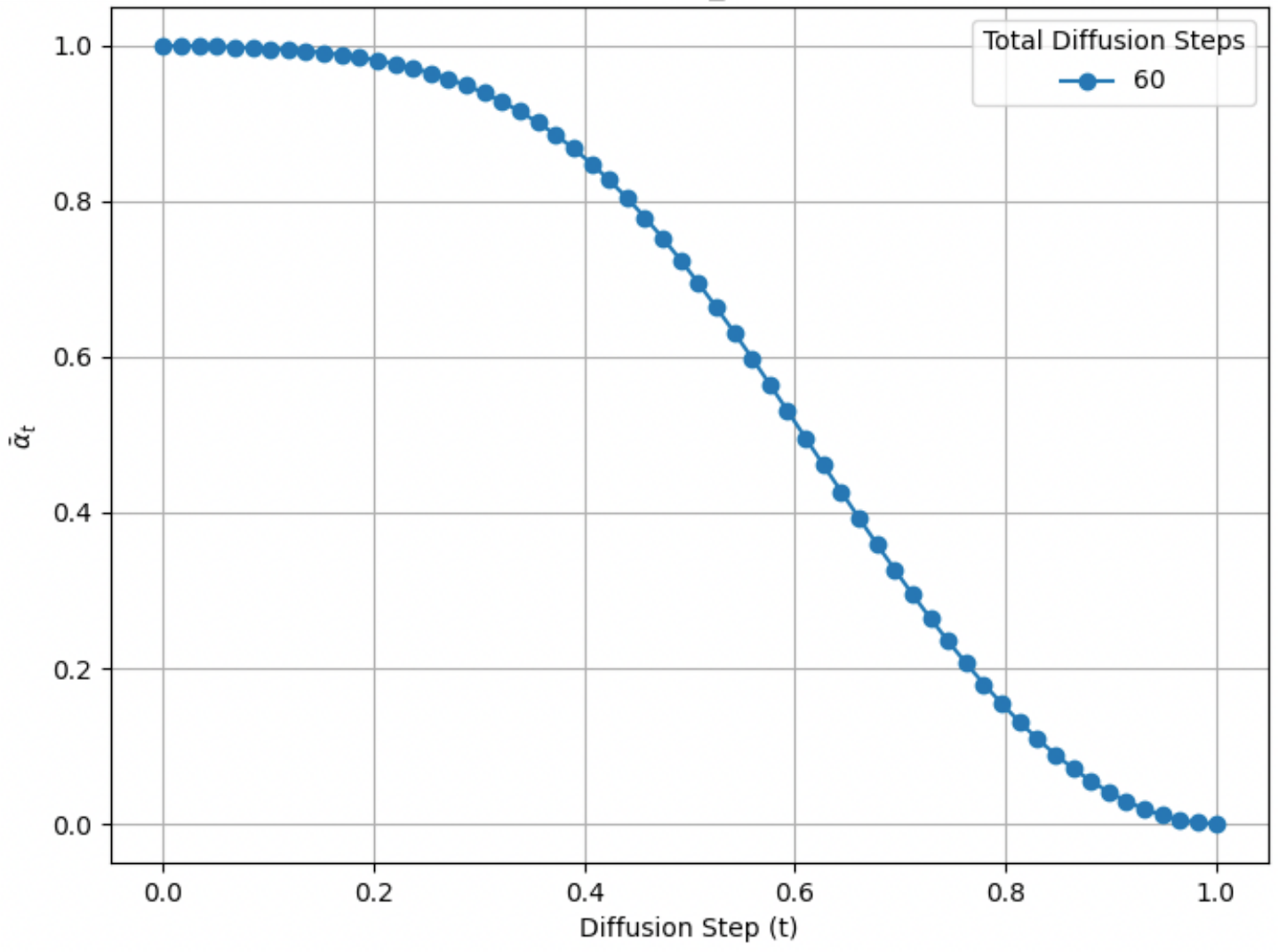}
        % \caption{$\bSigma$}
                \caption{}
        % \label{subfig:cov_mat_1_mul_20}
    \end{subfigure}

    \caption{Various covariance matrices (first column) along with their eigenvalues (second column) and the corresponding optimized noise schedules (third column). The first row presents the matrix from \ref{subsec:Scenario_1} with $l=0.05$, $\bmu = 0.01\cdot\mathbf{1}_d$. The second row shows the same matrix scaled by a factor of $20$ while keeping the same $\bmu$. The third row displays a covariance matrix incorporating a Cosine function in the first row $a$ of the circulant covariance, with $\bmu = 0.3\cdot\mathbf{1}_d$. The fourth row features a circulant matrix derived from a sinusoidal signal with a frequency of $1000$ in the first row $a$ of the circulant covariance matrix, scaled by 0.01, and with $\bmu = 0.1\cdot\mathbf{1}_d$.}

    \label{fig:Scenario_1_different_cov_mtrices}
\end{figure}

In \ref{subsec:Scenario_1}, we designed a specific covariance matrix $\bSigma_0$ and a mean vector $\bmu$ with the intention of resembling characteristics observed in real signals, such as a centered signal with $\bmu \approx \mathbf{0}$. However, the optimization process is not restricted to these particular choices and can be generalized to accommodate various alternative decisions.
Figure \ref{fig:Scenario_1_different_cov_mtrices} displays different covariance matrices along with their corresponding $\bmu$ vectors, followed by the resulting spectral schedules computed using the \emph{Wasserstein-2} distance for $60$ diffusion steps.
% Figure \ref{fig:Scenario_1_different_cov_mtrices} shows a series of covariance matrices (first column) along with their corresponding $\bmu$ vectors, followed by the resulting spectral schedules, computed using the \emph{Wasserstein-2} distance, for $60$ diffusion steps (third column).

% Figure \ref{fig:Scenario_1_different_cov_mtrices} presents various covariance matrices and corresponding $\bmu$ vectors, followed by the resulting spectral schedules based on the \emph{Wasserstein-2} distance for $60$ diffusion steps.

\begin{figure}[H]
    \centering
\vskip 0.2in
    \begin{subfigure}{0.45\textwidth}
        \centering
     \includegraphics[width=\columnwidth]{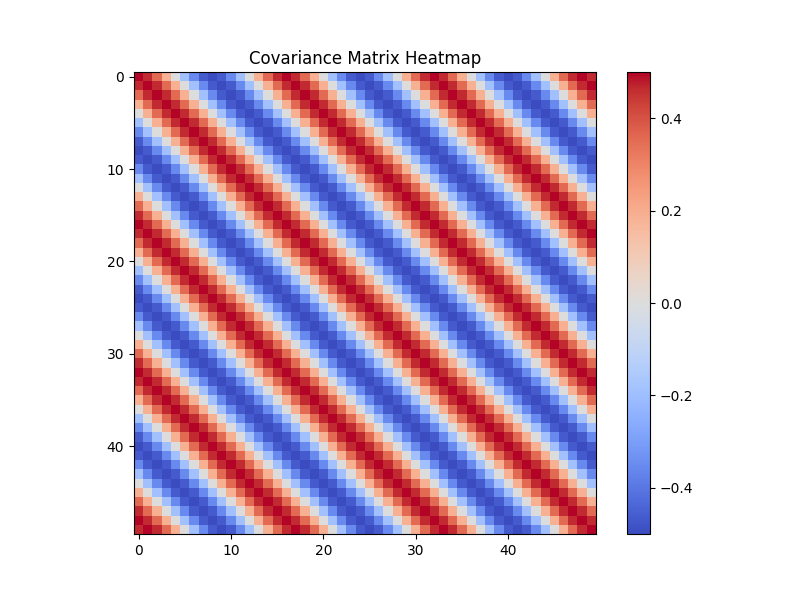}  
        \caption{$\bSigmaZ$}
        \label{fig:matrix_9_no_mul_Sigma_appendix}
        \end{subfigure}
    \begin{subfigure}{0.35\textwidth} % Define the width of the 
            \centering
\includegraphics[width=\columnwidth]{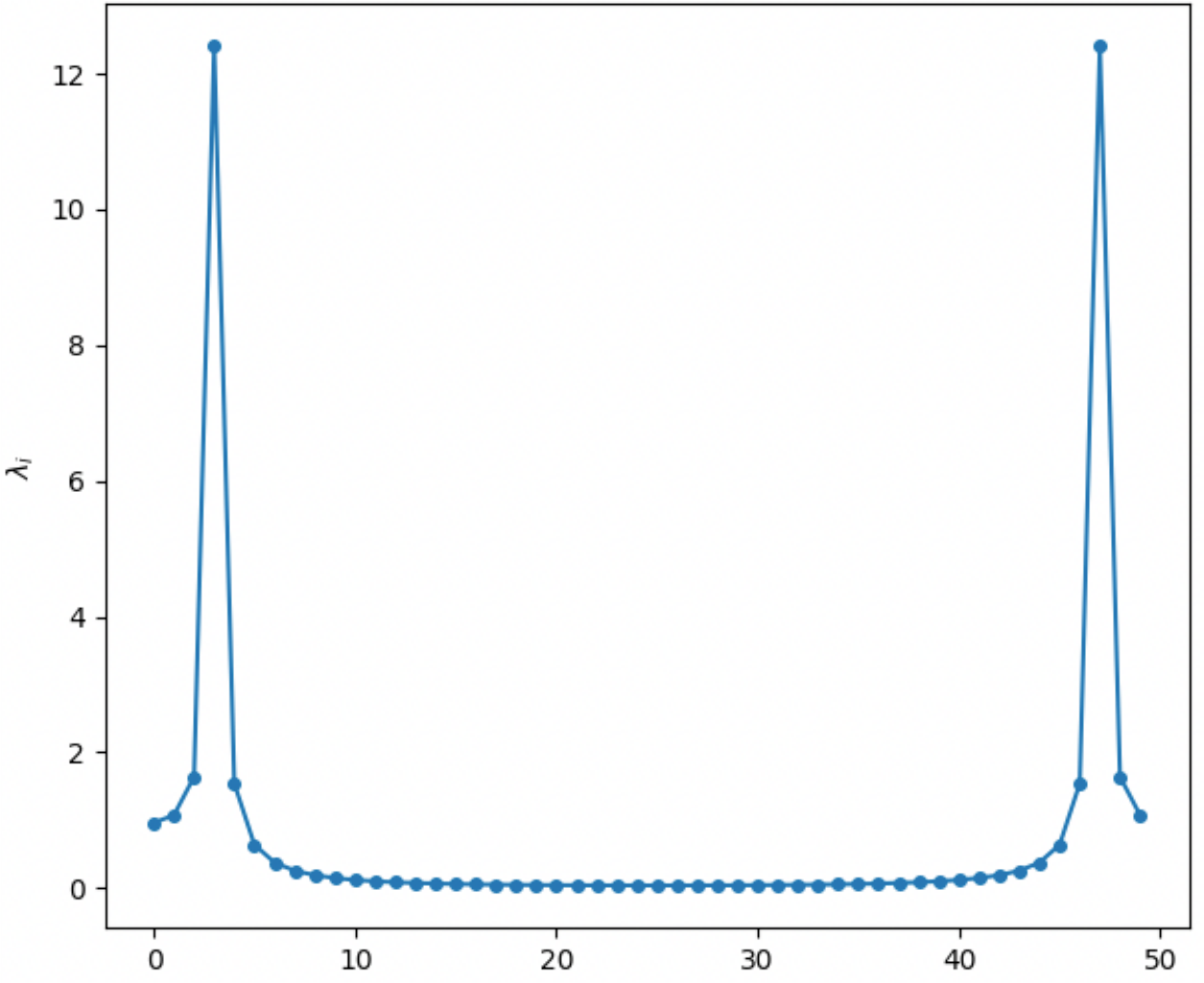}
 \caption{$tr(\mathbf{\Lambda}_0)$}
 \label{fig:matrix_9_no_mul_lambda_appendix}
    \end{subfigure}
     \centering
    \begin{subfigure}{0.45\textwidth}
  \centering
  \includegraphics[width=\textwidth]{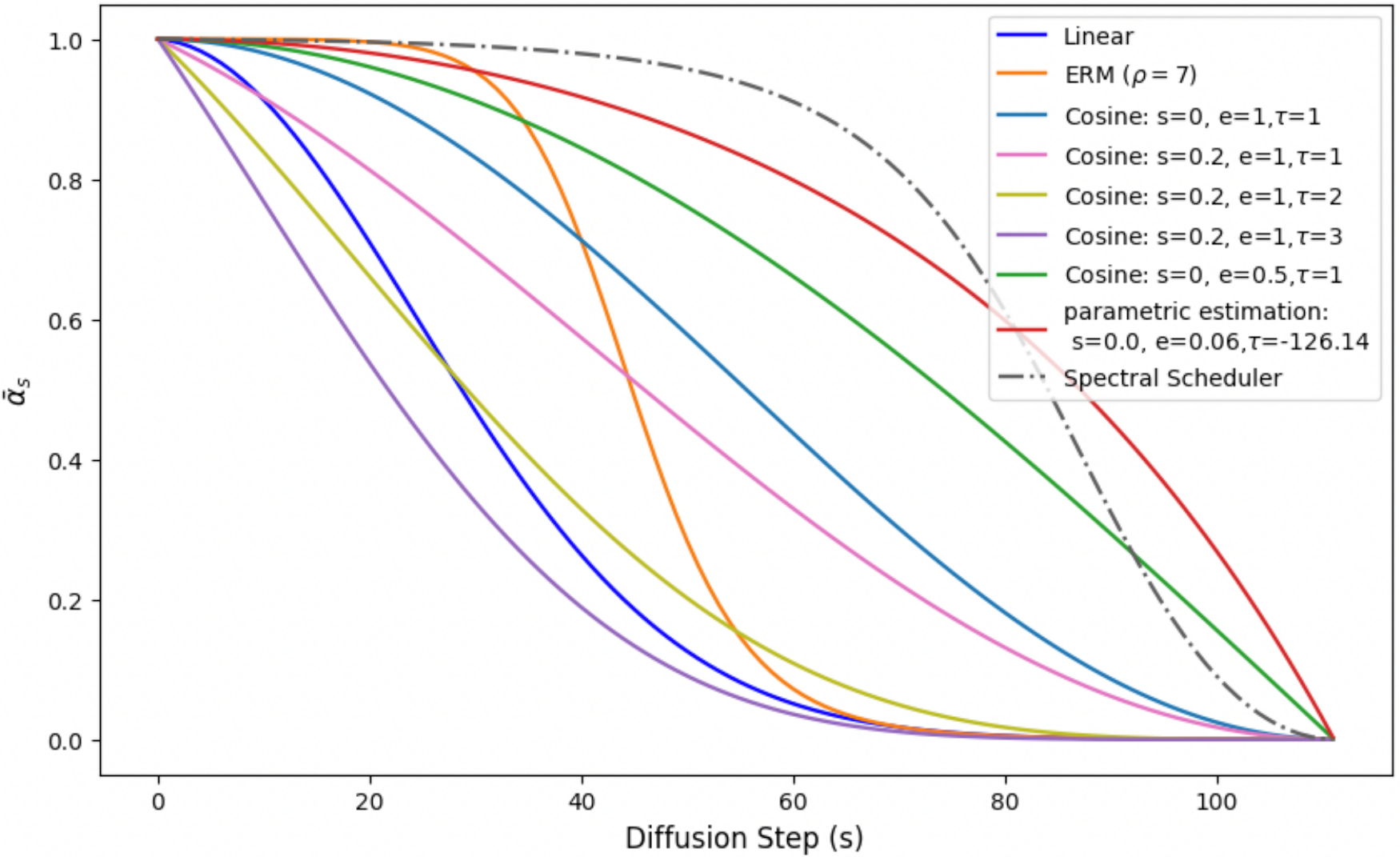}
  \caption{}
\label{subfig:matrix_9_Cosine_Comparison_wasserstein_appendix_dkl}
  \end{subfigure}
 \centering
    \hspace{2mm}
\begin{subfigure}{0.45\textwidth}
  \centering
  \includegraphics[width=\textwidth]{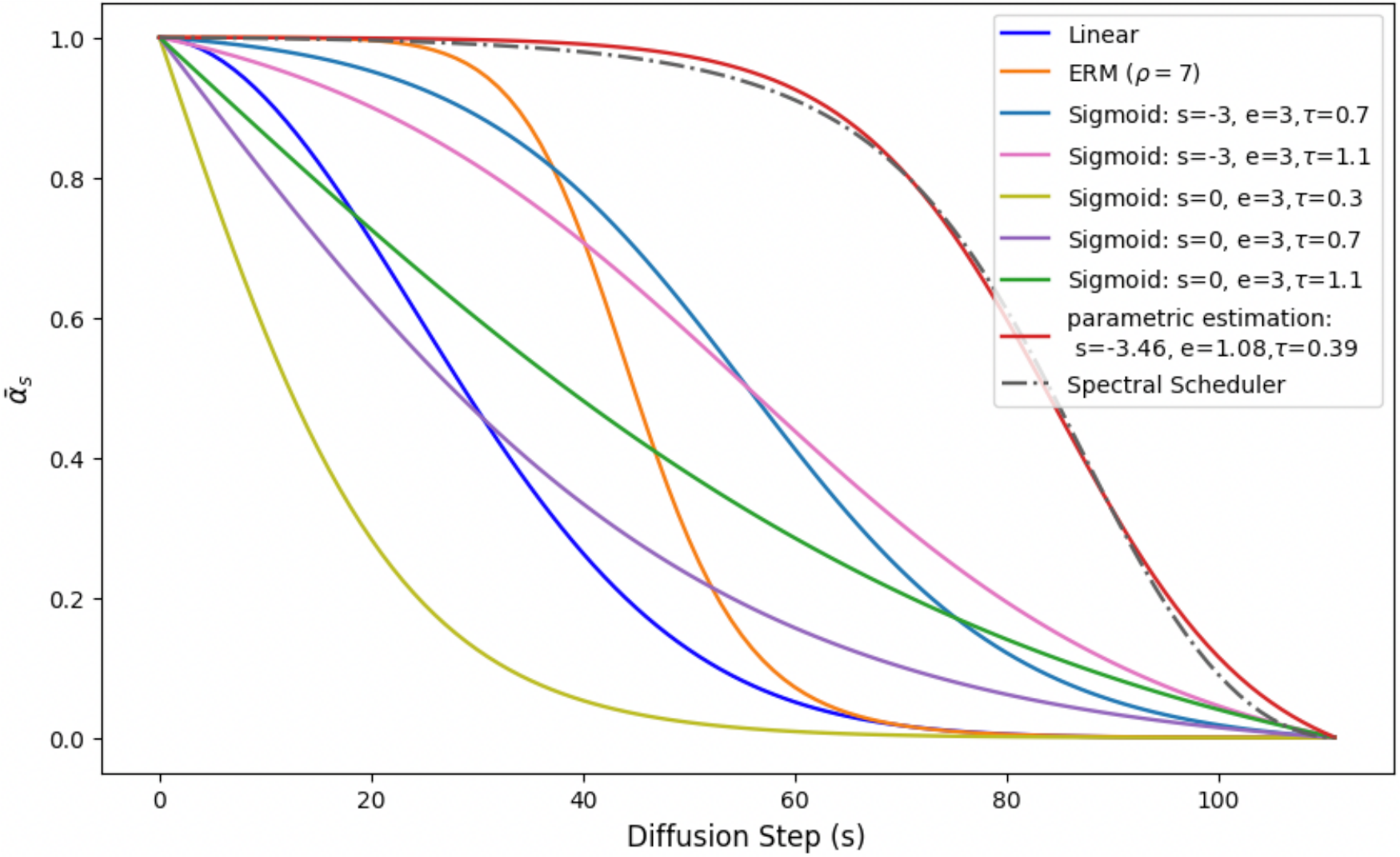}
 \caption{}
\label{subfig:matrix_9_Sigmoid_Comparison_wasserstein_appendix_dkl}
\end{subfigure}
% \vskip -0.2in
\caption{
Figure \ref{fig:matrix_9_no_mul_Sigma_appendix} shows the circulant covariance matrix, $\bSigma_0$, whose first row is derived from a sinusoidal signal with a frequency of $1000$ Hz. Figure \ref{fig:matrix_9_no_mul_lambda_appendix} displays the trace of the corresponding $\bLambda$ matrix. Figures \ref{subfig:matrix_9_Cosine_Comparison_wasserstein_appendix_dkl} and \ref{subfig:matrix_9_Sigmoid_Comparison_wasserstein_appendix_dkl} compare the spectral recommendations for $d=50$, $112$ diffusion steps, using the \emph{KL-divergence}, with various noise schedule heuristics including Cosine and Sigmoid, respectively. The parametric estimations for the Cosine and Sigmoid are highlighted in red.}
\label{fig:exp_1_Sigma_0_and_Lambda_0_matrix_9}
\end{figure}

\newpage
While we cannot cover all possible choices for the covariance matrix $\bSigma_0$ and the vector $\bmu$, we aim to provide a broader perspective on the \emph{KL-divergence} loss. Figure \ref{fig:exp_1_Sigma_0_and_Lambda_0_matrix_9} illustrates a circulant covariance matrix whose first row is derived from a sinusoidal signal with a frequency of $1000$ Hz, along with the corresponding spectral recommendation based on \emph{KL-divergence}, where $\bmu = 0.1\cdot\mathbf{1}_d$.

The results above show that modifying the dataset properties, such as the covariance matrix and expectation, along with altering the loss function, leads to noise schedules with a similar overall structure but varying details. In Appendix \ref{sec:appendix_further_discussion}, we explore the connection between the dataset properties, the loss function, and the resulting noise schedules.

\newpage
\section{Supplementary Experiments for Empirical Gaussian Distribution}\label{sec:Scenario_2_experiments}

We present additional details on the CIFAR-10, Gaussian MUSIC piano and SC09 datasets, along with the spectral noise schedules derived from them \cite{moura2020music,warden2018speech}.
\subsection{CIFAR-10 Dataset}

% The covariance matrix dimention on the CIFAR-10 is $d=3072$ while we used the original resolution of $32\times32$ with 3 colored channels.

\begin{figure}[H]
    \centering
    \begin{subfigure}{0.45\textwidth}
        \centering
        \includegraphics[width=\textwidth]{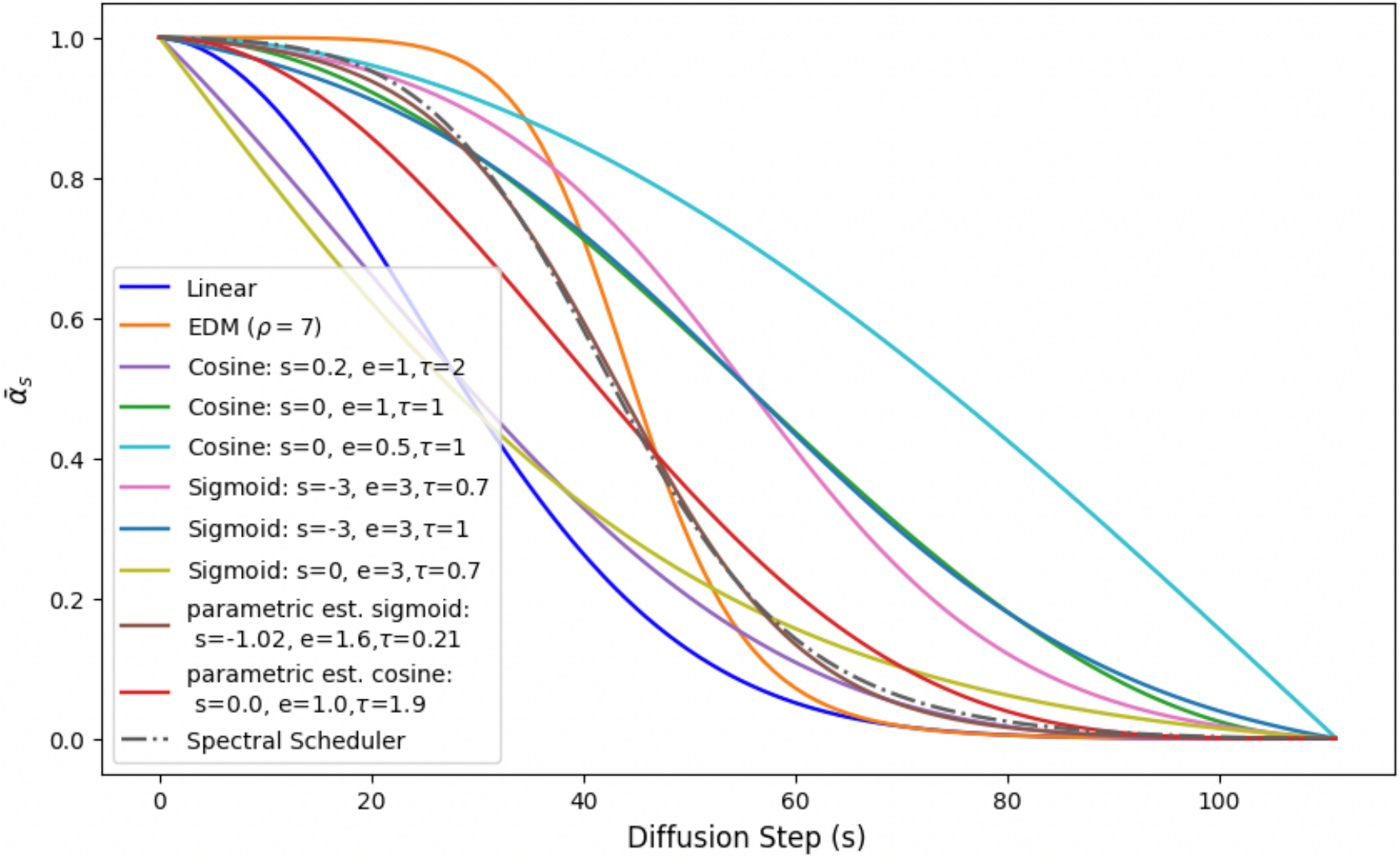}
        \caption{}
        \label{subfig:CIFAR-10_optimized_schedule}
    \end{subfigure}
    \begin{subfigure}[b]{0.45\textwidth}
        \centering
     \includegraphics[width=\textwidth]{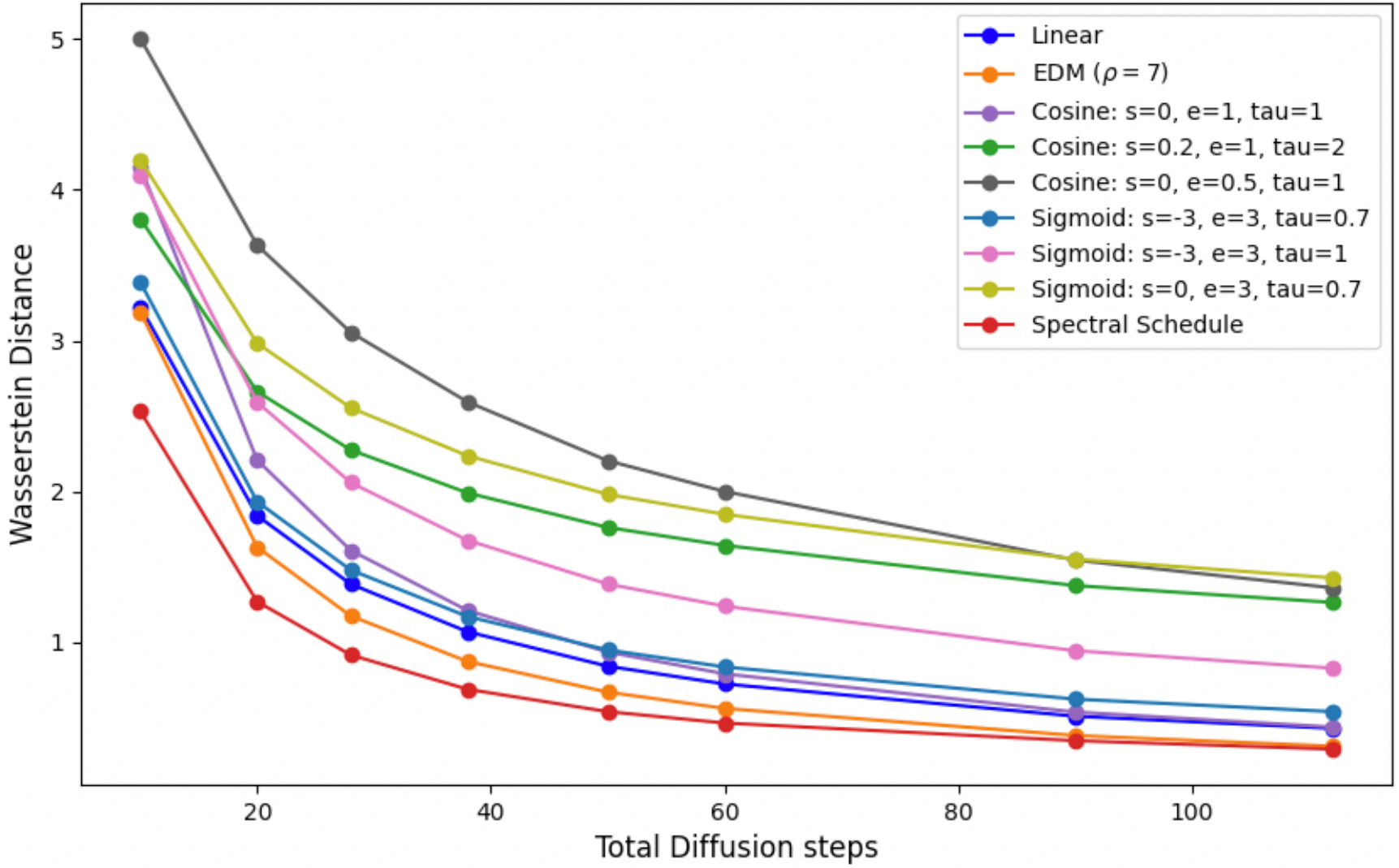}
        \caption{}
        \label{subfig:Cifar_10_EXP_2_wasserstein}
    \end{subfigure}
   \hspace{1mm}

\caption{Figure \ref{subfig:CIFAR-10_optimized_schedule} compares the spectral schedule with several heuristic noise schedules for a 112-step diffusion process. The best-fitting parametric approximations for the Cosine and Sigmoid schedules are shown in red and brown, respectively. Figure \ref{subfig:Cifar_10_EXP_2_wasserstein} demonstrates the optimality of the spectral recommendation in the Gaussian case, revealing higher effectiveness in diffusion processes with fewer steps, where discretization error is more pronounced.}

% \caption{Figure \ref{subfig:CIFAR-10_optimized_schedule} compares the spectral schedule with various heuristic noise schedules for $S=112$ diffusion steps. The best fitted parametric estimations for the Cosine and Sigmoid schedules are presented in red and brown respectively. Figure \ref{subfig:Cifar_10_EXP_2_wasserstein} demonstrate the optimality of the the spectral recommendation for the gaussian case. reaviling a higher effectivenes in diffusion process with lower number of steps where discritization error is more }

\label{fig:CIFAR-10_scheduling_results_appendix_exp2}
\end{figure}

The estimated covariance matrix for CIFAR-10 has a dimension of $d = 3072$, corresponding to the original image resolution of $32 \times 32$ with three color channels. Figure \ref{subfig:CIFAR-10_optimized_schedule} presents the spectral schedule derived for this covariance matrix using a 112-step diffusion process. Notably, its structure resembles that of the EDM schedule. Figure \ref{subfig:Cifar_10_EXP_2_wasserstein} illustrates the Wasserstein-2 distance evaluated on the Gaussian distribution associated with CIFAR-10, demonstrating the optimality of the spectral schedule across all diffusion steps. This effect is especially pronounced at lower step counts, where discretization error is more significant. Interestingly, heuristic schedules whose structures closely align with the spectral schedule, such as EDM, also achieve comparable performance, further supporting the connection between schedule structure and diffusion quality.

\subsection{MUSIC Dataset}

% For the MUSIC dataset, The covariance matrix is estimated by using a sliding window of length $d=400$ ($0.025$ seconds) from the original dataset, excluding those with an $L_1$ energy below a specified threshold ($th=0.05$) so as to mitigate the influence of silent regions in the covariance estimation. 
% The resulting covariance matrix is symmetric and nearly a Toeplitz matrix.

To estimate the covariance matrix for the MUSIC dataset, we apply a sliding window of length $d = 400$ ($0.025$ seconds) to the original audio, excluding segments with $L_1$ energy below a threshold of $th = 0.05$ to reduce the influence of silent regions. The resulting covariance matrix is symmetric and exhibits an approximately Toeplitz structure.

% Figure \ref{fig:Exp_2_original_characteristics_MUSIC_dataset_dim_400_appendix} provides a visual representation of the estimated covariance matrix $\bSigma_0$ and its corresponding $\bLambda$, as discussed in Sec. \ref{subsec:scenario_2}.

% \begin{figure}[h]
%     \centering
%     \begin{subfigure}{0.4\textwidth}
%         \centering
%         \includegraphics[width=\columnwidth]{figures_spectral/Exp_2/Dim_400_silence_threshold_0_05_music_piano/MUSIC_piano_ALL_est_circulant_covariance_mat_th_0.05_lambda_cir_1.pdf}
%         \caption{$\bSigmaZ$}
%         \label{subfig:Estimated_circulant_cov_400}
%     \end{subfigure}
%     % \hspace{1mm}
%     % \hfill
%     \begin{subfigure}{0.4\textwidth}
%         \centering
%         \includegraphics[width=\columnwidth]{figures_spectral/Exp_2/Dim_400_silence_threshold_0_05_music_piano/lambda_0.pdf}
%         \caption{$\text{tr}\left(\bLambda\right)$}
%         \label{subfig:original_lambda_0_400}
%     \end{subfigure}

%     \caption{\autoref{subfig:Estimated_circulant_cov_400} illustrates $\bSigma_0$, the Circulant approximation of the Covariance matrix for the MUSIC piano dataset with $d=400$ and $th=0.05$, while \autoref{subfig:original_lambda_0_400} displays its DFT coefficients (the eigenvalues). $\bmu \approx \0$ is also calcualted from the dataset.}
% \label{fig:Exp_2_original_characteristics_MUSIC_dataset_dim_400_appendix}
% \end{figure}

\begin{figure}[H]
    \centering
    \begin{subfigure}{0.45\textwidth}
        \centering
        \includegraphics[width=\textwidth]{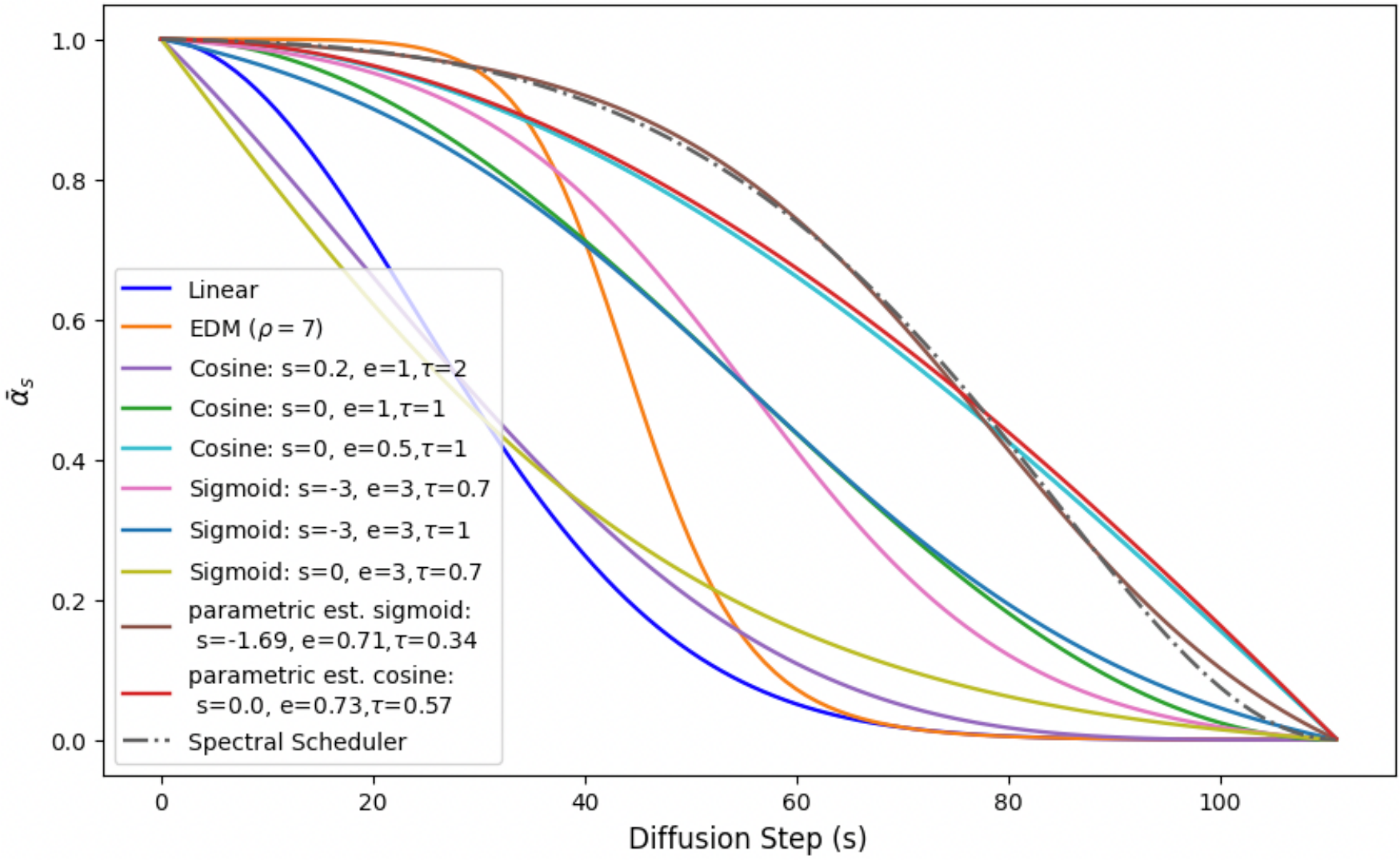}
        \caption{}
       \label{fig:Exp_2_Spectral_scheduler_comparison_MUSIC_400}
    \end{subfigure}
    \begin{subfigure}[b]{0.45\textwidth}
        \centering
     \includegraphics[width=\textwidth]{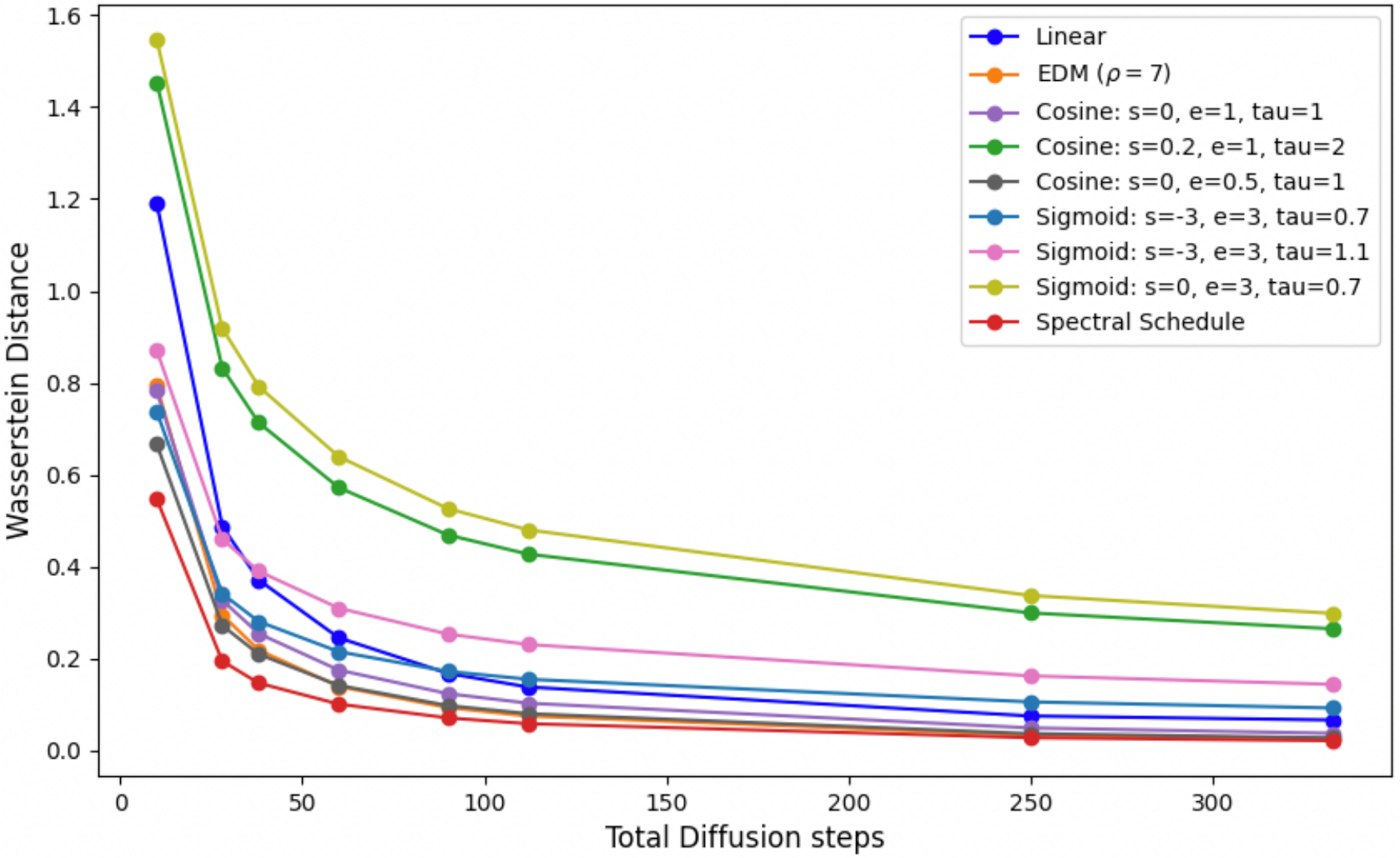}
        \caption{}
        \label{subfig:wasserstein_distence_Exp_2}
    \end{subfigure}
    \hspace{1mm}
    % \hfill

\caption{Figure \ref{fig:Exp_2_Spectral_scheduler_comparison_MUSIC_400} compares the spectral schedule with several heuristic noise schedules for a 112-step diffusion process. The best-fitting parametric approximations for the Cosine and Sigmoid schedules are shown in red and brown, respectively. Figure \ref{subfig:wasserstein_distence_Exp_2} demonstrates the optimality of the spectral recommendation in the Gaussian case, revealing higher effectiveness in diffusion processes with fewer steps, where discretization error is more pronounced.}
\label{fig:MUSIC_400_scheduling_results_app_exp_2}
\end{figure}

Figure \ref{fig:Exp_2_Spectral_scheduler_comparison_MUSIC_400} presents the spectral schedule derived from the estimated covariance matrix using a 112-step diffusion process, while Figure \ref{subfig:wasserstein_distence_Exp_2} shows the Wasserstein-2 distance comparison for the MUSIC Piano Gaussian dataset. Similar conclusions regarding optimality and the alignment with heuristic schedules hold here as well, with the Cosine $(0, 0.5, 1)$ schedule being the closest in both structure and performance to the spectral recommendation.

% \caption{Comparison of the spectral schedule with various heuristic noise schedules for $S=112$ diffusion steps. The best fitted parametric estimations for the Cosine and Sigmoid schedules are presented in red and brown respectively.}

% \red{Figures \ref{subfig:Exp_3_frobenius_norm_app} and \ref{subfig:Exp_3_wasserstein_2_distance_app} visualize the \emph{Frobenius norm} and the \emph{Wasserstein distance} of the spectral noise schedule (in red) compared to existing heuristics, considering the following diffusion steps: $\{10, 20, 28, 30, 38, 60, 90, 112, 200, 250\}$.
%  Furthermore, for Cifar-10, the approximation error appears to be less pronounced, as the results demonstrate greater stability.}

\newpage
\subsection{SC09 Dataset}

In this section, we apply our method to a different dataset, SC09.  SC09 is a subset of the \emph{Speech Commands Dataset} \cite{warden2018speech} and consists of spoken digits ($0$–$9$). Each audio sample has a duration of one second and is recorded at a sampling rate of $16$ kHz.

\begin{figure}[H]
 \centering
    \begin{subfigure}{0.3\textwidth}
  \centering
  \includegraphics[width=\textwidth]{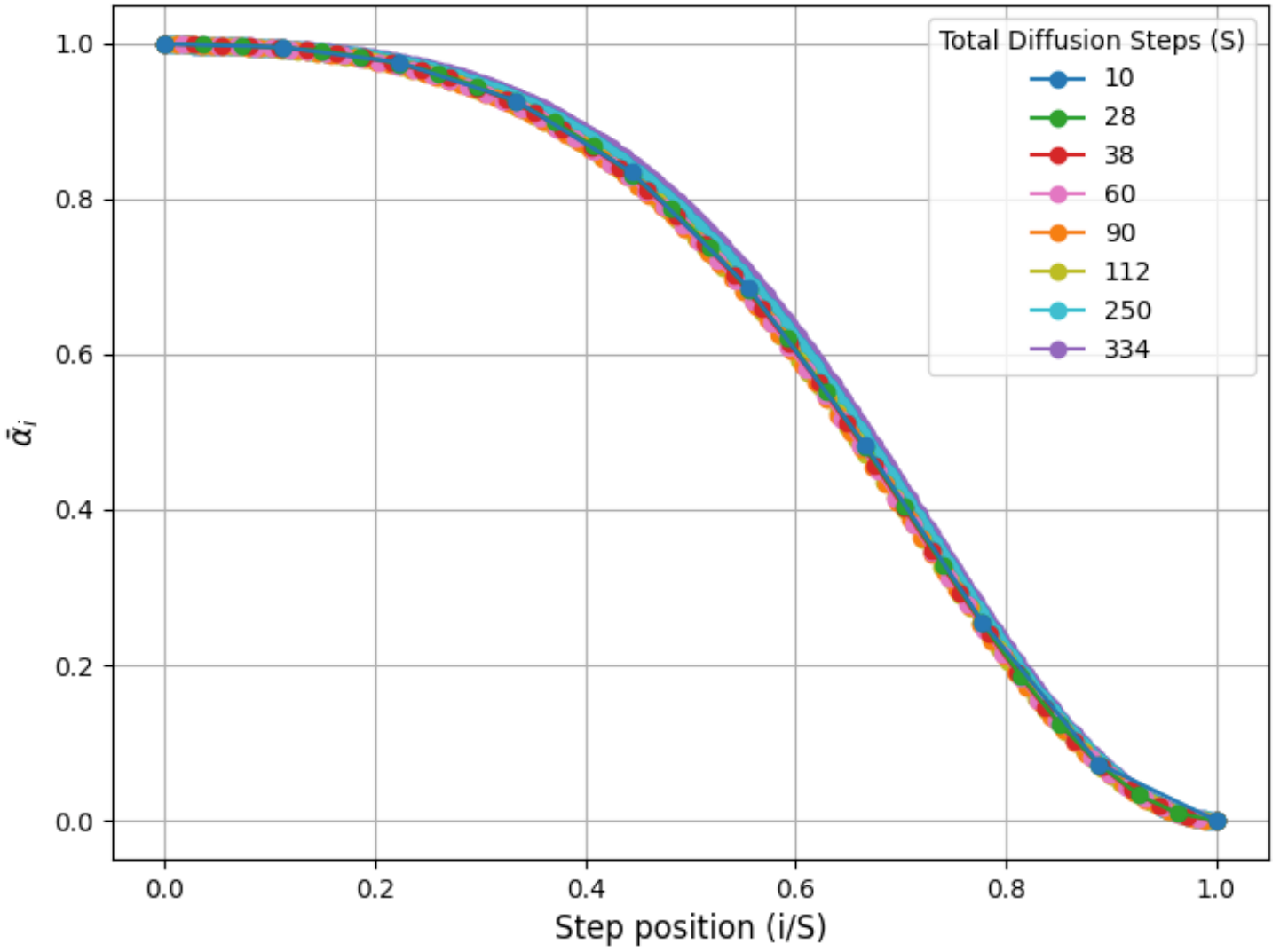}
  \caption{}
  \label{fig:Exp_2_Spectral_reccomandation_wasserstein_0_05}
\end{subfigure}
 \centering
    \begin{subfigure}{0.3\textwidth}
  \centering
  \includegraphics[width=\textwidth]{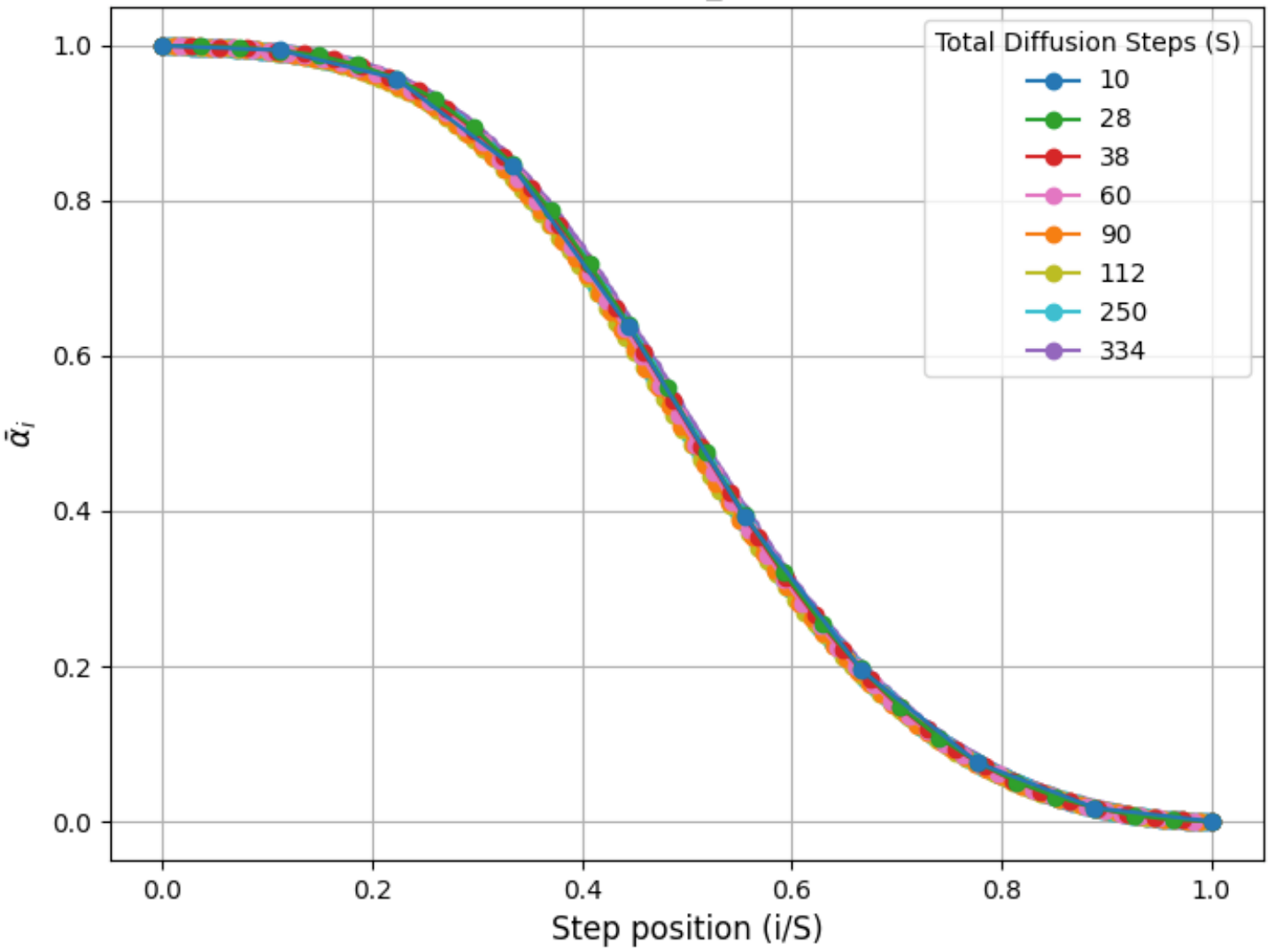}
  \caption{}
  \label{fig:Exp_2_Spectral_reccomandation_wasserstein_0_1}
\end{subfigure}
 \centering
    % \hfill
\begin{subfigure}{0.4\textwidth}
  \centering
  \includegraphics[width=\textwidth]{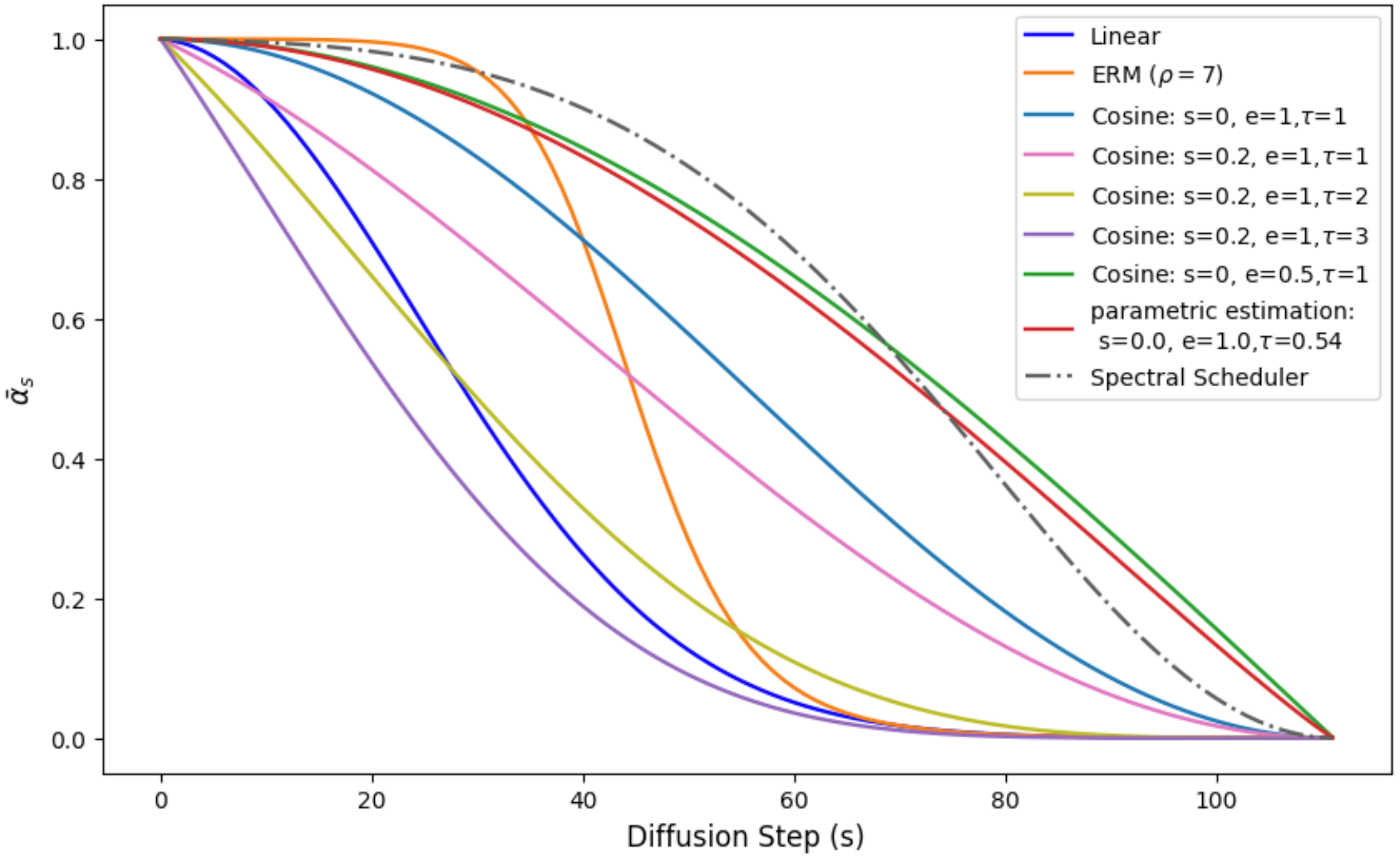}
 \caption{}
  \label{subfig:Exp_2_sigmoid_Comparison_wasserstein_0_05}
\end{subfigure}
    \centering
    \begin{subfigure}{0.4\textwidth}
        \centering
        \includegraphics[width=\columnwidth]{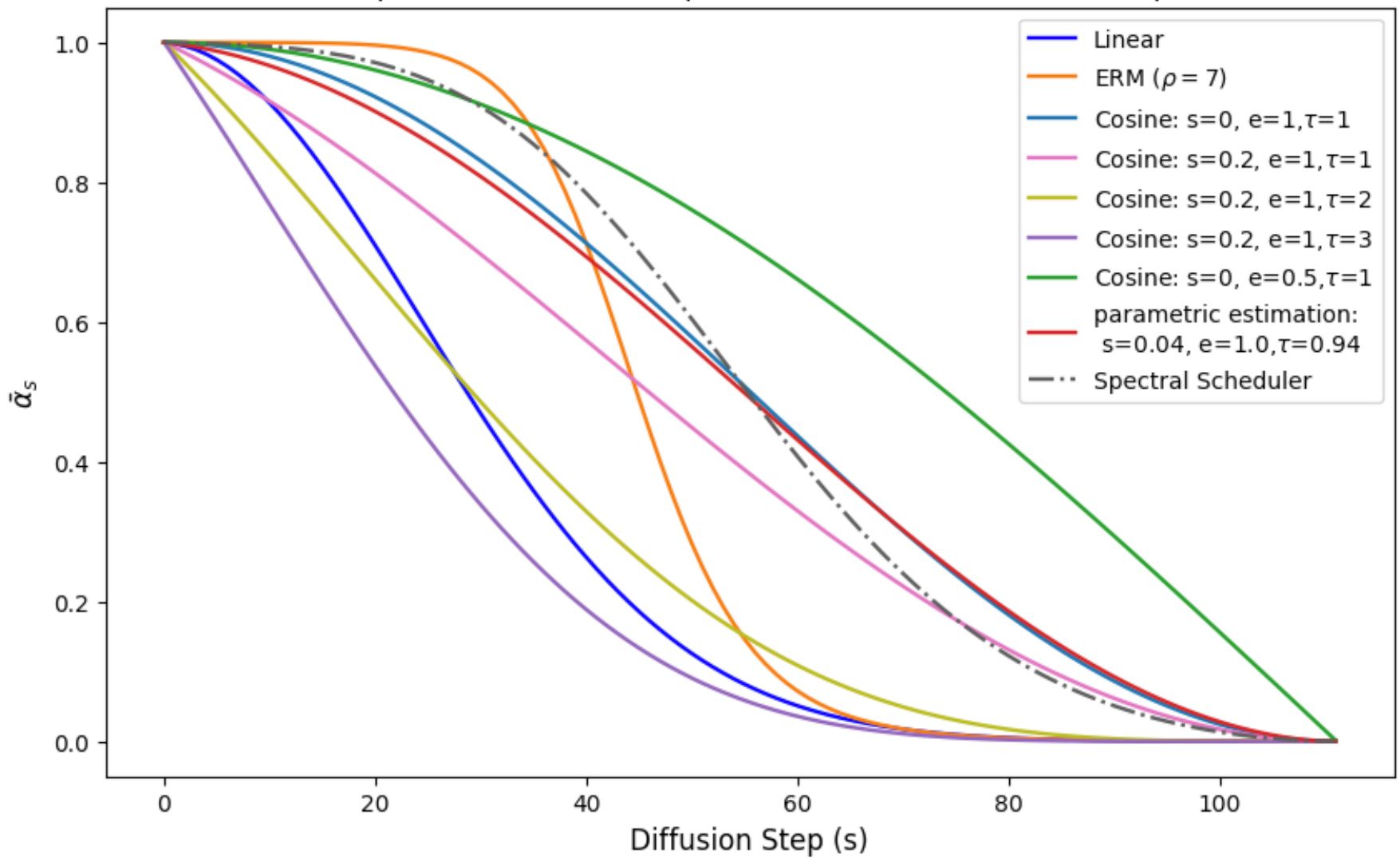}
        \caption{}
        \label{subfig:Exp_2_Cosine_Comparison_wasserstein_0_1}
    \end{subfigure}
       \centering
    \begin{subfigure}{0.4\textwidth}
        \centering
        \includegraphics[width=\columnwidth]{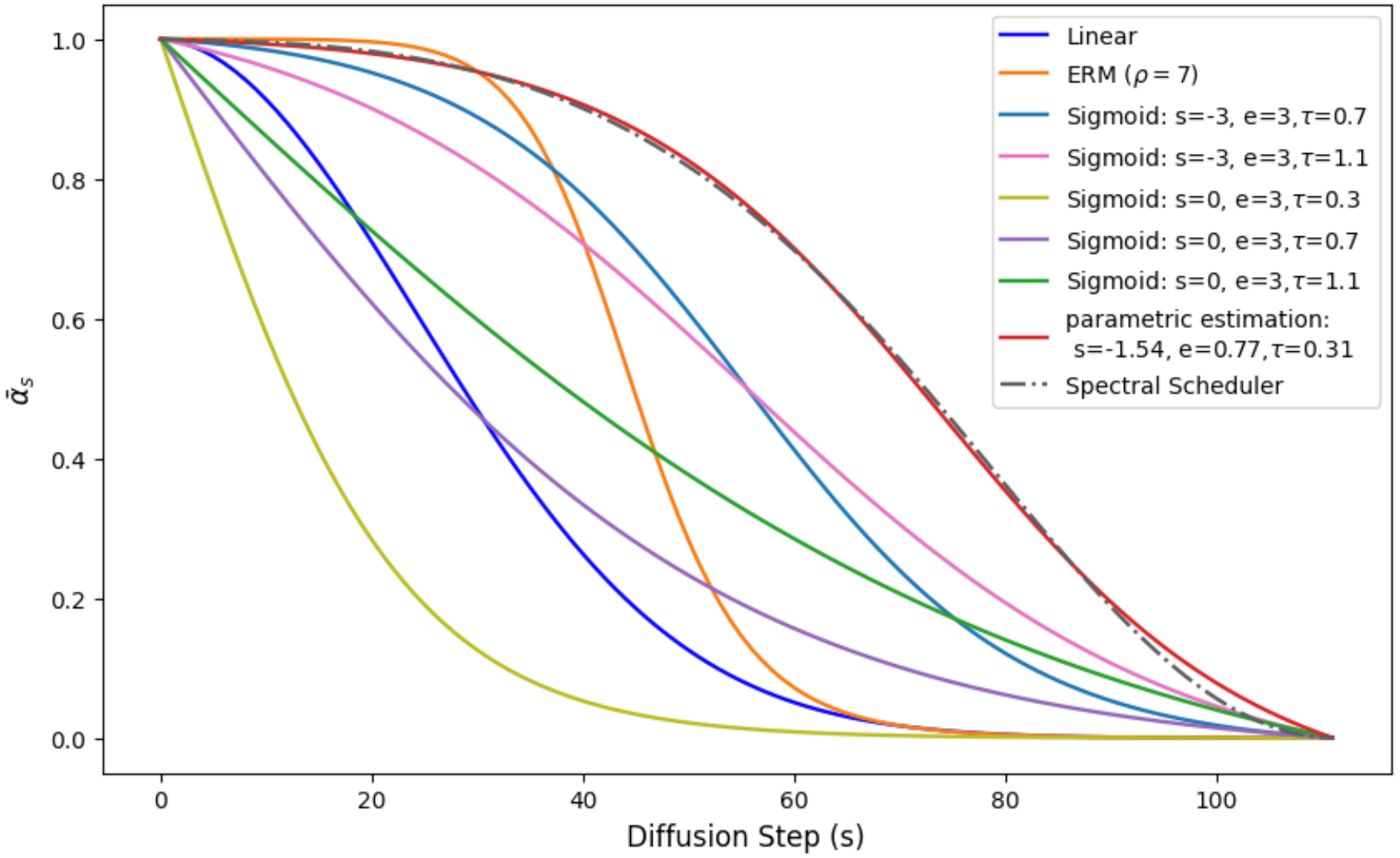}
        \caption{}
        \label{subfig:Exp_2_Cosine_Comparison_wasserstein_0_05}
    \end{subfigure}
 \centering
    % \hfill
    \hspace{1mm} 
\begin{subfigure}{0.4\textwidth}
  \centering
  \includegraphics[width=\textwidth]{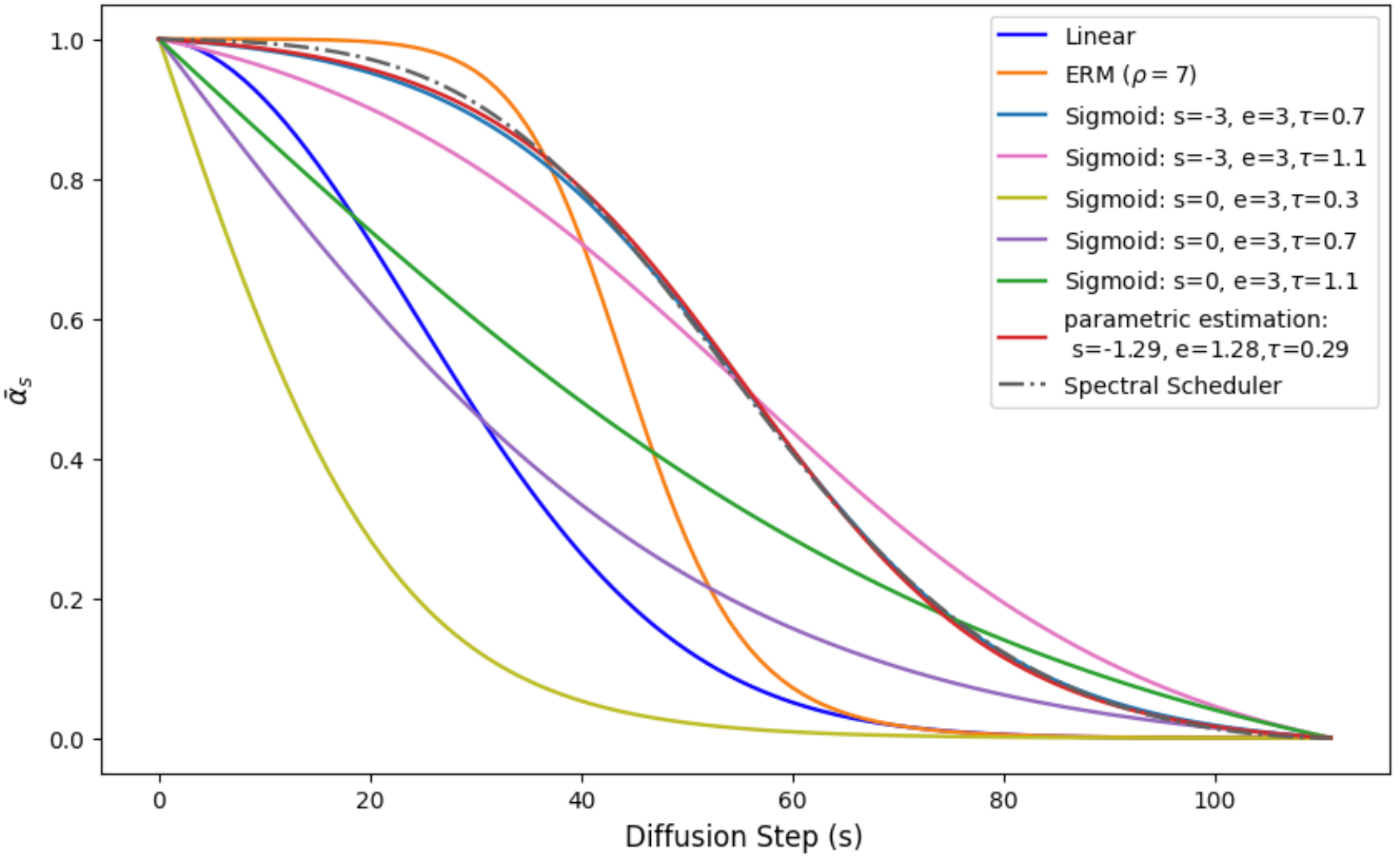}
 \caption{}
  \label{subfig:Exp_2_sigmoid_Comparison_wasserstein_0_1}
\end{subfigure}
\caption{Spectral noise schedules for the $SC09$ dataset with $d = 16,000$ and thresholds $th = 0.05$ (left column) and $th = 0.1$ (right column). The first row shows spectral recommendations for various diffusion steps, while the second and third rows compare the spectral recommendation for $112$ steps with heuristic noise schedules, including the Cosine and Sigmoid schedules. The parametric estimations for the Cosine and Sigmoid functions are shown in red, respectively.}

\label{fig:Exp_2_Spectral_scheduler_comparison_SC09_th_0_1_th_0_05}
\end{figure}

Differing from Sec. \ref{subsec:scenario_2}, here we use segments of length $d=16000$ samples (one second) and set $th=0.05$ in one setting and $0.1$ in another. Figure \ref{fig:Exp_2_Spectral_scheduler_comparison_SC09_th_0_1_th_0_05} presents the spectral recommendations for $th = 0.05$ and $th = 0.1 $ in the left and right columns, respectively.

% ###################

Figures \ref{fig:Exp_2_Spectral_reccomandation_wasserstein_0_05} and \ref{fig:Exp_2_Spectral_reccomandation_wasserstein_0_1} demonstrate that the spectral recommendation for $th=0.05$ exhibits a more concave behavior compared to $th=0.1$. A more detailed discussion on the influence of the $th$ and $d$ parameters on the covariance matrix and the resulting noise schedule is provided in Appendices \ref{sec:scenario_2_Analysis_of_Different_Aspects} and \ref{sec:appendix_further_discussion}.

\begin{figure}[H]
    \centering
        \centering
        \includegraphics[width=0.5\textwidth]{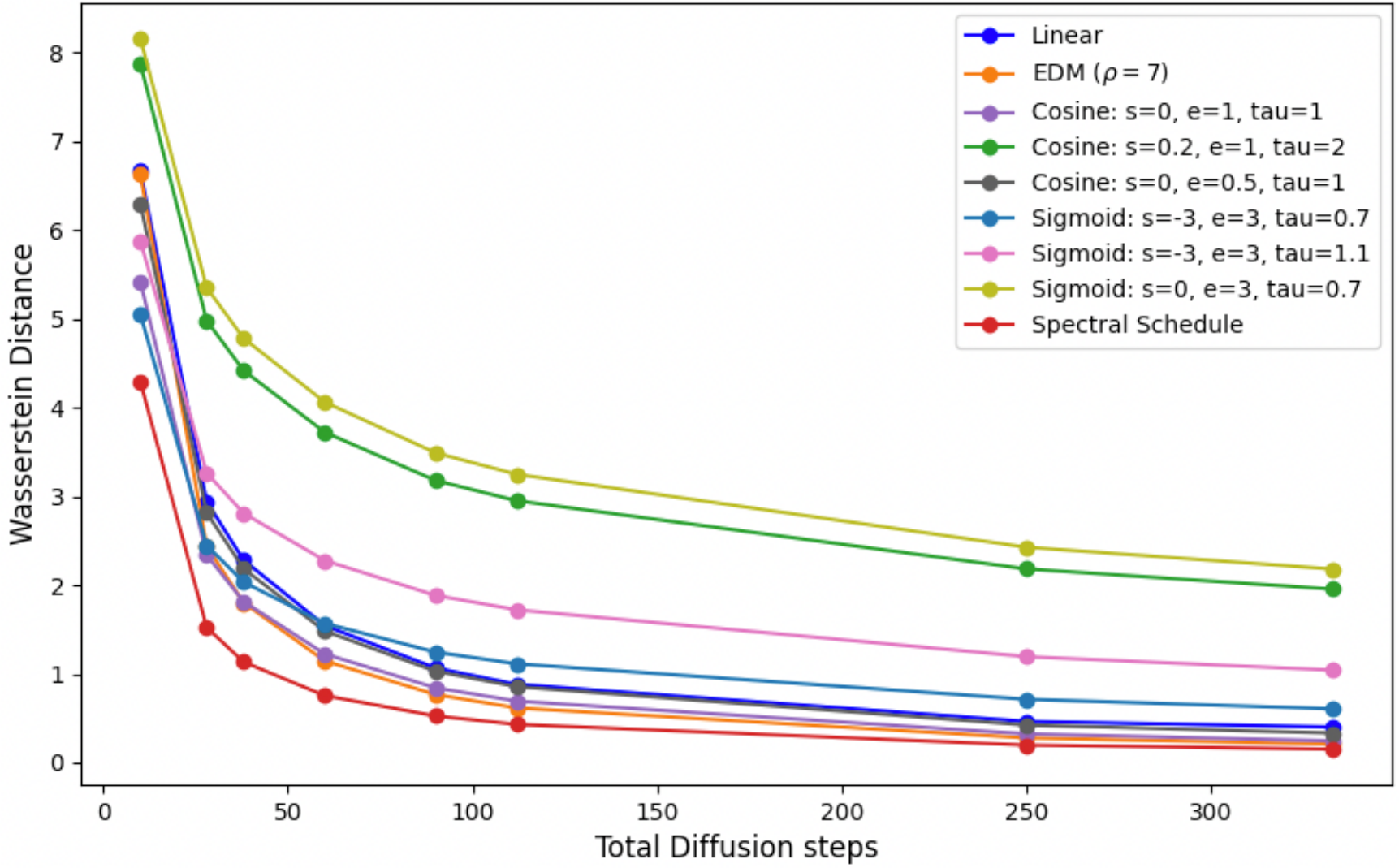}
\caption{Wasserstein-2 distance comparison across different numbers of diffusion steps and various heuristic noise schedules for the SC09 Gaussian dataset. The results highlight the optimality of the spectral recommendation, showing higher effectiveness in diffusion processes with fewer steps, where discretization error is more pronounced.}
\label{fig:Exp_2_wasserstein_distance_SC09}
\end{figure}

We also evaluate the Wasserstein-2 distance for the spectral recommendation on the SC09 Gaussian dataset, using a silence threshold of $th=0.1$. As shown in Figure \ref{fig:Exp_2_wasserstein_distance_SC09}, similar conclusions hold: the spectral schedule remains optimal, with Cosine $(0, 1, 1)$ and EDM heuristics most closely matching its structure and performance.

\newpage

\subsection{Analysis of Different Aspects}
\label{sec:scenario_2_Analysis_of_Different_Aspects}

The estimation of the covariance matrix, which is essential for finding the spectral recommendation for a real dataset, relies on the choice of two key parameters: $th$ and $d$.

The parameter $d$ represents the dimension of the signals and controls the frequency resolution, which affects the eigenvalues. A smaller $d$ may result in a more generalized eigenvalue spectrum, reducing accuracy by averaging energy across neighboring eigenvalues. In contrast, a larger $d$ improves the precision in capturing frequency details but increases computational time for both estimation and optimization.

\begin{figure}[H]
    \centering
    \begin{subfigure}{0.45\textwidth}
        \centering
        \includegraphics[width=\textwidth]{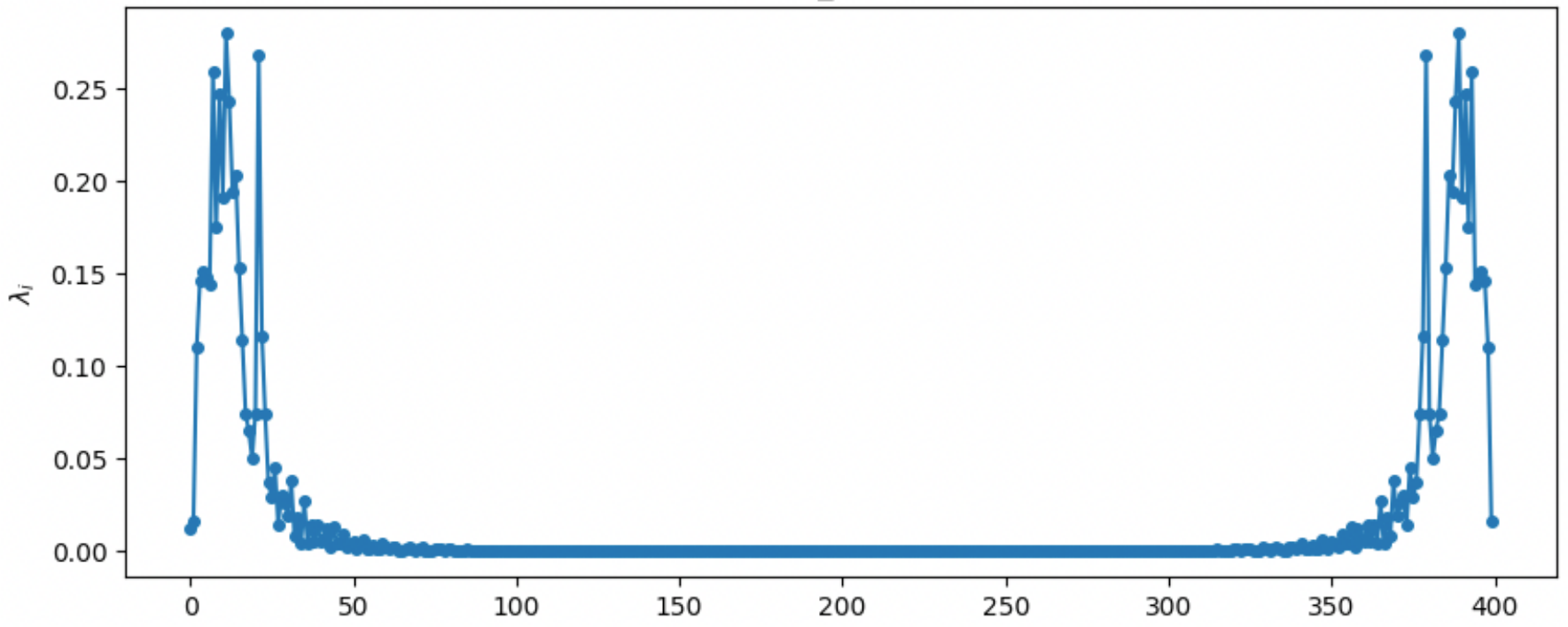}
        \caption{$d=400$}
        \label{subfig:Resolution_d_400}
    \end{subfigure}
    
      \hspace{1mm}
    \begin{subfigure}[b]{0.45\textwidth}
        \centering
     \includegraphics[width=\textwidth]{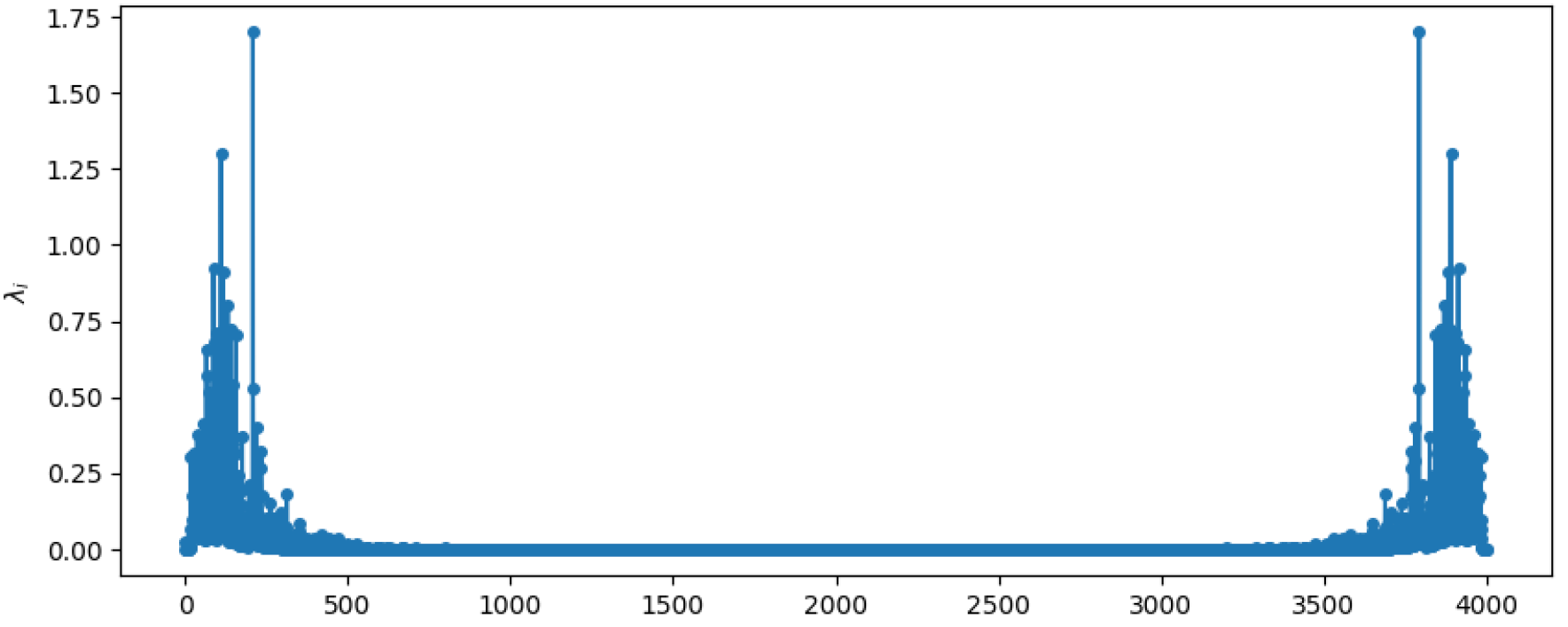}
        \caption{$d=4000$}
        \label{subfig:Resolution_d_4000}
    \end{subfigure}
    
   \hspace{1mm}
    \begin{subfigure}[b]{0.45\textwidth}
        \centering
        \includegraphics[width=\textwidth]{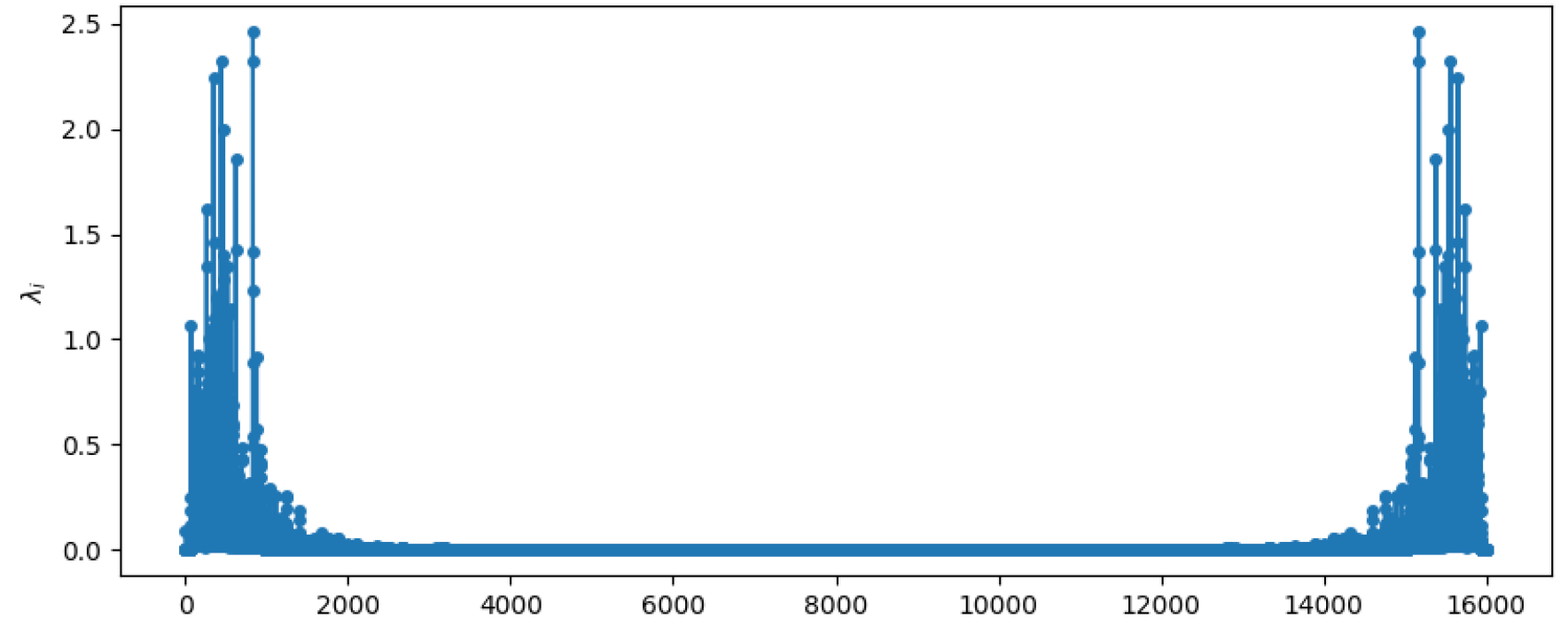}
        \caption{$d=16000$}
        \label{subfig:Resolution_d_16000}
    \end{subfigure}
\caption{Eigenvalues of the MUSIC dataset using $d=[400,4000,16000]$}
\label{fig:Eigenvalues_for_MUSIC_dataset_different_resolutions}
\end{figure}

Figure \ref{fig:Eigenvalues_for_MUSIC_dataset_different_resolutions} shows that as $d$ increases, the eigenvalue structure becomes more precise, with the maximum eigenvalues magnitude growing larger. Conversely, as $d$ decreases, the eigenvalue structure becomes more generalized, exhibiting a monotonic decrease, as discussed in Appendix  \ref{sec:Noise_schedule_loss_function}.

An additional consideration is the choice of the threshold value $th$. This threshold helps prevent the covariance matrix estimation from being overly influenced by silent regions in the signal, which are characterized by low $L_1$ energy. Adjusting $th$ affects both the covariance matrix values and the eigenvalues, i.e. $ C \cdot A v = C \cdot \lambda v$, thereby influencing the resulting noise schedule, as shown in Appendix \ref{sec:Relationship_Noise_Schedules_Eigenvalues}.

% To satisfy the Circulancy assumption, it is approximated as a circulant matrix using the approach discussed in Appendix \ref{sec:appendix_Circulant_from_toplitz}.  
% A detailed discussion on the influence of selecting $d$ and $th$ is provided in Appendix \ref{sec:scenario_2_Analysis_of_Different_Aspects}. }

\section{Supplementary Experiments for Empirical Distribution}\label{sec:Scenario_3_experiments}

In the following section, we present a more comprehensive analysis of the results for the general empirical distribution, where the Gaussianity assumption no longer holds and a neural denoiser is required. 

We begin by providing technical details of the denoiser architectures used for each dataset. For CIFAR-10, we employed a pretrained denoiser from \cite{karras2022elucidating}, trained with a continuous noise schedule. This approach facilitates a better comparison across noise schedules by reducing approximation error and ensuring alignment between training and testing conditions.

For the MUSIC dataset, we trained a model based on the architecture proposed in \cite{kong2020diffwave, benita2023diffar}, using a linear noise schedule with $T = 1000$ diffusion steps. Training was performed in an unconditional setting on raw waveforms with a batch size of $64$.
For the SC09 dataset, we adopted the architecture from \href{https://github.com/philsyn/DiffWave-Vocoder}{ GitHub} , and similarly trained it with a linear schedule of $T = 1000$ steps.

In the both cases, during synthesis, we mapped the tested noise schedule to the closest matching step in the linear schedule to enable use of the trained denoiser. While this approach is suboptimal compared to training with a continuous noise schedule, introducing additional approximation error, it reflects a practical scenario where a specific denoiser is given, and is therefore worthwhile to examine.

% For the MISUC dataset, we trained a model based on the architecture presented in \cite{kong2020diffwave, benita2023diffar}
% which employs a linear noise schedule with $T = 1000$ diffusion steps during training. we used a batch size of 64 and traind the denoiser in unconditional setting using the waveform representation itself.

% For the SC09 DATASET, we used the architectrue  from \href{https://github.com/philsyn/DiffWave-Vocoder}{DiffWave-Vocoder on GitHub} and traind it on the SC09 Dataset using a linear schedule with T=1000 steps as well.

% For the last two cases, during synthesis, given a tested noise scheule, we look for the most closer pair from the linear schedule for using the trained denoiser. While this approach is suboptimal compared to training with a continuous noise schedule—introducing additional approximation error, it reflects a practical scenario that may arise in real applications, and is therefore worthwhile to examine.

\begin{figure}[H]
    \centering
       \begin{subfigure}{0.45\textwidth}
        \centering
        \includegraphics[width=\columnwidth]{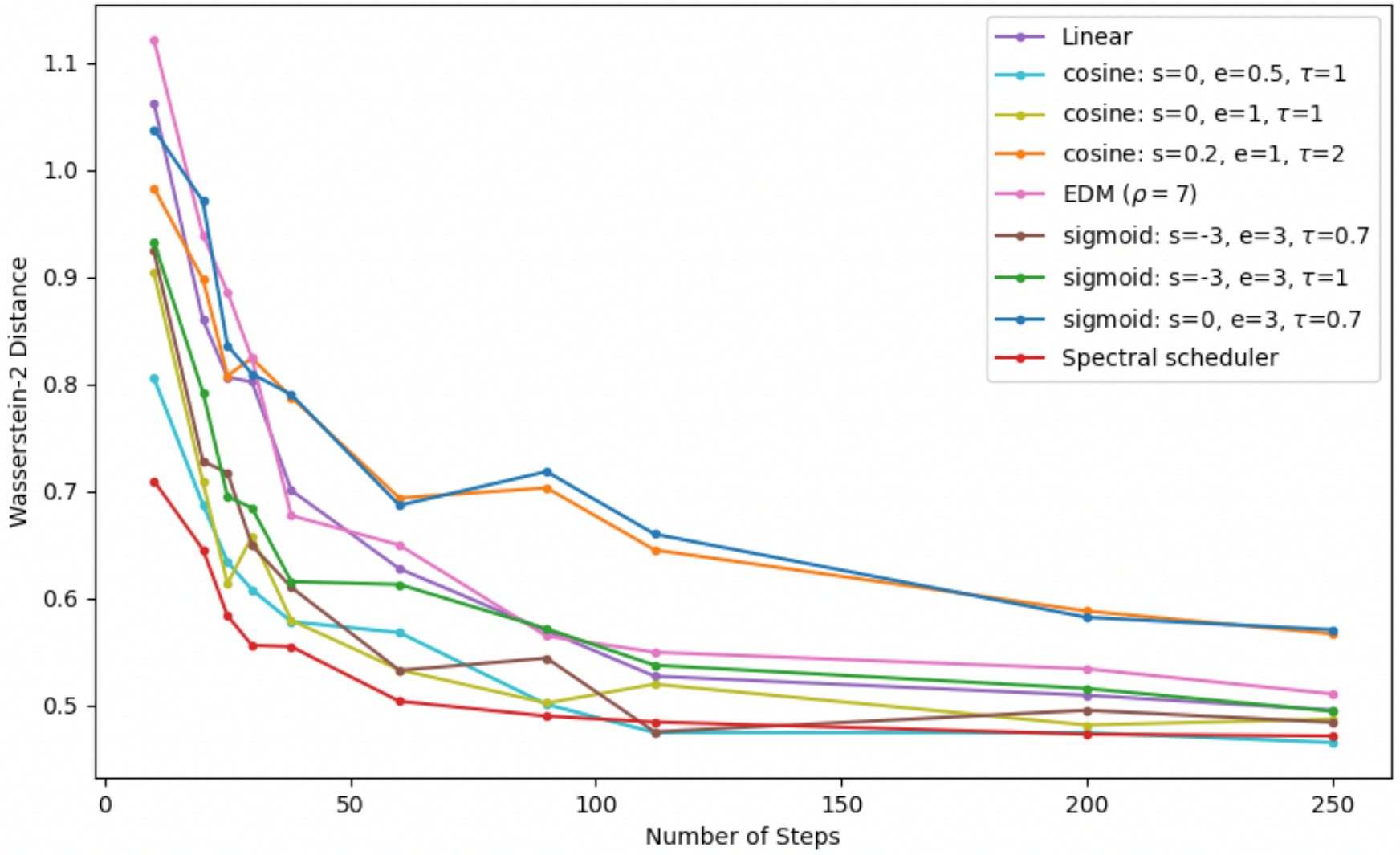}
        \caption{}
        \label{subfig:Exp_3_wasserstein_2_distance_app}
    \end{subfigure}
    \begin{subfigure}[b]{0.45\textwidth}
        \centering
     \includegraphics[width=\textwidth]{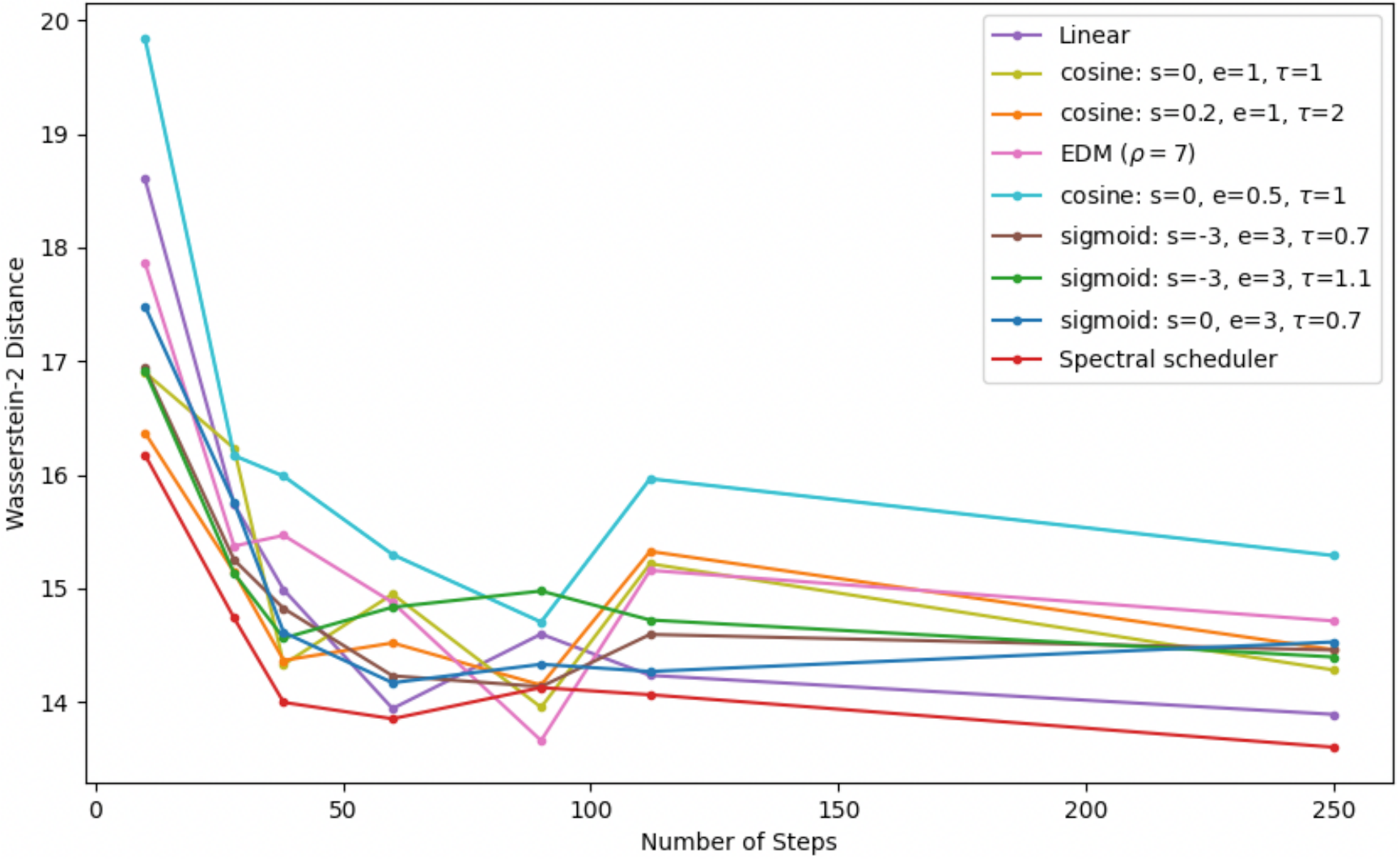}
        \caption{}
        \label{subfig:wasserstein_distence_Exp_3_SC09}
    \end{subfigure}
    \hspace{1mm}
    % \hfill
\caption{\emph{Wasserstein-2} distance  on the MUSIC-piano dataset (Figure \ref{subfig:Exp_3_wasserstein_2_distance_app}) and SC09 dataset (Figure \ref{subfig:wasserstein_distence_Exp_3_SC09}), evaluated over diffusion steps: $\{10, 20, 28, 38, 60, 90, 112, 250\}$.}
\label{fig:MUSIC_and_SC09_scheduling_results_app_exp_3}
\end{figure}

% \cite{philsyn_diffwave}  
 % and the \emph{Frobenius} 
% \subsection{CIFAR-10 Dataset}
We assess the performance of the spectral recommendation on the MUSIC and SC09 datasets using second-order metric. In particular, we compute the \emph{Wasserstein-2} distance between the covariance matrices estimated from $20,000$ generated samples at each diffusion step and those estimated from the real data. This allows us to evaluate how well each noise schedule preserves the core properties of the distribution during synthesis (Figure \ref{fig:MUSIC_and_SC09_scheduling_results_app_exp_3})

\begin{figure}[H]

    \centering
    \begin{subfigure}[b]{0.45\textwidth}
        \centering
        \includegraphics[width=\textwidth]{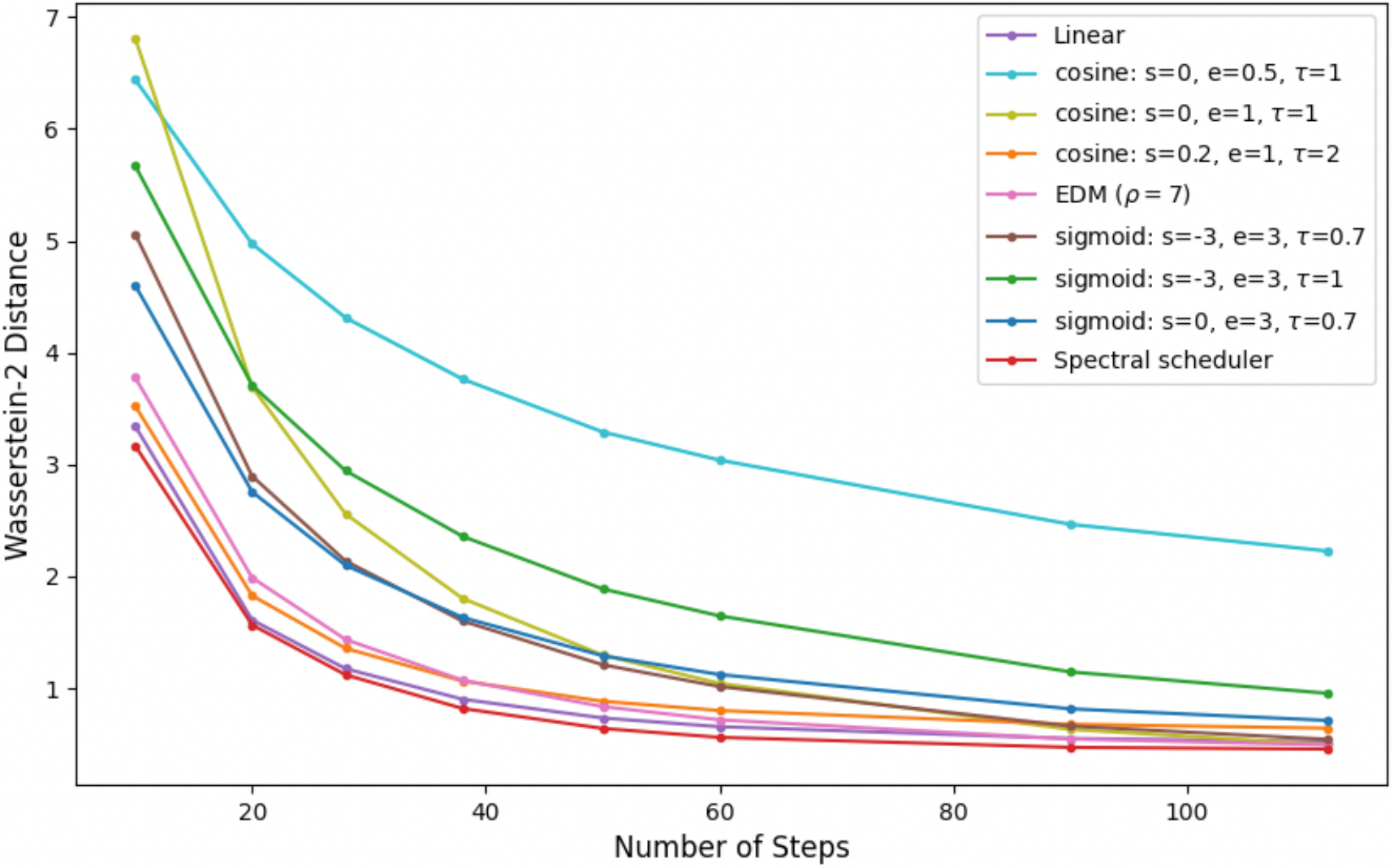}
        \caption{}
        \label{subfig:Exp_3_Wasserstein_Cifar_appendix}
    \end{subfigure}
        \begin{subfigure}[b]{0.45\textwidth}
        \centering
        \includegraphics[width=\textwidth]{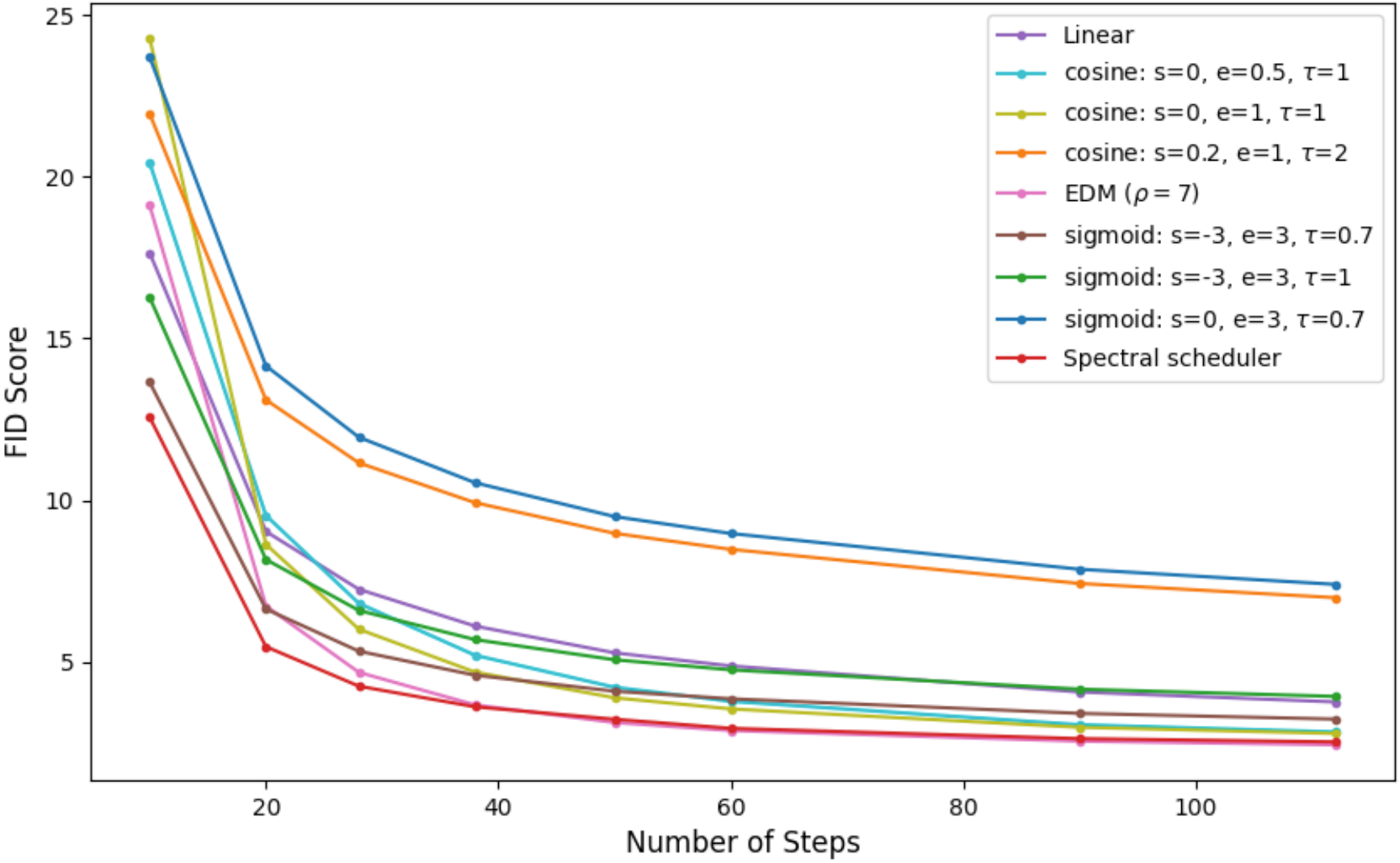}
        \caption{}
        \label{subfig:Exp_3_FID_Cifar_appendix}
    \end{subfigure}
    \caption{\emph{Wasserstein-2} distance (Figure~\ref{subfig:Exp_3_Wasserstein_Cifar_appendix}) and FID (Figure~\ref{subfig:Exp_3_FID_Cifar_appendix}) results on the CIFAR-10 dataset.}
\label{fig:CIFAR-10_scheduling_results_appendix_exp3}
\end{figure}

Additionally, for CIFAR-10, we use the standard FID metric to assess whether perceptual properties are preserved with our recommendation (see Figures~\ref{fig:CIFAR-10_scheduling_results_appendix_exp3}). The FID score is calculated following the method outlined in \cite{karras2022elucidating}, using $50,000$ samples for each computation, with seeds ranging from $0$ to $49,999$. As for FAD \cite{kilgour2018fr} calculation, it is less reliable for the MUSIC dataset due to its short frame lengths, and it provides limited insight for SC09, given its nature as a speech dataset.

% Additionally, for CIFAR-10, we use the standard FID metric to evaluate whether perceptual properties are maintained using our reccomendation. (Figures~\ref{fig:CIFAR-10_scheduling_results_appendix_exp3}). The callculation of the FID based on the code on \cite{karras2022elucidating} while using 50,000 samples, and using 0-49999 seeds. As for FAD calculation is less reliable for the MUSIC dataset due to its short frame lengths, and less informative for SC09, given its nature as a speech dataset

% \subsection{MUSIC Dataset}

The results support the relevance of the spectral recommendation for a broader range of distributions. The improved stability observed on CIFAR-10 may be attributed to factors such as the model architecture, the training procedure, or differences between discrete and continuous noise schedules. Interestingly, heuristics that achieve similar performance, such as EDM for FID on the CIFAR-10 dataset and cosine ($0$, $0.5$, $1$) schedules for the Wasserstein distance on the MUSIC dataset, tend to share structural similarities with the spectral recommendation. This suggests a connection between the structure of the noise schedule and overall performance, and points to the potential of the proposed method to inform the design of effective noise schedules.

\begin{figure}[H]
    \centering
        \begin{subfigure}[b]{0.45\textwidth}
        \centering
     \includegraphics[width=\textwidth]{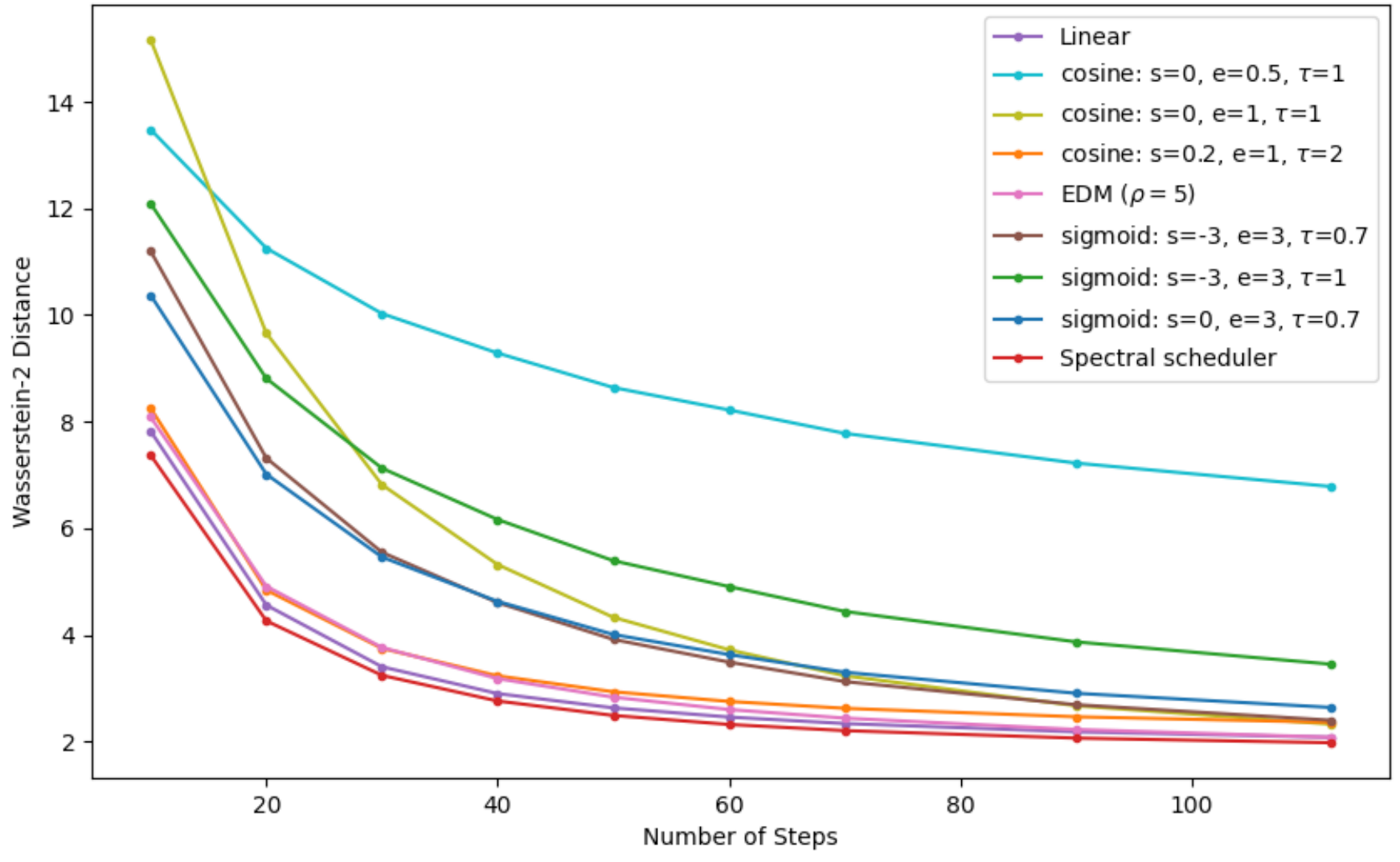}
        \caption{}
        \label{subfig:wasserstein_distence_AFHQv2}
    \end{subfigure}
           \begin{subfigure}{0.45\textwidth}
        \centering
        \includegraphics[width=\columnwidth]{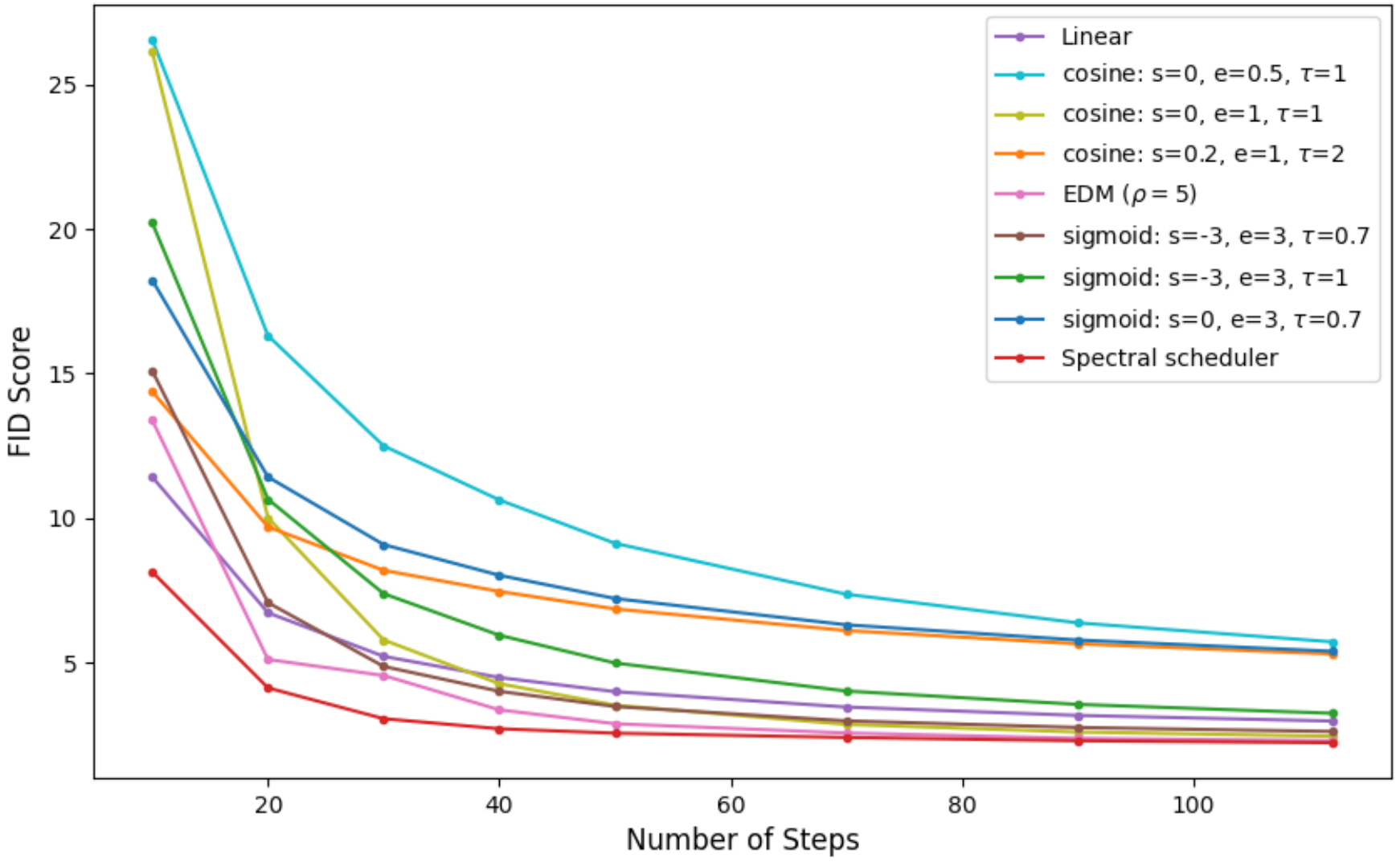}
        \caption{}
        \label{subfig:Exp_3_fid_AFHQv2}
    \end{subfigure}
    \hspace{1mm}
    % \hfill
        \caption{\emph{Wasserstein-2} distance (Figure~\ref{subfig:wasserstein_distence_AFHQv2}) and FID results (Figure~\ref{subfig:Exp_3_fid_AFHQv2}) on the AFHQv2 dataset, evaluated over diffusion steps: $\{10, 20, 30, 40, 50, 70, 90, 112\}$.}
\label{fig:AFHQv2_scheduling_results_app_exp_3}
\end{figure}

 % we conducted an additional experiment on the higher-resolution AFHQv2 dataset [1], which has a covariance matrix of size 12,288 × 12,288 (corresponding to 64×64×3 images). Applying our method, we derived the optimal noise schedule and evaluated its performance on this real-world dataset using the pretrained denoiser from [2]. Tables 1 and 2 show a comparison of Wasserstein distance and FID scores between our optimal noise schedule and various heuristic baselines on AFHQv2. Tables 1 and 2 demonstrate that the spectral recommendation outperforms all other heuristics across both metrics. As noted in the paper, the performance gap is especially pronounced at low number of diffusion steps, where discretization error tends to be higher.
 % A heuristic that is closest to the spectral schedule in both structure and FID performance is EDM with ρ = 5 (a figure illustrating this will be included in the paper, as it cannot be shown here). This differs from the results observed on CIFAR-10, MUSIC and SC09, highlighting the unique spectral characteristics of each dataset. 

We conducted an additional experiment on the higher-resolution AFHQv2 dataset \cite{choi2020starganv2}, with a covariance matrix of size $12,288 \times 12,288$ (corresponding to $64\times64\times3$ images). Figure~\ref{fig:AFHQv2_scheduling_results_app_exp_3} presents Wasserstein distance and FID scores for the spectral recommendation and heuristic baselines, demonstrating the superior performance of the spectral schedule. In this case, EDM with $\rho = 5$ most closely approximates the spectral schedule in both structure and FID, reflecting dataset-specific adaptation.

\textbf{Experiment statistical significance}

We compute the FID and Wasserstein-2 distance for CIFAR-10 using three independent runs, each consisting of $50{,}000$ samples generated with different random seeds ([$0$–$4999$], [$5000$–$9999$], and [$10000$–$14999$]), to capture random variability. The results in Tables  \ref{tab:wasserstein_score} and \ref{tab:fid_CI} report the Wasserstein-2 distance and FID (respectively), along with a $95\%$  confidence interval, calculated using the standard error and the t-distribution.

% $\bar{x} \pm t_{\alpha/2,\,n-1} \cdot \frac{s}{\sqrt{n}}$

\begin{table}[htbp]
\centering
\scriptsize
\caption{Wasserstein scores (mean $\pm$ 95\% CI) across selected diffusion steps for different noise schedules.}
\begin{tabular}{lcccccc}
\toprule
 & $\mathbf{10}$ & $\mathbf{20}$ & $\mathbf{28}$ & $\mathbf{38}$ & $\mathbf{50}$ & $\mathbf{90}$ \\
\midrule
Linear & $ \underline{3.28 \pm 0.28}$ & $\underline{1.61 \pm 0.01}$ & $\underline{1.17 \pm 0.01}$ & $\underline{0.90 \pm 0.02}$ & $\underline{0.73 \pm 0.02}$ & $\underline{0.54 \pm 0.04}$ \\
EDM & $3.77 \pm 0.02$ & $1.99 \pm 0.01$ & $1.43 \pm 0.01$ & $1.07 \pm 0.01$ & $0.83 \pm 0.02$ & $0.53 \pm 0.04$ \\
Cosine ($0$, $1$, $1$) & $6.81 \pm 0.01$ & $3.69 \pm 0.01$ & $2.56 \pm 0.01$ & $1.80 \pm 0.01$ & $1.30 \pm 0.01$ & $0.63 \pm 0.01$ \\
Cosine ($0.2$, $1$, $2$) & $3.52 \pm 0.01$ & $1.83 \pm 0.01$ & $1.35 \pm 0.01$ & $1.06 \pm 0.01$ & $0.88 \pm 0.02$ & $0.67 \pm 0.03$ \\
Cosine ($0$, $0.5$, $1$) & $6.44 \pm 0.01$ & $4.97 \pm 0.02$ & $4.31 \pm 0.02$ & $3.76 \pm 0.01$ & $3.29 \pm 0.01$ & $2.46 \pm 0.01$ \\
Sigmoid ($0$, $3$, $0.7$) & $4.60 \pm 0.01$ & $2.75 \pm 0.01$ & $2.10 \pm 0.01$ & $1.63 \pm 0.01$ & $1.29 \pm 0.01$ & $0.81 \pm 0.02$ \\
Sigmoid ($-3$, $3$, $1$) & $5.67 \pm 0.01$ & $3.71 \pm 0.01$ & $2.94 \pm 0.01$ & $2.36 \pm 0.01$ & $1.89 \pm 0.01$ & $1.15 \pm 0.00$ \\
Sigmoid ($-3$, $3$, $0.7$) & $5.06 \pm 0.01$ & $2.89 \pm 0.01$ & $2.13 \pm 0.01$ & $1.60 \pm 0.01$ & $1.21 \pm 0.00$ & $0.66 \pm 0.02$ \\
Spectral & $\mathbf{3.21 \pm 0.24}$ & $\mathbf{1.59 \pm 0.05}$ & $\mathbf{1.12 \pm 0.01}$ & $\mathbf{0.81 \pm 0.02}$ & $\mathbf{0.63 \pm 0.02}$ & $\mathbf{0.46 \pm 0.04}$ \\
\bottomrule
\end{tabular}
\label{tab:wasserstein_score}
\end{table}

\begin{table}[htbp]
\centering
\scriptsize
\caption{FID scores (mean $\pm$ 95\% CI) across selected diffusion steps for different noise schedules.}
\begin{tabular}{lcccccc}
\toprule
 & $\mathbf{10}$ & $\mathbf{20}$ & $\mathbf{28}$ & $\mathbf{38}$ & $\mathbf{50}$ & $\mathbf{90}$ \\
\midrule
Linear & $17.59 \pm 0.26$ & $9.00 \pm 0.17$ & $7.19 \pm 0.25$ & $6.04 \pm 0.26$ & $5.22 \pm 0.25$ & $4.00 \pm 0.26$ \\
EDM & $19.04 \pm 0.25$ & $6.70 \pm 0.17$ & $\underline{4.64 \pm 0.19}$ & $\underline{3.63 \pm 0.22}$ & $\mathbf{3.09 \pm 0.21}$ & $\mathbf{2.52 \pm 0.19}$ \\
Cosine ($0$, $1$, $1$) & $24.31 \pm 0.14$ & $8.63 \pm 0.16$ & $6.00 \pm 0.16$ & $4.65 \pm 0.18$ & $3.86 \pm 0.20$ & $2.94 \pm 0.21$ \\
Cosine ($0.2$, $1$, $2$) & $21.92 \pm 0.29$ & $13.06 \pm 0.21$ & $11.10 \pm 0.20$ & $9.85 \pm 0.25$ & $8.90 \pm 0.25$ & $7.33 \pm 0.29$ \\
Cosine ($0$, $0.5$, $1$) & $20.49 \pm 0.27$ & $9.60 \pm 0.21$ & $6.85 \pm 0.07$ & $5.23 \pm 0.12$ & $4.22 \pm 0.14$ & $3.05 \pm 0.16$ \\
Sigmoid ($0$, $3$, $0.7$) & $23.66 \pm 0.17$ & $14.10 \pm 0.31$ & $11.90 \pm 0.23$ & $10.48 \pm 0.26$ & $9.44 \pm 0.27$ & $7.78 \pm 0.29$ \\
Sigmoid ($-3$, $3$, $1$) & $16.28 \pm 0.06$ & $8.16 \pm 0.23$ & $6.56 \pm 0.23$ & $5.66 \pm 0.21$ & $5.03 \pm 0.22$ & $4.12 \pm 0.24$ \\
Sigmoid ($-3$, $3$, $0.7$) & $\underline{13.67 \pm 0.06}$ & $\underline{6.60 \pm 0.19}$ & $5.31 \pm 0.20$ & $4.55 \pm 0.24$ & $4.05 \pm 0.25$ & $3.36 \pm 0.22$ \\
Spectral & $\mathbf{12.50 \pm 0.22}$ & $\mathbf{5.44 \pm 0.20}$ & $\mathbf{4.21 \pm 0.21}$ & $\mathbf{3.56 \pm 0.24}$ & $\underline{3.20 \pm 0.23}$ & $\underline{2.60 \pm 0.21}$ \\
\bottomrule
\end{tabular}
\label{tab:fid_CI}
\end{table}

The results presented in Tables  \ref{tab:wasserstein_score} and \ref{tab:fid_CI} align with the experimental observations in Figure \ref{fig:CIFAR-10_scheduling_results_appendix_exp3}, supporting the spectral recommendation and its relation to alternative noise schedules. The optimality with respect to the Wasserstein loss remains robust, while for the FID metric, as the number of diffusion steps increases and discretization error becomes less significant, schedules such as EDM achieve comparable performance.

\newpage
\textbf{Experimental result reproducibility}

To facilitate easier reproduction of our results, we provide examples of the derived spectral schedules obtained for the CIFAR-10 dataset by minimizing the Wasserstein-2 loss for [10, 20, 28] diffusion steps. The values are rounded to four decimal places. 

\textbf{10 Diffusion Steps:} 
   $[0.9999,\ 0.9899,\ 0.9177,\ 0.6796,\ 0.3398,\ 0.1271,\ 0.0429,\ 0.0135,\ 0.0030,\ 0.00004]$
   
\textbf{20 Diffusion Steps:}
    $[0.9999,\ 0.9981,\ 0.9914,\ 0.9735,\ 0.9340,\ 0.8591,\ 0.7389,\ 0.5814,\ 0.4162,\ 0.2751,$\\
    $0.1724,\ 0.1048,\ 0.0627,\ 0.0371,\ 0.0216,\ 0.0121,\ 0.0063,\ 0.0028,\ 0.0008,\ 0.00004]$
    
\textbf{28 Diffusion Steps:}
    $[0.9999,\ 0.9989,\ 0.9961,\ 0.9899,\ 0.9778,\ 0.9562,\ 0.9204,\ 0.8656,\ 0.7882,\ 0.6889,$\\
    $0.5747,\ 0.4572,\ 0.3487,\ 0.2570,\ 0.1849,\ 0.1309,\ 0.0917,\ 0.0638,\ 0.0442,\ 0.0304,$\\
    $0.0207,\ 0.0138,\ 0.0089,\ 0.0055,\ 0.0031,\ 0.0015,\ 0.0005,\ 0.00004]$

\newpage
\section{Further Discussion}\label{sec:appendix_further_discussion}

\subsection{Relationship Between Noise Schedules and Eigenvalues}\label{sec:Relationship_Noise_Schedules_Eigenvalues}

\begin{figure}[h]
    \centering      
    \begin{subfigure}{0.45\textwidth}
        \centering
        \includegraphics[width=\textwidth]{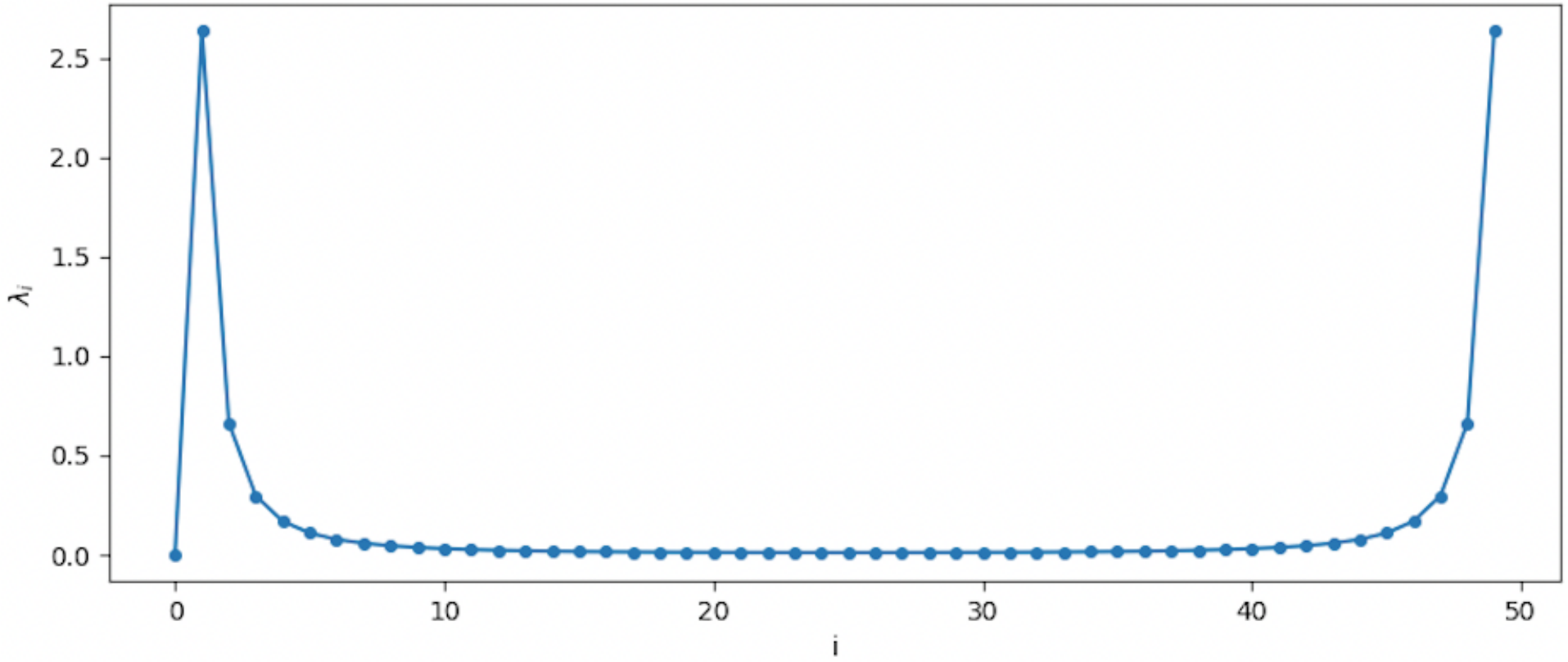}
        \caption{}
        \label{subfig:scenario_1_lambda_0_shorter}
    \end{subfigure}
   \hfill
   
    \begin{subfigure}{0.45\textwidth}
    \centering
    \includegraphics[width=\textwidth] {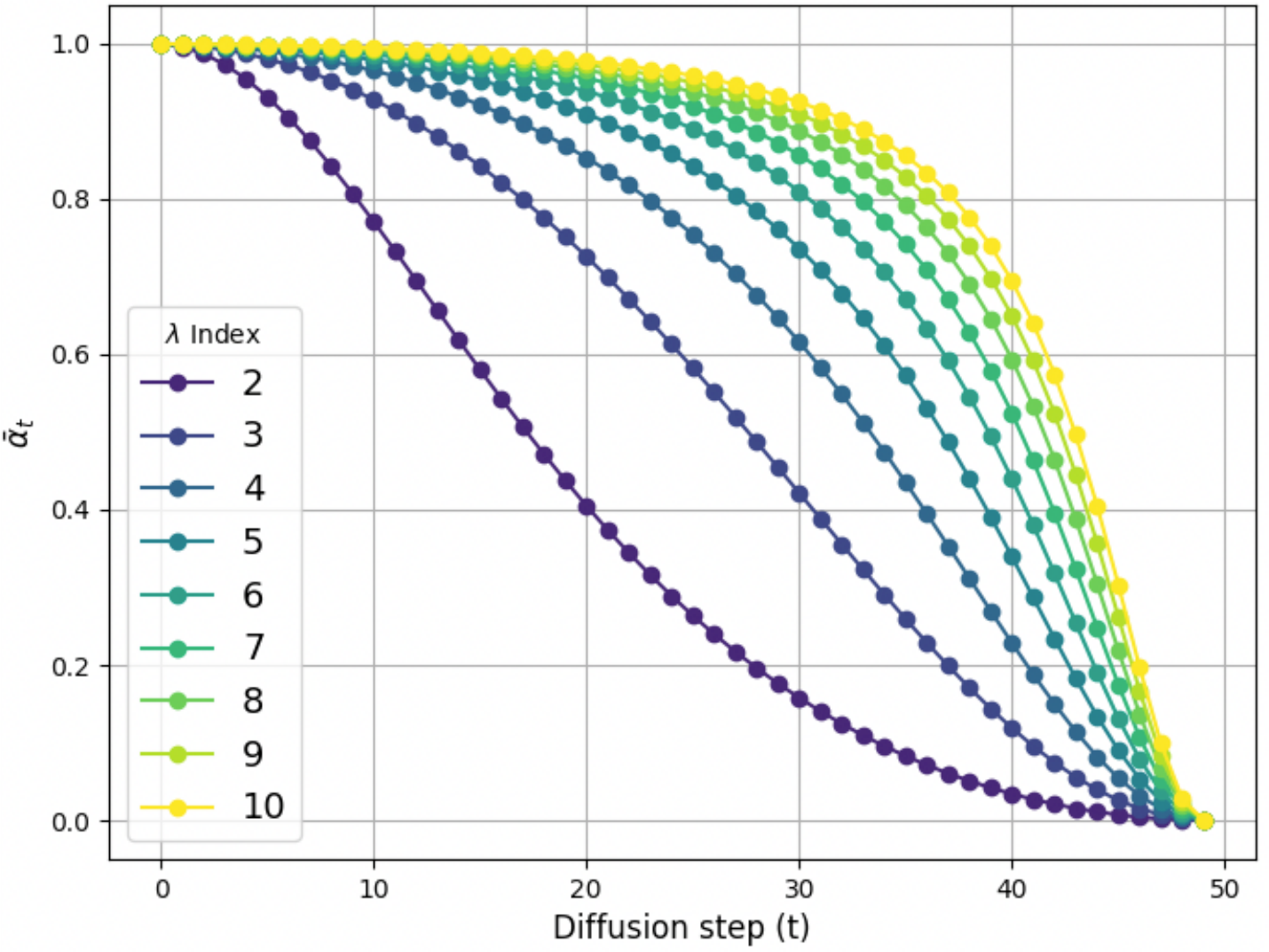}
\caption{}
\label{subfig:Each_eigenvalue_at_a_time}
\end{subfigure}
\caption{Figure \ref{subfig:scenario_1_lambda_0_shorter} shows the eigenvalues for the covariance matrix from Section \ref{subsec:Scenario_1}. Figure \ref{subfig:Each_eigenvalue_at_a_time} presents the results of solving the optimization problem for each eigenvalue individually, with the contributions from the other eigenvalues set to zero.}
\end{figure}

To explore the relationship between the optimal spectral noise schedule and the data characteristics, we solved the optimization problem
for each eigenvalue individually, with the contributions from the other eigenvalues set to zero.
% by focusing on one eigenvalue at a time, while setting the contributions of the other eigenvalues to zero. 
Using the eigenvalues of the covariance matrix from \ref{subsec:Scenario_1}, Figure \ref{subfig:scenario_1_lambda_0_shorter} shows these eigenvalues, while \ref{subfig:Each_eigenvalue_at_a_time} presents the optimal solutions for $50$ diffusion steps, computed using the \emph{Wasserstein-2} distance in the optimization problem. Each solution corresponds to a single eigenvalue (considering only positions $2$ to $10$ for clarity)\footnote{The first eigenvalue is excluded as it disrupts the monotonicity.}

It can be observed from Figure \ref{subfig:Each_eigenvalue_at_a_time} that the solution becomes more concave as the magnitude of the eigenvalue decreases (yellow) and more convex as the magnitude increases (blue). Notably, this behavior is determined by the magnitude of the eigenvalues ($\{ \lambda_i \}_{i=1}^{d}$) rather than their indices ($i$), as the objective functions are independent of the index itself.

 When the eigenvalues, or equivalently the DFT coefficients, decrease monotonically, a direct relationship emerges between the eigenvalue magnitude and its corresponding frequency (for example, the $1/f$ behavior observed in speech \citep{voss1975f}). In such cases, the first eigenvalues
correspond to the low frequencies, having larger amplitudes,
while the last correspond to high frequencies and smaller
amplitudes. This pattern, along, with the previous observations \footnote{Low magnitude eigenvalues relate with concave schedule and high magnitude eigenvalues correspond to convex schedule.}, aligns with the well-known coarse-to-fine signal construction behavior of diffusion models.\footnote{Higher-frequency components are emphasized by allocating more steps toward the end of the diffusion process, while lower-frequency components are emphasized earlier.}

\begin{figure}[H]
    \centering      
    \begin{subfigure}{0.45\textwidth}
        \centering
        \includegraphics[width=\textwidth]{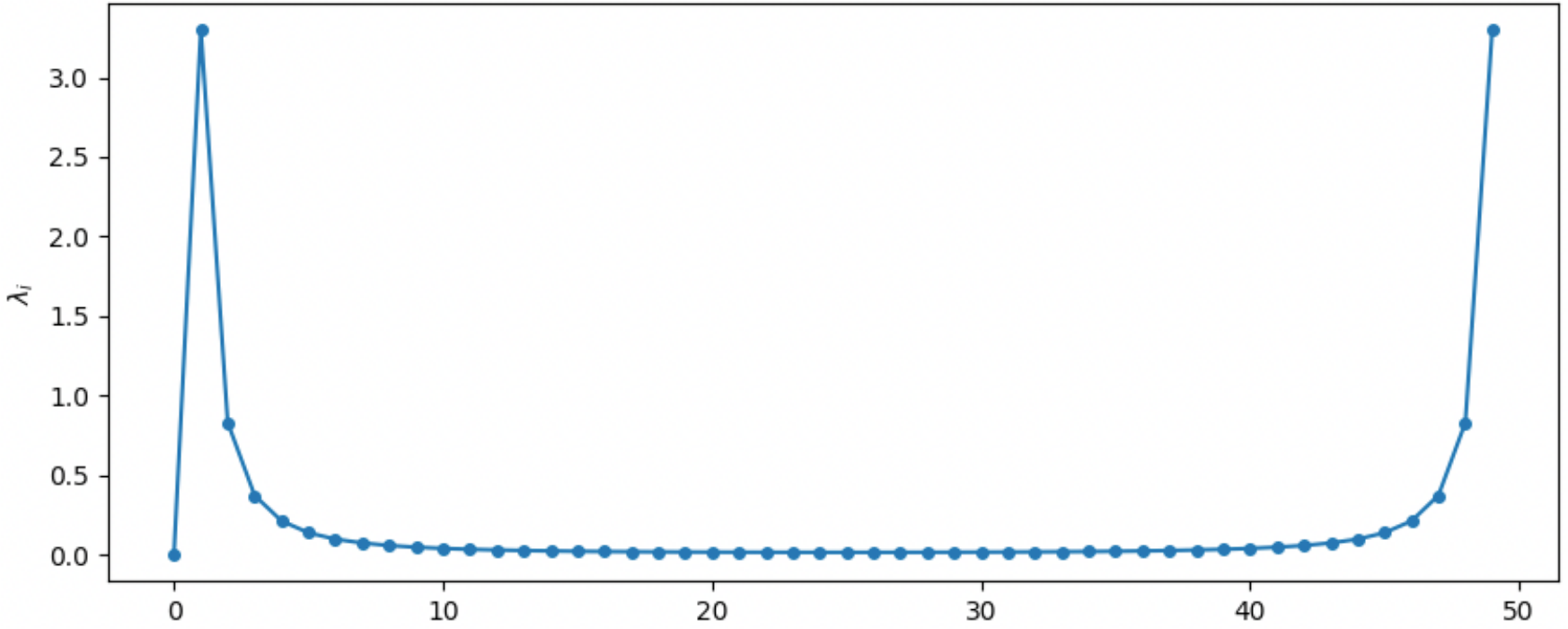}
        \caption{}
        \label{subfig:cov_1_original_lambda}
    \end{subfigure}
   \hfill
    \begin{subfigure}{0.45\textwidth}
    \centering
    \includegraphics[width=\textwidth]{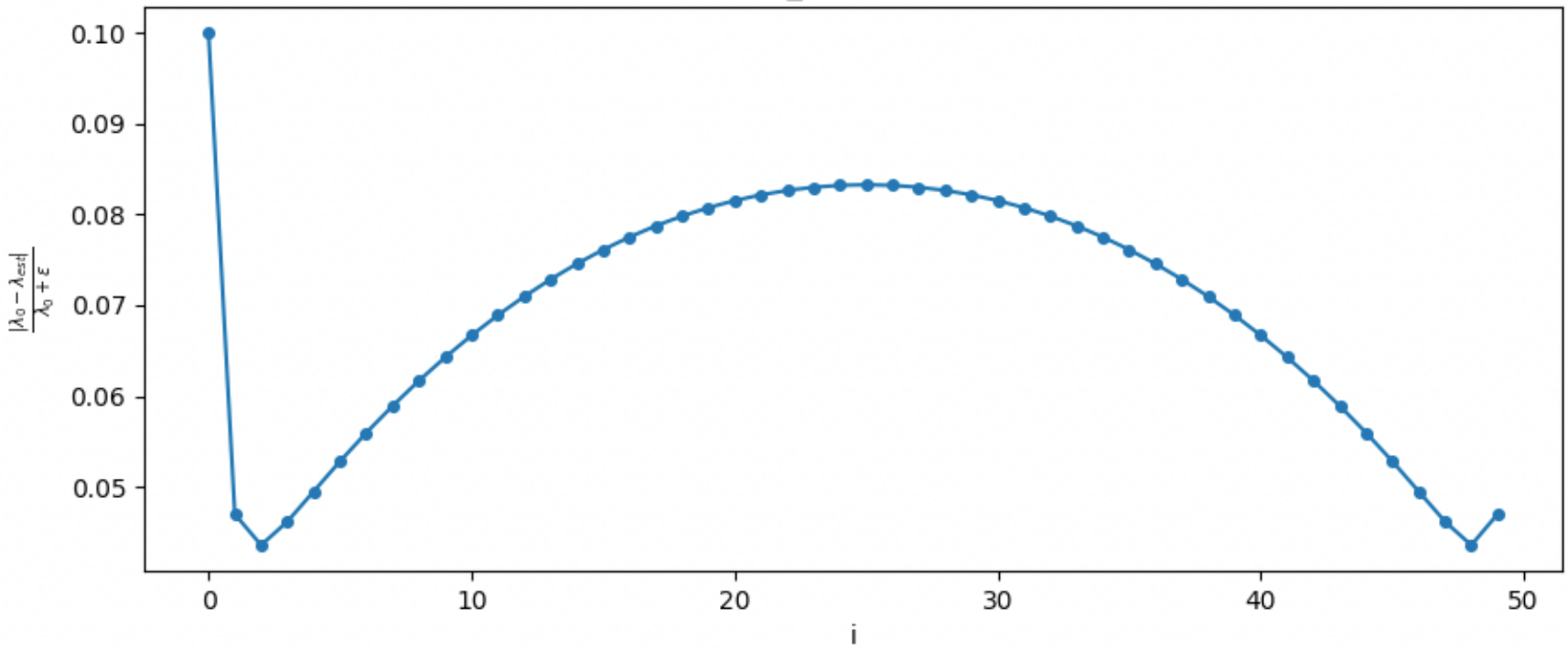}
\caption{}
\label{fig:Relative_loss_Diff_lambda_0}
\end{subfigure}
    \hspace{1mm}
    \centering
    \begin{subfigure}{0.45\textwidth}
        \centering
        \includegraphics[width=\textwidth]{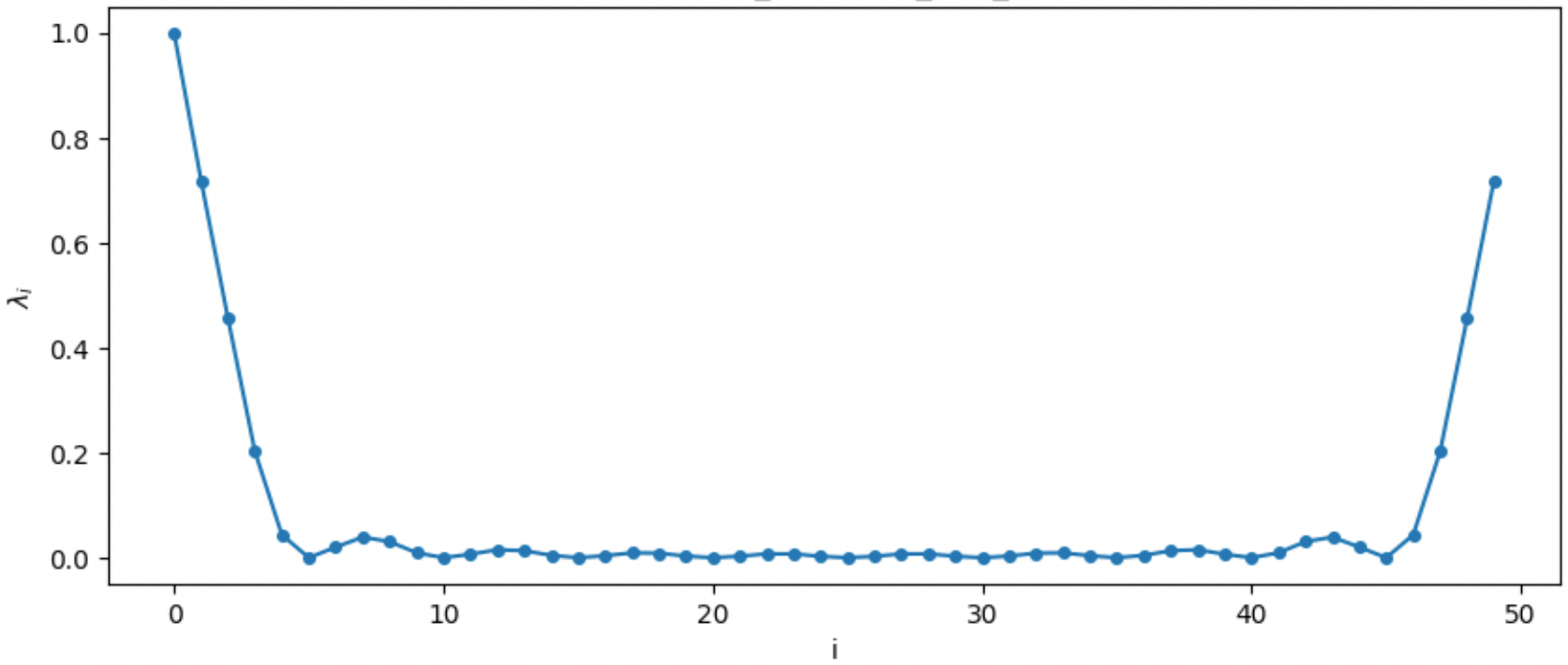}
        \caption{}
\label{subfig:original_lambda_zoom_in_cov_5}
    \end{subfigure}
    % \hspace{1mm}
    \hfill
    \begin{subfigure}{0.45\textwidth}
        \centering
        \includegraphics[width=\textwidth]{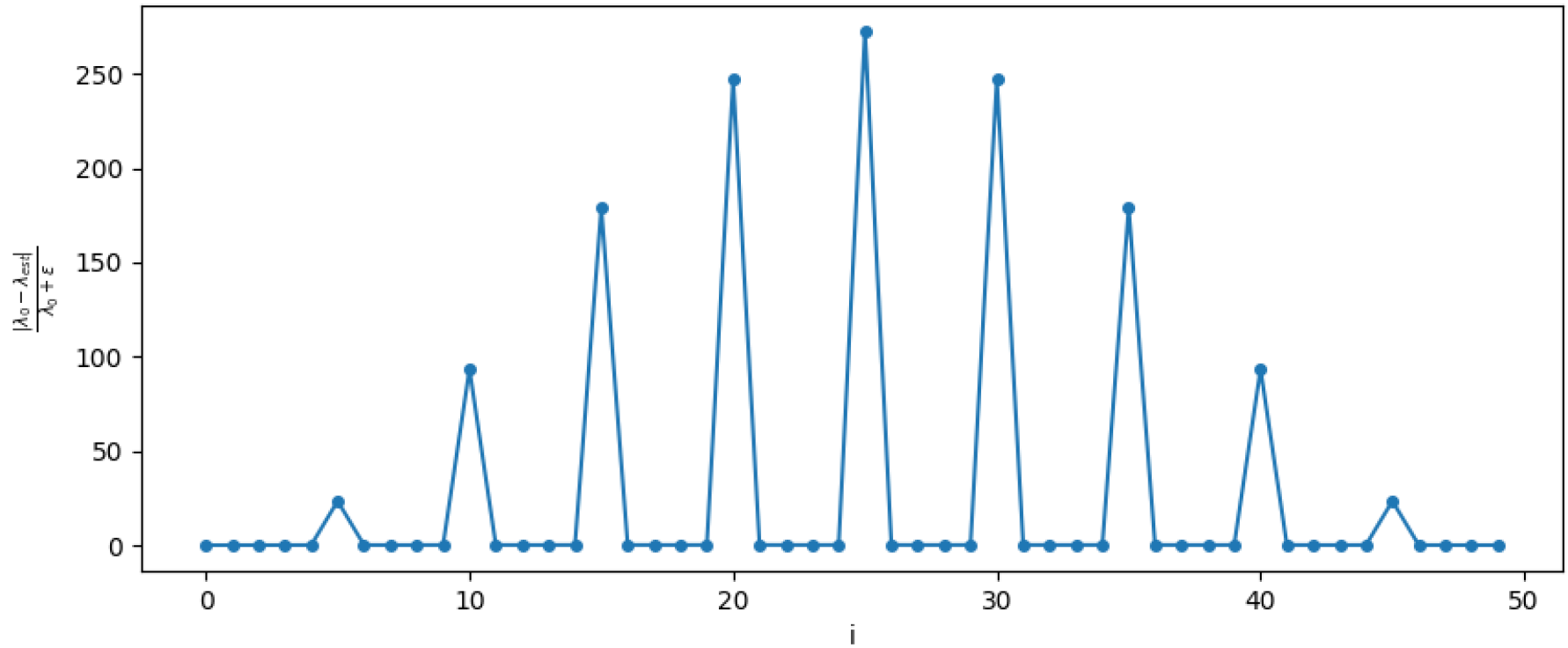}
        \caption{}
        \label{subfig:cov_5_relative_loss}
    \end{subfigure}
\caption{Figures \ref{subfig:cov_1_original_lambda} and \ref{subfig:original_lambda_zoom_in_cov_5} present the eigenvalues of matrices \ref{subfig:cov_mat_1_none} and \ref{subfig:cov_mat_5_mul_none}, respectively. Figures \ref{fig:Relative_loss_Diff_lambda_0} and \ref{subfig:cov_5_relative_loss} illustrate the relative error of the spectral recommendation, using $60$ diffusion steps and the \emph{Wasserstein-2} distance. Notably, larger eigenvalues exhibit smaller relative errors. In Figures \ref{fig:Relative_loss_Diff_lambda_0} and \ref{subfig:original_lambda_zoom_in_cov_5}, the first elements were manually chosen, as they originally had extreme values ($1000$ and $10$ respectively), and we aimed to keep the figure within a reasonable scale.}
    \label{fig:cov_1_5_relative_loss}
\end{figure}

Interestingly, by examining the spectral recommendation from Figure \ref{fig:Exp_1_Spectral_reccomandation_wasserstein}, it closely resembles the solutions obtained by emphasizing the highest eigenvalue (\ref{subfig:Each_eigenvalue_at_a_time}). This suggests that using the \emph{Wasserstein-2 loss} tends to favor larger magnitude eigenvalues. This behavior is also reflected in the relative error,
$(|\lambda_i-\lambda_{est}|)/(\lambda_i)$, shown in Figure \ref{fig:cov_1_5_relative_loss}, where larger eigenvalues exhibit smaller relative errors.

The influence of the eigenvalue magnitude, especially the dominant ones, on the resulting schedule is further illustrated through additional examples. Figure \ref{subfig:cov_mat_1_mul_20} displays the covariance matrix from Figure \ref{subfig:cov_mat_1_none}, scaled by a factor of $C = 20$, which amplifies the dominant eigenvalues, as shown by the relation $C \cdot A v = C \cdot \lambda v$. Consequently, the spectral recommendation in Figure \ref{subfig:alpha_bar_cov_mat_1_mul_20} appears more convex than in Figure \ref{subfig:alpha_bar_mat_1_None}. A similar trend is observed in Figure \ref{fig:Exp_2_Spectral_scheduler_comparison_SC09_th_0_1_th_0_05}, where the spectral recommendation for $th = 0.05$ exhibits a more concave shape compared to $th = 0.1$. This relationship open up a possibility of designing loss functions which focus on specific frequency ranges of interest.  Further discussion is provided in \ref{sec:Noise_schedule_loss_function}.

% As mentioned in \ref{sec:migrating_to_the_spectral_domain}, assuming circularity, the eigenvalues correspond to the DFT coefficients of first row of $\bSigma_0$.
% When the eigenvalues, or equivalently the coefficients of the Discrete Fourier Transform, decrease monotonically, there is a direct relationship between the magnitude of the eigenvalue and its corresponding frequency (for example, $1/f$ behavior observed in speech \citep{voss1975f}). 

Note: We used the \emph{Wasserstein-2} loss. However, alternative measures, such as \emph{KL divergence}, could also yield similar results.
% with the well-known coarse-to-fine signa

\newpage
\subsection{Temporal Dynamics of the Diffusion Process}
\label{sec:Dynamics_throughout_the_diffusion_process}

Our analysis has focused so far on the final distribution resulting from the diffusion process, as described by \eqref{eq:D_1,D_2}. However, the analytical formulation also enables the examination of phenomena occurring throughout the diffusion process. In particular, Equation \eqref{eq:itermidate_step_diffusion_process} describes each intermediate state $\ve_s$ as a function of the preceding noise schedule components and the initial noise $\ve_0$. 
This enables the study of spectral properties  and  dynamics across all diffusion steps, rather than being limited to the final output.

\begin{figure}[H]
    \centering
    \begin{subfigure}[b]{0.3\textwidth}
        \centering
     \includegraphics[width=\textwidth]{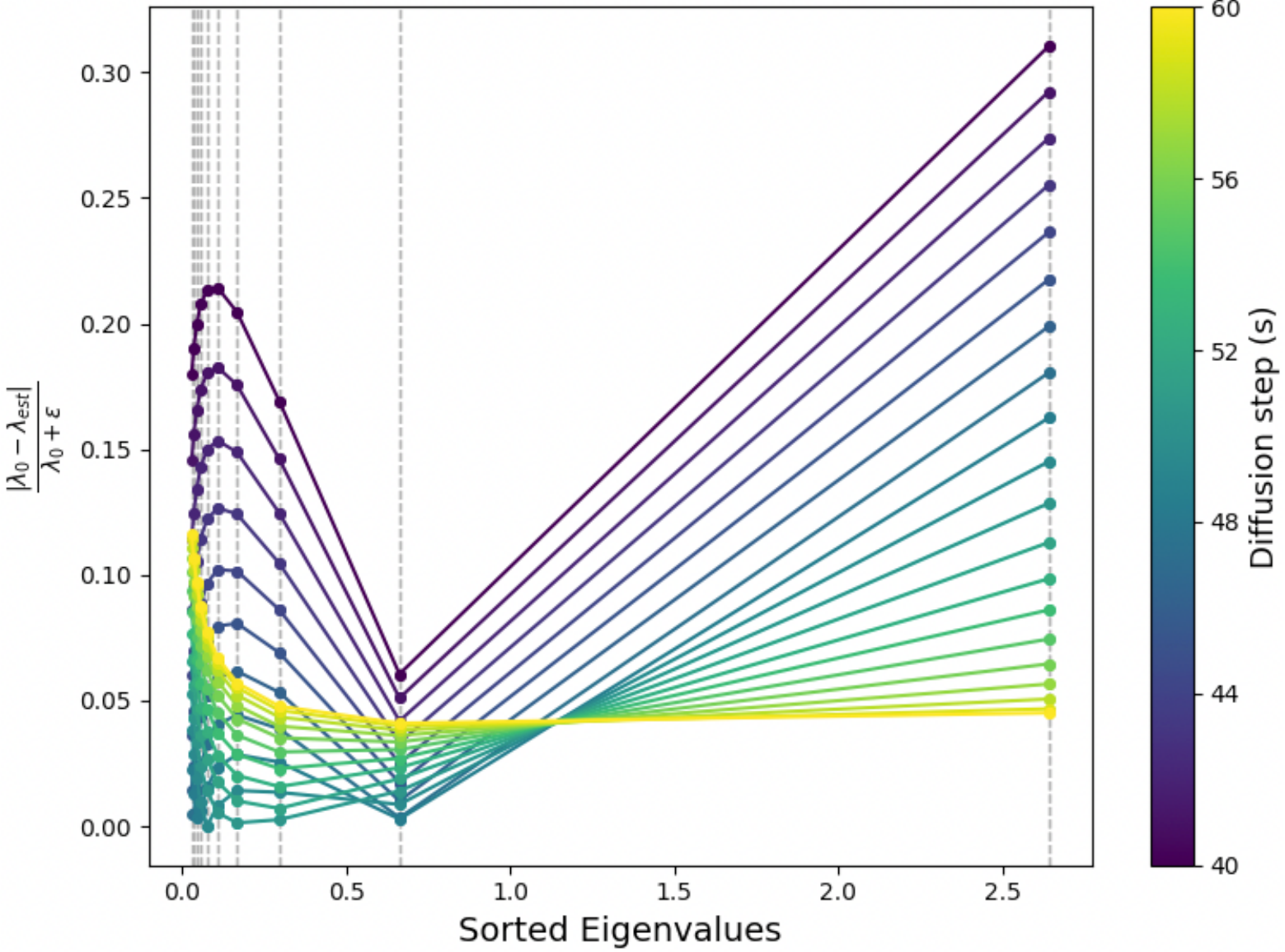}
        % \caption{Cosine ($s=0$, $e=1$, $\tau=1$)}
        \caption{}
        \label{subfig:temporal_relative_error_cosine}
    \end{subfigure}
    \begin{subfigure}{0.3\textwidth}
        \centering
        \includegraphics[width=\textwidth]{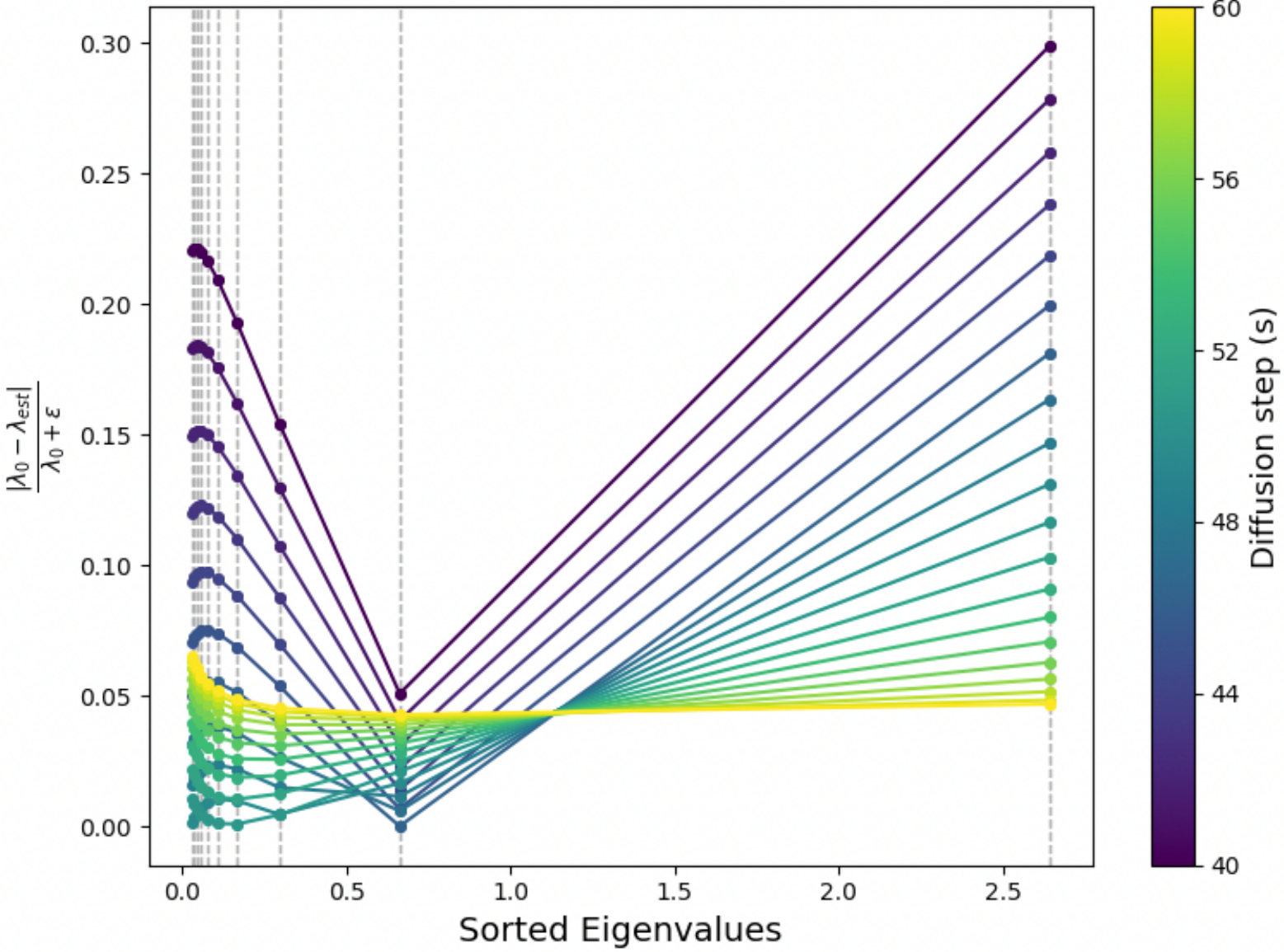}
        % \caption{Spectral}
        \caption{}
       \label{subfig:temporal_relative_error_spectral}
    \end{subfigure}
\caption{The relative errors of the the Cosine ($s=0$, $e=1$, $\tau=1$) (\ref{subfig:temporal_relative_error_cosine})  and the spectral (\ref{subfig:temporal_relative_error_spectral}) schedules for the 10 largest eigenvalues (sorted, largest on the right) over the final 20 steps of a 60-step diffusion process. }
\label{fig:temporal_relative_error}
\end{figure}

Figures \ref{subfig:temporal_relative_error_cosine} and \ref{subfig:temporal_relative_error_spectral} show the relative error $|\lambda_i - \lambda_{\text{est}}| / \lambda_i + \epsilon$ for the 10 largest eigenvalues (sorted, largest on the right) over the final 20 steps of a 60-step diffusion process, using the \emph{Cosine} schedule ($s=0$, $e=1$, $\tau=1$) and the spectral schedule respectively. In both figures, eigenvalues are sorted in ascending order, allowing a comparison between high-frequency components (lower-magnitude eigenvalues) and low-frequency components (higher-magnitude eigenvalues).

% final for each eigenvalue over the 
In both figures, We observe that the error decreases consistently for large eigenvalues, but increases in the final steps for smaller ones. In addition, the final relative error (yellow curve) is notably lower for high-magnitude eigenvalues (i.e., low frequencies). This behavior aligns with the observed bias toward mid-to-high frequencies reported in prior works \cite{yang2023diffusion}. Notably, this effect is more pronounced under the Cosine schedule; while both schedules yield similar errors for large eigenvalues, the Cosine schedule exhibits much higher errors for small eigenvalues (i.e., high-frequency components). This suggests the effectiveness of the proposed optimization procedure in reducing such errors. 

\begin{figure}[H]
    \centering
    \begin{subfigure}[b]{0.3\textwidth}
        \centering
     \includegraphics[width=\textwidth]{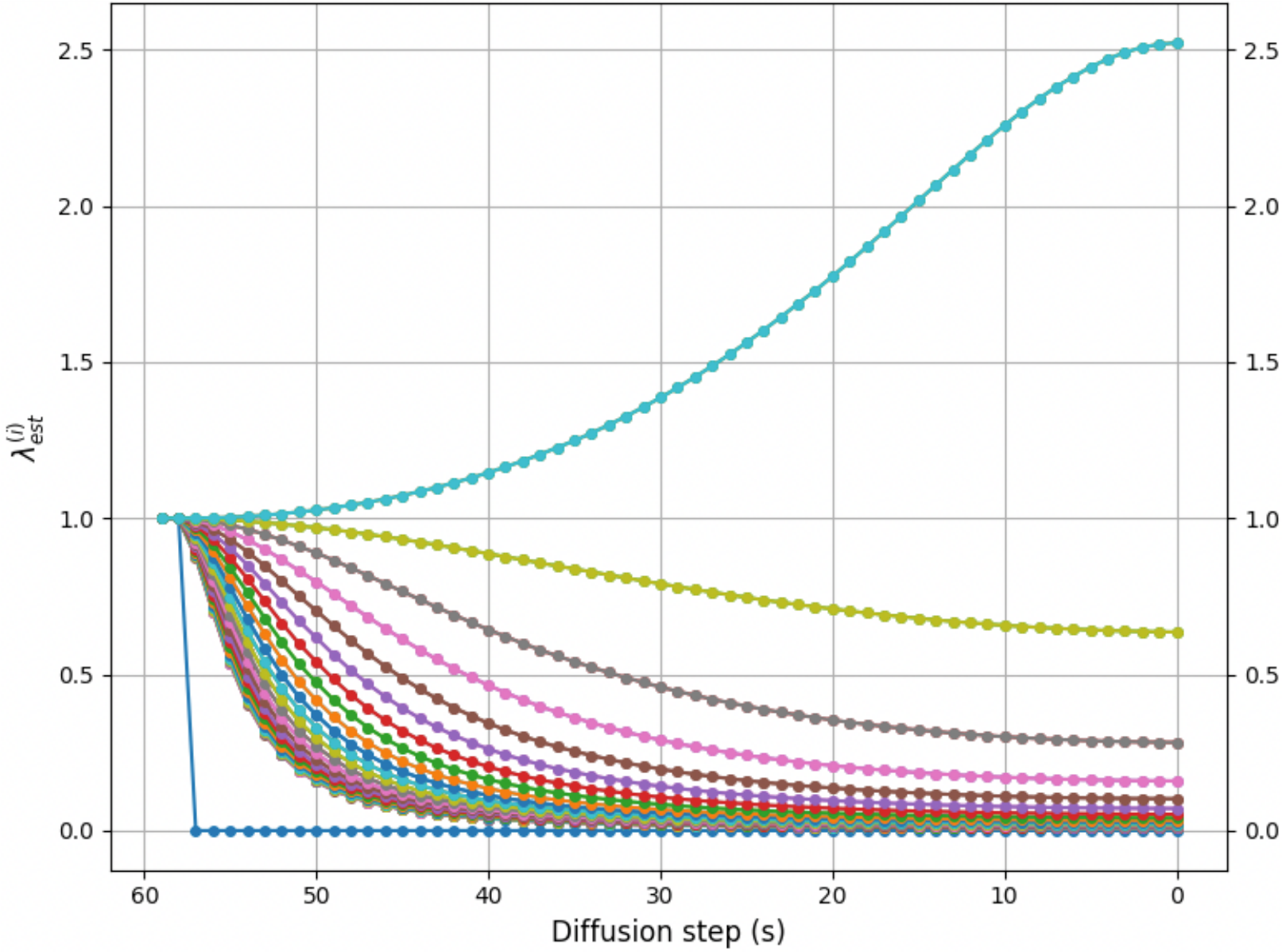}
        % \caption{Cosine ($s=0$, $e=1$, $\tau=1$)}
        \caption{}
        \label{subfig:temporal_eigenvalues_cosine}
    \end{subfigure}
    \begin{subfigure}{0.3\textwidth}
        \centering
        \includegraphics[width=\textwidth]{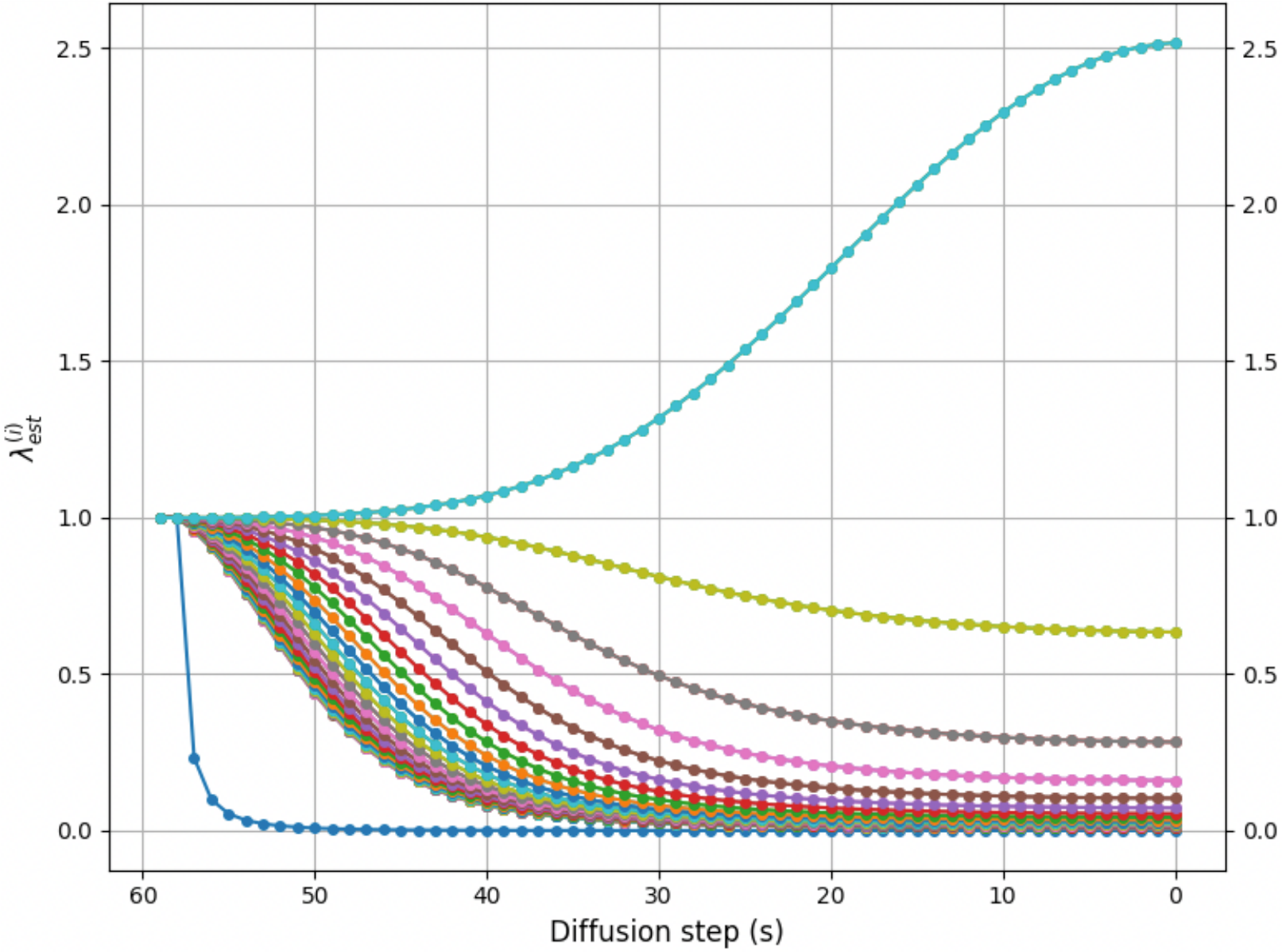}
        % \caption{Spectral}
        \caption{}
       \label{subfig:temporal_eigenvalues_spectral}
    \end{subfigure}
\caption{The dynamics of eigenvalues through a 60-step diffusion process  for the Cosine ($s=0$, $e=1$, $\tau=1$) (\ref{subfig:temporal_eigenvalues_cosine}) and the spectral (\ref{subfig:temporal_eigenvalues_spectral})schedules. }
\label{fig:temporal_eigenvalues}
\end{figure}

An additional aspect is the evolution of eigenvalues throughout the diffusion process. Figures \ref{subfig:temporal_eigenvalues_cosine} and \ref{subfig:temporal_eigenvalues_spectral} illustrate this for the \emph{Cosine} ($s=0$, $e=1$, $\tau=1$) and spectral schedules for a 60-step diffusion process. An interesting observation is that low-magnitude eigenvalues tend to converge to their final values faster than high-magnitude ones. Notably, the convergence rate does not necessarily reflect the accuracy of the final values, as discussed in Figure \ref{fig:temporal_relative_error}. Comparing convergence rates across different noise schedules or leveraging the intermediate expressions to design improved schedules are additional aspects that can be explored; however, we leave them for future work.

\begin{figure}[H]
    \centering
    \begin{subfigure}[b]{0.3\textwidth}
        \centering
     \includegraphics[width=\textwidth]{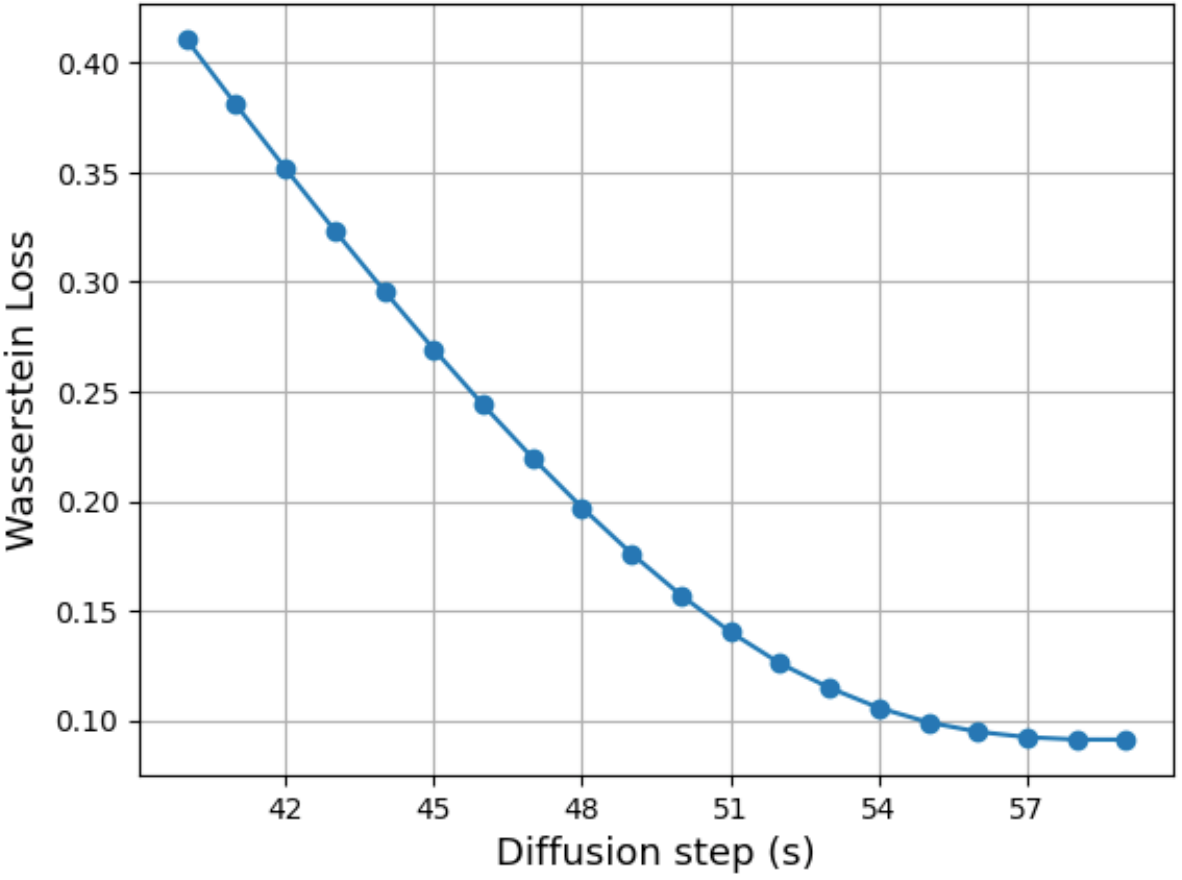}
        % \caption{Cosine ($s=0$, $e=1$, $\tau=1$)}
        \caption{}
        \label{subfig:temporal_wasserstein_cosine}
    \end{subfigure}
    \begin{subfigure}{0.3\textwidth}
        \centering
        \includegraphics[width=\textwidth]{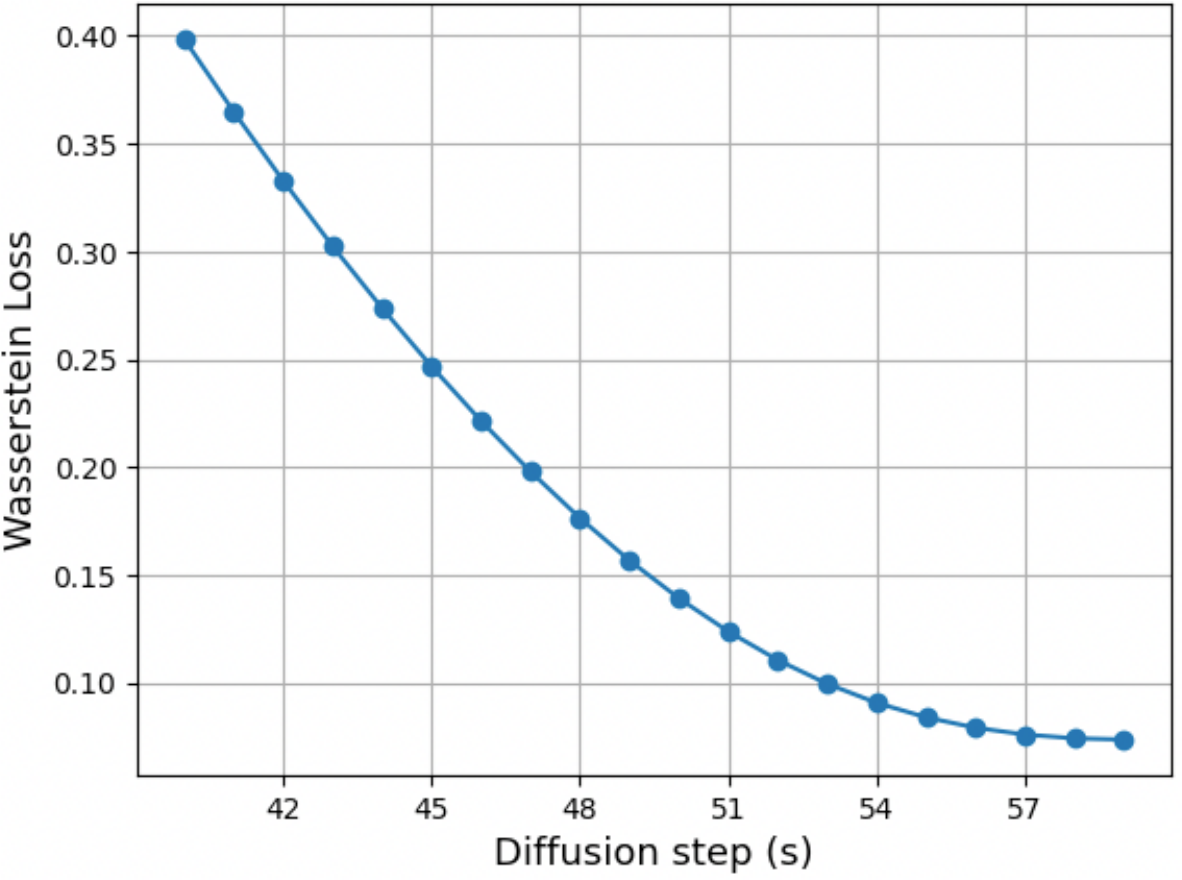}
        % \caption{Spectral}
        \caption{}
       \label{subfig:temporal_wasserstein_spectral}
    \end{subfigure}
\caption{The dynamics of the \emph{Wasserstein-2} distance over the final 20 steps of a 60-step diffusion process for the Cosine ($s=0$, $e=1$, $\tau=1$) (\ref{subfig:temporal_wasserstein_cosine}) and the spectral (\ref{subfig:temporal_wasserstein_spectral})  schedules.}
\label{fig:temporal_wasserstein}
\end{figure}

% \red{\textbf{Eigenvalues dynamics:} \autoref{fig:Eagenvalues_convergence} illustrates the dynamics of the eigenvalues across $60$ diffusion steps using the spectral schedule. While lower eigenvalues tend to stabilize earlier in the process, though this does not necessarily indicate the values they ultimately reach (see Appendix \ref{} for details).}
Figure \ref{fig:temporal_wasserstein} illustrates the evolution of the \emph{Wasserstein-2} error over the final $20$ steps of a $60$-step diffusion process. It can be observed that the \emph{Wasserstein-2} error consistently decreases for both the cosine ($0$,$1$,$1$) (Figure \ref{subfig:temporal_wasserstein_cosine} ) and the spectral noise schedules (Figure \ref{subfig:temporal_wasserstein_spectral} ), with the spectral schedule converging to a lower final value

\newpage
\subsection{Relationship Between Noise Schedules and the loss functions}
\label{sec:Noise_schedule_loss_function}

% As mentioned in \ref{sec:migrating_to_the_spectral_domain}, assuming circularity, the eigenvalues correspond to the DFT coefficients of first row of $\bSigma_0$.
% When the eigenvalues, or equivalently the coefficients of the Discrete Fourier Transform, decrease monotonically, there is a direct relationship between the magnitude of the eigenvalue and its corresponding frequency (for example, $1/f$ behavior observed in speech \citep{voss1975f}). In such cases, the first eigenvalues
% correspond to the low frequencies, having larger amplitudes,
% while the last correspond to high frequencies and smaller
% amplitudes. This pattern, along, with the observations in Appendix \ref{sec:appendix_further_discussion}\footnote{Low magnitude eagenvalues relate with concave schedule and high magnitude eigenvalues correspond to convex schedule.}, aligns with the well-known coarse-to-fine signal construction behavior of diffusion models.\footnote{Higher-frequency components are emphasized by allocating more steps toward the end of the diffusion process, while lower-frequency components are empahsized erallier.} 

Building on the spectral monotonic behavior, the loss function can be adjusted to weight different spectral regions in various ways, shaping the noise schedule based on specific objectives.

% We propose a \emph{weighted l1} loss for the first and the second moments of two Gaussian distributions $P(\mathbf{\hat{x}_{0}^{\mathcal{F}}}; \bbalpha)$ and $P(\mathbf{{x}_{0}^{\mathcal{F}}})$. 

% \begin{equation}
%  \mathcal{D}_{L_1}\big(P(\mathbf{\hat{x}_{0}^{\mathcal{F}}}; \bbalpha), P(\mathbf{{x}_{0}^{\mathcal{F}})}\big) = 
% \sum_{i=1}^{d} \frac{\mathbf{\lambda}_i}{\sum_{j}\mathbf{\lambda}_j}
% \left| [\D_1]_i^2 - \mathbf{\lambda}_i \right| + \sum_{i=1}^{d}\frac{[\bmu^{\mathcal{F}}]^2_i}{\sum_{j} [\bmu^{\mathcal{F}}]^2_j} \left( [\D_2]_i-1 \right)^2   
% \end{equation}

We propose a \emph{weighted l1} loss for the first and the second moments of two Gaussian distributions $P(\projectedxhat; \bbalpha)$ and $P(\projectedxz)$. 

\begin{equation}
 \mathcal{D}_{L_1}\big(P(\projectedxhat; \bbalpha), P(\projectedxz)\big) = 
\sum_{i=1}^{d} \frac{\mathbf{\lambda}_i}{\sum_{j}\mathbf{\lambda}_j}
\left| [\D_1]_i^2 - \mathbf{\lambda}_i \right| + \sum_{i=1}^{d}\frac{[\projectedmu]^2_i}{\sum_{j} [\projectedmu]^2_j} \left( [\D_2]_i-1 \right)^2   
\end{equation}

The first term applies a \emph{weighted l1} loss to the eigenvalues, while the second term computes a \emph{weighted l2} norm of the mean vectors.\footnote{We aim to maintain the relationship between both components similar to the \emph{Wasserstein-2} distance.} This design ensures that eigenvalues with larger magnitudes and mean components with higher values have greater influence on the overall loss.

\begin{figure}[h]
    \centering
        \includegraphics[width=0.4\columnwidth]{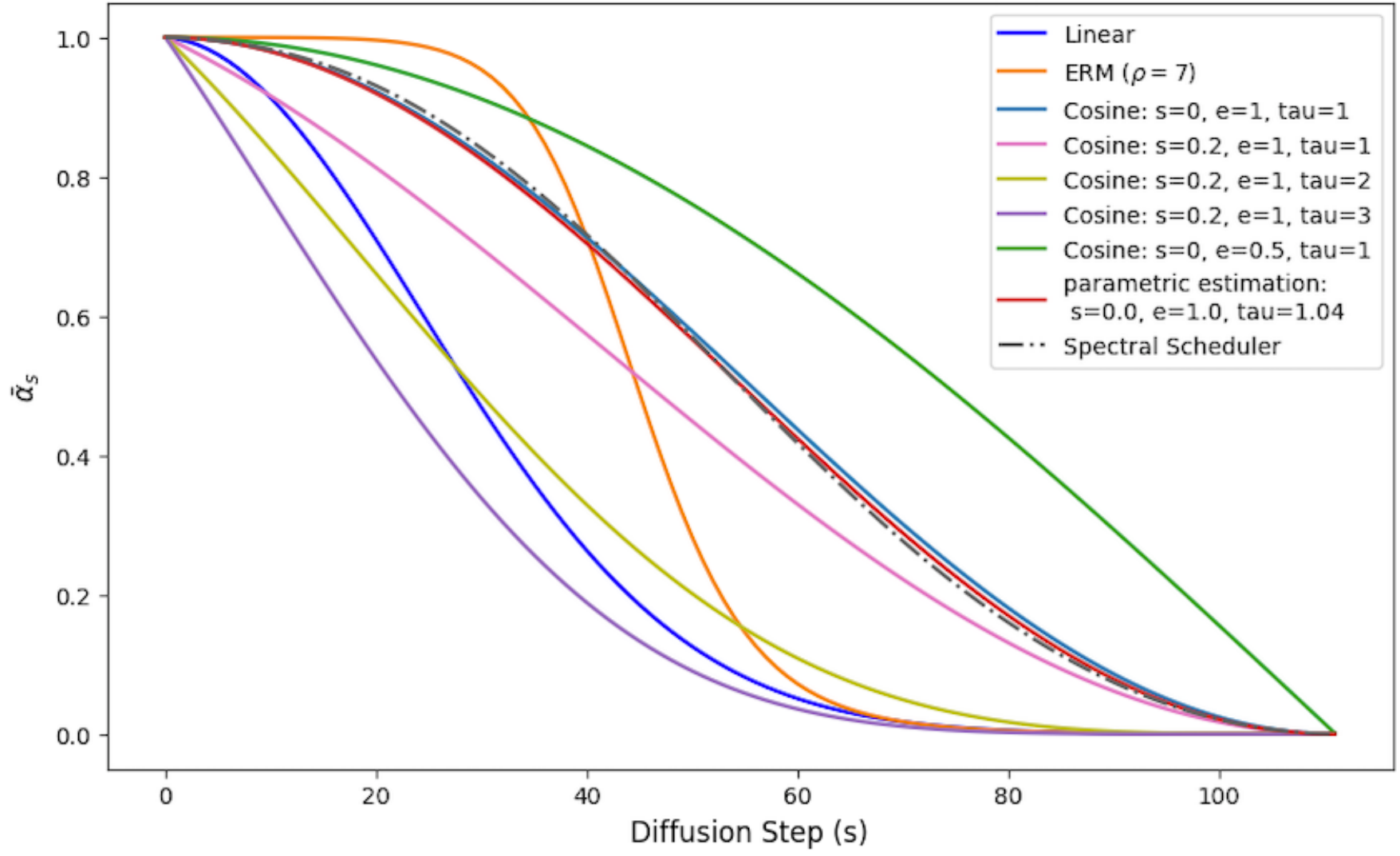}

\caption{Comparison of the spectral schedule with various heuristic noise schedules for $S=112$ diffusion steps. The figure presents the linear, EDM ($\rho = 7$), Cosine-based schedules, including \emph{Cosine} ($s=0$, $e=1$, $\tau=1$) from \citep{nichol2021improved, chen2023importance}. The parametric estimation of the cosine function is shown in red.}
    \label{fig:Exp_2_Spectral_scheduler_comparison_weighted_l1}
\end{figure}

Figure \ref{fig:Exp_2_Spectral_scheduler_comparison_weighted_l1} illustrates the spectral recommendation obtained by solving the optimization problem in \ref{eq:optimization_problem} using the \emph{Weighted l1} loss. The results are based on the Gaussian MUSIC-Piano dataset described in \ref{subsec:scenario_2} where $d=16,000$ and $th=0.05$.

% Interestingly, using the \emph{weighted l1} loss results in a spectral recommendation that naturally aligns with established heuristic approaches. Specifically, it aligns with the manually designed \emph{cosine (0,1,1)} schedule proposed in \cite{nichol2021improved}.
% This result may raise questions about the potential connection between the design of widely-used noise schedule heuristics and a bias against high-frequency generation, which has been noted in previous studies \cite{yang2023diffusion}.

Interestingly, using the \emph{weighted l1} loss results in a spectral recommendation that aligns with established heuristic methods. Specifically, it corresponds to the manually designed \emph{cosine (0,1,1)} schedule proposed in \cite{nichol2021improved}. This observation could indicate a potential link between the design of widely used noise schedule heuristics and a bias against high-frequency regions, which has been observed in previous research \cite{yang2023diffusion}.

Note: The relationship between the magnitude of the eigenvalues and their corresponding frequencies holds tight only when monotonic behavior is present. In real-world scenarios, as shown in Figure \ref{subfig:Resolution_d_16000}, the eigenvalues' magnitudes generally decrease, but the function is not strictly monotonic. In such cases, an alternative approach is required, one that either analyzes broader spectral regions or considering both the values and indices of the eigenvalues.
\newpage
\section{Analysis of Mean Bias}\label{sec:appendix_Mean_Bias}

In the following section, we analyze the mean bias expression.
To maintain consistency and preserve the connection to the frequency domain, we conduct the analysis on the circulant matrices used in Section~\ref{subsec:Scenario_1}, which can be diagonalized using the discrete Fourier transform (DFT), denoted by \( \mathcal{F}\{\cdot\} \). Naturally, the same approach can be extended to any covariance matrix, $\mathbf{\Sigma}_0$, using its eigenbasis~$\textbf{U}$.

% In the following section, We analyze the mean bias expression.
% To maintain consistency and the relation to frequencies, we perform the on a circulant matrices uses in section \ref{subsec:Scenario_1}, which can be diagonalized using the discrete Fourier transform (DFT), denoted by \( \mathcal{F}\{\cdot\} \). Naturally, the same approach can be extended to any covariance matrix using its eigenbasis $U$.

The mean bias expression $(\D_2 - \I)\boldsymbol{\mu_0^{\mathcal{F}}}$ arises from the difference between  $\mathbb{E} \left[ \mathbf{\x}_{0}^{\mathcal{F}} \right]$ and $\mathbb{E}  \left[\mathbf{\hat{\x}}_{0}^{\mathcal{F}} \right]$. In particular, We focus on the absolute magnitude of the expression $|\D_2 - \I||\boldsymbol{\mu_0^{\mathcal{F}}}|$.
The term $\D_2$, as defined in \eqref{eq:D_1,D_2}, depends on $\bLambda$ and on $\bbalpha$, the chosen noise schedule. Notably, for stationary signals, $\bmu$ is deterministic, resulting in the vector $\boldsymbol{\mu_0^{\mathcal{F}}}$ where all entries are zero except for the first element.
specifically, for a mean-centered signal where 
$\boldsymbol{\mu_0^{\mathcal{F}}} = \0$, the DDIM process remains unbiased, regardless of $|\D_2 - \I|$ expression. In other cases, for a given
$\boldsymbol{\mu_0^{\mathcal{F}}}$, the primary source of bias originates from the main diagonal of $|\D_2 - \I|$.

\begin{figure}[H]
    \centering
        \begin{subfigure}[b]{0.4\textwidth}
        \centering
        \includegraphics[width={\textwidth}]{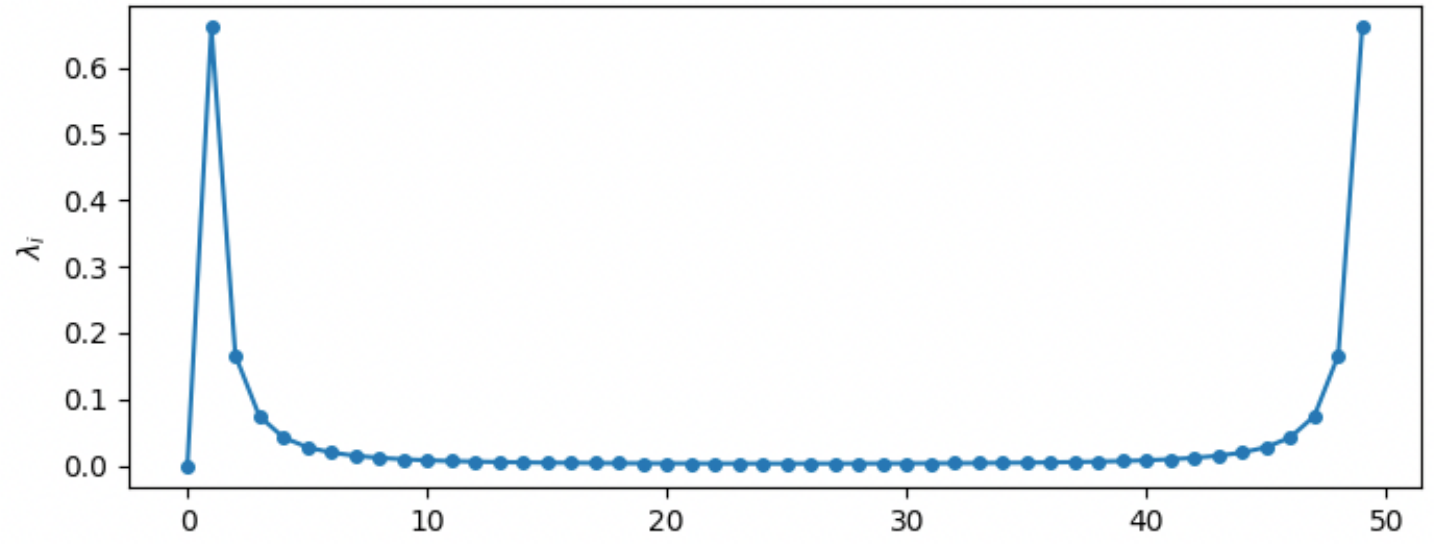}
         \caption{}
        % \caption{$\bSigma$}
        \label{subfig:mu_drift_lambda_0_matrix_1}
    \end{subfigure}
     % \hfill
    %     \begin{subfigure}[b]{0.3\textwidth}
    %     \centering
    %     \includegraphics[width={\textwidth}]{figures_spectral/MU_Drift/EXP1_New_figures/matrix_2/original_dft_coef.pdf}
    %      \caption{}
    %     % \caption{$\bSigma$}
    %      \label{subfig:mu_drift_lambda_0_matrix_2}
    % \end{subfigure}
         % \hfill
    \begin{subfigure}[b]{0.4\textwidth}
        \centering
        \includegraphics[width={\textwidth}]{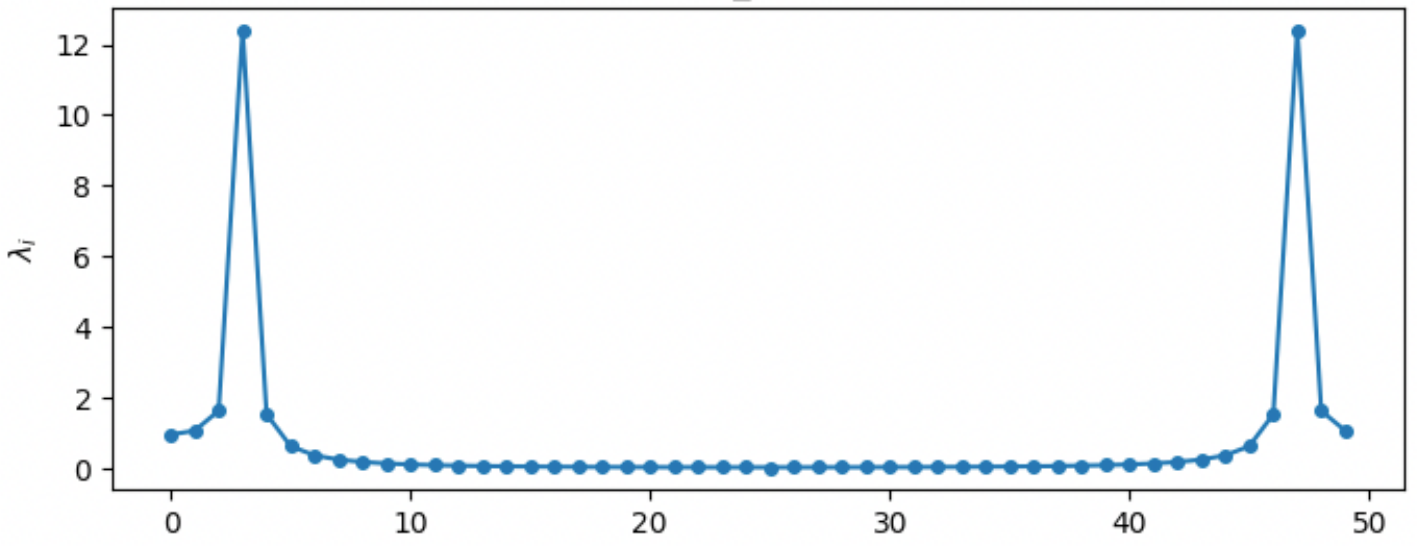}
         \caption{}
        % \caption{$\bSigma$}
         \label{subfig:mu_drift_lambda_0_matrix_8}
    \end{subfigure}
\hfill
    \hspace{1mm}
    \begin{subfigure}[b]{0.4\textwidth}
        \centering
        \includegraphics[width=\textwidth]{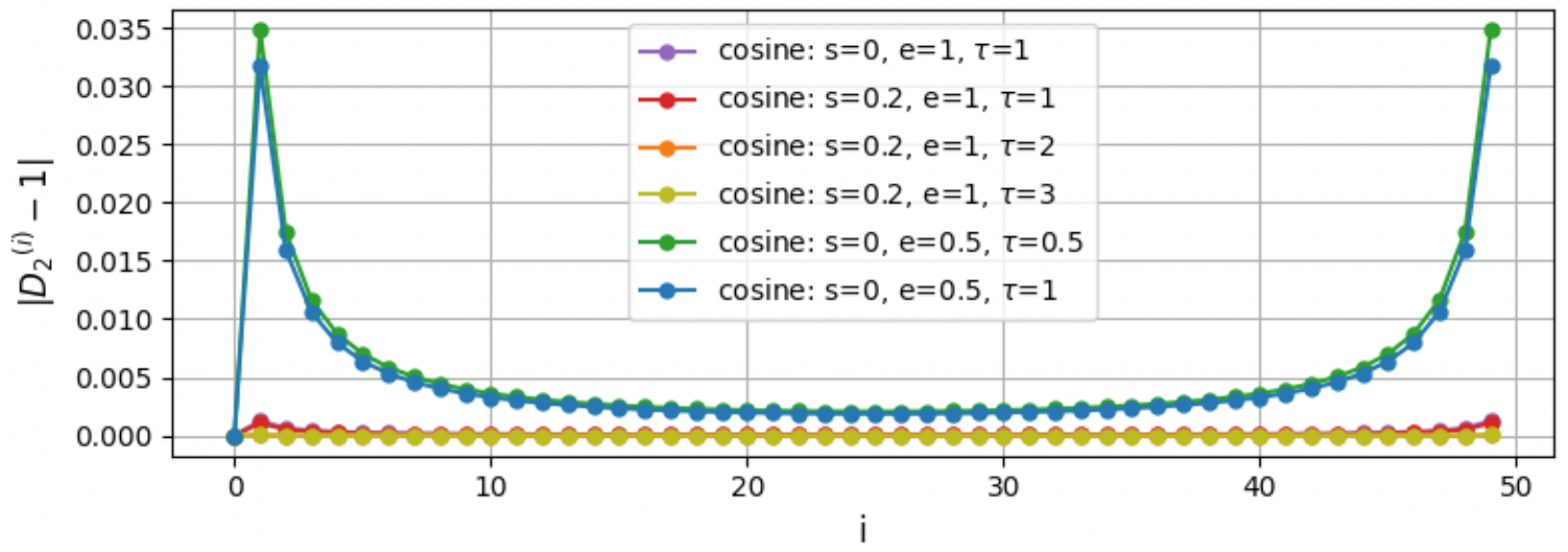}
         \caption{}
        % \caption{$\bSigma$}
         \label{subfig:mu_drift_noise_sched_matrix_1}
    \end{subfigure}
         % \hfill
    % \begin{subfigure}[b]{0.3\textwidth}
    %     \centering
    %     \includegraphics[width=\textwidth]{figures_spectral/MU_Drift/EXP1_New_figures/matrix_2/100.pdf}
    %      \caption{}
    %     % \caption{$\bSigma$}
    %     \label{subfig:mu_drift_noise_sched_matrix_2}
    % \end{subfigure}
      % \hfill
        \begin{subfigure}[b]{0.4\textwidth}
        \centering
        \includegraphics[width=\textwidth]{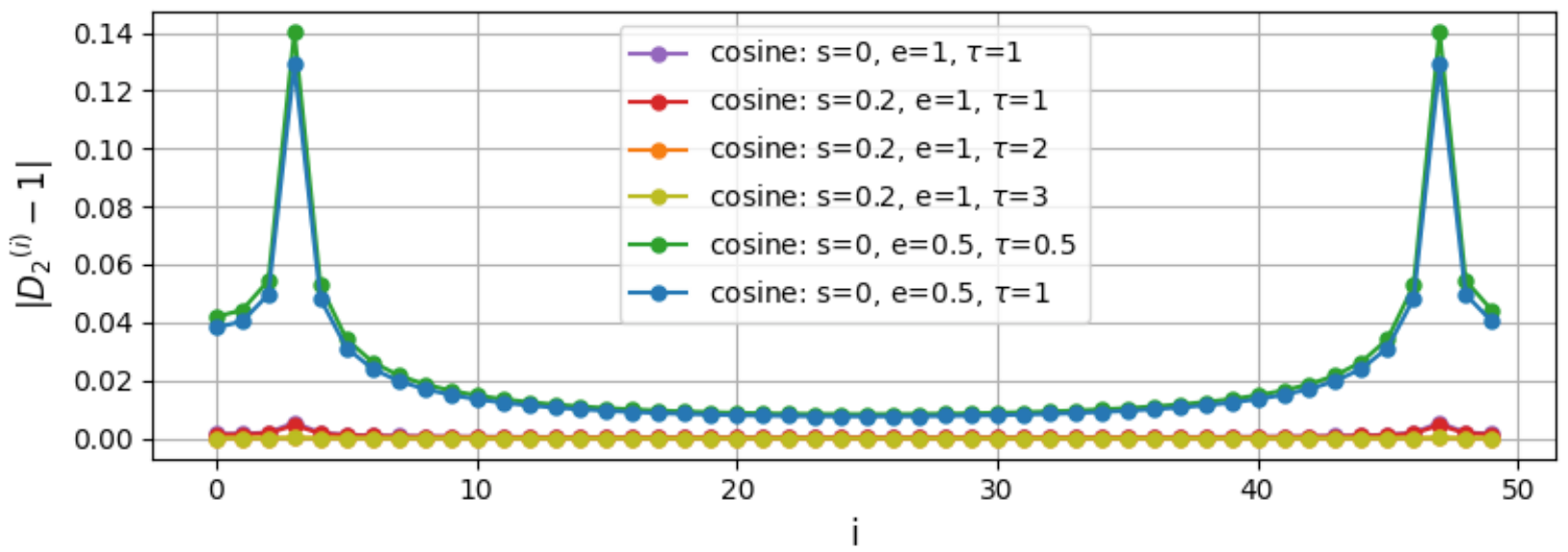}
         \caption{}
        % \caption{$\bSigma$}
        \label{subfig:mu_drift_noise_sched_matrix_8}
    \end{subfigure}
        \hspace{1mm}
    \begin{subfigure}[b]{0.4\textwidth}
        \centering
        \includegraphics[width=\textwidth]{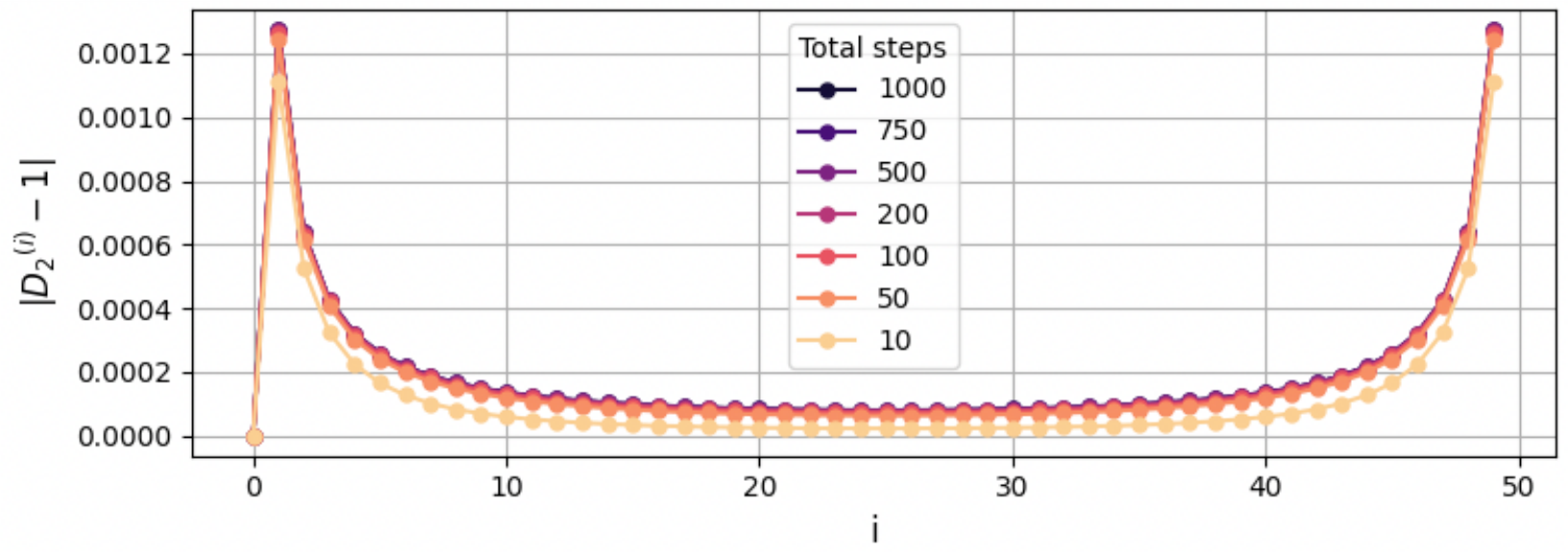}
         \caption{}
        % \caption{$\bSigma$}
        \label{subfig:mu_drift_cosine_0_1_1_matrix_1}
    \end{subfigure}
    %     \hfill
    %     \begin{subfigure}[b]{0.3\textwidth}
    %     \centering
    %     \includegraphics[width=\textwidth]{figures_spectral/MU_Drift/EXP1_New_figures/matrix_2/Diff_D_2_(0, 1, 1).pdf}
    %      \caption{}
    %     % \caption{$\bSigma$}
    %     \label{subfig:mu_drift_cosine_0_1_1_matrix_2}
    % \end{subfigure}
      % \hfill
        \begin{subfigure}[b]{0.4\textwidth}
        \centering
        \includegraphics[width=\textwidth]{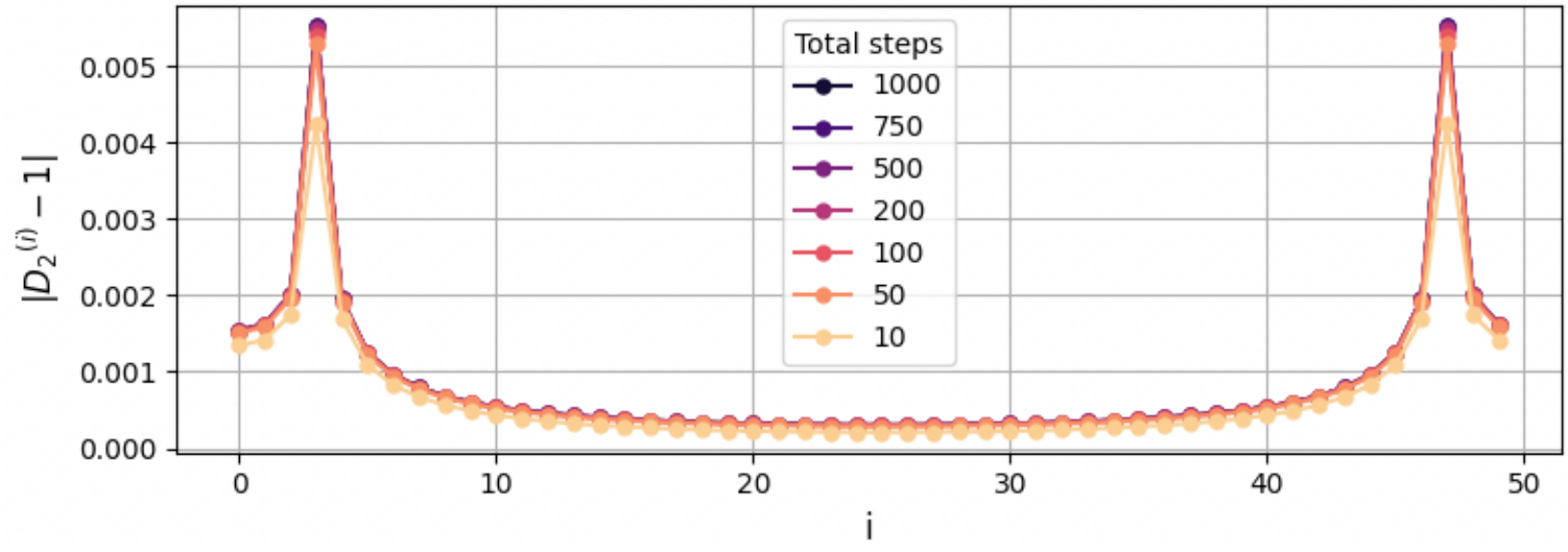}
         \caption{}
        % \caption{$\bSigma$}
        \label{subfig:mu_drift_cosine_0_1_1_matrix_8}

    \end{subfigure}

    \hspace{1mm}
    \begin{subfigure}[b]{0.4\textwidth}
        \centering
        \includegraphics[width=\textwidth]{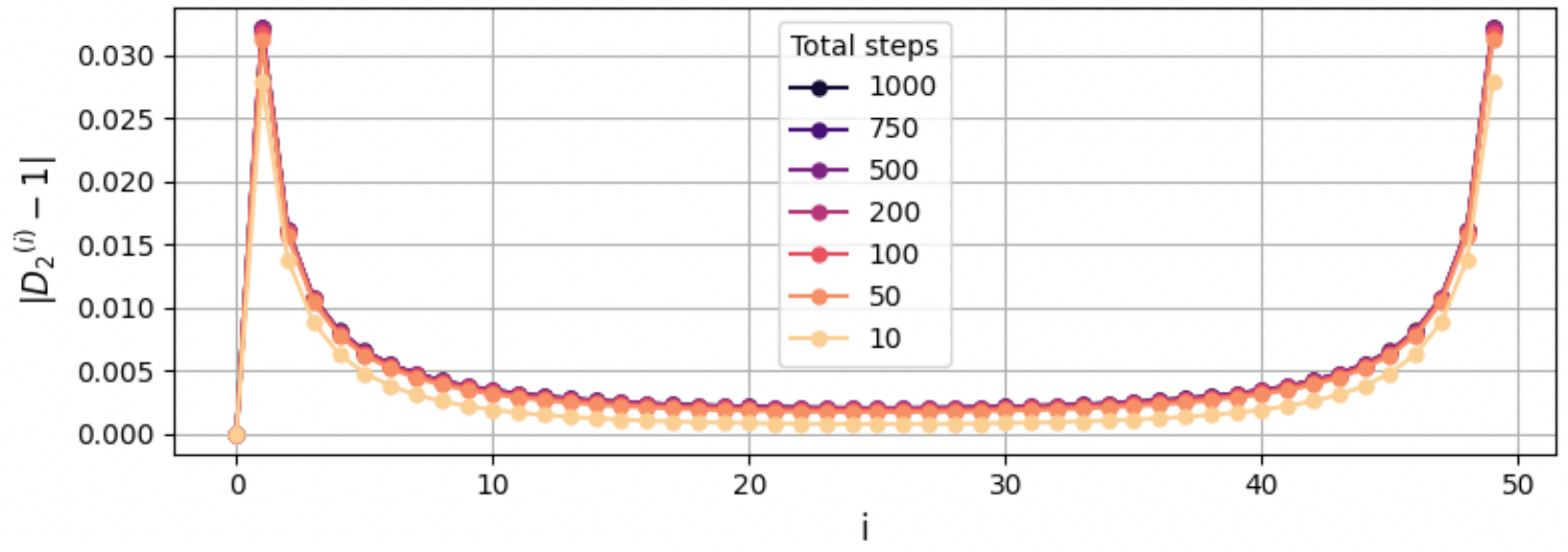}
         \caption{}
        % \caption{$\bSigma$}
        \label{subfig:mu_drift_cosine_0_0_5_1_matrix_1}
    \end{subfigure}
    %     \hfill
    %     \begin{subfigure}[b]{0.3\textwidth}
    %     \centering
    %     \includegraphics[width=\textwidth]{figures_spectral/MU_Drift/EXP1_New_figures/matrix_2/Diff_D_2_(0, 0.5, 1).pdf}
    %      \caption{}
    %     % \caption{$\bSigma$}
    %     \label{subfig:mu_drift_cosine_0_0_5_1_matrix_2}
    % \end{subfigure}
      % \hfill
        \begin{subfigure}[b]{0.4\textwidth}
        \centering
        \includegraphics[width=\textwidth]{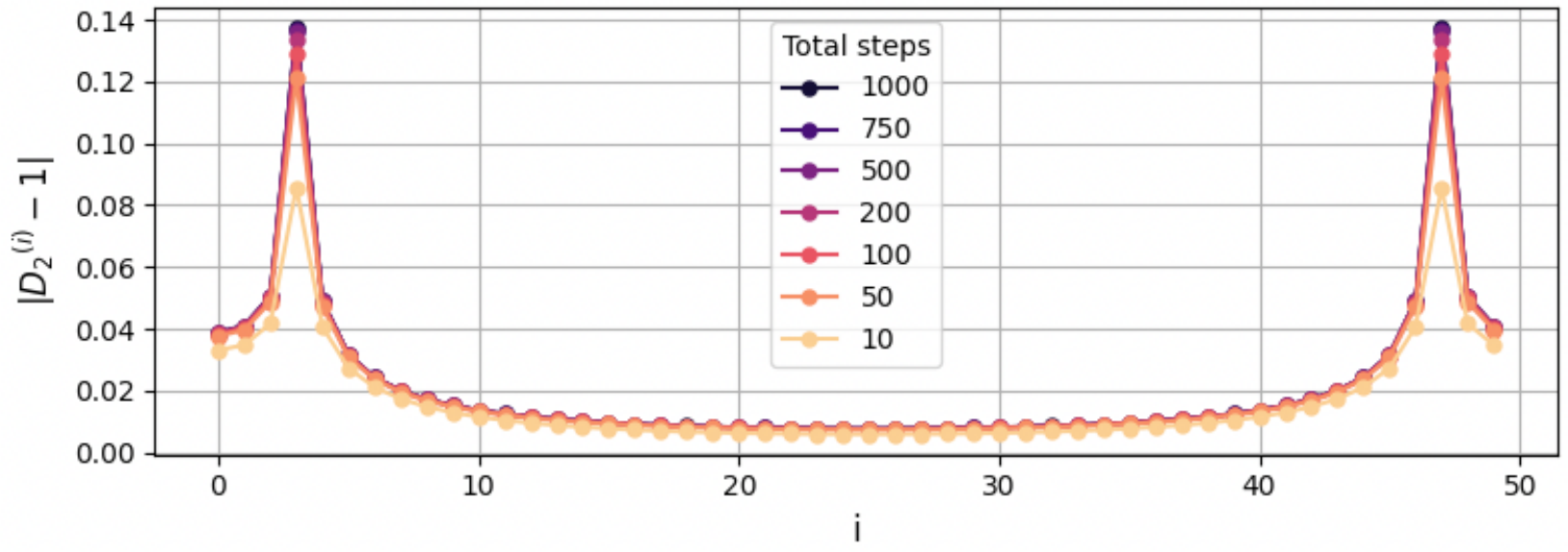}
         \caption{}
        % \caption{$\bSigma$}
        \label{subfig:mu_drift_cosine_0_0_5_1_matrix_8}
    \end{subfigure}
   \caption{Figures \ref{subfig:mu_drift_lambda_0_matrix_1} and \ref{subfig:mu_drift_lambda_0_matrix_8} display the eigenvalues for two choices of $\bLambda$ with $d=50$. Figures \ref{subfig:mu_drift_noise_sched_matrix_1} and \ref{subfig:mu_drift_noise_sched_matrix_8} compare the mean bias values across different parametrization of the Cosine noise schedule using $112$ diffusion steps. Figures \ref{subfig:mu_drift_cosine_0_1_1_matrix_1} and \ref{subfig:mu_drift_cosine_0_1_1_matrix_8} show the bias for varying numbers of diffusion steps with the Cosine ($0$,$1$,$1$) noise schedule, while Figures \ref{subfig:mu_drift_cosine_0_0_5_1_matrix_1} and \ref{subfig:mu_drift_cosine_0_0_5_1_matrix_8} illustrate the same for the Cosine ($0$,$0.5$,$1$) schedule.}
    \label{fig:Appendix_mu_drift}
\end{figure}

Figure \ref{fig:Appendix_mu_drift} analyzes the mean bias for two choices of $\bLambda$ with $d=50$.
It compares the values of $|\D_2 - \I|$ across several cosine noise schedule heuristics and illustrates how its behavior depends on the number of diffusion steps.

% compare the elements of the vector $|\D_2 - \I|$ for several cosine noise schedule heuristics. 

Figures \ref{subfig:mu_drift_noise_sched_matrix_1} and \ref{subfig:mu_drift_noise_sched_matrix_8} reveal that for certain heuristics, such as Cosine ($0$,$1$,$1$), the bias is negligible, while for others, like Cosine ($0$,$0.5$,$1$), the bias increases. Additionally, the magnitude of the eigenvalues
 $\{{\lambda}_i\}_{i=1}^d$ plays a significant role in determining the bias; as the eigenvalues grow larger, the bias also tends to increase.

Figures \ref{subfig:mu_drift_cosine_0_1_1_matrix_1} and \ref{subfig:mu_drift_cosine_0_1_1_matrix_8} illustrate the bias across various numbers of diffusion steps $\{10,50,100,200,500,750,1000\}$  for the Cosine ($0$,$1$,$1$) noise schedule. Similarly, Figures \ref{subfig:mu_drift_cosine_0_0_5_1_matrix_1} and \ref{subfig:mu_drift_cosine_0_0_5_1_matrix_8} show the bias for the Cosine ($0$,$0.5$,$1$). In both cases, increasing the number of diffusion steps leads to a gradual rise in the mean bias $|\D_2 - \I|$.
\section{DDPM vs DDIM}\label{sec:appendix_DDPM_vs_DDIM}

\begin{figure}[H]
  \centering
\includegraphics[width=0.5\textwidth]{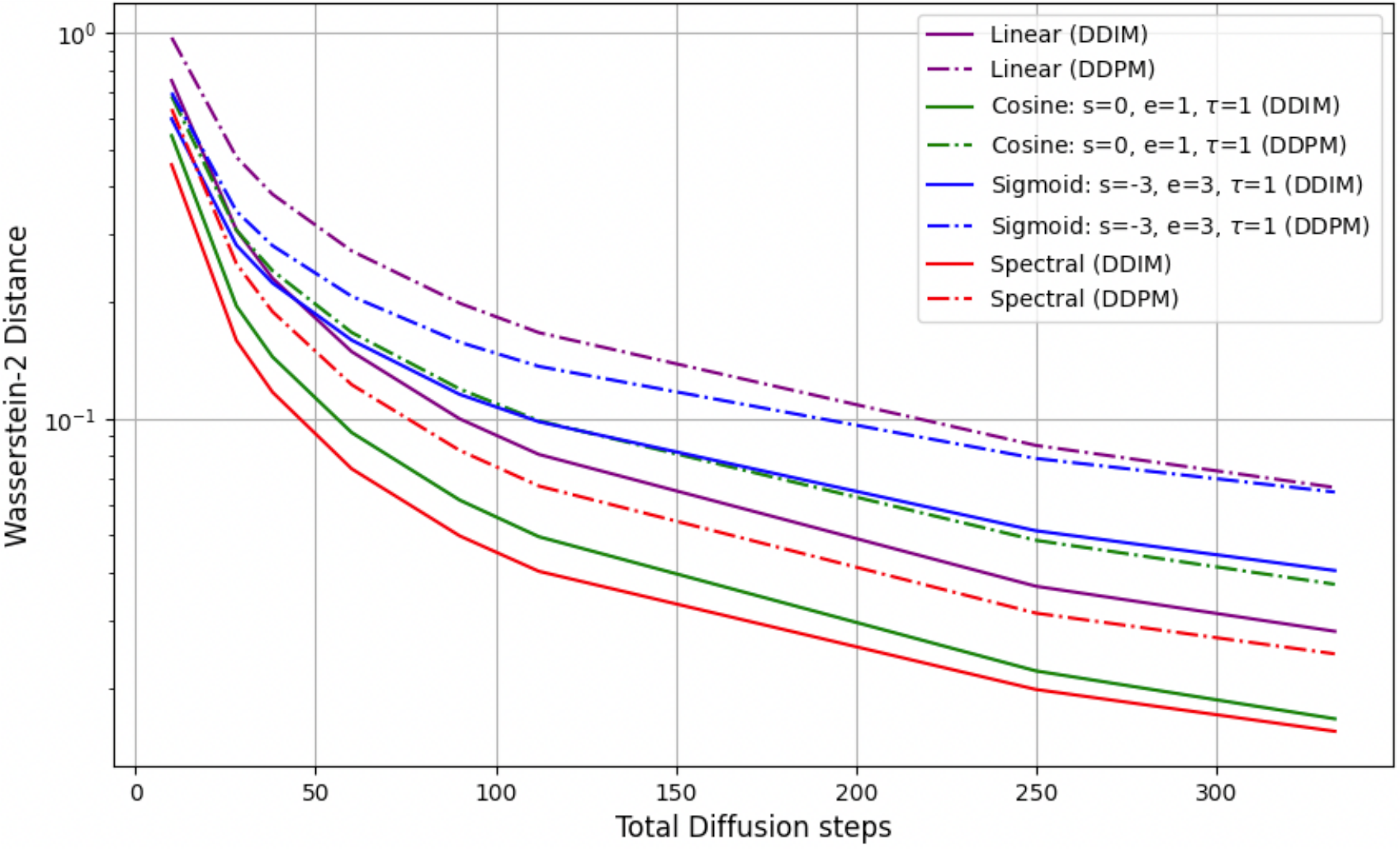}
  \caption{Comparison of the \emph{Wasserstein-2} distance between DDIM and DDPM for different noise schedules, including the spectral recommendation, across various diffusion steps.} 
  \label{fig:DDPM_DDIM_several_schedules_334}
\end{figure}

% \subsubsection{DDPM VS DDIM}
The explicit formulations of DDPM  and DDIM in \eqref{eq:D1D2-DDPM} and 
\eqref{eq:D_1,D_2} also enable their comparison in terms of loss across varying diffusion depths and noise schedules. Figure \ref{fig:DDPM_DDIM_several_schedules_334}
presents such a comparison using the \emph{Wasserstein-2} distance on a logarithmic scale. The results show that DDIM sampling is faster and yields lower loss values than DDPM for each noise schedule, aligning with the empirical observations in \cite{song2020denoising}. In addition, it can be seen that the spectral schedule with DDIM consistently maintains optimality with respect to all other experiments.

% The results clearly show that DDIM sampling is faster and yields lower loss values, aligning with the empirical observations in \cite{song2020denoising}. 

\begin{figure}[H]
  \centering
\includegraphics[width=0.5\textwidth]{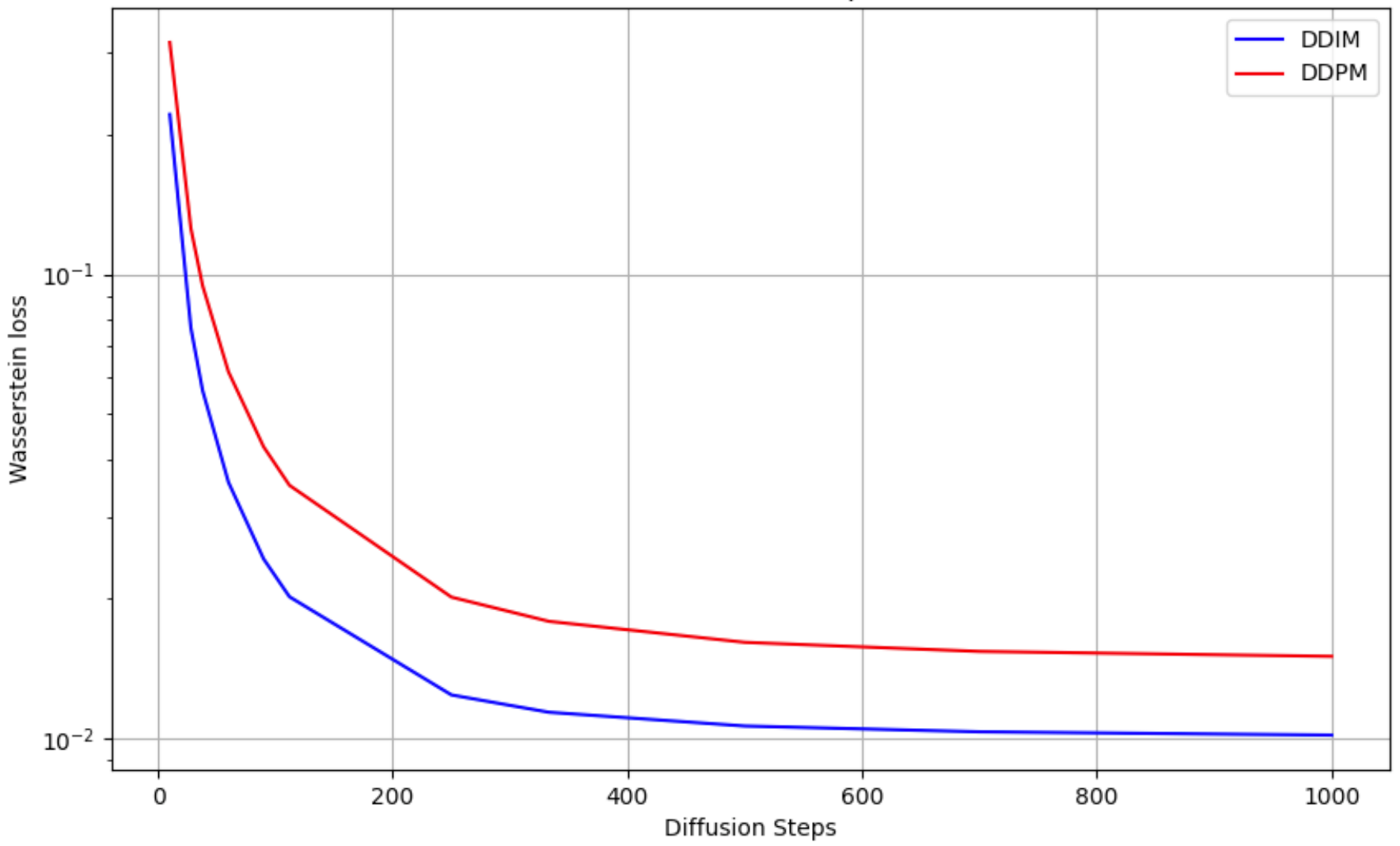}
 \caption{Comparison of the spectral noise schedules in DDPM and DDIM, derived from the covariance matrix outlined in Sec. \ref{subsec:Scenario_1}, with parameters d=50, l=0.1, and total diffusion steps set to $\{10, 28, 38, 60, 90, 112, 250, 334, 500, 750, 1000\}$.}
    \label{fig:DDPM_DDIM_Spectral_1000}
\end{figure}

% \begin{wrapfigure}{t}{0.4\textwidth}
%   \centering
% \includegraphics[width=0.4\textwidth]{figures_spectral/DDPM_DDIM/loss_DDPM_DDIM_log_short_8_0_1_0_5_with_sigma_1_total_of_diffusion_steps.pdf}
% \caption{Comparison of the \emph{Wasserstein-2} distance between DDPM and DDIM for different noise schedules, including the spectral recommendation, across various diffusion steps.}\label{fig:DDPM_DDIM_several_schedules_334}
% \end{wrapfigure}

\section{Optimization Time Analysis}
\label{sec:appendix_optimization_time_analysis}

% Here, we demonstrate the computation time for the $400\times400$ covariance matrices for MUSIC dataset and the $3072\times3072$ covariance matrices for CIFAR10 dataset (compared to $50\times50$ in Section $6.1$). for the MUSIC dataset we leveraged a  stopping conditions of $1500$ optimization steps or a function tolerance of $10^-6$ while for the CIFAR10 we used a stopping conditions of $2000$ optimization steps or a function tolerance of $10^-6$ for enabling converege also in variable diffusion steps. all running on a standard CPU.  A more lenient choice may yield the same optimal schedule in less time.
The computational cost of solving the optimization problem for time-dependent signals depends on several factors. As discussed in Appendix \ref{sec:scenario_2_Analysis_of_Different_Aspects}, increasing the resolution of temporal signals improves both accuracy and adaptation to the data characteristics, but also increases computational cost (larger covariance matrices allow for higher frequency accuracy). Additionally, factors like the sampling rate and stationary regions influence the results. 

We report the computation times for the covariance matrices of different datasets, with dimensions $400\times400$ for MUSIC, $3072\times3072$  for CIFAR10, and $12288\times12288$ for AFHQv2. For the MUSIC dataset, we applied stopping criteria of $1500$ optimization steps or a function tolerance of $10^{-6}$. For CIFAR10 and AFHQv2, we used $2000$ steps or the same tolerance to ensure convergence for a large number of diffusion steps. All computations were performed on a standard CPU. Notably, using slightly more relaxed stopping criteria could achieve the same optimal schedule in less time.

% Here, we demonstrate the computation time for the $400\times400$ covariance matrices for the MUSIC dataset, $3072\times3072$ covariance matrices for the CIFAR10 dataset (compared to $50\times50$ in Section \ref{subsec:Scenario_1}) and $12288\times12288$ for the AFHQv2 dataset. For the MUSIC dataset, we leveraged stopping conditions of $1500$ optimization steps or a function tolerance of $10^{-6}$, while for the CIFAR10 and AFHQv2 datasets, we used stopping conditions of $2000$ optimization steps or a function tolerance of $10^{-6}$ to ensure the optimization problem converged successfully for a large number of diffusion steps. All computations were performed on a standard CPU. Notably, using slightly more lenient stopping criteria could yield the same optimal schedule in less time.

% \begin{table}[h]

%   \caption{Comparison of the optimization time (in seconds) for a $400 \times 400$ covariance matrix (e.g., MUSIC dataset) and a $3072 \times 3072$ covariance matrix (e.g., CIFAR10 dataset).}
%   \label{table:optimizarion_time_comparison}
%   \centering
%   \begin{tabular}{ccc}
%     \toprule
%     Diffusion steps     & $400\times400$     & $3072 \times 3072$ \\
%     \midrule
%     $10$ & $0.09$  & $0.36$    \\
%     $50$     & $2.94$ & $8.19$      \\
%     $90$     & $18.31$       & $25.4$  \\
%     $130$ & $55.21$  & $321.76$     \\
%     $170$     & $69.5$ & $558.76$      \\
%     $210$     & $172.02$    &   $594.51$  \\
%     $250$     & $261.69$  &    $949.00$  \\
%     \bottomrule
%   \end{tabular}
% \end{table}

% Camera ready: %

\begin{table}[h]

  \caption{Comparison of the optimization time (in seconds) for a $400 \times 400$ covariance matrix (e.g., MUSIC dataset), a $3072 \times 3072$ covariance matrix (e.g., CIFAR10 dataset) and a $12288 \times 12288$ covariance matrix  (e.g., AFHQv2 dataset)}
  \label{table:optimizarion_time_comparison}
  \centering
  \begin{tabular}{cccc}
    \toprule
    Diffusion steps     & $400\times400$     & $3072 \times 3072$ & $12288 \times 12288$\\
    \midrule
    $10$ & $0.09$  & $0.36$  & $3.88$    \\
    $50$     & $2.94$ & $8.19$  & $93.09$      \\
    $90$     & $18.31$       & $25.4$  & $292.48$   \\
    $130$ & $55.21$  & $321.76$    & $1555.87$ \\
    $170$     & $69.5$ & $558.76$  & $2316.73$    \\
    $210$     & $172.02$    &   $594.51$ & $3884.07$ \\
    $250$     & $261.69$  &    $949.00$ & $7417.46$ \\
    \bottomrule
  \end{tabular}
\end{table}

As shown in the Table \ref{table:optimizarion_time_comparison}, for a low number of steps, where discretization error is significant and the spectral scheduler greatly improves performance, the computation time ranges from less than a second to a few tens of seconds.  Over approximately 130 diffusion steps, the computation time exceeds one minute for the CIFAR10; however, in this regime, discretization errors become negligible, the overall structure of the scheduler remains unchanged, and its density increases with minimal change.

% \textcolor{red}{To leverage this structural similarity, we adopt a gradual optimization strategy: 
To accelerate the optimization process, we leverage this structural similarity by adopting a gradual optimization strategy: we begin with a small number of diffusion steps (e.g., 10) which requires relatively low computational effort and use the resulting solution (interpolated to the appropriate length) to initialize optimization for a larger number of steps. This approach significantly reduces computation time by reusing information across step counts.

Table \ref{table:accelerating_optimization_time} compares optimization times on the AFHQv2 dataset using random initialization (left column) versus iterative initialization. For the iterative approach, the third column reports the time for each individual run, while the forth shows the total cumulative time to reach the solution for a given diffusion step count.
As observed in Table \ref{table:accelerating_optimization_time}, iterative initialization consistently reduces optimization time while achieving noise schedules and loss values comparable to random initialization. Further improvements may be possible by refining the strategy for transitioning between diffusion step counts during optimization.

An additional observation from Table \ref{table:optimizarion_time_comparison} is the significant convergence time required for large covariance matrices, especially when a high number of diffusion steps is used. While covariance matrices in audio signals are typically considered in the stationary domain with moderate dimensions, image data often involve higher resolutions, resulting in larger matrices and increased computational cost. A possible approach to reduce computational cost is to assume a low-rank covariance matrix, selecting only the $d$ most significant directions and ignoring near-zero eigenvalues.

% Specifically, we apply PCA to the original dataset and perform optimization in a reduced-dimensional subspace. On the AFHQv2 dataset, we reduce the original dimensionality from $64 \times 64 \times 3$ $(12,288)$ to $32 \times 32 \times 3$ $(3,072)$ using PCA. The corresponding results are reported in the rightmost column of Table \ref{table:accelerating_optimization_time}. As expected, optimization in the reduced subspace is faster than in the original space ($64 \times 64 \times 3$, left column). The subspace dimensionality serves as a hyperparameter; we chose $d = 3072$, which provides a substantial reduction while preserving the dominant spectral components.

Specifically, we apply PCA to the original dataset and conduct optimization within a reduced-dimensional subspace. For the AFHQv2 dataset, the dimensionality is reduced from $64 \times 64 \times 3$ $(12,288)$ to $32 \times 32 \times 3$ $(3,072)$ using PCA, with results shown in the rightmost column of Table \ref{table:accelerating_optimization_time}. As expected, optimization in this reduced subspace is faster than in the original space ($64 \times 64 \times 3$, left column). The subspace dimensionality acts as a hyperparameter; here, we set $d = 3072$, achieving a significant reduction while preserving the dominant spectral components.

% \textcolor{red}{Specifically, we apply PCA to the original dataset and perform the optimization in a reduced-dimensional subspace. We demonstrate our method on the AFHQv2 dataset by reducing the original dimensionality from $64 \times 64 \times 3$ $(12288)$ to $32 \times 32 \times 3$ $(3072)$ using PCA. The results are shown in the rightmost column of Table 4.}

\begin{table}[h]

  \caption{Comparison of optimization times (seconds) across proposed acceleration methods on AFHQv2 dataset}
  \label{table:accelerating_optimization_time}
  \centering
  \begin{tabular}{ccccc}
    \toprule
    Diffusion steps & Random init. & Iterative init. & Iterative init. (summed) & PCA   \\
    &  $(12288 \times 12288)$     & $(12288 \times 12288)$ & $(12288 \times 12288)$ & $(3072 \times 3072)$\\
    
    \midrule
    $10$ & $3.88$  & $3.92$  & $3.92$  & $0.49$   \\
    $50$     & $93.09$ & $39.60$  & $43.52$  & $12.71$     \\
    $90$     & $292.48$       & $21.61$  & $65.13$  & $32.80$  \\
    $130$ & $1555.87$  & $46.05$    & $111.18$ & $350.27$ \\
    $170$     & $2316.73$ & $20.59$  & $131.78$   & $548.57$  \\
    $210$     & $3884.07$    &   $32.00$ & $163.78$ & $605.66$ \\
    $250$     & $74117.47$  &    $68.23$ & $232.02$ & $919.99$ \\
    \bottomrule
  \end{tabular}
\end{table}

% This trade-off between spectral resolution and optimization time may allows an efficient design of noise schedules which captures the essential features of the signal while reducing optimization time.

% for designing efficient noise schedules that capture the essential features of the signal in less optimization time.

% While in audio signals, covariance matrices are typically considered in the stationary domain at reasonable sampling rates, in images, the resolution is often much higher, resulting in larger matrices and increased computational demands.
% In the context of audio signals, we typically work with stationary covariance matrices at reasonable sampling rates. However, in image processing, the resolution can be considerably higher.
% As seen in Table \ref{table:optimizarion_time_comparison} results, for a low number of steps, where discretization error is significant and the spectral scheduler greatly improves performance, the time scale ranges from under a second to a few tens of seconds.  Over approximately $130$ diffusion steps, discretization errors become small, the overall structure of the scheduler remains unchanged, while its density increases.

% \input{Appendices/Appendix_Estimating_a_circulant_matrix}

\newpage

\end{document}